%% file: bmvc_final.tex
\documentclass{bmvc2k}

\usepackage{graphicx}
\usepackage{booktabs}

\usepackage[accsupp]{axessibility}  
\usepackage{graphicx}
\usepackage{amsmath}
\usepackage{amssymb}
\usepackage{booktabs}
\usepackage{indentfirst}
\usepackage{pifont}
\usepackage[capitalize]{cleveref}
\crefname{section}{Sec.}{Secs.}
\Crefname{section}{Section}{Sections}
\Crefname{table}{Table}{Tables}
\crefname{table}{Tab.}{Tabs.}

\usepackage{epsfig}
\usepackage{multirow}
\usepackage{makecell}
\usepackage{amsfonts}
\usepackage{xcolor}
\usepackage{enumitem}
\usepackage{array}
\usepackage{algorithm}
\usepackage{algpseudocode}
\usepackage[symbol]{footmisc}
\usepackage{blindtext}
\usepackage{graphicx}
\usepackage{mathtools}
\usepackage{amssymb}
\usepackage{dsfont}

\algrenewcommand\algorithmicindent{0.75em}

\def\eqref#1{(\ref{eq:#1})}
\def\eqlabel#1{\label{eq:#1}}
\def\figref#1{\ref{fig:#1}}
\def\figlabel#1{\label{fig:#1}}

\def\eqref#1{(\ref{eq:#1})}
\def\eqlabel#1{\label{eq:#1}}
\def\figref#1{\ref{fig:#1}}
\def\figlabel#1{\label{fig:#1}}

\def\xcomment#1{\textcolor[rgb]{.3,.3,.1}{\text{$/\!\!/$ {\em #1}}}}
\def\comment#1{\kern-1cm\xcomment{#1}}
\def\eqcomment#1{\kern-1cm\xcomment{#1}}

\def\m#1{\ensuremath{\mathtt{#1}}}

\def\v#1{\ensuremath{\mathbf{#1}}}

\def\Rmx#1#2{{\mathbb R}^{{#1}\times{#2}}}

\def\real{\mathbb{R}}

\def\tr{^\top}

\def\norm#1{\left\lVert#1\right\rVert}

\def\l2#1{\norm{#1}_2}

\newcommand{\centered}[1]{\begin{tabular}{l} #1 \end{tabular}}

\title{Depth-Guided Privacy-Preserving Visual Localization Using 3D Sphere Clouds}

\addauthor{Heejoon Moon}{wilko97@hanyang.ac.kr}{1}
\addauthor{Jongwoo Lee}{sanngu5@hanyang.ac.kr}{1}
\addauthor{Jeonggon Kim}{drgon22@hanyang.ac.kr}{2}
\addauthor{Je Hyeong Hong}{jhh37@hanyang.ac.kr}{$\ast$1,2}

\addinstitution{
 Department of Artificial Intelligence\\
 Hanyang Univeristy\\
 Seoul, Republic of Korea
}
\addinstitution{
 Department of Electronic Engineering\\
 Hanyang Univeristy\\
 Seoul, Republic of Korea
}

\runninghead{H. Moon, J. Lee, J. Kim, J.H. Hong}{3D Sphere Clouds}

\def\eg{\emph{e.g}\bmvaOneDot}

\def\etal{\emph{et al}\bmvaOneDot}

\usepackage{algorithm}
\usepackage{algpseudocode}
\usepackage{xparse}
\makeatletter
\NewDocumentCommand{\LeftComment}{s m}{%
  \Statex \IfBooleanF{#1}{\hspace*{\ALG@thistlm}}\(\triangleright\) #2}
\makeatother 

\begin{document}

\maketitle

\begin{abstract}
The emergence of deep neural networks capable of revealing high-fidelity scene details from sparse 3D point clouds has raised significant privacy concerns in visual localization involving private maps.
Lifting map points to randomly oriented 3D lines is a well-known approach for obstructing undesired recovery of the scene images, but these lines are vulnerable to a density-based attack that can recover the point cloud geometry by observing the neighborhood statistics of lines.
With the aim of nullifying this attack, we present a new privacy-preserving scene representation called \emph{sphere cloud}, which is constructed by lifting all points to 3D lines crossing the centroid of the map, resembling points on the unit sphere.
Since lines are most dense at the map centroid, the sphere cloud mislead the density-based attack algorithm to incorrectly yield points at the centroid, effectively neutralizing the attack. 
Nevertheless, this advantage comes at the cost of i) a new type of attack that may directly recover images from this cloud representation and ii) unresolved translation scale for camera pose estimation.
To address these issues, we introduce a simple yet effective cloud construction strategy to thwart new attack and 
 propose an efficient localization framework to guide the translation scale by utilizing absolute depth maps acquired from on-device time-of-flight (ToF) sensors.
Experimental results on public RGB-D datasets demonstrate sphere cloud achieves competitive privacy-preserving ability and localization runtime while not excessively compensating the pose estimation accuracy compared to other depth-guided localization methods.
\end{abstract}

\begin{figure}[t]
\centering
    \vspace{-5mm}
    \setlength{\tabcolsep}{0.1pt}
    \begin{tabular}{cccccccc} \\
    \subfigure{
    \setlength{\fboxsep}{0pt}%
    \setlength{\fboxrule}{0.1pt}%
    \centered{\fbox{\includegraphics[width=0.21\linewidth]{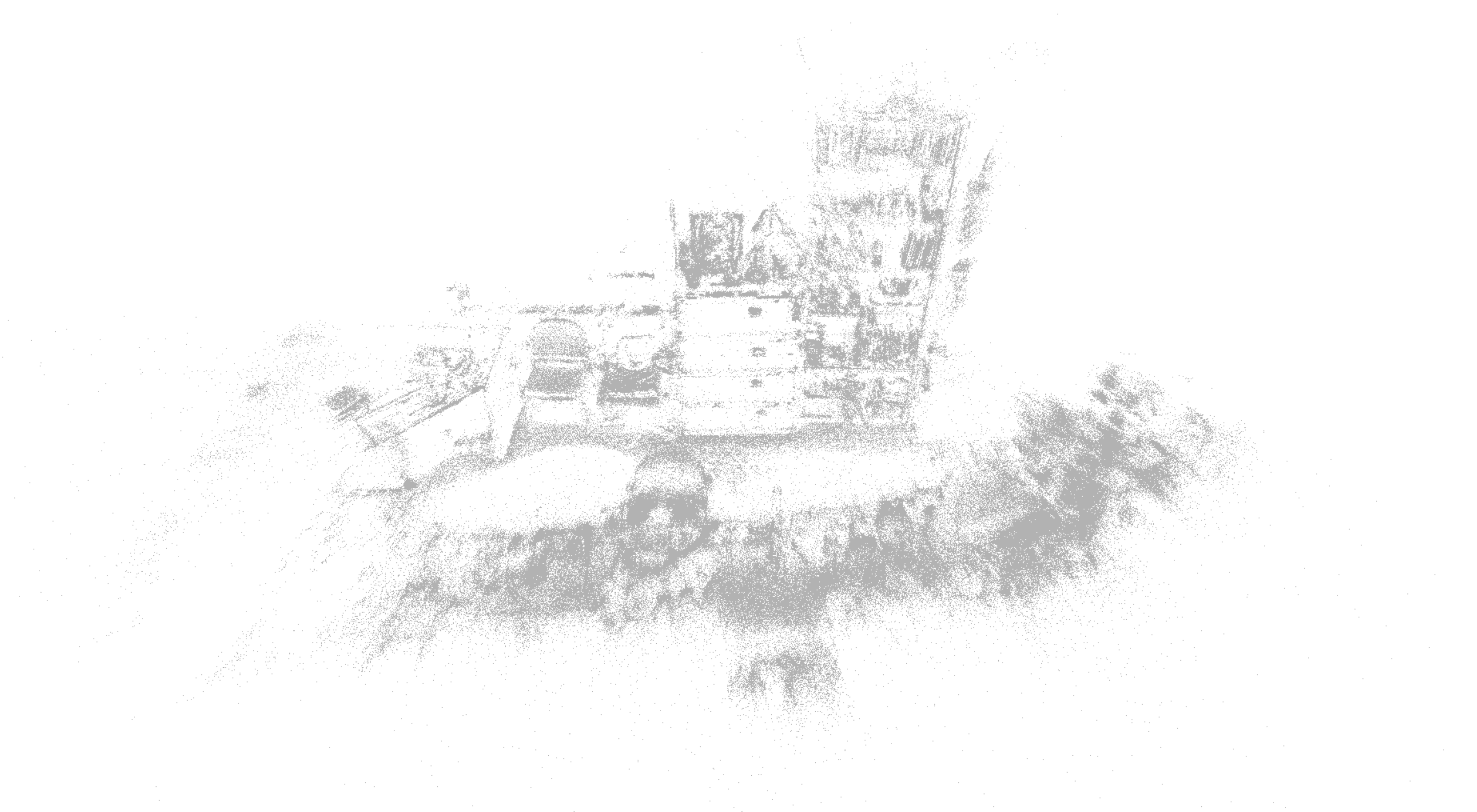}}}
    }
    &
    \subfigure{
    \setlength{\fboxsep}{0pt}%
    \setlength{\fboxrule}{0.1pt}%
    \centered{\fbox{
    \includegraphics[width=0.208\linewidth]{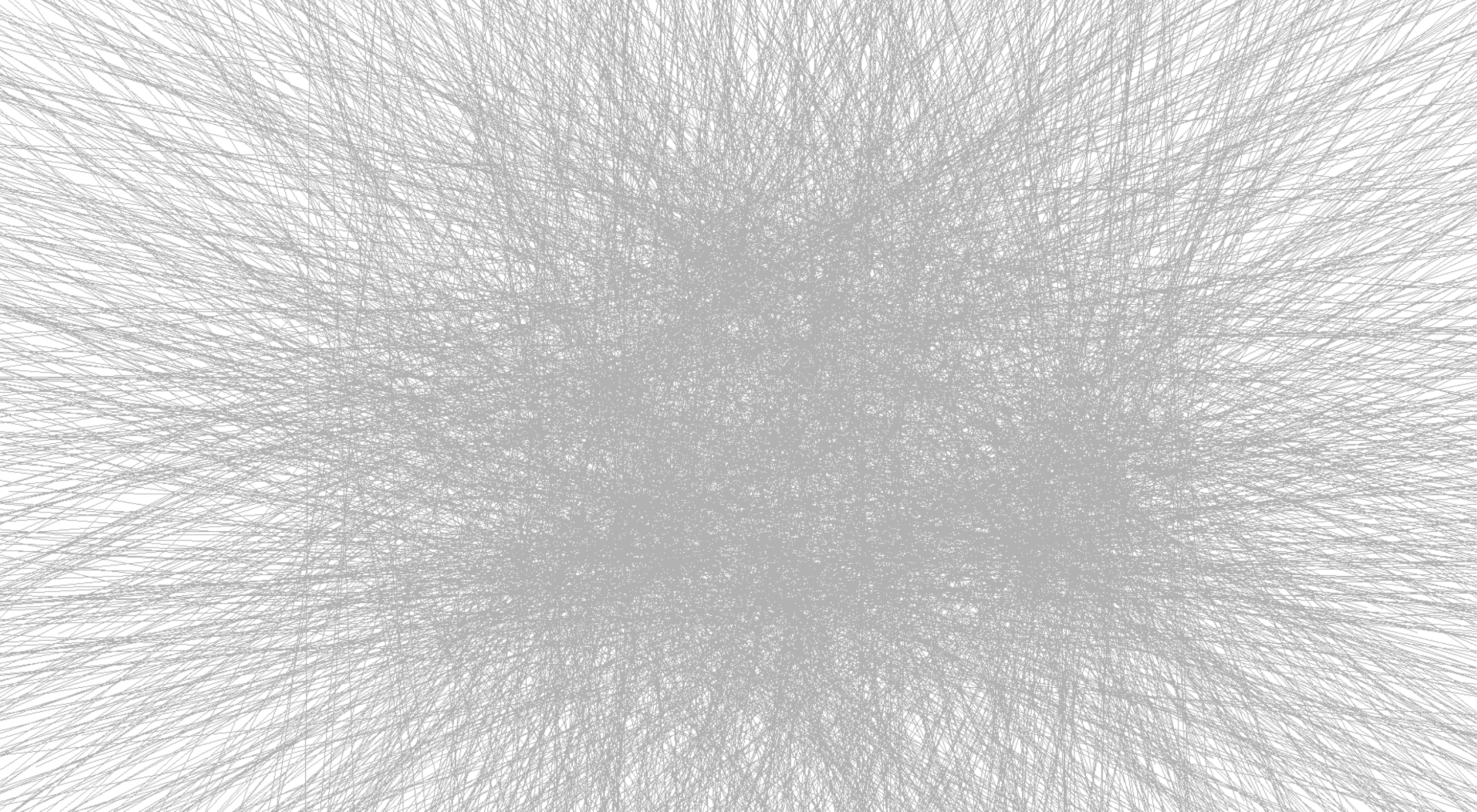}}}
    }
    &
    \subfigure{
    \setlength{\fboxsep}{0pt}%
    \setlength{\fboxrule}{0.1pt}%
    \centered{\fbox{\includegraphics[width=0.21\linewidth]{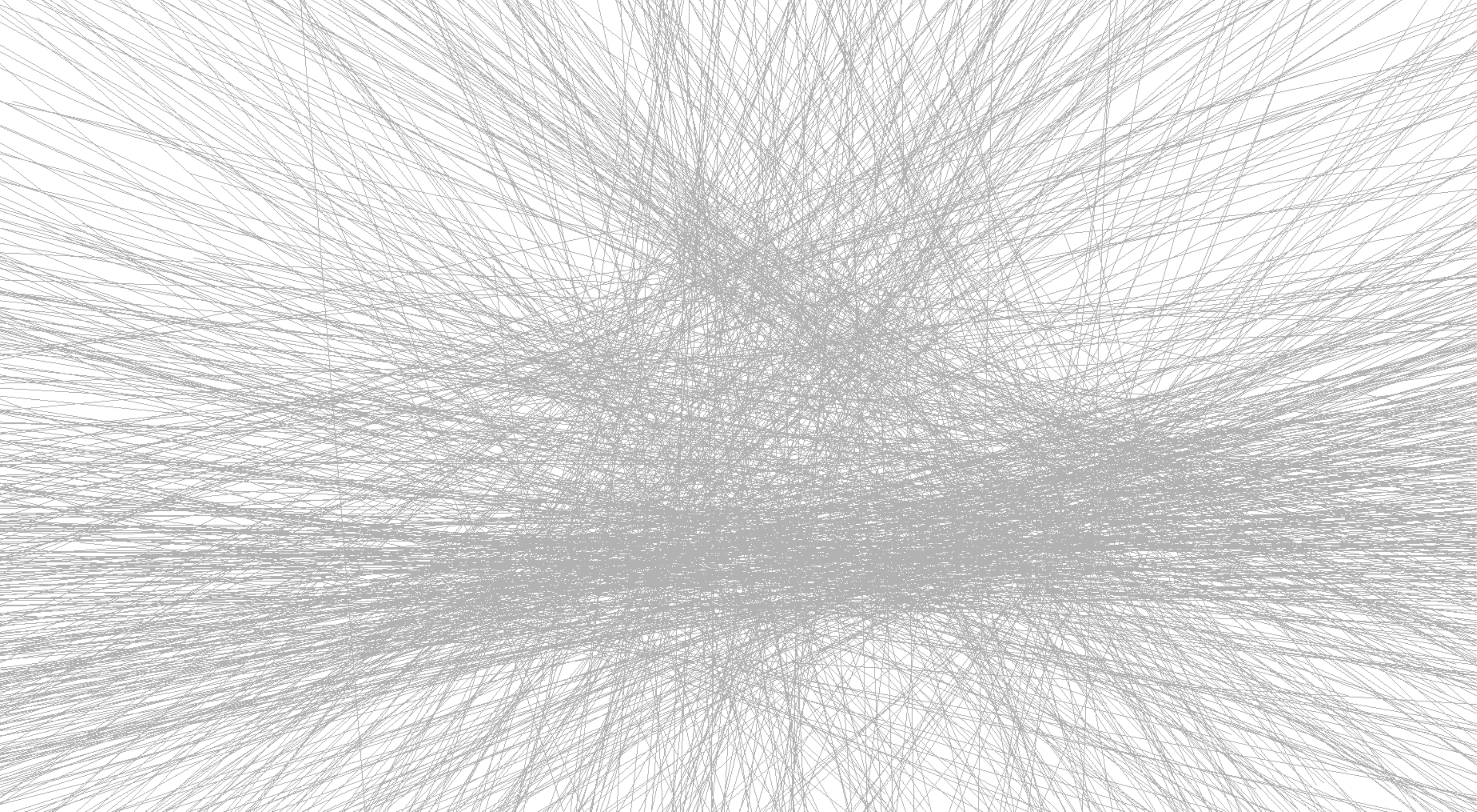}}}
    }
    &
    \subfigure{
    \setlength{\fboxsep}{0pt}%
    \setlength{\fboxrule}{0.1pt}%
    \centered{\fbox{\includegraphics[width=0.21\linewidth]{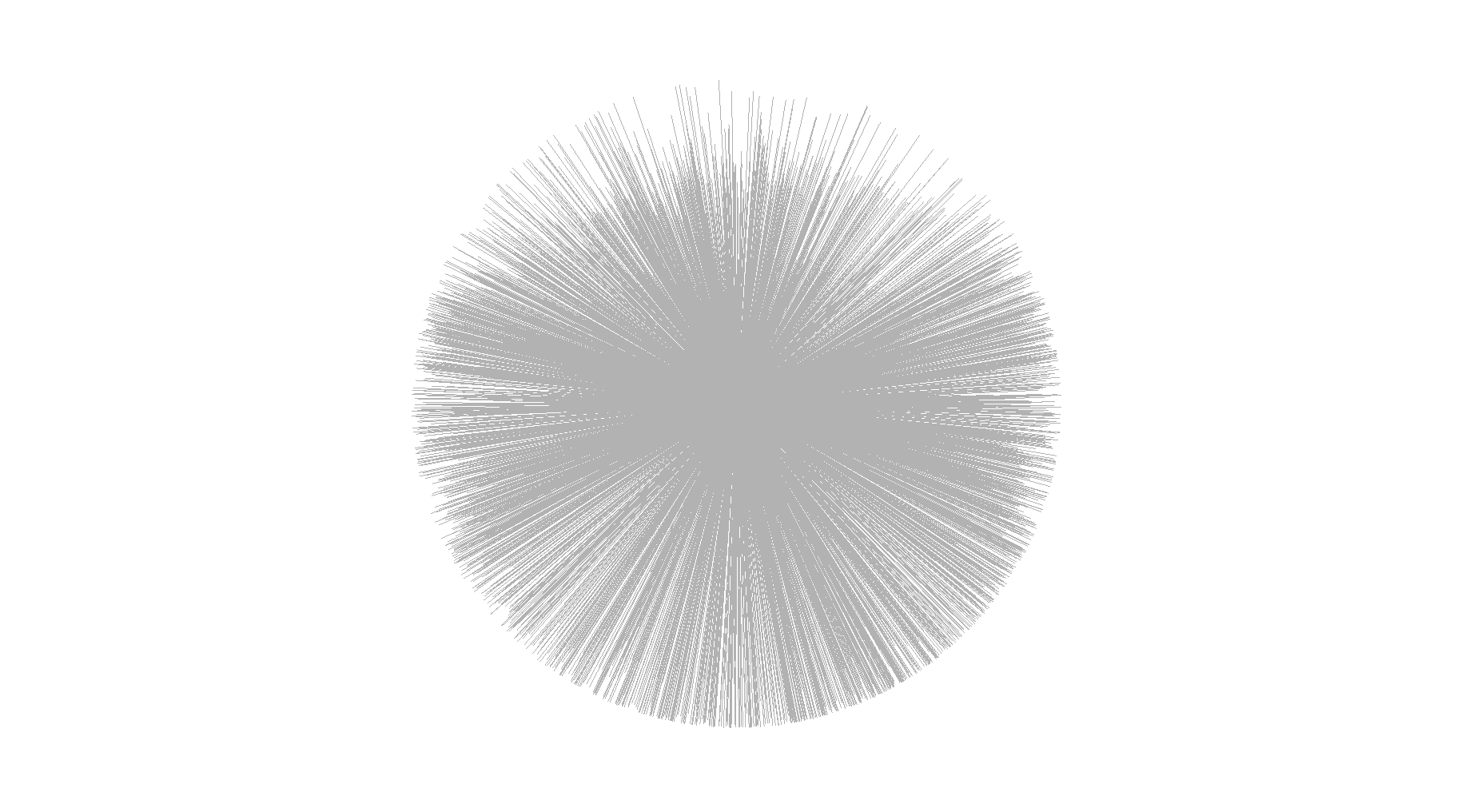}}}}
    \vspace{-4mm}
    \\
    \subfigure{
    \setlength{\fboxsep}{0pt}%
    \setlength{\fboxrule}{0.1pt}%
    \centered{\fbox{\includegraphics[width=0.21\linewidth]{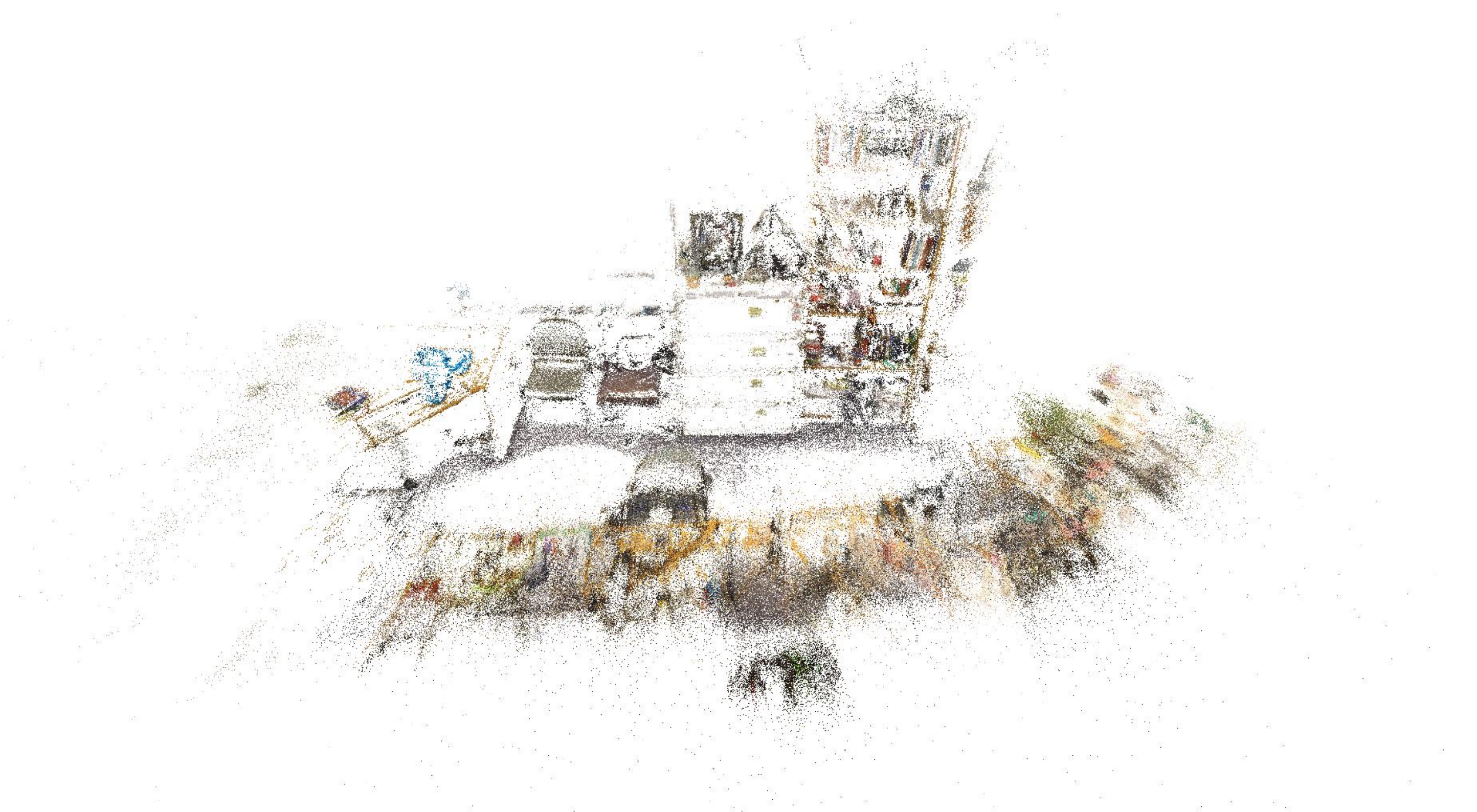}}}
    }
    &
    \subfigure{
    \setlength{\fboxsep}{0pt}%
    \setlength{\fboxrule}{0.1pt}%
    \centered{\fbox{\includegraphics[width=0.21\linewidth]{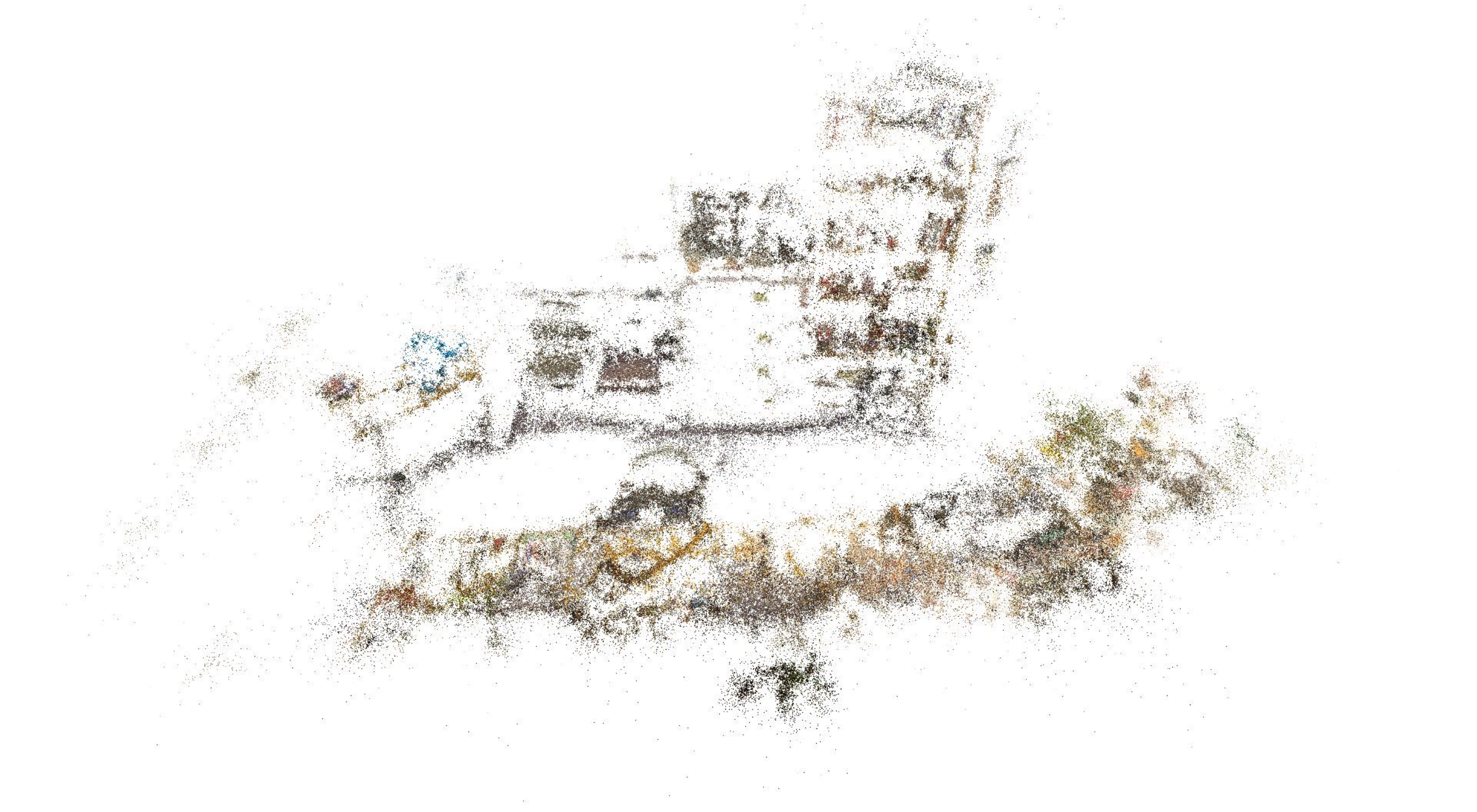}}}
    }
    &
    \subfigure{
    \setlength{\fboxsep}{0pt}%
    \setlength{\fboxrule}{0.1pt}%
    \centered{\fbox{\includegraphics[width=0.21\linewidth]{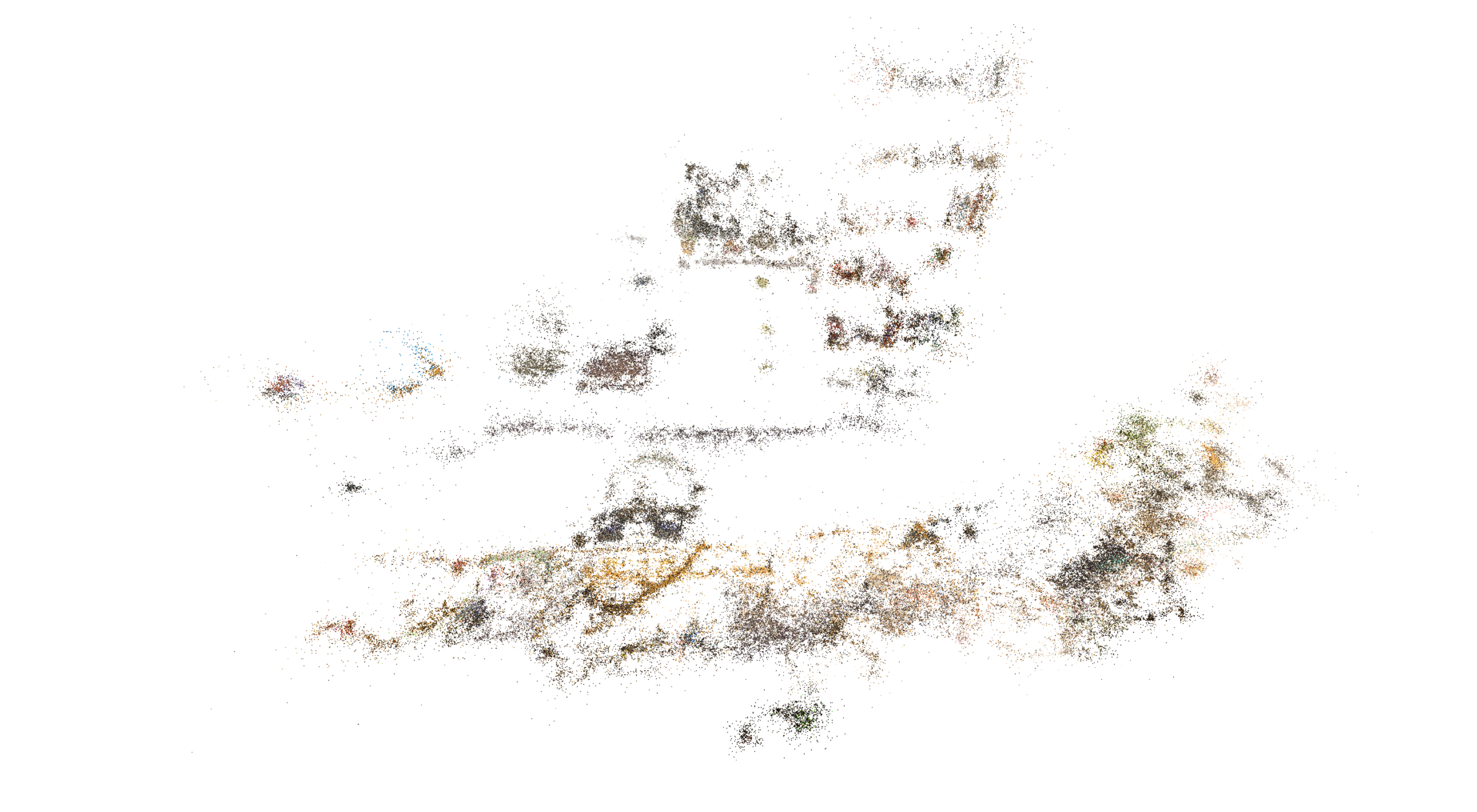}}}
    }
    &
    \subfigure{
    \setlength{\fboxsep}{0pt}%
    \setlength{\fboxrule}{0.1pt}%
    \centered{\fbox{\includegraphics[width=0.21\linewidth]{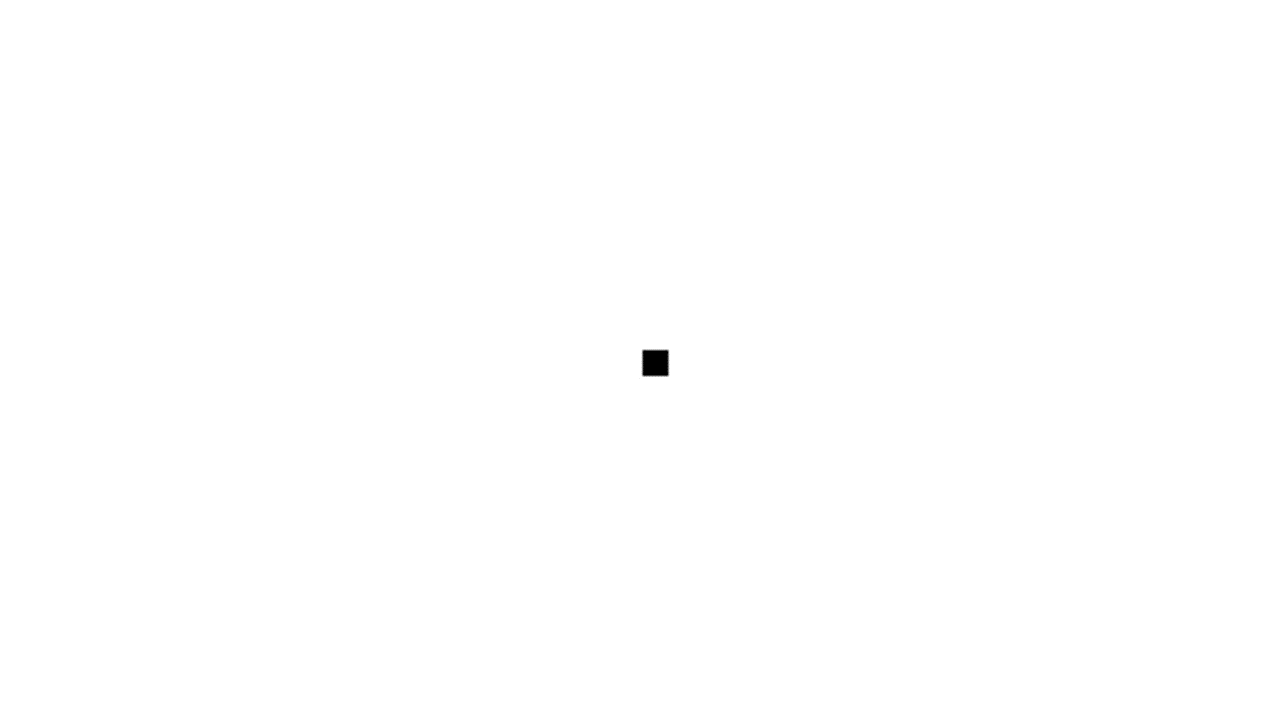}}}}
    \vspace{-4mm}
    \\
    \setcounter{subfigure}{0}    
    \subfigure[Ground Truth]{
    \setlength{\fboxsep}{0pt}%
    \setlength{\fboxrule}{0.1pt}%
    \centered{
    \hspace{-2mm}
    \fbox{\includegraphics[width=0.1\linewidth]{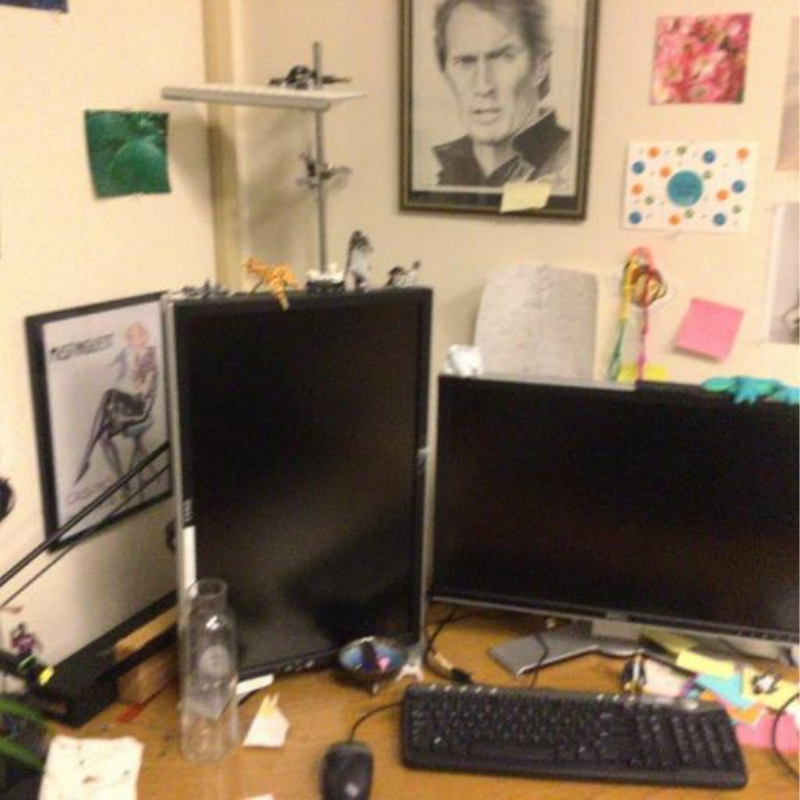}}
    \fbox{\includegraphics[width=0.1\linewidth]{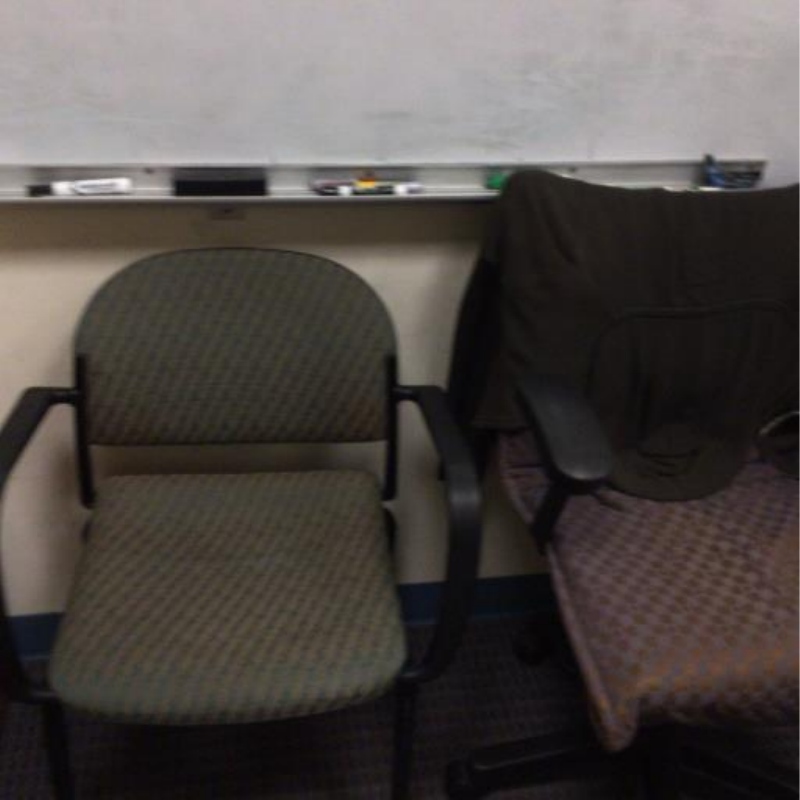}}
    }} 
    &
    \subfigure[ULC~\cite{speciale2019privacy}]{
    \setlength{\fboxsep}{0pt}%
    \setlength{\fboxrule}{0.1pt}%
    \centered{
    \fbox{\includegraphics[width=0.1\linewidth]{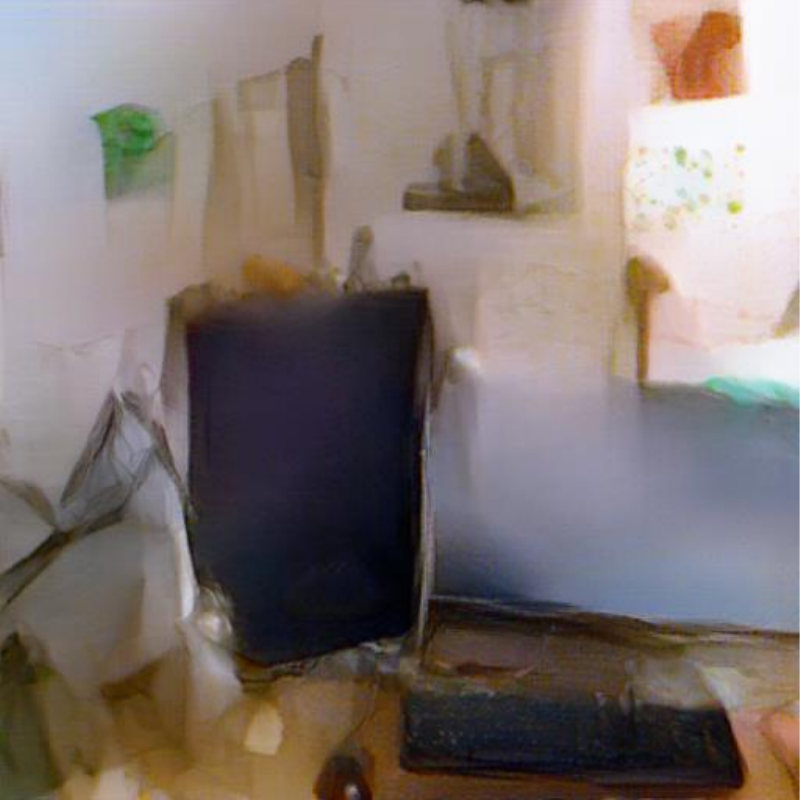}}
    \fbox{\includegraphics[width=0.1\linewidth]{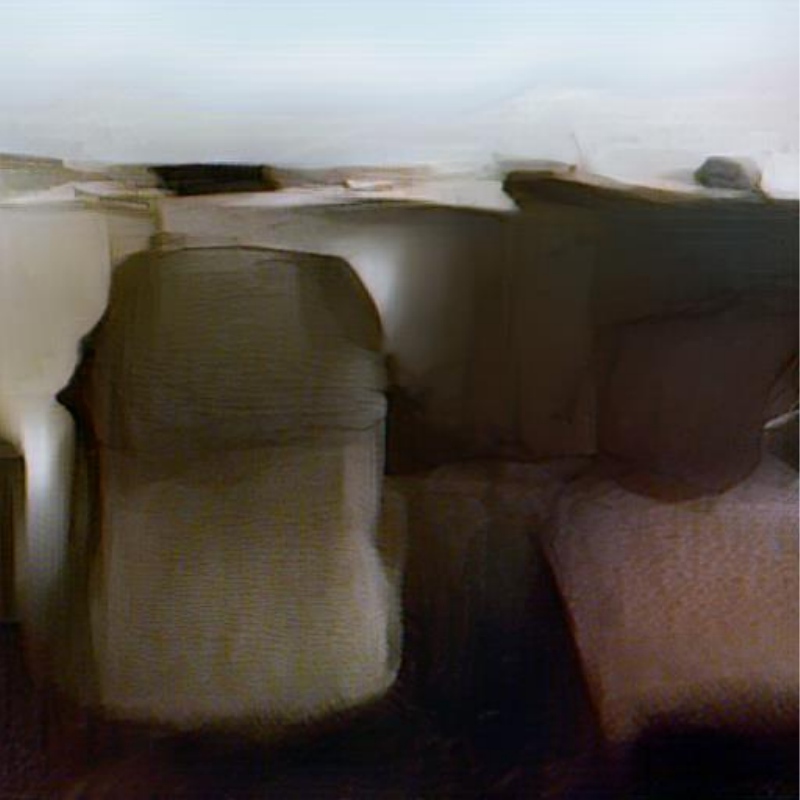}}
    }} 
    &
    \subfigure[PPL~\cite{lee2023ppl}]{
    \setlength{\fboxsep}{0pt}%
    \setlength{\fboxrule}{0.1pt}%
    \centered{
    \fbox{\includegraphics[width=0.1\linewidth]{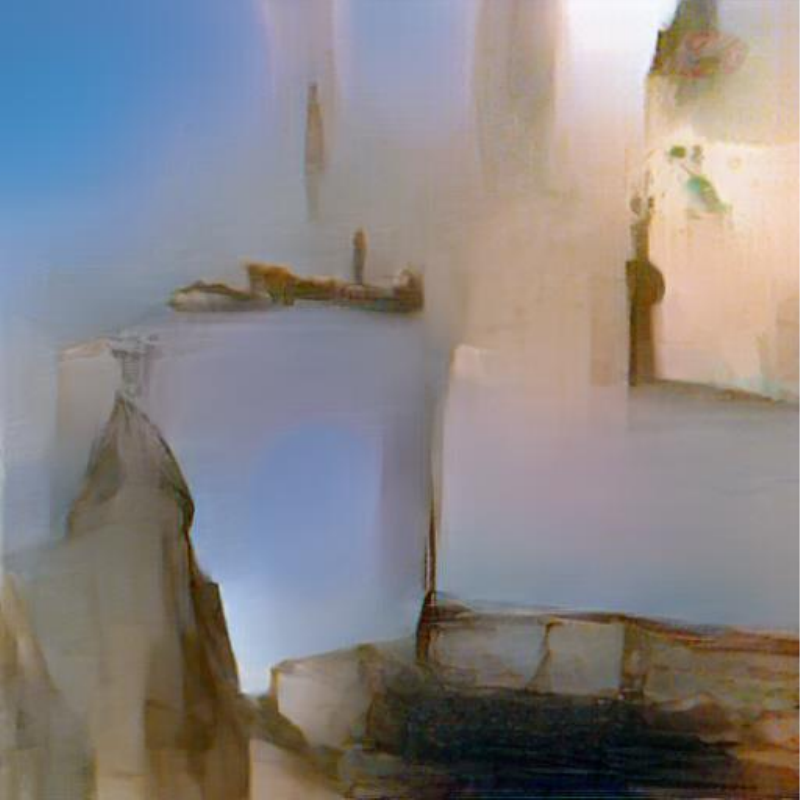}}
    \fbox{\includegraphics[width=0.1\linewidth]{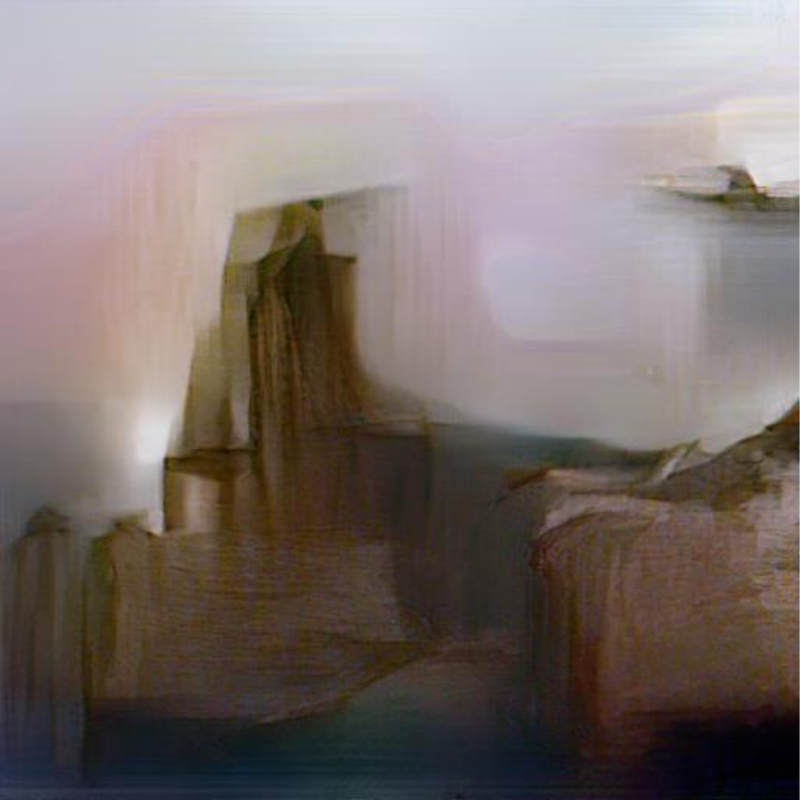}}
    }} 
    &
    \subfigure[\textbf{Sphere cloud (ours)}]{
    \setlength{\fboxsep}{0pt}%
    \setlength{\fboxrule}{0.1pt}%
    \centered{
    \fbox{\includegraphics[width=0.1\linewidth]{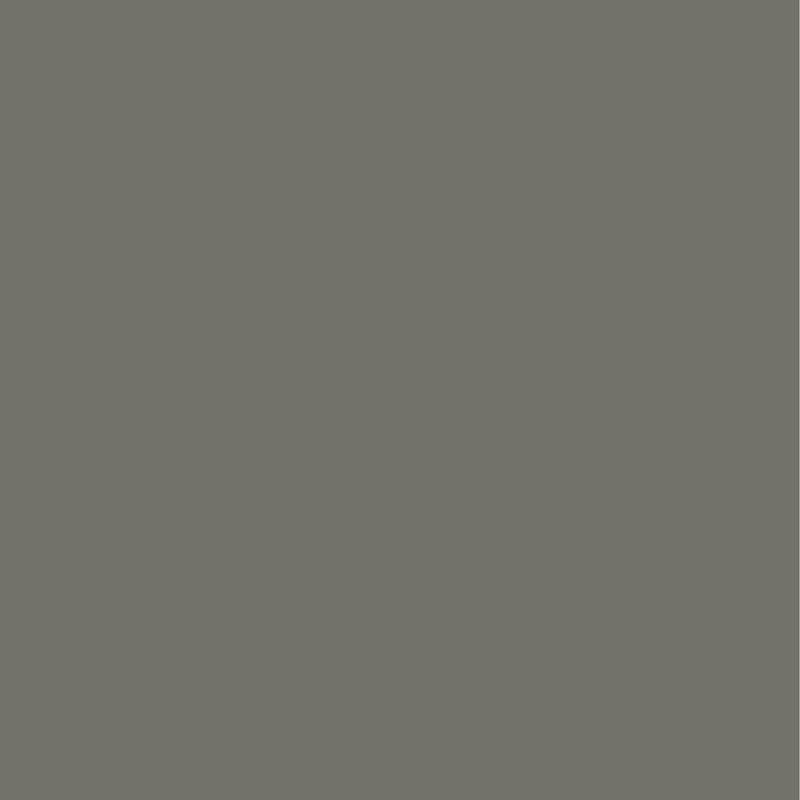}}
    \fbox{\includegraphics[width=0.1\linewidth]{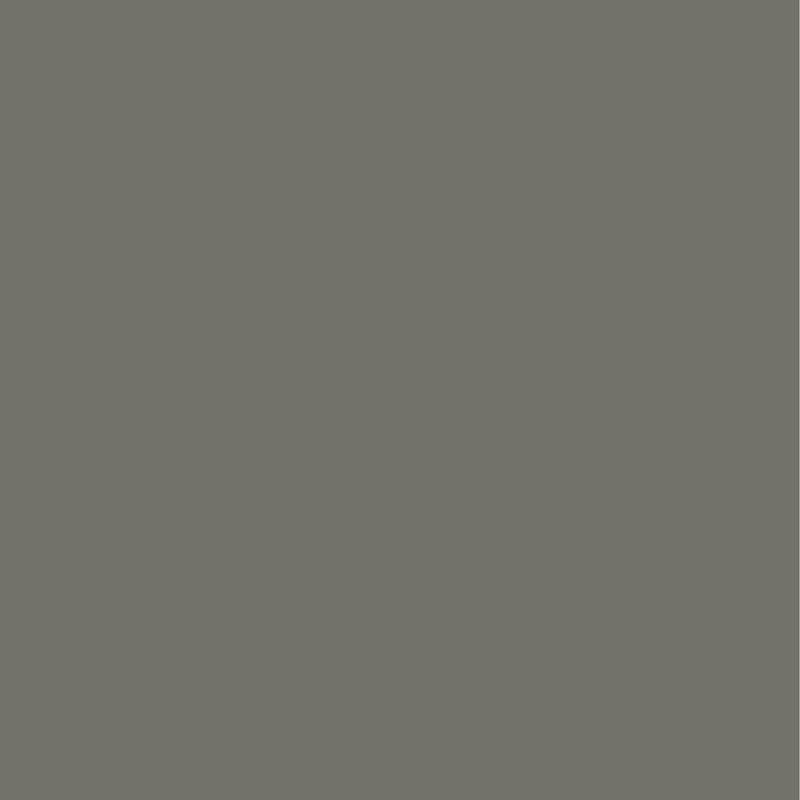}}
    }}
    \end{tabular}
    \vspace{-1mm}
    
    \caption{
    (\emph{Top}) visualization of different 3D scene representations (\emph{Office1 manolis}  from 12 Scenes~\cite{valentin2016energy}). ULC~\cite{speciale2019privacy} denotes uniform line cloud and PPL~\cite{lee2023ppl} denotes paired-point lifting.
    (\emph{Middle}) recovered 3D points from the geometry revealing attack~\cite{chelani2023privacy}.
    (\emph{Bottom}) images reconstructed via InvSfM~\cite{pittaluga2019revealing} using the recovered 3D points.
    Since our sphere cloud always results in points  recovered at the sphere centre, the recovered scene images are blank.
    }
    \figlabel{inversion_results}
    \vspace{-6mm}
\end{figure}

\section{Introduction}
\label{sec:intro}
\vspace{-2mm}
Visual localization, which refers to the task of estimating the 6-DOF camera pose from an input image, is a key computation in autonomous driving, extended reality (XR)~\cite{arth2009wide,castle2008video} and robotics~\cite{castle2008video,lynen2015get,mur2015orb}.
While a full taxonomy of localization algorithms exists in the literature, the mainstream pipeline to this date comprises the following steps: i) build a sparse 3D point cloud of the scene via structure-from-motion (SfM)~\cite{schoenberger2016sfm}, ii) match keypoints of the query image against the features in the point map and iii) perform perspective-$n$-points to obtain the camera pose.
The point cloud and descriptors are either stored on the server for cloud-based localization or distributed to the client (\eg a robot or XR device) for real-time localization.

Until recently, it was perceived that these point maps, which may often comprise a private or confidential area/objects, are usually sparse enough to discourage any attempt by curious intruders or malicious clients to reveal scene details from the 3D points. 
Nevertheless, the work of Pittaluga~\etal~\cite{pittaluga2019revealing} called InvSfM showed possibility of recovering high-fidelity scene images from the sparse point cloud, raising significant privacy concerns when using the barebone point maps for localization.
Currently, one of the most widely known approaches to mitigating this issue is to conceal the point map as a \emph{line cloud}, which is constructed by lifting each point to a 3D line~\cite{speciale2019privacy,lee2023ppl}, subsequently hiding the point locations and disabling direct image synthesis using InvSfM.
Unfortunately, this line of works is potentially vulnerable to the density-based attack~\cite{chelani2021privacy} (see Fig.~\figref{inversion_results} for an example), which can effectively reverse the 3D lines back to points using the neighbourhood statistics of the lines.
Providing a full defense against such attack is yet an unaccomplished goal and serves as our main motivation.

In this work, we present a new privacy-preserving scene representation called \emph{sphere cloud} in an effort to nullify  aforementioned geometry-revealing attack~\cite{chelani2021privacy}.
The sphere cloud, which is simply constructed by lifting points to 3D lines passing through the centroid of the point cloud (which can be viewed as points on the unit sphere centered at the map centroid), has the advantage of completely disabling the geometry-revealing attack~\cite{chelani2021privacy} by forcing the neighbourhood line statistics to lead to a degenerate point recovery (see Sec.~\ref{sec:proposed_method} for details).
Unfortunately, employing a sphere cloud for privacy-preserving visual localization is not straightforward due to two issues, that i) a new type of attack (discussed in Sec.~\ref{sec:proposed_method}) may partly reveal scene details about the map centroid and ii) the camera pose can only be retrieved up to unknown scale.
We tackle the first issue by proposing a simple effective strategy to hinder new attack and address the second issue by utilizing calibrated depth maps that can be easily acquired from an on-device time-of-flight (TOF) sensor to resolve the translation scale.

Our contributions in this work are summarized as follows:
\begin{itemize}[noitemsep, nolistsep]
\item a novel privacy-preserving scene representation called \emph{sphere cloud} which completely avoids known density-based attack and disables recovery of the point cloud geometry,
\item a simple yet effective strategy based on cloud sparsification and descriptor augmentation to thwart a new type of attack from breaching the sphere cloud, and
\item to the best of our knowledge, the first privacy-preserving framework to leverage raw depth information from a ToF sensor for efficient camera pose estimation.
\end{itemize}

\section{Related work}
\label{sec:related_work}
\vspace{-2mm}
\paragraph{Revealing private scene details from sparse point cloud}
The first method that succeeded in revealing high-fidelity scene details from a sparse point cloud was proposed by Pittaluga~\etal~\cite{pittaluga2019revealing}, in which a network called InvSfM based on cascaded U-Net~\cite{ronneberger2015u} is employed to reconstruct a scene image from a set of inputs including 2D locations of the projected 3D points as well as corresponding depths, RGB values and SIFT descriptors.
As noted in~\cite{speciale2019privacy}, this raised alarms as any confidential maps (\eg inside a factory) or public maps with temporary private objects inadvertently obtained by a user can now be revealed in detail.
While extensions of this work have been proposed to reconstruct images without keypoint descriptors~\cite{song2020deep} or with different types of descriptors~\cite{dangwal2021mitigating}, the pretrained InvSfM model is still widely used as the baseline for analyzing the privacy-preserving capability~\cite{chelani2021privacy,lee2023ppl,pan2023privacy,pietrantoni2023segloc,moon2024raycloud}.

\vspace{-2mm}
\paragraph{Privacy-preserving 3D scene representations}
With the aim of obstructing use of InvSfM for scene image reconstruction, Speciale~\etal~\cite{speciale2019privacy} proposed line cloud in which each point is represented as a randomly oriented 3D line passing through the original point, intending to conceal the scene geometry by introducing ambiguities in the point locations.
While this was initially perceived as an effective strategy to block attempts for revealing scene details and extended to simultaneous localization and mapping (SLAM)~\cite{shibuya2020privacy}, 
it was later shown by Chelani~\etal~\cite{chelani2021privacy} that line clouds with uniformly distributed line directions are vulnerable to a density-based geometry-inversion attack that can accurately recover the scene points (more details at the end of Sec.~\ref{sec:related_work}), from which the scene images can subsequently be revealed (see Fig.~\figref{inversion_results} (b)).
While this weakness was addressed in~\cite{lee2023ppl} by drawing 3D lines through random pairs of 3D points to induce combinatorial complexity in point cloud recovery, it is not fully impervious to the geometry-revealing attack~\cite{chelani2021privacy} as observed in the second row of Fig.~\figref{inversion_results} (c).
The most similar work to our approach is \cite{moon2024raycloud}, in which all 3D lines intersect through one of two pre-defined 3D locations to reduce the effectiveness of the density-based attack~\cite{chelani2021privacy}.
However, this method does not theoretically guarantee full defense against~\cite{chelani2021privacy} and moreover, it can be vulnerable to another type of attack involving direct image synthesis at the intersections (see Sec.~\ref{sec:center_inversion}).

\begin{figure}[t]
    \centering
    \subfigure[][Uniform line cloud (ULC)~\cite{speciale2019privacy}\figlabel{ulc_peak}]{
        \includegraphics[width=0.375\linewidth]{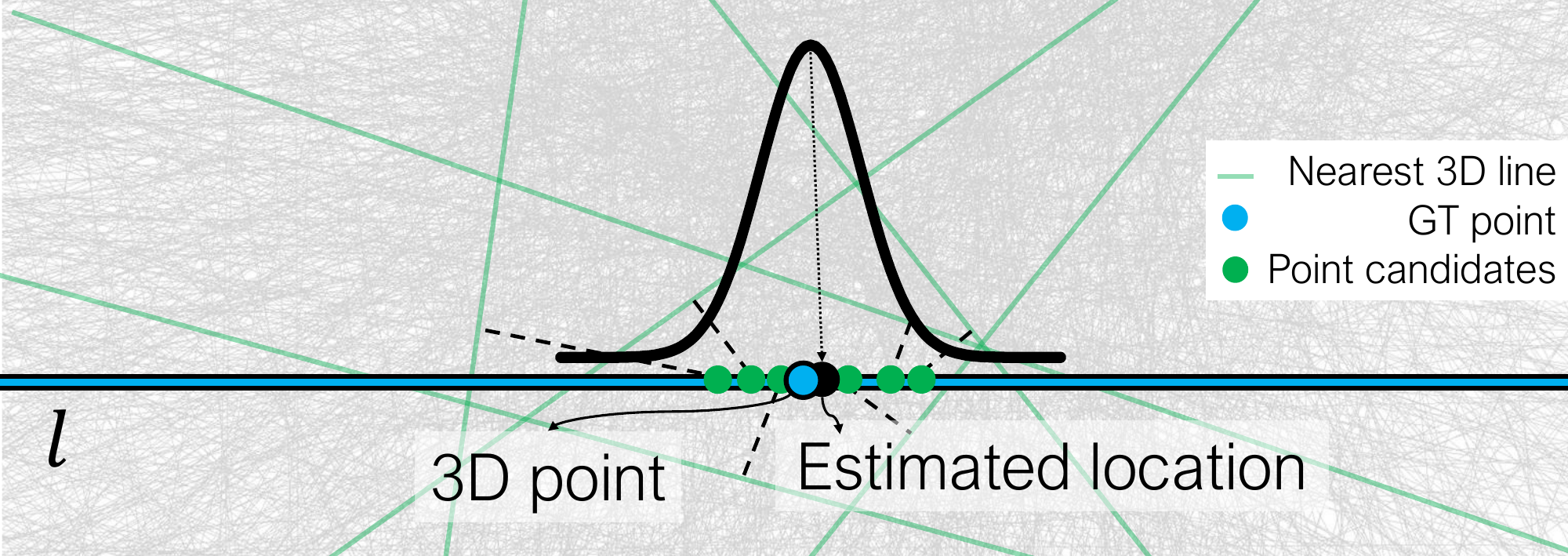}
    }
    \subfigure[][Sphere cloud (ours)\figlabel{sphere_peak}]{
        \includegraphics[width=0.375\linewidth]{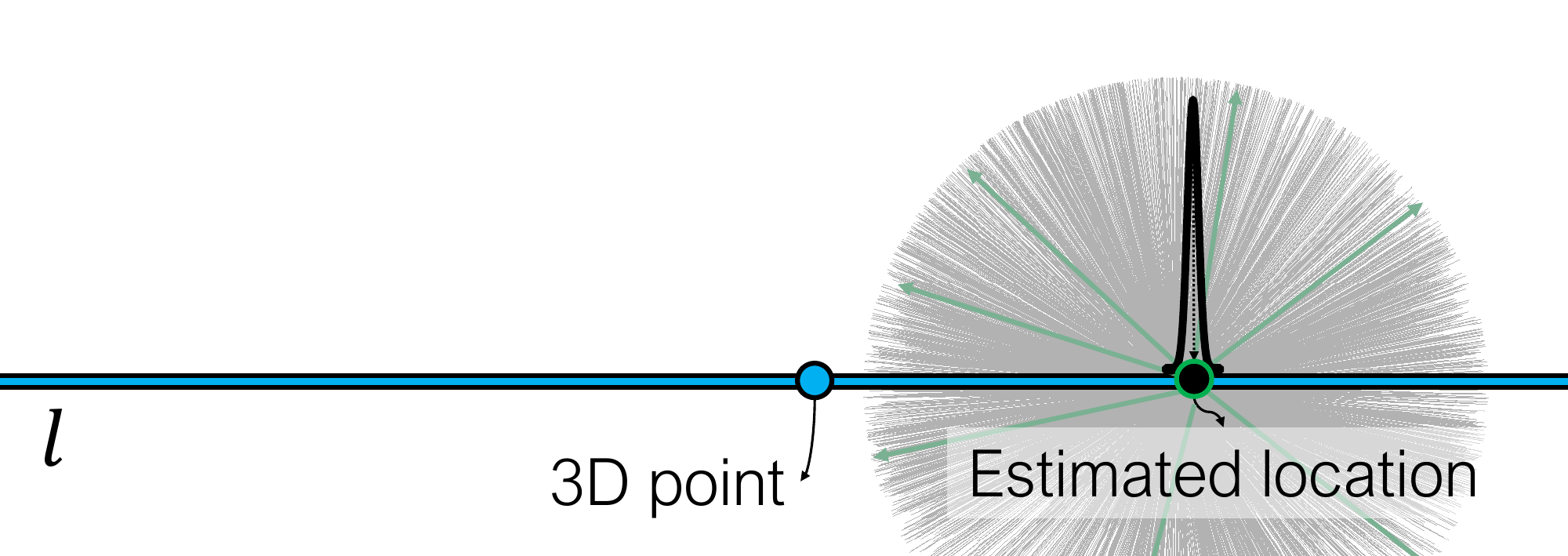}
    }
    \vspace{-1mm}    
    \caption{
    A motivating illustration for the sphere cloud.
    As shown in (a), the density-based geometry-revealing attack in~\cite{chelani2021privacy} recovers the point location of each line by constructing a histogram of point candidates on the line that are closest to $K$-nearest neighbouring lines and finding the peak of the histogram.
    While this method often yields good point estimates for uniform line clouds, all lines from the sphere cloud in (b) intersect at the map centroid, and consequently the points estimated via peak finding are incorrectly recovered at the centroid. 
    }
    \figlabel{peak_finding}
    \vspace{-6mm}
\end{figure}
Other types of scene representations include the work of Geppert~\etal~\cite{geppert2022privacy}, in which the sparse point cloud is divided into three 1D partial maps stored in separate servers for enhanced security at the cost of reduced off localization accuracy and runtime.
Pan~\etal~\cite{pan2023privacy} proposed to pair up 3D points and permute coordinates between each pair of points to disallow meaningful reconstruction of the scene while enabling accurate localization, but the permutation process incurs combinatorial search over the correct camera pose, drastically slowing the localization speed. 
Overall, these approaches are not susceptible to the geometry-inversion attack~\cite{chelani2021privacy} but they are computationally much more involved than line cloud-based approaches, impeding their practical use for efficient real-time localization.
Currently, no representation can fully bypass above attack while maintaining real-time localization speed.

\vspace{-4mm}
\paragraph{Localization using 3D line clouds} 
The classic absolute camera pose estimation problem involving a 3D point cloud can be solved with an efficient perspective-$n$-point (pnP) solver~\cite{persson2018lambda, ding2023revisiting, ke2017efficient} derived from the 2D-3D point-to-point constraints.
In contrast, line clouds can only introduce weaker constraints between 2D points and 3D lines.
Speciale~\etal~\cite{speciale2019privacy} noted absolute pose estimation with line clouds is identical to the problem of generalized relative pose estimation, and proposed a perspective-6-lines (p6L) algorithm based on the minimal solver for generalized relative pose estimation~\cite{henrikstewenius2005solutions}.
Due to the intrinsic flexibility of generalized cameras, p6L yields 64 pose candidates for each of six 2D point-3D line correspondences from which the correct solution needs to be identified via  geometric verification.
Hence, employing p6L contributes to much increased runtime when compared with p3P that only yields 4 pose candidates for each of three 2D-3D point correspondences.

\vspace{-4mm}
\paragraph{Geometry-revealing density-based attack for line clouds}
\label{sec:chelani_explanation}
Chelani \etal~\cite{chelani2021privacy} proposed an algorithm for recovering the original points from a uniform 3D line cloud~\cite{speciale2019privacy}.
This work is motivated by the empirical observation that for any two distinct 3D points and their lifted 3D lines, the points on the lines which are closest to the counterpart lines are likely to be in the proximity of the original 3D points.
As shown in Fig.~\figref{ulc_peak}, this result is extended to consider the closest points to multiple neighbouring lines as point candidates (green) for each line.
The final point location is estimated by finding the peak (black) of the histogram of these point candidates which is usually close to ground truth (blue) as long as the line directions are uniformly distributed.
We neutralize this attack by essentially breaking this assumption as will be described in Sec.~\ref{sec:proposed_method}, leading to an incorrect recovery (see Fig.~\figref{sphere_peak}).

\begin{figure}[t]
    \centering
    \includegraphics[width=0.7\linewidth]{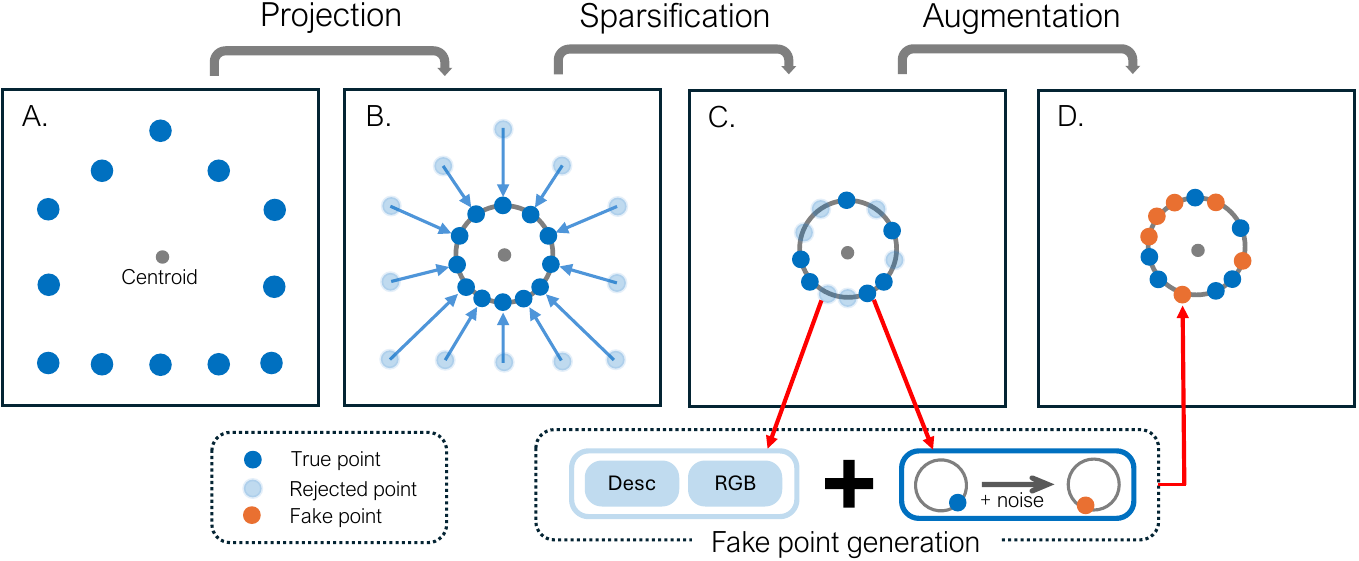}
    \vspace{-3mm}
    \caption{
    An overview of our complete strategy for constructing a sphere cloud.
    (A) we find the centroid of the sparse 3D point map.
    (B) we create a basic sphere cloud by projecting the 3D points onto the unit sphere centred at the map centroid.
    (C)  we discard a portion of sphere points from the sphere cloud but keep their RGB values and SIFT descriptors.
    (D) we generate fake points around the remaining points with their RGB values and SIFT descriptors recycled from the rejected points. 
    Since the basic construction from (A)+(B) may be prone to a new attack based on direct image synthesis, we enhance the strategy through (C)+(D).
    }
    \figlabel{overall_pipeline}
    \vspace{-6mm}
\end{figure}

\section{Sphere cloud}
\label{sec:proposed_method}
\paragraph{Motivations}
As shown in Fig.~\figref{ulc_peak}, \cite{chelani2021privacy} recovers the 3D points by constructing a histogram of point candidates for each 3D line (i.e. a set of points on the line each of which is closest to one of the neighboring lines) and finding the peak of this histogram.
Now, if all lines are lifted to meet at a single point $\v c\in\real^3$, then the point candidates for each line will always be located at the intersection point as any two lines are the closest at $\v c$.
Consequently, the peak of the histogram is always at $\v c$, leading to a degenerate recovery and thereby voiding the attack (see Fig.~\figref{sphere_peak}).
This motivates us to have all lines lifted to meet at a single point.

Unfortunately, the above representation is not sufficient to yield a unique camera rotation.
Since the 3D lines intersecting at a single point can be viewed as rays from a virtual camera centered at the intersection point (see Fig.~\figref{vp3p}), estimating camera pose from these lines resembles the relative pose estimation problem between the query camera and virtual camera (also noted in~\cite{chelani2021privacy}).
Out of 4 possible configurations~\cite{hartley1998cheirality} between the two cameras, we can choose the correct solution only if the cheirality is enforced on the lifted 3D lines (more details in~\cite{supmat}).
This serves as motivation for storing each line $\v l_i$ as a point $\hat{\v x}_i \in S^2$ on the unit sphere centered at $\v c$ such that the original point is always along the positive direction of\,$\hat{\v x}_i$.

\vspace{-2mm}
\subsection{Basic construction procedure and limitations}
\label{sec:naive_construction}
\vspace{-2mm}
Constructing a basic 3D sphere cloud involves two straightforward steps (steps A and B in Fig.~\figref{overall_pipeline}).
First, we set the intersection point as the mean centroid of the 3D point cloud to ensure the resulting line directions are roughly evenly distributed for stable localization (see~\cite{supmat} for discussions).
Second, we project all 3D points onto the unit sphere centred at the map centroid to create a basic 3D sphere cloud. 
Unfortunately, there are two major issues with deploying this basic implementation for privacy-preserving visual localization.

\vspace{-5mm}
\paragraph{Possible attack based on direct image synthesis}
\label{sec:center_inversion}
While the sphere cloud does not leak any scene geometry, an intruder may seek to directly reveal images from the sphere cloud.
The simplest approach is to project the sphere points to a virtual image plane and feed the projected points and their descriptors to InvSfM~\cite{pittaluga2019learning}.
Although the intruder is confined to viewpoints about the map centroid, Fig.~\figref{pseudo_gt} shows this attack can partly reveal the scene.
We aim to thwart this attack through an enhanced construction strategy in Sec.~\ref{sec:enhanced_sphere_cloud}.

\vspace{-2mm}
\begin{figure}[t]
    \centering
    \subfigure{
        \includegraphics[width=0.12\linewidth]{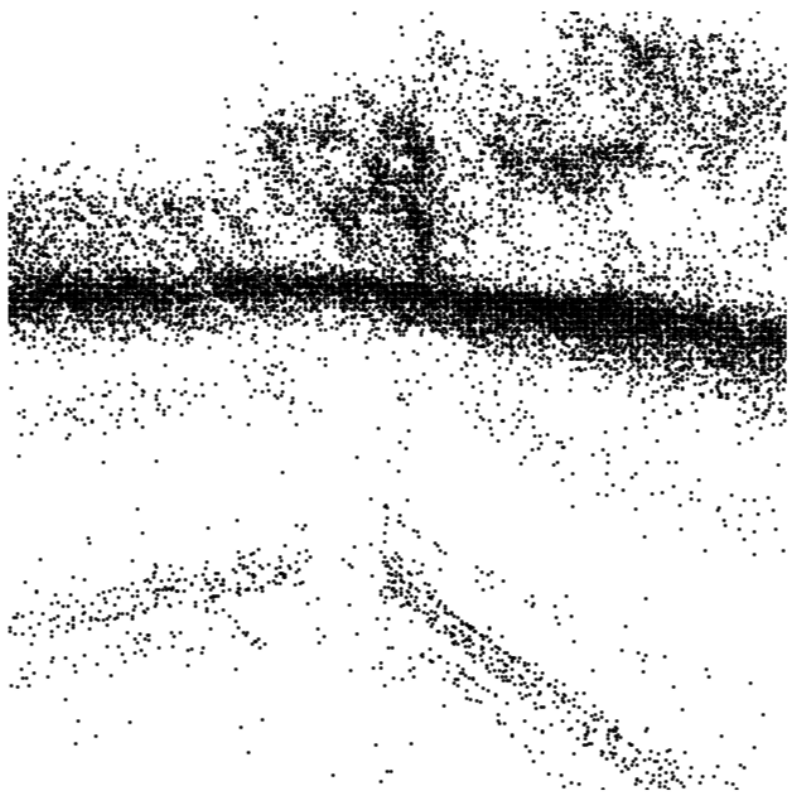}
    }
    \subfigure{
        \includegraphics[width=0.12\linewidth]{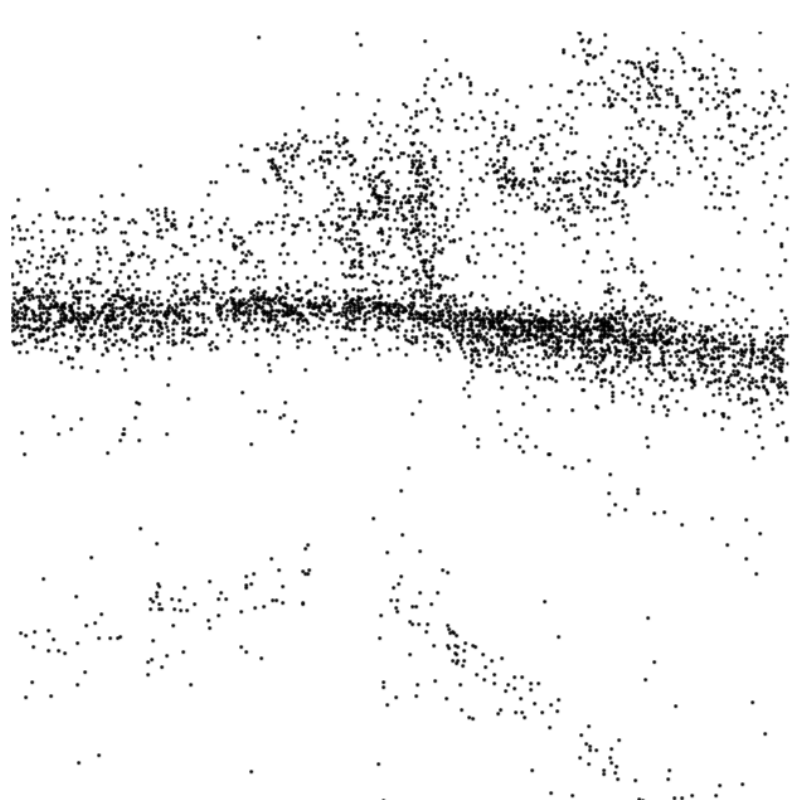}
    }
    \subfigure{
        \includegraphics[width=0.12\linewidth]{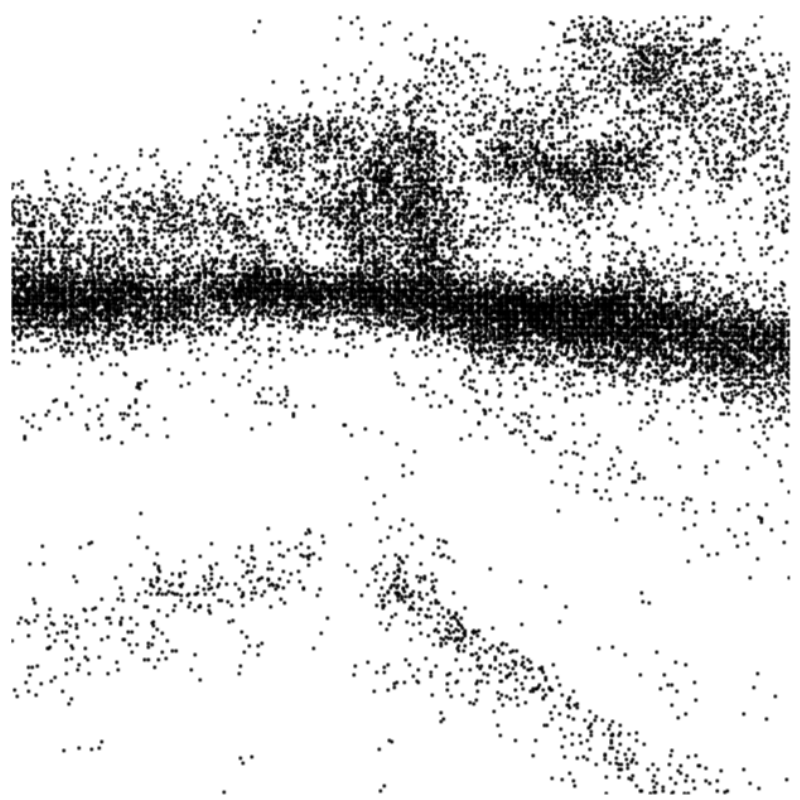}
    }
    \subfigure{
        \includegraphics[width=0.12\linewidth]{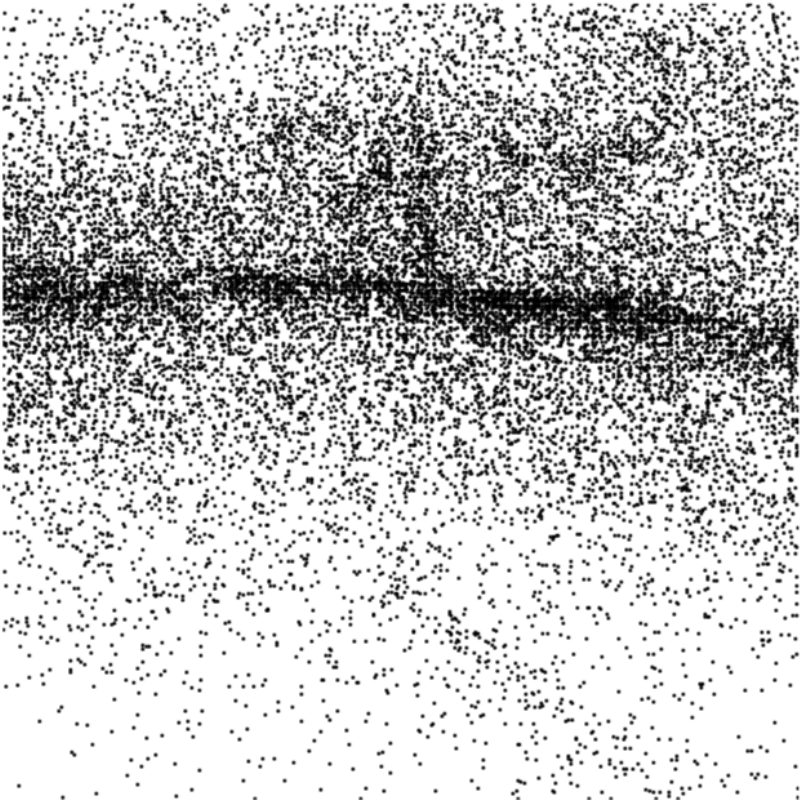}
    }
    \subfigure{
        \includegraphics[width=0.12\linewidth]{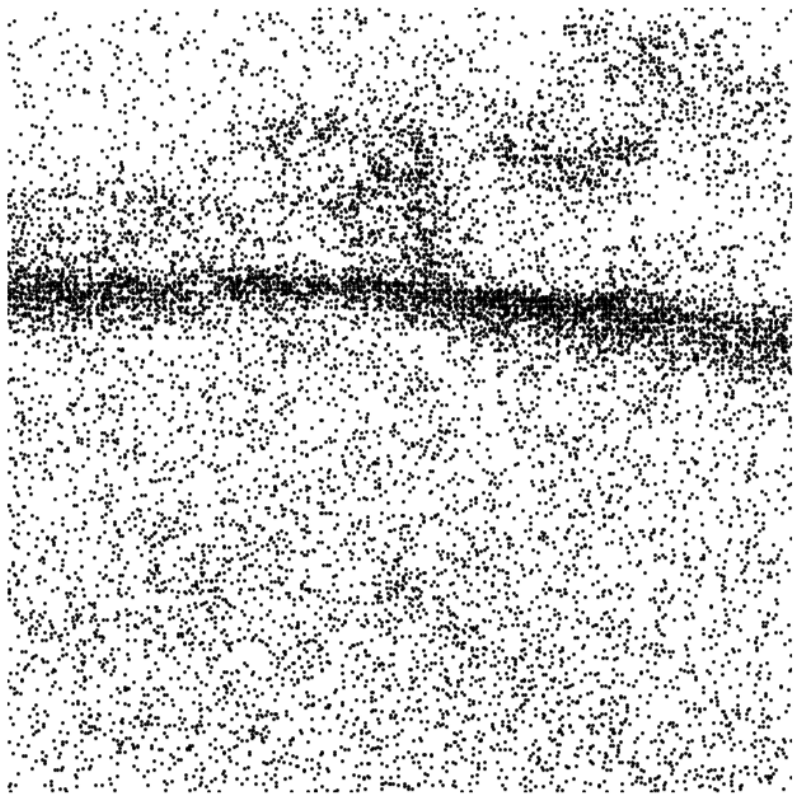}
    }
    \\
    \vspace{-4mm}
    \setcounter{subfigure}{0}
    \subfigure[][Pseudo-GT\figlabel{pseudo_gt}]{
        \includegraphics[width=0.12\linewidth]{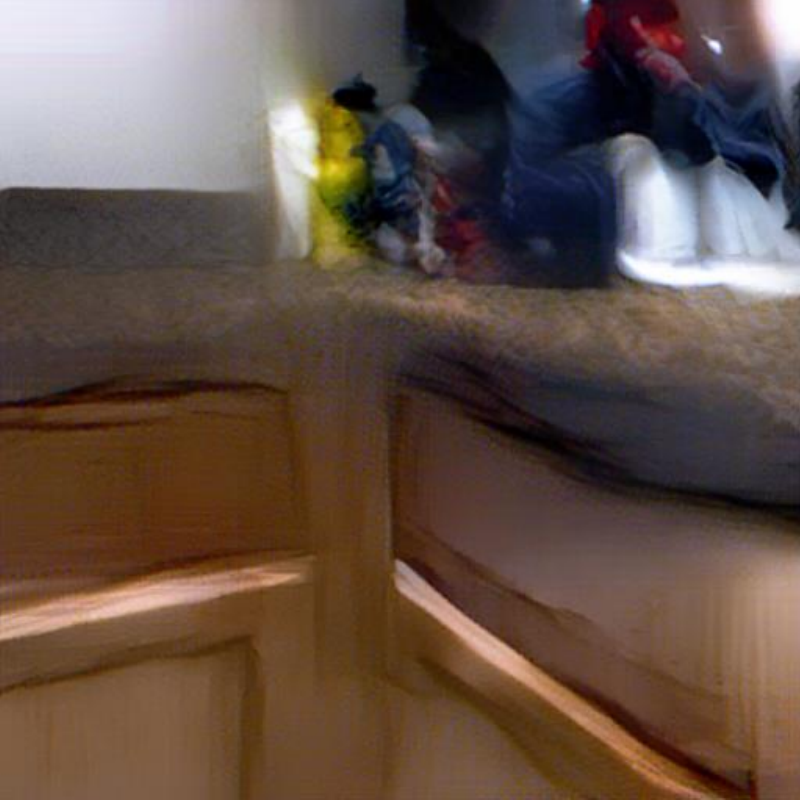}
    }
    \subfigure[][$\sigma^2=0$]{
        \includegraphics[width=0.12\linewidth]{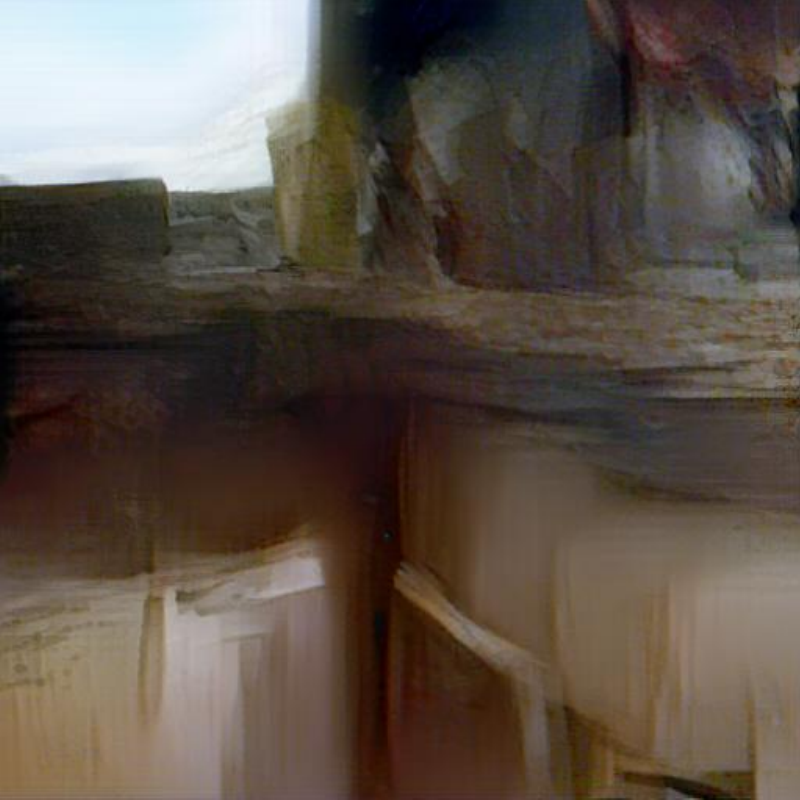}
    }
    \subfigure[][$\sigma^2=0.01$]{
        \includegraphics[width=0.12\linewidth]{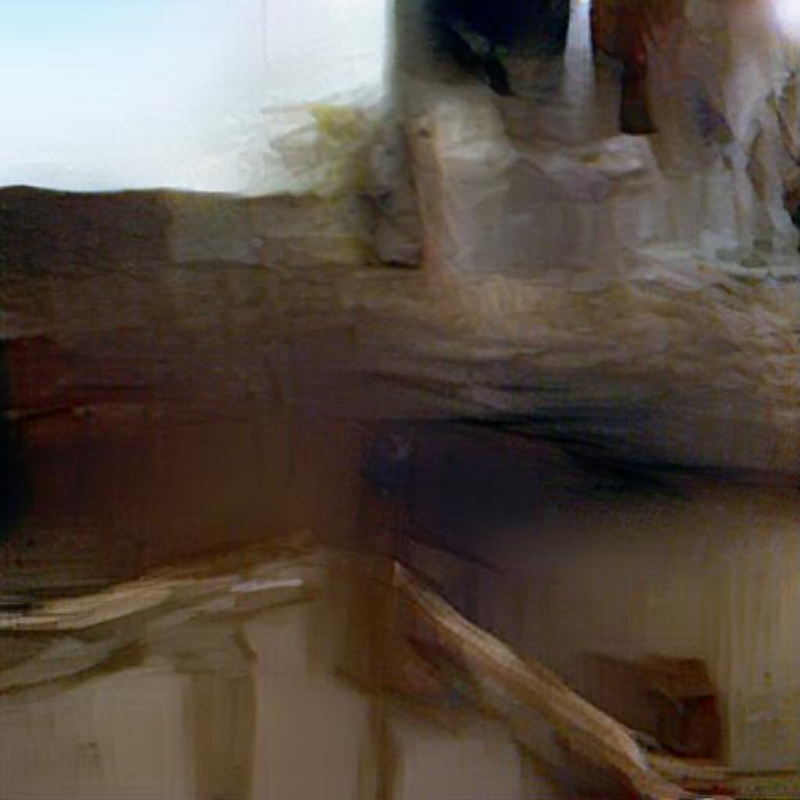}
    }
    \subfigure[][$\sigma^2=0.1$]{
        \includegraphics[width=0.12\linewidth]{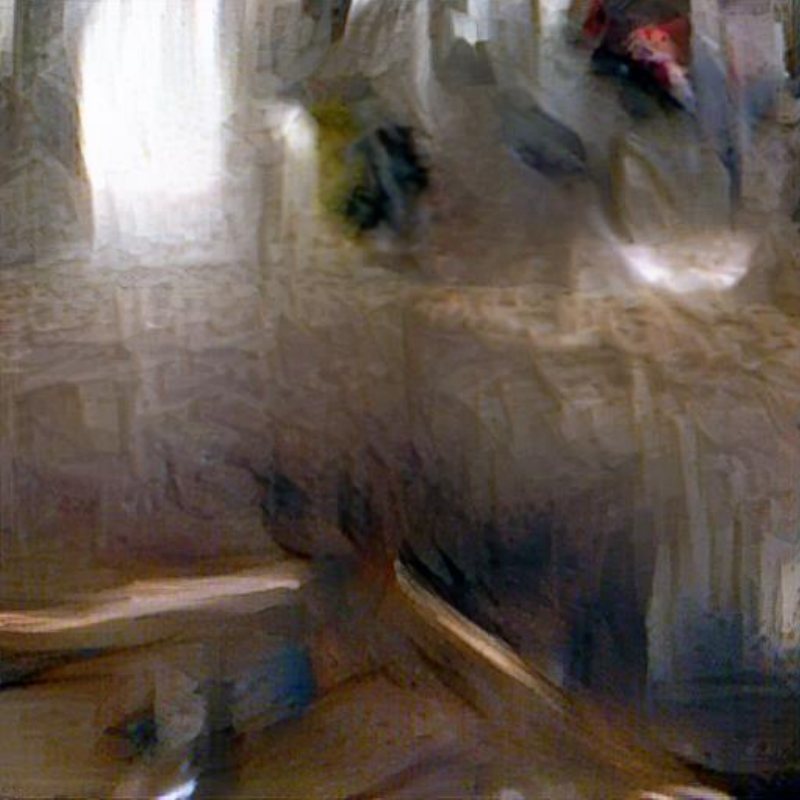}
    }
    \subfigure[][$\sigma^2=1$]{
        \includegraphics[width=0.12\linewidth]{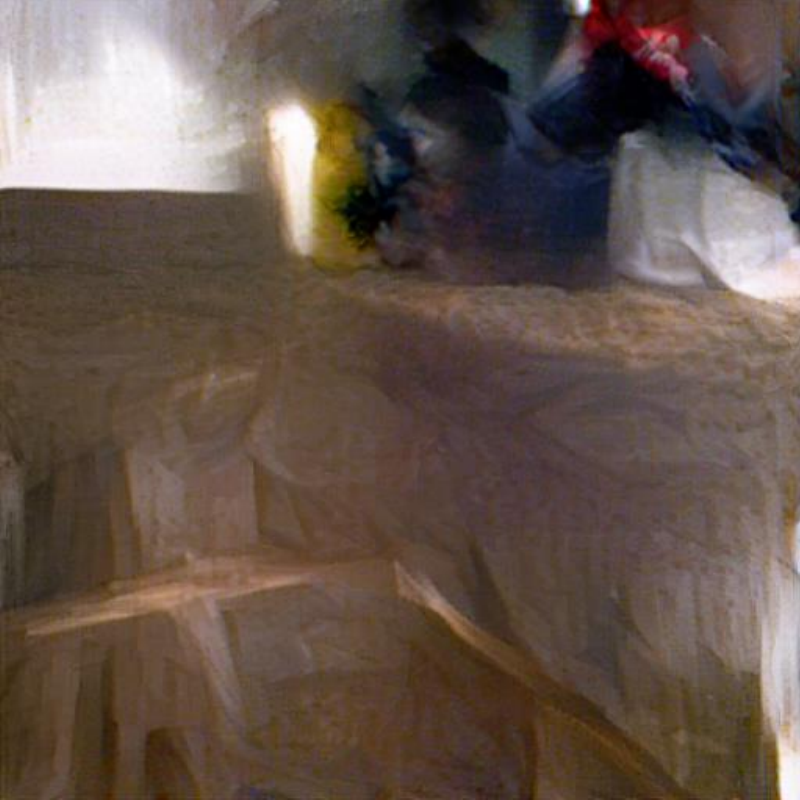}
    }
    \vspace{-1mm}
    \caption{
    Effect of our fake point generation strategy on the direct image-synthesis attack.
    We use the $\eta=0.33$ setting whereby 67\% sphere points are discarded and replaced by fake points recycling the SIFT descriptors of the rejected points.
    Pseudo-GT stands for an image reconstructed via InvSfM~\cite{pittaluga2019revealing} about the sphere centre using the original points.
    $\sigma$ denotes the standard deviation of Gaussian noise injected to generate fake points (see Sec.~\ref{sec:enhanced_sphere_cloud}).
    We determine $\sigma^2=0.1$ as the ``sweet'' spot as it hides both the scene geometry and image details.
    }
    \figlabel{variance_qualitative}
    \vspace{-2mm}
\end{figure}

\vspace{-3mm}
\paragraph{Unresolved translation scale}
As mentioned earlier in Sec.~\ref{sec:proposed_method} camera pose estimation using the sphere cloud boils down to the perspective relative pose estimation problem, meaning the translation scale is unknown~\cite{nister2004efficient}.
As many modern commercial devices such as iPad or  HoloLens 2 comprise depth sensors, we attempt to efficiently leverage calibrated raw depth maps from the on-device time-of-flight (ToF) sensor to retrieve absolute scale (see Sec.~\ref{sec:localization_procedure}).

\vspace{-2mm}
\subsection{Enhanced construction strategy}
\label{sec:enhanced_sphere_cloud}
\vspace{-2mm}
To hinder direct image synthesis from the sphere points, we add fake points to the sphere cloud.
This is inspired by the observation that embedding fake points between real keypoints degrades the quality of scene images reconstructed via InvSfM (see Fig.~\figref{variance_qualitative}).

\vspace{-4mm}
\paragraph{Cloud sparsification}
As excess number of fake points can make the sphere dense and slow the localization speed, we avoid this by keeping the total number of sphere points fixed.
If the desired proportion of true positive sphere points is $\eta$, then we discard $1 - \eta$ of all points.

\vspace{-4mm}
\paragraph{Generating fake point locations}
We employ a simple approach of adding Gaussian noise to the coordinates of existing sphere points, i.e.
where $\hat{\v x}_i\in\,S^2$ is the $i$-th sphere point, $\v z_{ij}$ is the $j$-th fake point generated in the proximity of $\hat{\v x}_i$ and $\varepsilon \sim N(\v 0,\sigma^2 \m I)$ is Gaussian noise with $\sigma^2$ set to 0.1.
The number of fake points generated per sphere point is constant.
Since the total number of sphere points remains constant, we generate $(1-\eta)/\eta$ fake points for each remaining sphere point (\eg if $\eta=33\%$, then we discard 67\% points and create two new fake points per remaining sphere point.)

\vspace{-4mm}
\paragraph{Assigning fake point descriptors via  recycling}
After assigning the fake point locations, we need to designate a realistic feature descriptor to each fake point.
We refrain from using keypoint descriptors extracted from a large database of images~\cite{dusmanu2021privacy} as there is a potential risk of this database being hijacked in which case the fake points can be easily pruned.
We also do not adopt a learning-based scheme as the generated descriptors may potentially be detected by training a discriminator network.
Instead, we resort to a simple strategy of recycling the descriptors of discarded sphere points.
Since the number of fake points is equal to the number of rejected points, this amounts to a simple permutation of descriptors from the rejected points followed by assignment of these features to the fake point locations.

As shown in Fig.~\figref{center_inversion}, this strategy effectively mitigates the issue of direct image synthesis for the sphere cloud.
Additionally, adjusting $\eta$ controls the trade-off between localization accuracy and privacy-preserving ability as shown in Fig.~\figref{center_inversion} and Table~\ref{tab:localization}.

\begin{figure}[t]
    \centering
    \subfigure[][p6L solver for line clouds~\cite{speciale2019privacy}]{
        \includegraphics[width=0.3\linewidth]{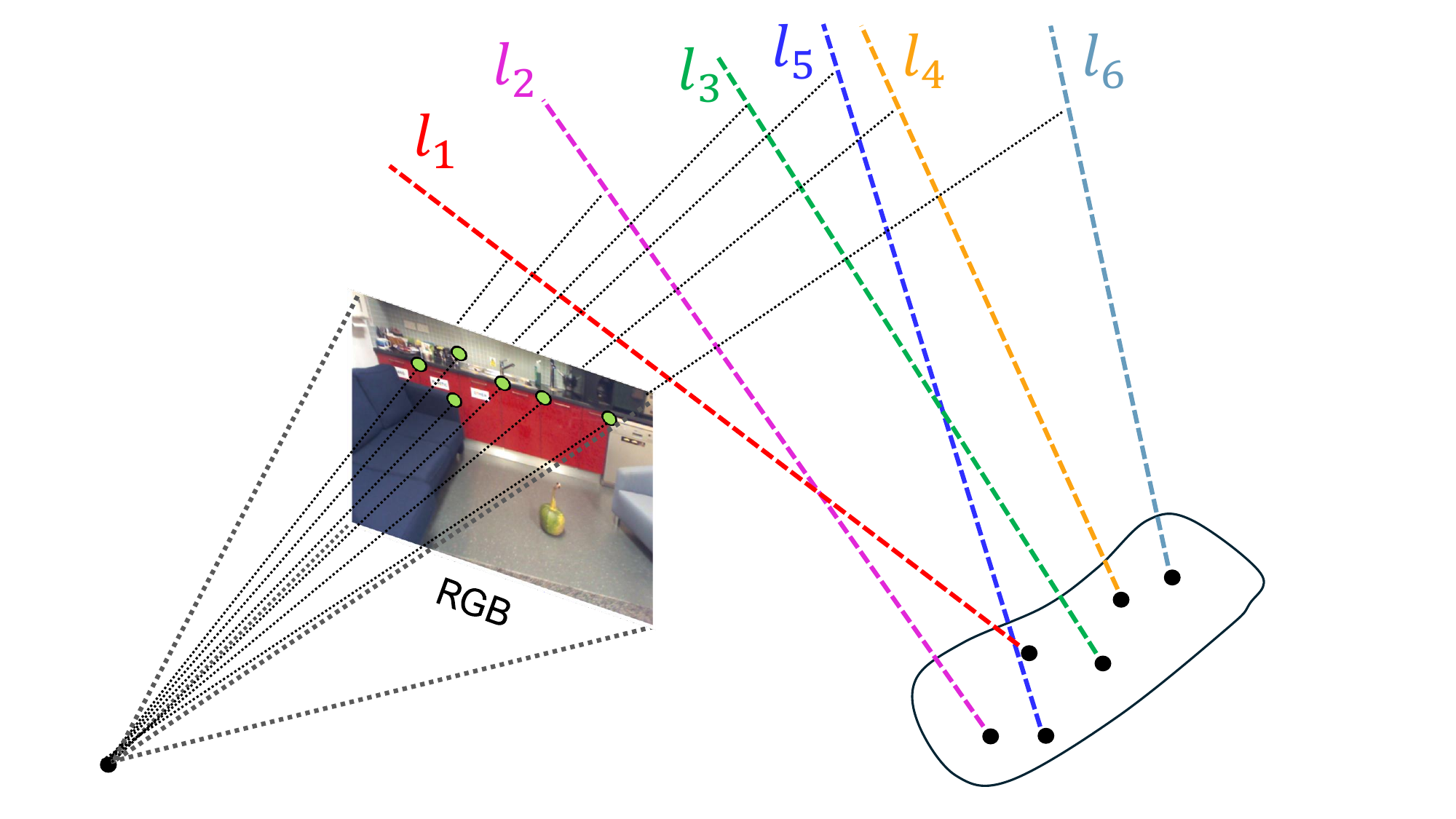}
    }
    \subfigure[][p3P solver for the sphere cloud \figlabel{vp3p}]{
        \includegraphics[width=0.3\linewidth]{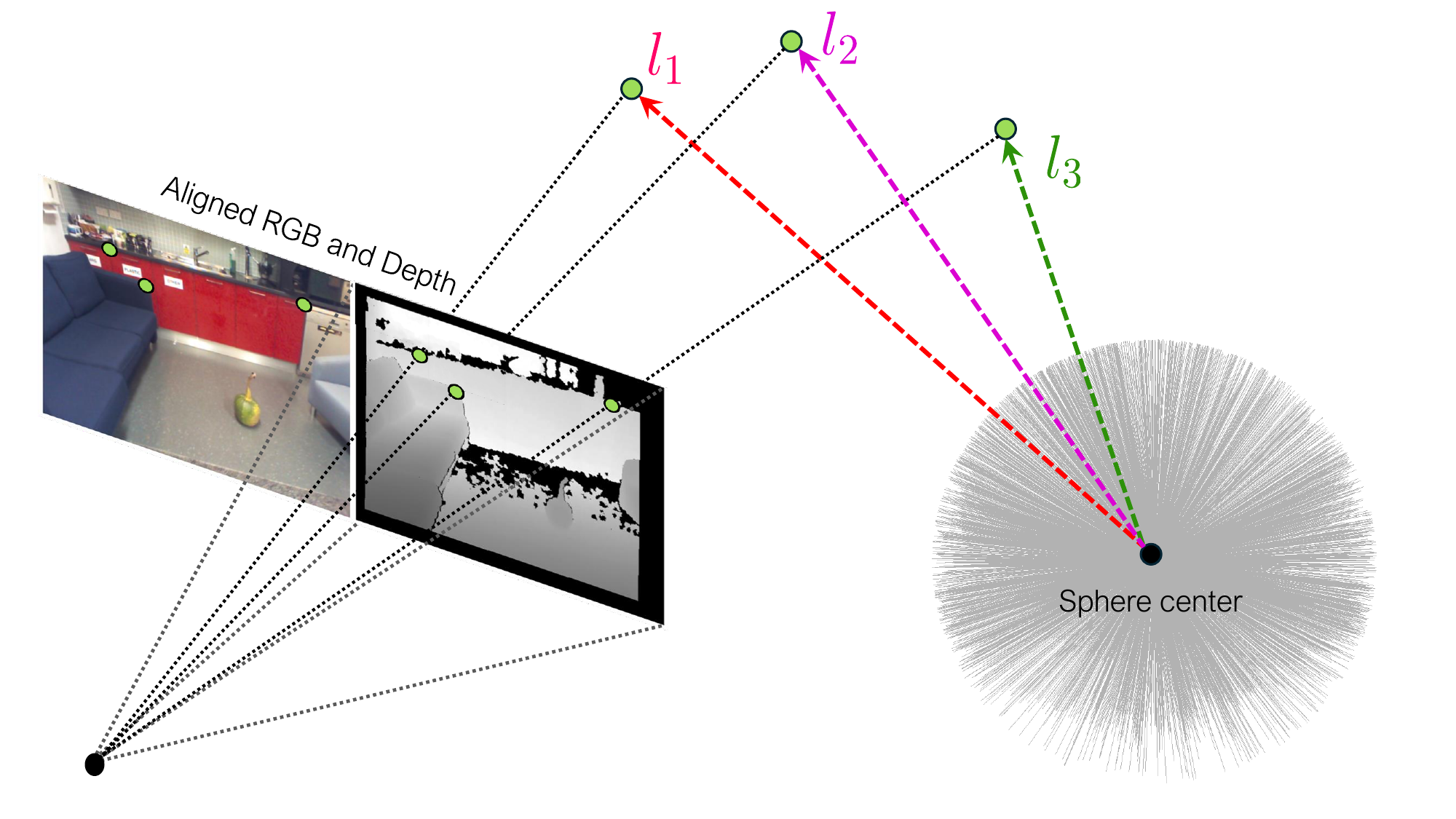}
    }
    \vspace{-1mm}
    \caption{
    Comparison of minimal solvers for privacy-preserving localization. 
    In (b), the depth map allows lifting keypoints to 3D and estimating pose using an efficient p3P solver. 
    }
    \figlabel{minimal_solver}
    \vspace{-6mm}
\end{figure}

\vspace{-2mm}
\subsection{Camera pose estimation using RGB image and  depth map}
\label{sec:localization_procedure}
\vspace{-2mm}

We illustrate a framework for absolute pose estimation using the sphere cloud assuming the query has an aligned pair of RGB image and absolute depth map with known intrinsic.

\vspace{-2mm}
\paragraph{Efficient initial pose estimation via  perspective-$n$-points}
Fusing the query RGB image with its aligned depth map allows us to lift each 2D keypoint $\v u_i\in\real^2$ to the 3D space as $\v p_i = z_i^{TOF} \m {K}^{-1} [\v u_i\tr, 1]\tr$, where $z_i^{TOF}\in\real$ is the depth of $\v u_i$ obtained from the ToF sensor and $\m K\in\Rmx{3}{3}$ is the camera intrinsics.
We note aligning these 3D keypoints $\{ \v p_i \}$ with the matched 3D lines $\{ \v l_i \}$ from the sphere cloud is nearly identical to the perspective-$n$-points problem except that the 3D points are on the query side and not on the map side (see Fig.~\figref{vp3p}).

Ideally, the 3D keypoint $\v p_i$ should lie along the positive direction of the vector shooting out from the sphere centre and passing through the sphere point $\hat{\v x}_i$, i.e. where $[\m R | \v t]$ defines the query-to-world (sphere cloud) transformation.
This geometric constraint can be efficiently solved using a combination of LO-RANSAC~\cite{mishchuk2017working,PoseLib} and the p3P solver~\cite{persson2018lambda}, which only needs 4 correspondences compared to 6 required in the absence of depth maps~\cite{speciale2019privacy,lee2023ppl,pan2023privacy}.
The final (world-to-query) pose is obtained as $[\m R\tr | -\m R\tr \v t ]$, and heavy outliers are pruned by checking the epipolar distance in Eq.~\eqref{epipolar_error} and the depth error in Eq.~\eqref{depth_error}. 
The threshold values used in our implementation can be found in supplementary material.

\vspace{-2mm}
\paragraph{Pose refinement with depth regularization} 
After an initial pose is obtained, we refine the pose via nonlinear optimization~\cite{PoseLib}.
Since the sphere cloud can be viewed as a special type of line cloud, we follow the direction of other line clouds~\cite{speciale2019privacy,lee2023ppl} and partly minimize the square of epipolar distance between the projection of 3D lines derived from the sphere cloud and the 2D query keypoints. The resulting loss function for the $i$-th keypoint, $L^e_i$, is
\vspace{-3mm}
\begin{align}
L^e_i = \frac{([\v u_i\tr, 1] \m {K}^{-\top}~ \m E ~ \tilde{\v x}_i)^2}
    {(\v e_{1}\tr \tilde{\v x}_i)^2 + (\v e_{2}\tr \tilde{\v x}_i)^2}       
\eqlabel{epipolar_error}
\end{align}

\vspace{-3mm}\noindent
where $\m E:=[\v e_{1}, \v e_{2}, \v e_{3}]\tr$ denotes the essential matrix between the sphere cloud and the query camera, 
and $\tilde {\v x}_i = \hat{\v x}_i / |\hat{x}_{iz}|$ is the $z$-normalization of the sphere point $\hat {\v x}_i$.

Since \eqref{epipolar_error} is oblivious to the translation scale, we additionally employ a depth regularization term to guide the camera pose to the correct scale.
For this purpose, we define another loss 
\vspace{-4mm}
\begin{align}
L_i^d &= (\beta_i -1)^2,
\eqlabel{depth_error}
\end{align}

\vspace{-3mm}\noindent
where $\beta_i = z_i / z_i^{TOF}$ is the proportional difference between the predicted depth $z_i(\m R, \v t)$ from the current pose and the sphere cloud and the observed depth $z_i^{TOF}$ 
(an analytic derivation of $z_i(\m R, \v t)$ can be found in~\cite{supmat}).

The overall cost function iteratively minimized over $\m R$ and $\v t$ is
\vspace{-2mm}
\begin{align}
    L &= 
    \sum_{i\in\Omega} \left( L_i^e + \lambda~ L_i^d \right),
    \eqlabel{eq:total_cost} 
\end{align}

\vspace{-3mm}\noindent
where $\lambda$ is the hyperparameter empirically set to $10^{-4}$ and $\Omega$ represents all of the 2D–3D correspondences on the query image. 
(more details in~\cite{supmat}).

\begin{figure}[t!]
    \vspace{-2mm}
    \centering
    \subfigure[3D point reconstruction errors~\figlabel{cdf_geo}]{
        \includegraphics[width=0.22\linewidth]{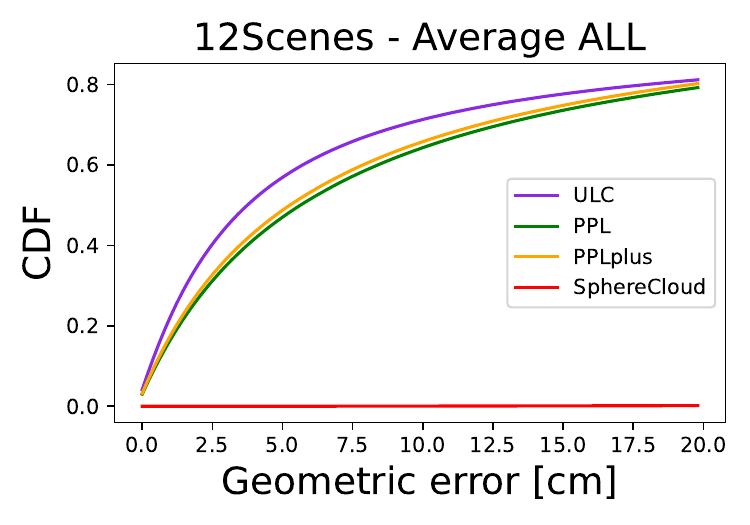}    
        \includegraphics[width=0.22\linewidth]{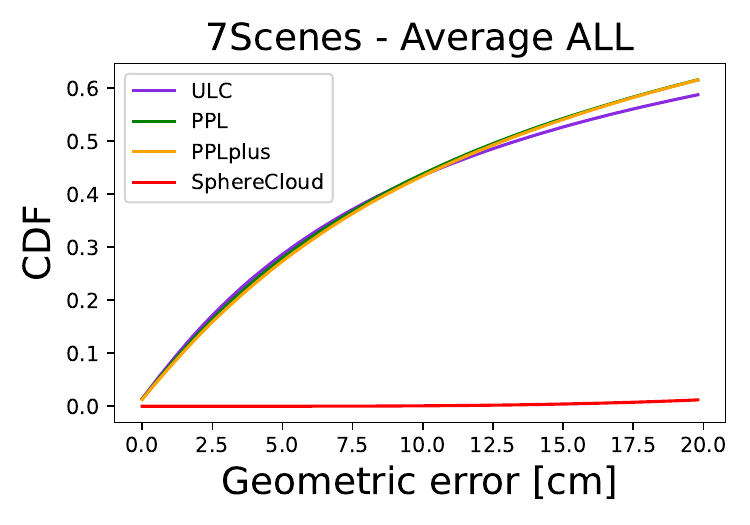}
    }    \subfigure[LPIPS~\cite{zhang2018lpips} of recovered images~\figlabel{cdf_LPIPS}]{
        \includegraphics[width=0.22\linewidth]{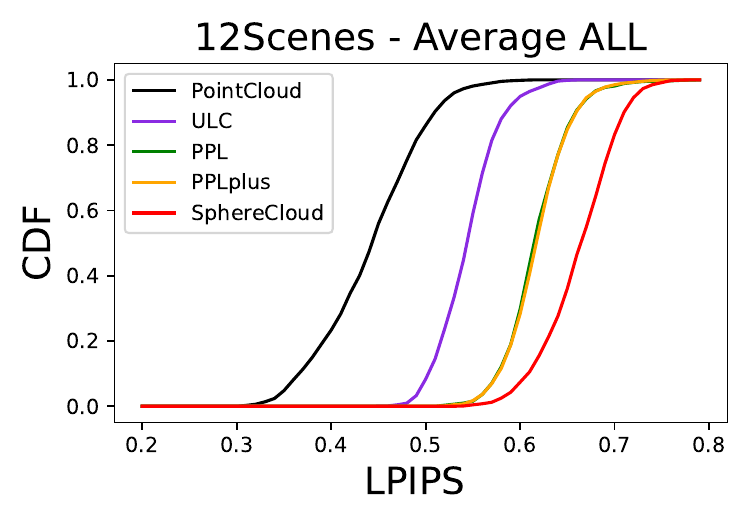}    
        \includegraphics[width=0.22\linewidth]{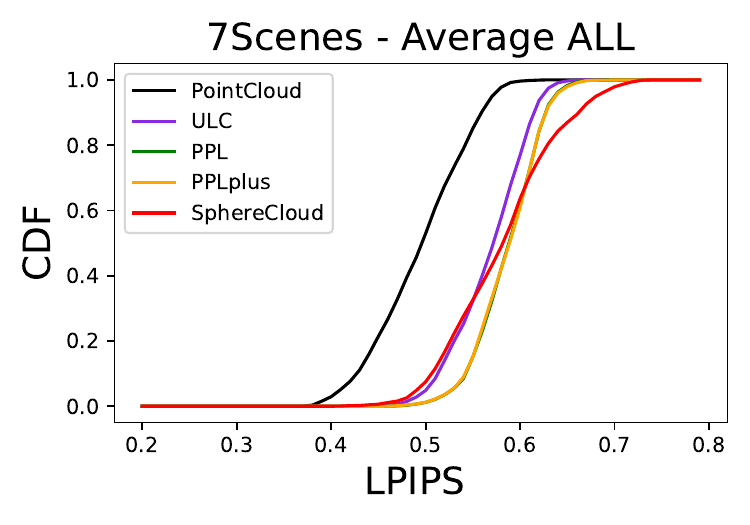}
    }
    \vspace{-1mm}
    \caption{Cumulative distributions of (a) the geometric error ($e_g$) of 3D points recovered using~\cite{chelani2021privacy} and (b) LPIPS of  reconstructed images from InvSfM~\cite{pittaluga2019revealing}.
    }
    \figlabel{fig:cdf}
    \vspace{-5mm}
\end{figure}
\begin{figure}[t]
    \centering

    \subfigure{
        \includegraphics[width=0.14\linewidth]{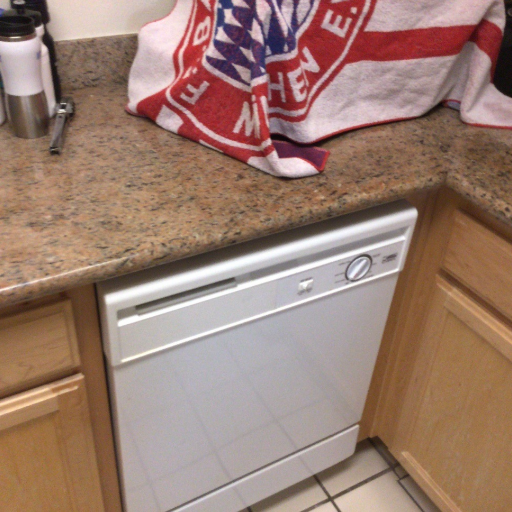}
    }
    \hspace{-3mm}
    \subfigure{
        \includegraphics[width=0.14\linewidth]{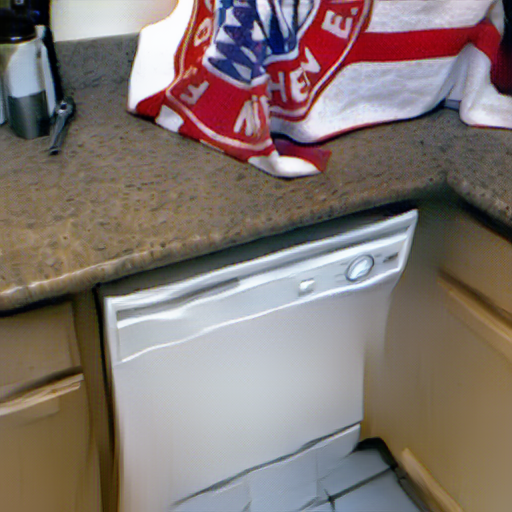}
    }
    \hspace{-3mm}
    \subfigure{
        \includegraphics[width=0.14\linewidth]{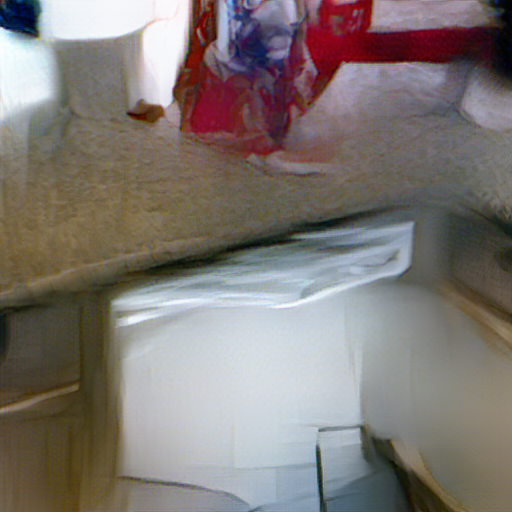}
    }
    \hspace{-3mm}
    \subfigure{
        \includegraphics[width=0.14\linewidth]{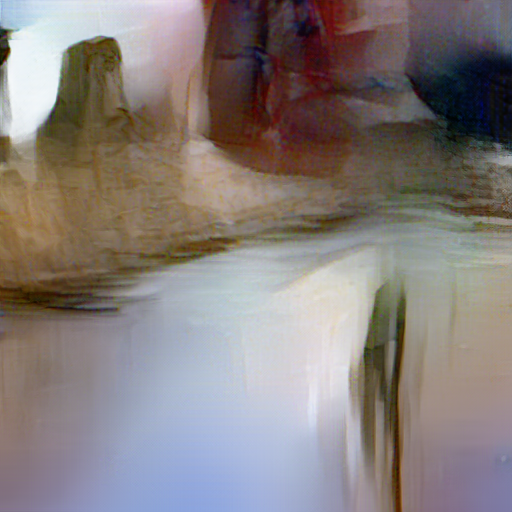}
    }
    \hspace{-3mm}
    \subfigure{
        \includegraphics[width=0.14\linewidth]{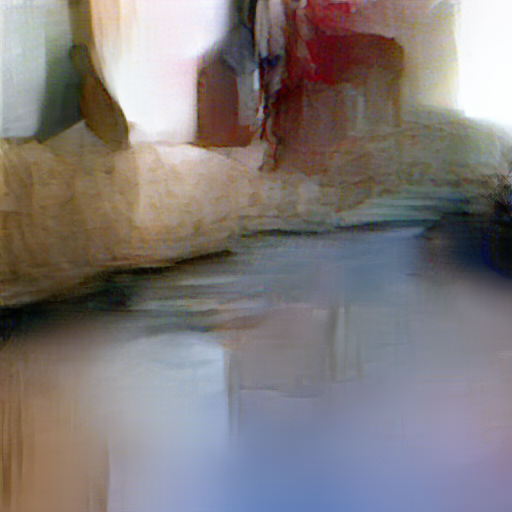}
    }
    \hspace{-3mm}
    \subfigure{
        \includegraphics[width=0.14\linewidth]{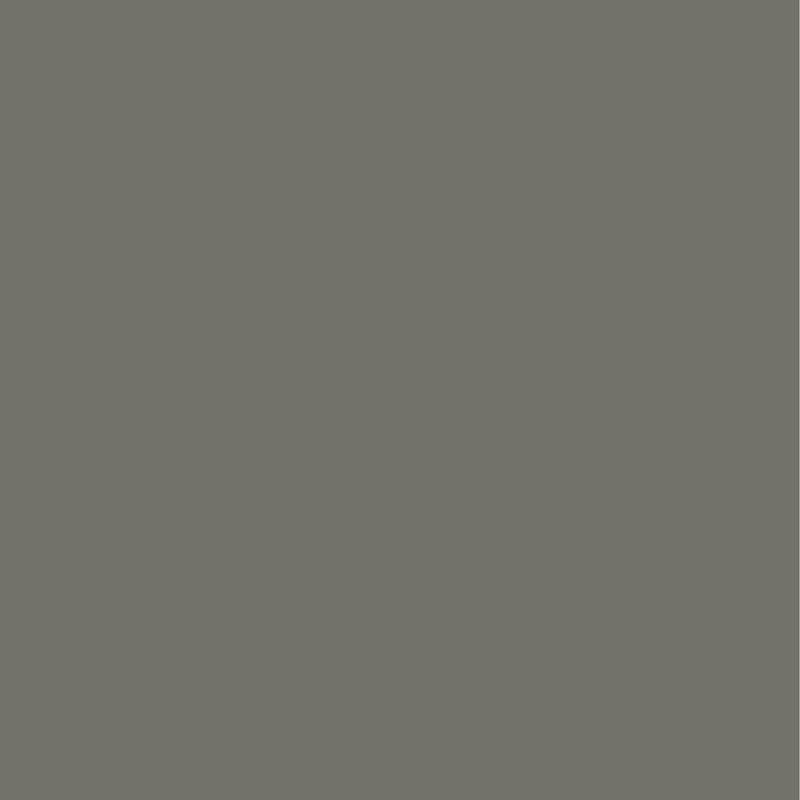}
    }
    \\
    \vspace{-3mm}
    \setcounter{subfigure}{0}
    \hspace{0.1mm}
    \subfigure[Ground truth]{
        \includegraphics[width=0.14\linewidth]{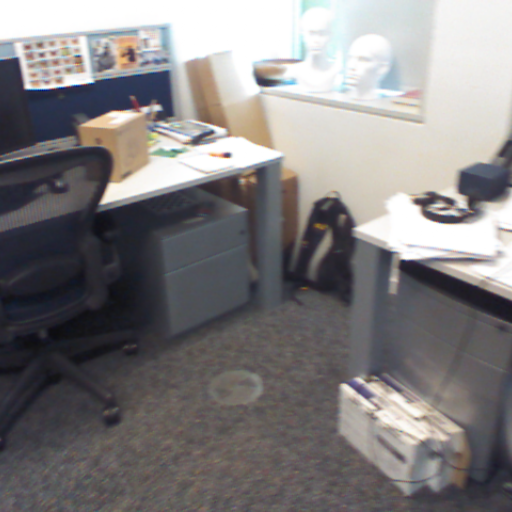}
    }
    \hspace{-3mm}
    \subfigure[Point cloud]{
        \includegraphics[width=0.14\linewidth]{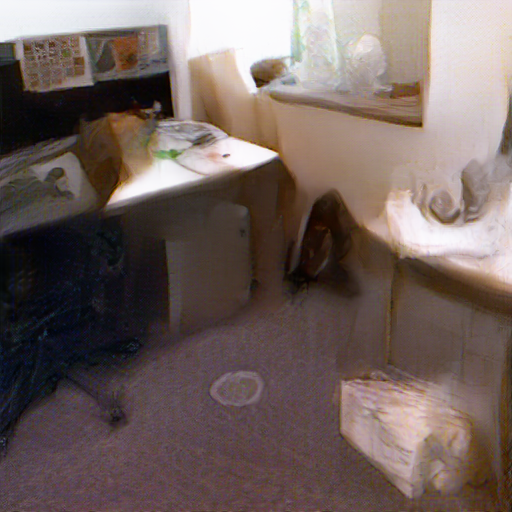}
    }
    \hspace{-3mm}
    \subfigure[ULC~\cite{speciale2019privacy}]{
        \includegraphics[width=0.14\linewidth]{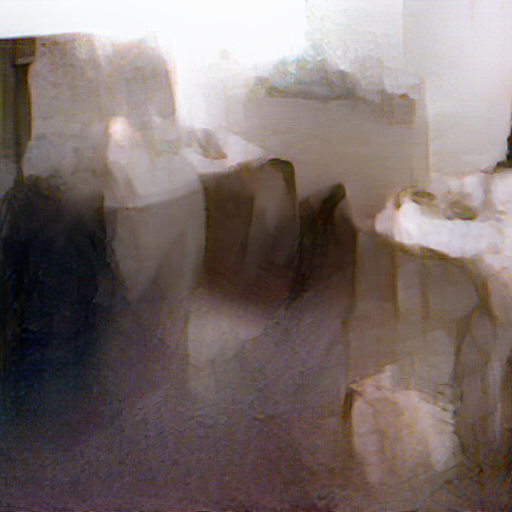}
    }
    \hspace{-3mm}
    \subfigure[PPL~\cite{lee2023ppl}]{
        \includegraphics[width=0.14\linewidth]{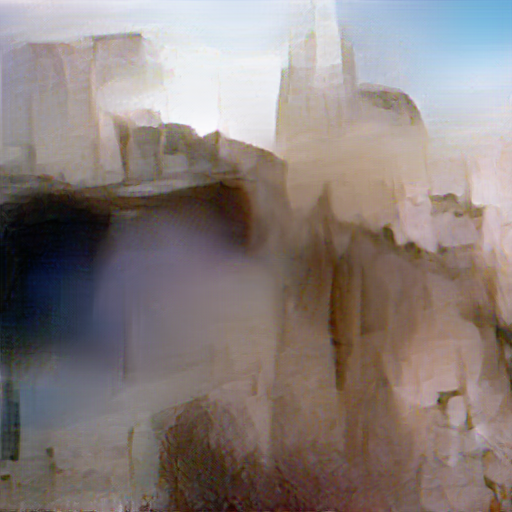}
    }
    \hspace{-3mm}
    \subfigure[PPL+~\cite{lee2023ppl}]{
        \includegraphics[width=0.14\linewidth]{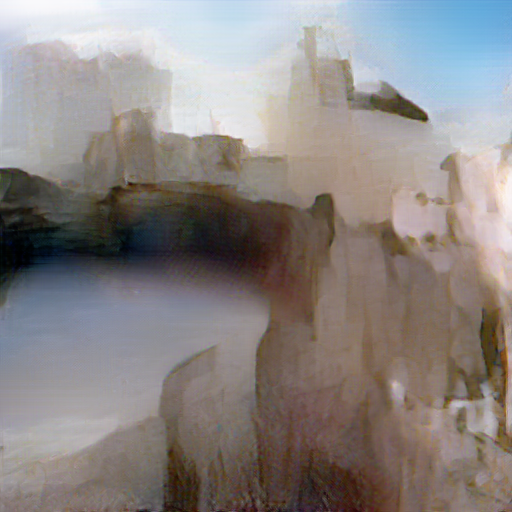}
    }
    \hspace{-3mm}
    \subfigure[Sphere cloud~\figlabel{sphere_invsfm}]{
        \includegraphics[width=0.14\linewidth]{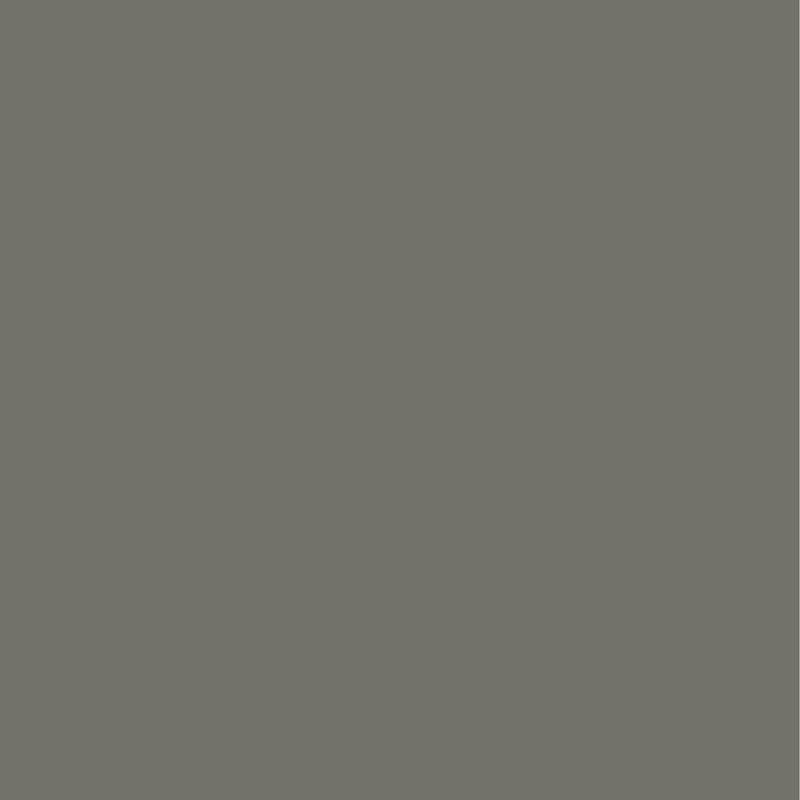}
    }    
    \vspace{-1mm}
    \caption{Images revealed from some test camera poses across different scene representation via 
    InvSfM~\cite{pittaluga2019revealing}.
    (Top) \emph{Apt2 kitchen} in 12-Scenes~\cite{valentin2016energy}.
    (Bottom) \emph{Office} in 7-Scenes~\cite{shotton2013scene}.
    }
    \figlabel{inversion}
\end{figure}

\vspace{-5mm}
\section{Experiments}
\label{sec:experiments}
\vspace{-2mm}

\paragraph{Datasets}
We used two public RGB-D camera re-localization datasets as presented in~\cite{shotton2013scene, valentin2016energy}.
7-Scenes~\cite{shotton2013scene} and 12-Scenes~\cite{valentin2016energy} consist of several RGB and depth frames of indoor spaces captured with multiple sequences.
For 7-Scenes~\cite{shotton2013scene}, we followed additional procedures in~\cite{brachmann2021dsacstar} to align depth maps to RGB images (not required for 12-Scenes).

\begin{table}[t]
\caption{
    Quantitative analysis of the direct image-synthesis attack (see Sec.~\ref{sec:naive_construction}) on sphere clouds.
    Each image is reconstructed from an arbitrary viewpoint at the sphere centre.
    Since no ground truth images are available for these viewpoints, the metrics are calculated using the images reconstructed from 3D point clouds as pseudo-ground truth, but these are often very noisy for the 7-Scenes dataset as shown in Fig.~\figref{center_inversion}. 
    For the sphere cloud, results are reported across different proportions of true positive sphere points ($\eta$).
}
\centering
\fontsize{7}{8}\selectfont
\renewcommand{\arraystretch}{1.3}
\setlength{\tabcolsep}{4pt} 
\label{tab: center inversion quantitative}
\begin{tabular}{c|c|cccccc}
\hline
\hline
Dataset & Metric & ULC~\cite{speciale2019privacy} & PPL~\cite{lee2023ppl} & PPL+~\cite{lee2023ppl} & \begin{tabular}{c}Sphere\\ ($\eta$$=$25\%) \end{tabular} & \begin{tabular}{c}Sphere\\ ($\eta$$=$33\%) \end{tabular} & \begin{tabular}{c} Sphere \\ ($\eta$$=$50\%) \end{tabular} \\ 
\hline
\multirow{5}{*}{$\begin{tabular}{c} 12-Scenes\\ \cite{valentin2016energy}\end{tabular}$} 
& PSNR ($\downarrow$) & 16.06 & 11.99 & \textbf{11.30} & 12.70 & 13.53 & 14.94 \\
& LPIPS ($\uparrow$) & 0.456 & 0.539 & 0.542 & \textbf{0.568} & 0.534 & 0.488 \\
& SSIM ($\downarrow$) & 0.519 & 0.440 & 0.436 & \textbf{0.372} & 0.429  & 0.493 \\
& MAE ($\uparrow$) & 31.82 & 55.91 & \textbf{56.88} & 47.19 & 42.53 & 35.39 \\ 
\hline
\multirow{5}{*}{ $\begin{tabular}{c} 7-Scenes\\ \cite{shotton2013scene}\end{tabular}$ }
& PSNR ($\downarrow$) & 13.11 & 11.04 & \textbf{10.84} & 13.41 & 14.01 & 15.29 \\
& LPIPS ($\uparrow$)  & 0.548 & 0.602 & \textbf{0.603} & 0.550 & 0.533 & 0.498 \\
& SSIM ($\downarrow$) & 0.417 & 0.390 & \textbf{0.380} & 0.393 & 0.417 & 0.471 \\
& MAE ($\uparrow$)  & 44.85 & 59.37 & \textbf{60.50} & 43.25 & 40.28 & 34.13 \\
\hline
\hline
\end{tabular}
\label{tab:reconstruction}
\vspace{-5mm}
\end{table}

\begin{figure*}[t]
    \centering
    \hspace{1.5mm}
    \subfigure{
        \includegraphics[width=0.13\linewidth]{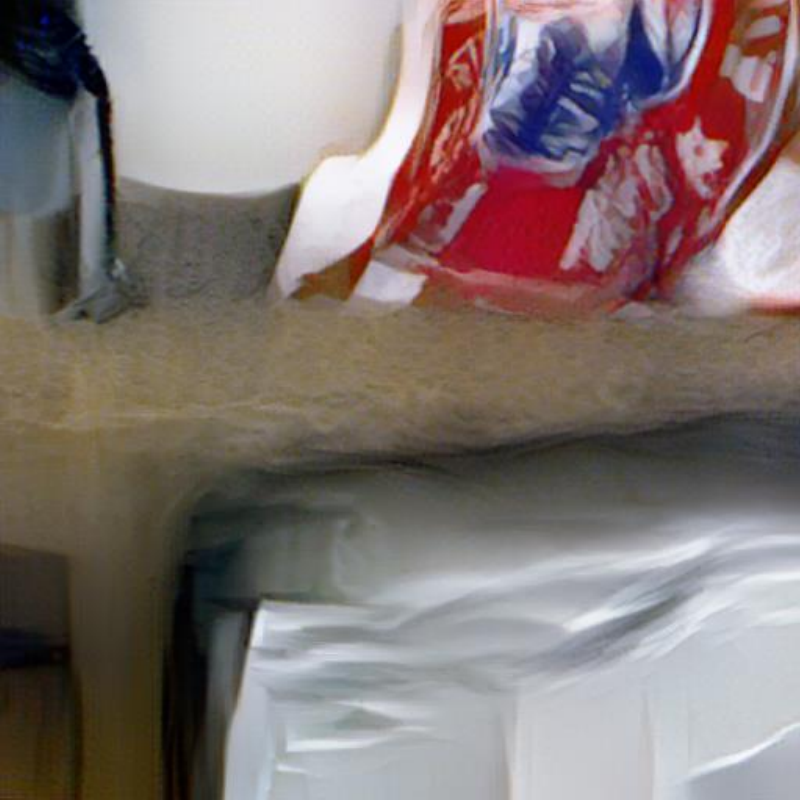}
    }
    \hspace{-3mm}
    \subfigure{
        \includegraphics[width=0.13\linewidth]{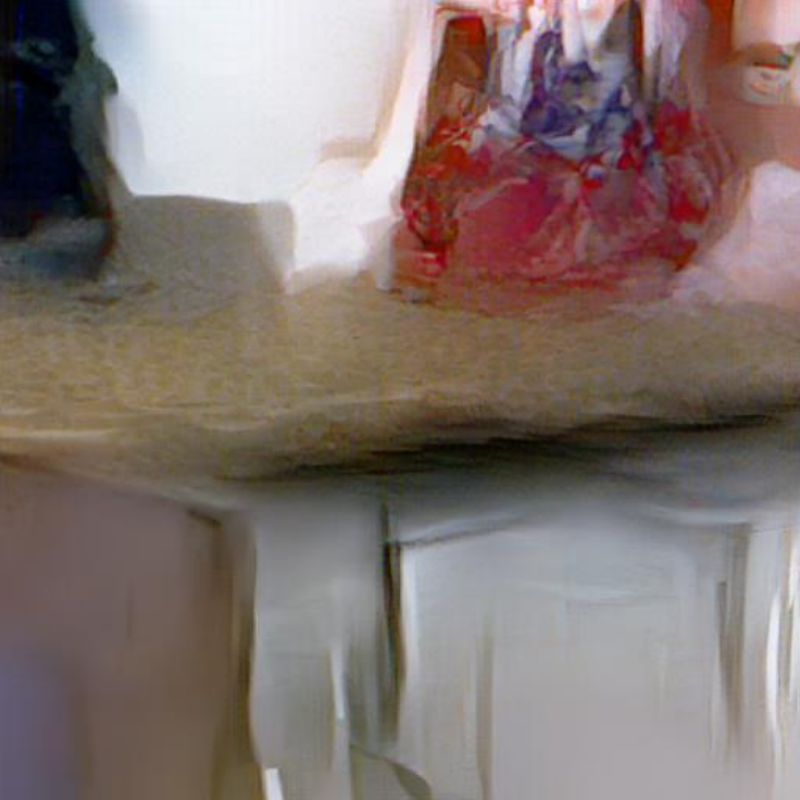}
    }
    \hspace{-3mm}
    \subfigure{
        \includegraphics[width=0.13\linewidth]{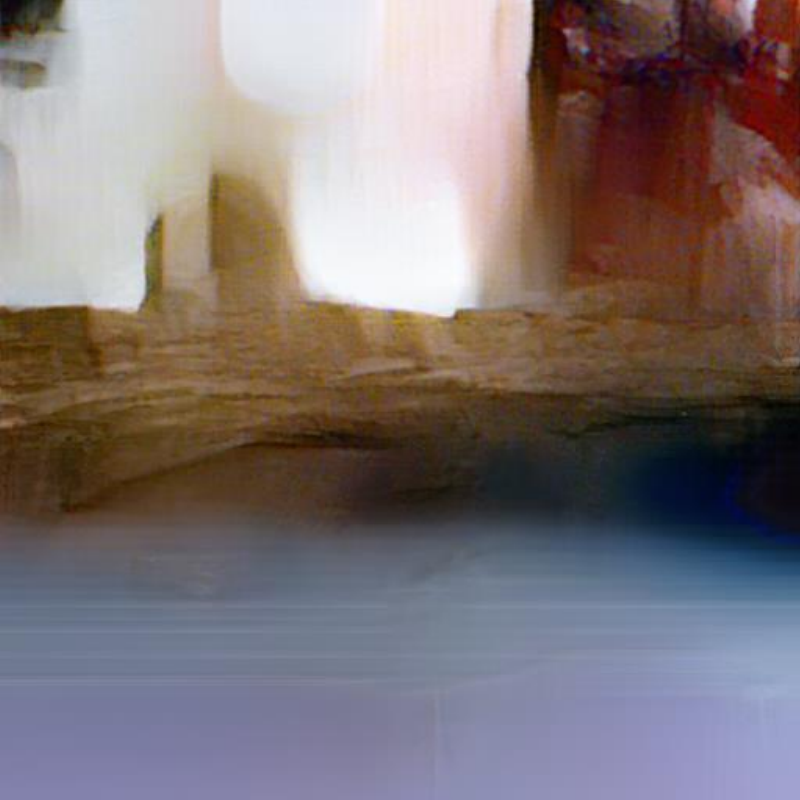}
    }
    \hspace{-3mm}
    \subfigure{
        \includegraphics[width=0.13\linewidth]{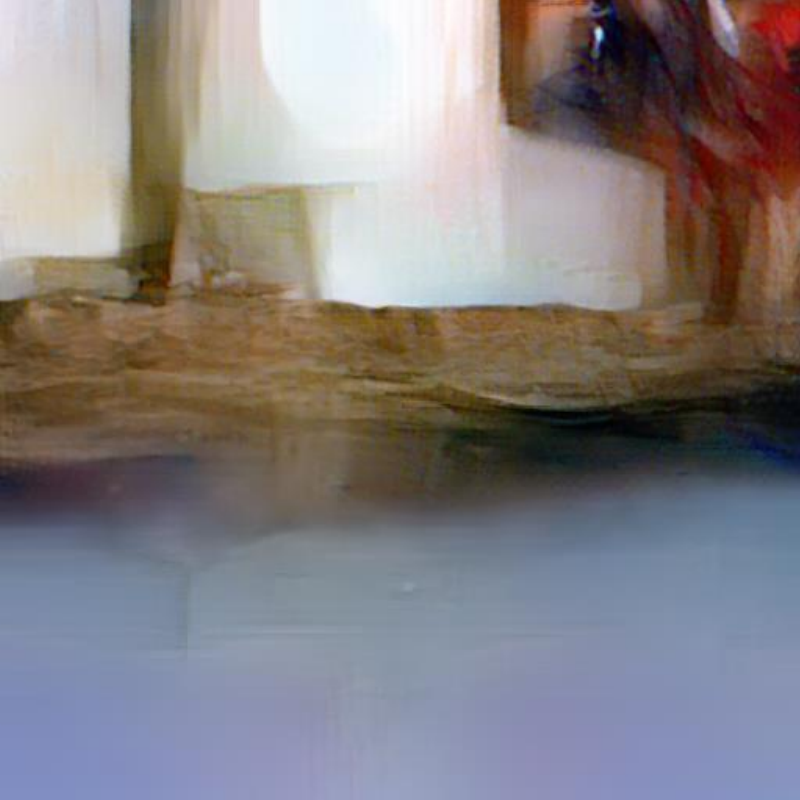}
    }
    \hspace{-3mm}
    \subfigure{
        \includegraphics[width=0.13\linewidth]{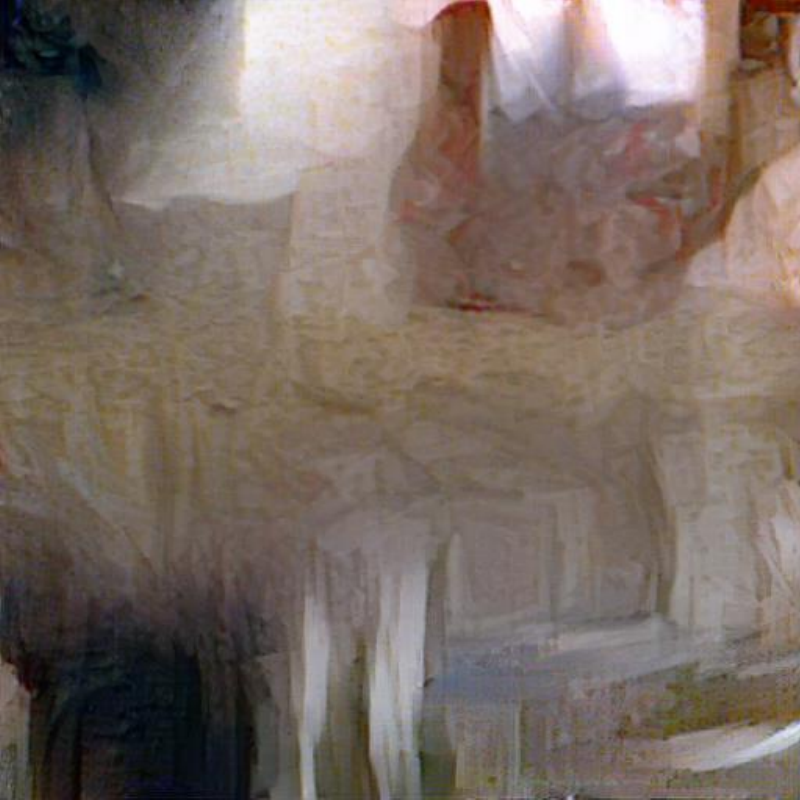}
    }
    \hspace{-3mm}
    \subfigure{
        \includegraphics[width=0.13\linewidth]{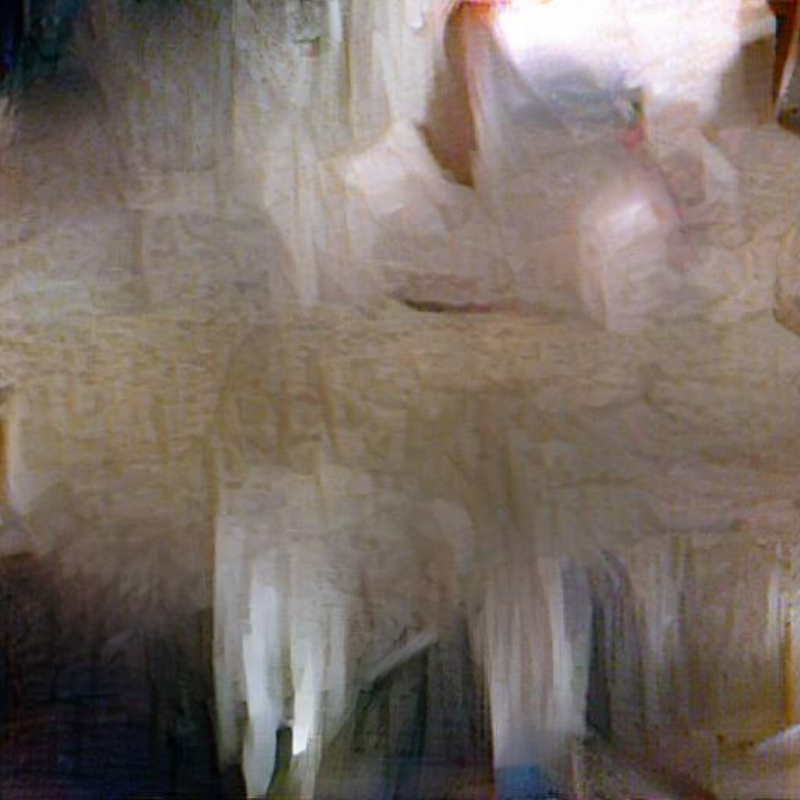}
    }
    \hspace{-3mm}
    \subfigure{
        \includegraphics[width=0.13\linewidth]{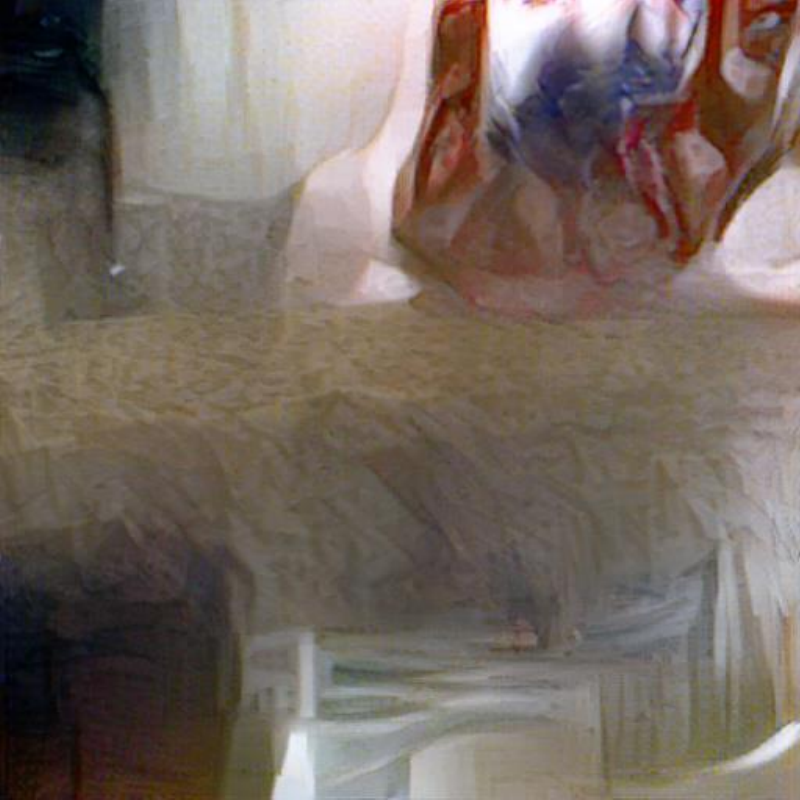}
    }
    \\

    \vspace{-3.5mm}
    \setcounter{subfigure}{0}
    \subfigure[Pseudo-GT~\figlabel{fig:pseudoGT}]{
        \includegraphics[width=0.13\linewidth]{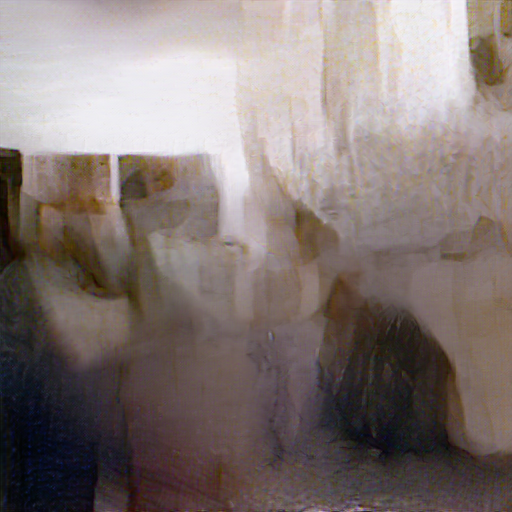}
    }
    \hspace{-3mm}
    \subfigure[ULC~\cite{speciale2019privacy}]{
        \includegraphics[width=0.13\linewidth]{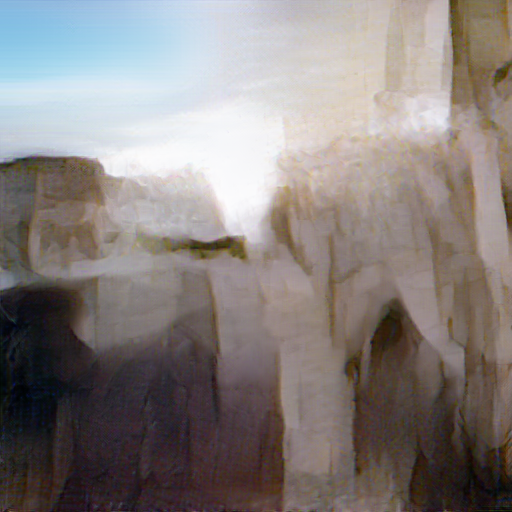}
    }
    \hspace{-3mm}
    \subfigure[PPL~\cite{lee2023ppl}]{
        \includegraphics[width=0.13\linewidth]{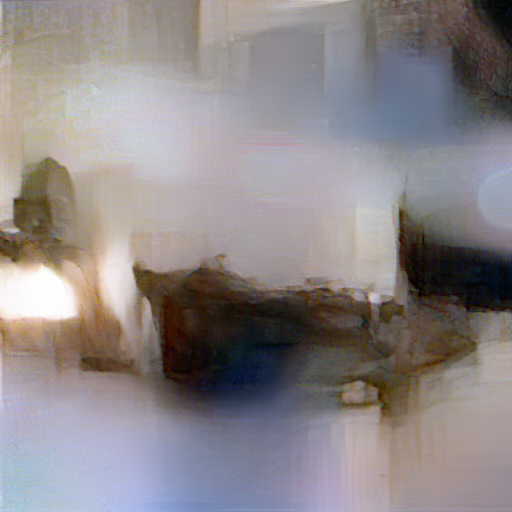}
    }
    \hspace{-3mm}
    \subfigure[PPL+~\cite{lee2023ppl}]{
        \includegraphics[width=0.13\linewidth]{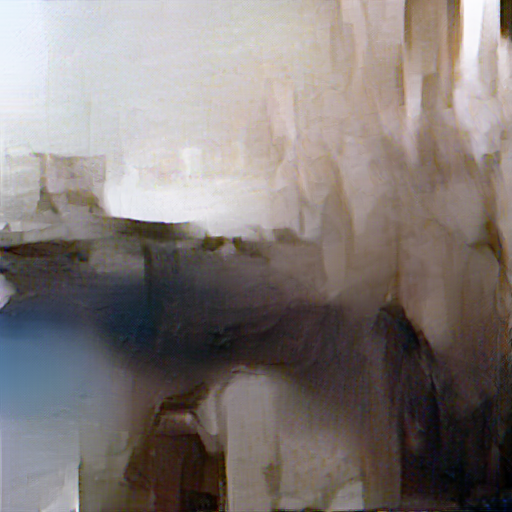}
    }
    \hspace{-3mm}
    \subfigure[Ours~(25\%)]{
        \includegraphics[width=0.13\linewidth]{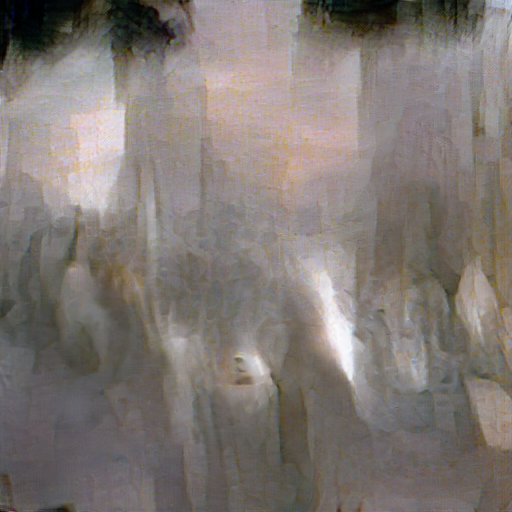}
    }
    \hspace{-3mm}
    \subfigure[Ours~(33\%)]{
        \includegraphics[width=0.13\linewidth]{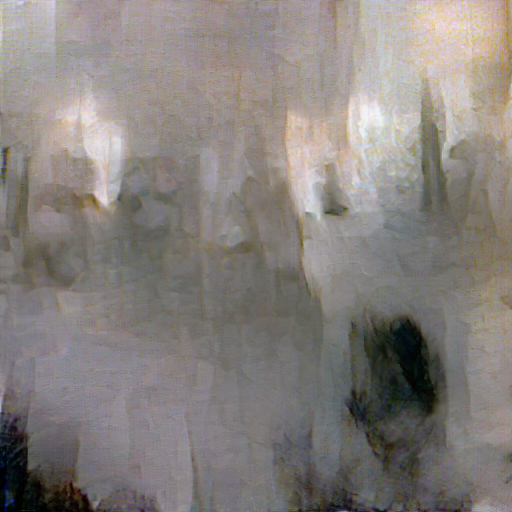}
    }
    \hspace{-3mm}
    \subfigure[Ours~(50\%)]{
        \includegraphics[width=0.13\linewidth]{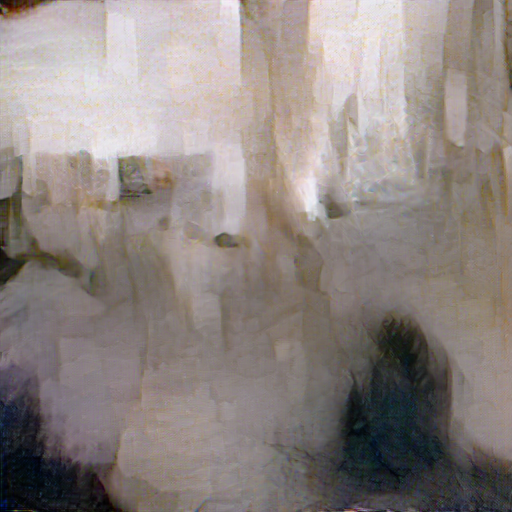}
    }
    \hspace{-4mm}
    \vspace{-1mm}
    \caption{Visualization of images directly reconstructed from sphere clouds about the sphere centre.
    (Top) \emph{apt2 kitchen} in 12-Scenes~\cite{valentin2016energy}.
    (Bottom) \emph{office} in 7-Scenes~\cite{shotton2013scene}.
    (a) is the result of applying InvSfM~\cite{pittaluga2019revealing} to the original point cloud.
    (e), (f), and (g) are the results of the sphere cloud across different proportions of true positive sphere points ($\eta$).
    Note the viewpoints are deliberately chosen to be close to the test poses in Fig.~\figref{inversion} for better comparison.
    }
    \vspace{-2mm}
    \figlabel{center_inversion}
\end{figure*}

\paragraph{Implementation details}
We implemented our RGB-D localization pipeline for sphere clouds using the PoseLib library~\cite{PoseLib} and brought the inversion pipeline from~\cite{lee2023ppl}.

\noindent
For our localization experiment, we used the official RGB-D benchmark released by~\cite{brachmann2021pseudoGT} based on the above two datasets~\cite{shotton2013scene,valentin2016energy}
which contains sparse 3D point clouds reconstructed using COLMAP~\cite{schoenberger2016sfm} and the lists of test images.
However, as \cite{brachmann2021pseudoGT} does not provide the SIFT descriptors~\cite{lowe2004sift} required for image recovery, we carefully reconstructed these point clouds ourselves using COLMAP following the same protocol of~\cite{brachmann2021pseudoGT} and used them along with the same set of test images for comparing the privacy-preserving capability of different methods.

\noindent    
All our experiments were carried out on a PC with Intel CPU i9-13900K running at 3.0 GHz and a single NVIDIA RTX 4090 graphics card.

\vspace{-2mm}
\paragraph{Evaluation metrics}
For the quantitative evaluation of 3D point recovery using~\cite{chelani2021privacy}, we reported the 3D point error $e_g = \l2{\v g - \v g^*}$ between the estimated point $\v g\in\real^3$ and the original  point $\v g^*\in\real^3$.
For comparing the image reconstruction quality, we used the
peak signal-to-noise ratio (PSNR), learned perceptual image patch similarity (LPIPS), structural similarity index measure (SSIM) and mean absolute error (MAE) metrics.
For evaluating the localization performance, we followed~\cite{brachmann2021pseudoGT} and reported the rotation error as $\Delta{\m R}= \angle (\m R \m R^{*\top})$ and the translation error as $\Delta{\v t} = \l2{\v t - \v t^*}$, where $\m R^*\in\,SO(3)$ and $\v t^*\in\real^3$ are the ground truth camera pose of the query image available in the benchmark~\cite{brachmann2021pseudoGT}.

\vspace{-2mm}
\paragraph{Results of 3D point recovery}
\label{sec:geometry_results}
\noindent
As shown in Fig.~\ref{fig:cdf_geo}, 
sphere cloud achieves significantly higher geometric errors compared to ULC~\cite{speciale2019privacy} and PPL/PPL+~\cite{lee2023ppl} due to its ability to neutralize the geometry-revealing attack~\cite{chelani2021privacy}.
This pattern is repeated in Fig.~\ref{fig:cdf_LPIPS} where the sphere cloud shows the lowest image quality due to large geometric errors.
Also,
Fig.~\figref{sphere_invsfm} shows that no content is revealed using the 3D points estimated from the sphere cloud.

\vspace{-2mm}
\paragraph{Direct image reconstruction about the map centroid}
\label{sec:sphere_center_inversion}

We tried to assess the sphere cloud's resilience to a new type of attack based on direct image-synthesis about the sphere centre.
For this purpose, we rotated the camera viewpoint about the sphere centre (map centroid) and projected sphere points to a virtual image plane for image reconstruction via InvSfM. 
Since no ground truth is available for these synthesized views, we used the pseudo-GT in Fig.~\figref{center_inversion} for evaluation as described in  Table~\ref{tab:reconstruction}.
For ULC~\cite{speciale2019privacy} and  PPL/PPL+~\cite{lee2023ppl}, we used the recovered 3D points using~\cite{chelani2021privacy} to reconstruct these images.
As shown in the same table, the sphere cloud achieves relatively high privacy-preserving ability against this new attack on 12-Scenes which is qualitatively verified in Fig.~\figref{center_inversion}.
However, sphere cloud surprisingly underperforms in the 7-Scenes~\cite{shotton2013scene} dataset.
While this requires further investigation, we anticipate this is partly due to noisy pseudo-GT images in 7-Scenes as observed in Fig.~\figref{center_inversion}.

\begin{table}[h!]
\centering
\fontsize{6}{9}\selectfont
\renewcommand{\arraystretch}{1.3}
\setlength{\tabcolsep}{1.0pt} 
\caption{
Comprehensive comparison of localization performance across different visual localization methods which are categorized into two groups: \emph{Image-based} and \emph{depth-guided}.
The median error of rotation ($\Delta{\m R}$) [$^{\circ}$], translation $(\Delta{\v t})$ [cm] and the ratio (\%) of the query images localized within each rotation and translation threshold are reported as in~\cite{brachmann2021pseudoGT}. Runtime [ms] includes the whole iterations of LO-RANSAC~\cite{chum2003loransac,PoseLib} and non-linear refinement. $\star$ indicates that results are from the official code in~\cite{brachmann2021pseudoGT} and runtime of \cite{humenberger2020robust} is not reported in~\cite{brachmann2021pseudoGT}.
Note, oracle denotes results of depth oracle (z-oracle) in sphere cloud .
\textbf{Bold} indicates the best result in each category.
}
\begin{tabular}{c|c|c|ccc|cccc|cc}
\hline\hline
&  & \multicolumn{4}{c|}{Image-based localization} & \multicolumn{6}{c}{Depth-guided localization} \\ \hline
Dataset & Metric & $\begin{tabular}{c} Point cloud\\\cite{persson2018lambda}\end{tabular}$ 
& $\begin{tabular}{c} ULC\\\cite{speciale2019privacy}\end{tabular}$ 
& $\begin{tabular}{c} PPL\\\cite{lee2023ppl}\end{tabular}$ 
& $\begin{tabular}{c} PPL+\\\cite{lee2023ppl}\end{tabular}$ 
& $\begin{tabular}{c} $\text{DVLAD}^{\star}$\\+R2D2(+D)\cite{humenberger2020robust}\end{tabular}$ 
& $\begin{tabular}{c} $\text{DSAC*}$\\(+D)\cite{brachmann2021dsacstar}\end{tabular}$
& $\begin{tabular}{c} Sphere\\($\eta$$=$25\%)\end{tabular}$
& $\begin{tabular}{c} Sphere\\($\eta$$=$33\%)\end{tabular}$ 
& $\begin{tabular}{c} Sphere (oracle)\\($\eta$$=$25\%)\end{tabular}$
& $\begin{tabular}{c} Sphere (oracle)\\($\eta$$=$33\%)\end{tabular}$ 
\\
\hline
\multirow{5}{*}{$\begin{tabular}{c}12-Scenes\\\cite{valentin2016energy}\end{tabular}$}

    & $\Delta{\m R} (^{\circ})$~($\downarrow$) & 0.139 & \textbf{0.159} & 0.170 & 0.168 & 0.389 & 0.397 & 0.300 & \textbf{0.288} & 0.240 & \textbf{0.232} \\
    & \textbf{$\Delta{\v t} $} (cm)~($\downarrow$) & 0.627 & \textbf{0.727} & 0.775 & 0.765 & 0.931 & \textbf{0.735} & 1.310 & 1.282 & 0.601 & \textbf{0.577} \\
    \cline{2-12}

    & $\Delta{\m R}$$<$$3^{\circ}$ (\%)~($\uparrow$) & 100.0 & \textbf{100.0} & \textbf{100.0} & \textbf{100.0} & 99.73 & \textbf{99.98} & 99.00 & 99.34 & 99.90 & \textbf{100.0} \\

    & $\Delta{\v t}$$<$3cm (\%)~($\uparrow$) & 97.94 & \textbf{95.88} & 95.16 & 95.13 & 97.06 & \textbf{99.21} & 86.97 & 87.86 & 97.22 & \textbf{97.60} \\
    \cline{2-12}

    & \textrm{Runtime(ms)} ($\downarrow$) & 3 & 96 & \textbf{91} & \textbf{91} & - & 84 & 48  & \textbf{24} & 22 & \textbf{13}\\    
    \cline{1-12} 

\multirow{5}{*}{ $\begin{tabular}{c} 7-Scenes\\\cite{shotton2013scene}\end{tabular}$}
    & $\Delta{\m R} (^{\circ})$~($\downarrow$) & 0.174 & \textbf{0.201} & 0.206 & 0.207 & 0.966 & 0.655 & 0.438 & \textbf{0.405} & 0.262 & \textbf{0.255}\\ 
    & $\Delta{\v t}$ (cm)~($\downarrow$) & 0.493 & \textbf{0.613} & 0.647 & 0.647 & 2.857 & \textbf{1.573} & 2.119 & 2.051 & 0.459 & \textbf{0.443}\\
    \cline{2-12}
    
    & $\Delta{\m R}$$<$$3^{\circ}$ (\%)~($\uparrow$) & 100.0 & \textbf{100.0} & \textbf{100.0} & \textbf{100.0} & 96.11 & \textbf{99.05} & 97.00 & 97.58 & 99.86 & \textbf{99.93} \\

    & $\Delta{\v t}$$<$3cm (\%)~($\uparrow$) & 99.85 & \textbf{99.32} & 99.12 & 98.96 & 55.90 & \textbf{82.81} & 69.75 & 70.93 & 98.21 & \textbf{98.51} \\
    \cline{2-12}

    & \textrm{Runtime (ms)} ($\downarrow$) & 3 & 82 & \textbf{78} & 79 & - & 80 & 52 & \textbf{25} & 31 & \textbf{16} \\
\hline\hline
\end{tabular}
\vspace{-2mm}
\label{tab:localization}
\end{table}

\paragraph{Localization results}
\label{sec:localization_results}
Table~\ref{tab:localization} presents the overall performance of different localization methods including DVLAD+R2D2(+D)~\cite{humenberger2020robust,revaud2019r2d2,torii201524} and $\text{DSAC}^{*}$(+D)~\cite{brachmann2021dsacstar} both of which serve as baselines for evaluating depth-guided approaches.
Notably, sphere cloud with $\eta$=33\% runs real-time unlike ULC and PPL/PPL+ as the result of being able to use the p3P solver~\cite{persson2018lambda}.
On the downside, we observe reduced camera localization accuracy compared to image-based methods, and a slight increase in translation errors when compared to other depth-guided methods.
Among the depth-guided methods, sphere cloud exhibits the lowest median rotational errors on both datasets while DSAC* (trained with rendered depth maps) achieves the lowest translation errors.
Overall, the sphere cloud shows efficient runtime with a slight reduction in translation accuracy compared to other depth-guided methods.

We also compared the localization performance of the \emph{depth-oracle} case of the sphere cloud.
Due to the reduced noise in depth measurements, we observed improvements in the localization accuracy of the sphere cloud in oracle cases, likely from the accuracy of the solutions obtained by the p3P solver~\cite{persson2018lambda}.
Surprisingly, the oracle cases of the sphere cloud show competitive localization accuracy compared to ULC~\cite{speciale2019privacy} and PPL/PPL+~\cite{lee2023ppl} and significantly outperform them in runtime due to the allowance of the efficient p3P solver.
Hence, we anticipate further research of reducing noises on depth measurements will improve the localization accuracy of the sphere cloud.

\vspace{-4mm}
\section{Conclusion and limitations}
\label{sec:conclusion}
\vspace{-2mm}
In this work, we presented a new privacy-preserving scene representation called \emph{sphere cloud} which can nullify the known geometry-revealing attack for line clouds.
We noted the main challenges in realizing this representation, namely the possibility of a new type of attack directly revealing images from the sphere points about the map centroid and the issue of unresolved translation scale.
We addressed these issues by introducing fake points with recycled real descriptors to thwart direct image reconstruction and presenting an efficient RGB-D privacy-preserving localization framework to guide the translation scale.
Experimental results showed that sphere cloud successfully neutralizes the known geometry attack and gains resilience to a new direct attack while reporting around 20--30 fps localization speed.
This demonstrates its potential as an efficient privacy-preserving scene representation.

Out of many limitations, our framework exhibits lower translation accuracy due to noisy depth measurements.
We also observe the trade-off between localization accuracy and privacy-preserving performance when the proportion of true positive sphere points ($\eta$) changes and we have not outlined a principled approach to setting $\eta$ along with other hyperparameters (\eg $\sigma$) to yield optimal performance.
We leave improvements to these for future work.

\clearpage
\begin{center}
    {\large \textbf{Supplementary Materials for Depth-Guided Privacy-Preserving Visual Localization Using 3D Sphere Clouds} \par}
\end{center}

\input{supp_camera_ready}

\clearpage

\paragraph{Acknowledgement} 
This work was supported by the NRF (National Research Foundation of Korea) grants funded by the Korea government (MSIT) (No. 2022R1C1C1004907).

\vspace{-2mm}
\bibliography{egbib}
\clearpage

\end{document}

%% file: supp_camera_ready.tex
\begin{abstract}
In this supplementary document, we present the extra explanations and experimental results advocating our research on sphere cloud.
In Sec.~\ref{sec:sphere_details}, we present details about sphere clouds including localization of sphere cloud (Sec.\ref{sec:cheirlatiy}, Sec.\ref{sec:location_centre}, Sec.\ref{sec:details_localization}, Sec.\ref{sec:details_depth_reguralization}), and the direct inversion attack at centre (Sec.~\ref{sec:details_inversion}).
In Sec.~\ref{sec:ablation}, we present the comprehensive results of ablation studies with sphere clouds, \eg the localization accuracy with different hyper-parameters in sphere clouds (Sec.\ref{sec:std_ratio}, Sec.\ref{sec:depth_reguralrization}, Sec.\ref{sec:eta_ablation}) and effect of noise on depth measurements (Sec.\ref{sec:depth_oracle}).
Our code is available at \url{https://github.com/PHANTOM0122/Sphere-cloud}.
\end{abstract}

\vspace{-4mm}
\section{More details about sphere clouds}
\label{sec:sphere_details}

\subsection{Visualization of need for cheirality constraints on sphere clouds}
\label{sec:cheirlatiy}
\begin{figure}[!ht]
    \centering
    \subfigure[][The correct solution with proper cheirality\figlabel{fig:vp3p_with_cheirality}]{
        \includegraphics[width=0.45\linewidth]{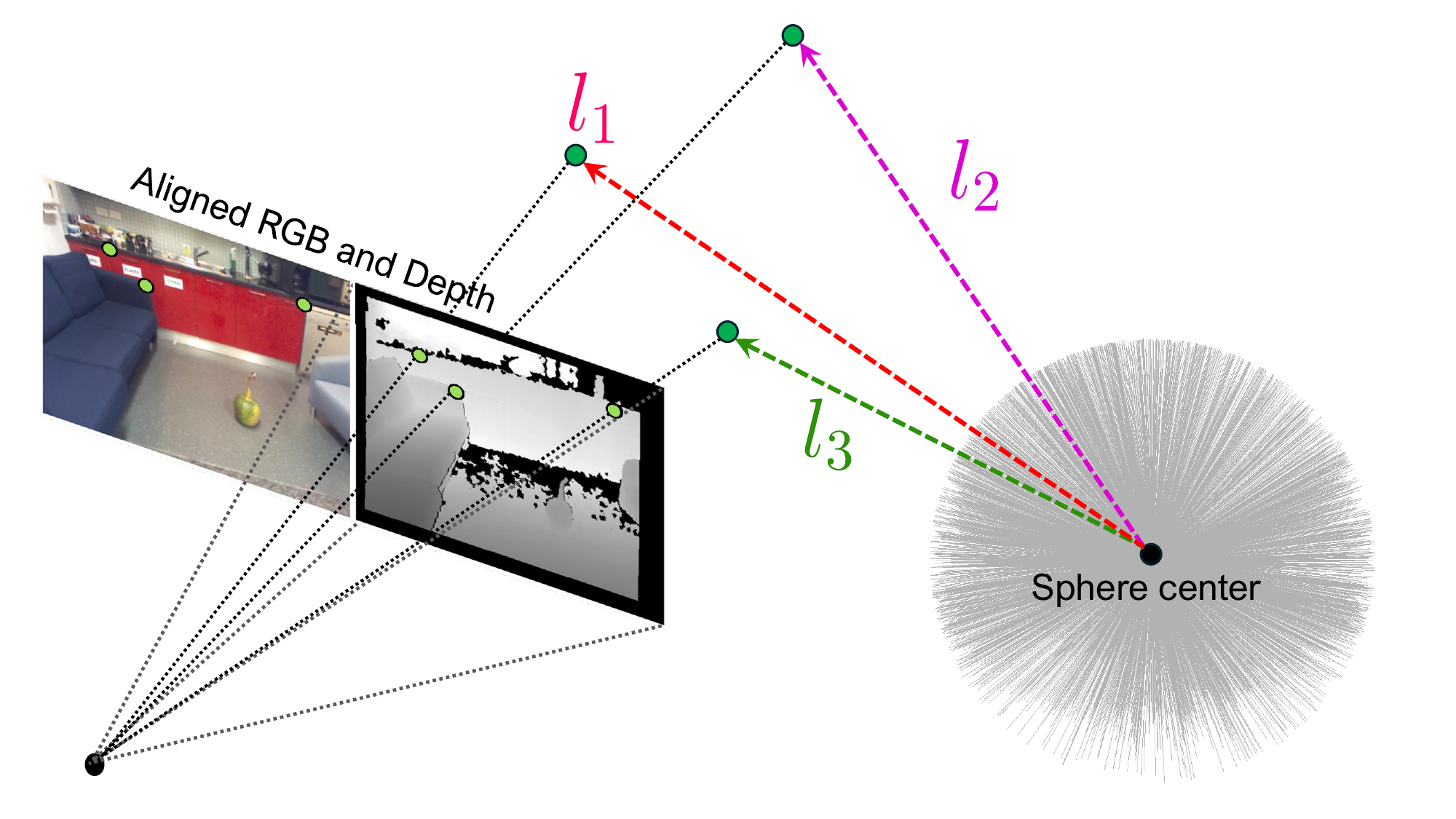}
    }    
    \subfigure[][A potential solution with incorrect cheirality\figlabel{fig:vp3p_no_cheirality}]{
        \includegraphics[width=0.45\linewidth]{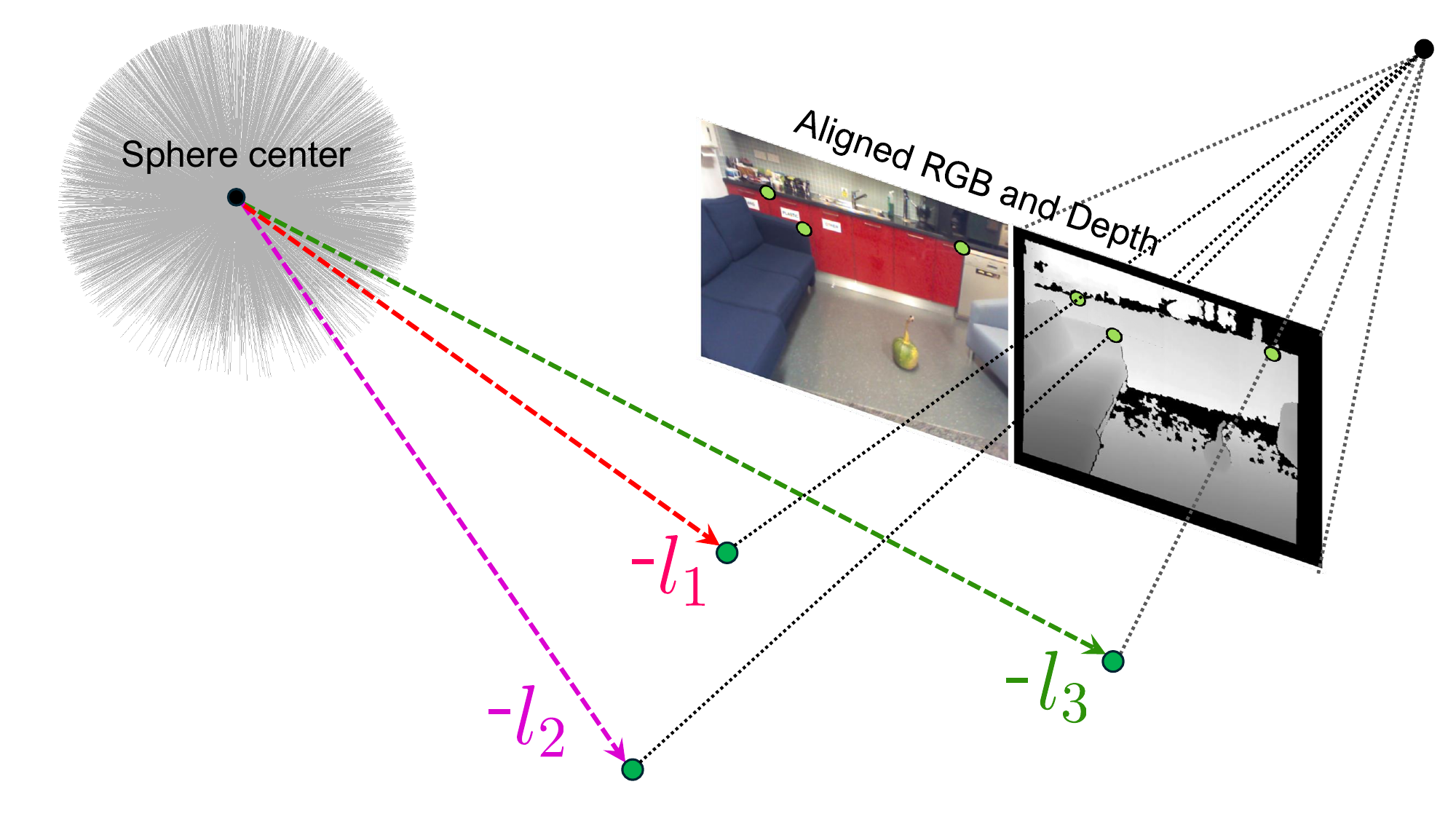}
    }
    \vspace{-1mm}
    \caption{
    Comparison of the solutions derived from the p3P minimal solver~\cite{persson2018lambda} in the sphere cloud (a) with cheirality constraint~\cite{hartley1998cheirality} and (b) without cheirality constraint~\cite{hartley1998cheirality}. If the cheirality constraint isn't enforced, the undesirable solution is induced.
    }
    \figlabel{fig:cheirality}
\end{figure}

In the sphere cloud, we applied the cheirality constraint~\cite{hartley1998cheirality} by ensuring that the direction of 3D lines is from the centre of the sphere cloud towards the 3D points.
Without the cheirality constraint, we may end up with a solution shown in Fig.~\figref{fig:vp3p_no_cheirality}, where the camera rotation is reversed due to the overall ambiguity in 3D line directions (points being triangulated on the opposite side of lines). 
Since this solution encompasses the same epipolar error as the correct pose, we enforced a simple cheirality check to yield correct camera pose from the initial LO-RANSAC~\cite{chum2003loransac} phase.
    
\subsection{Effect of sphere centre position on localization accuracy}
\label{sec:location_centre}

\begin{figure}[!ht]
    \centering
    \subfigure[][Centre at \textbf{centroid}~\figlabel{fig:centroid_loc}]{
        \includegraphics[width=0.45\linewidth]{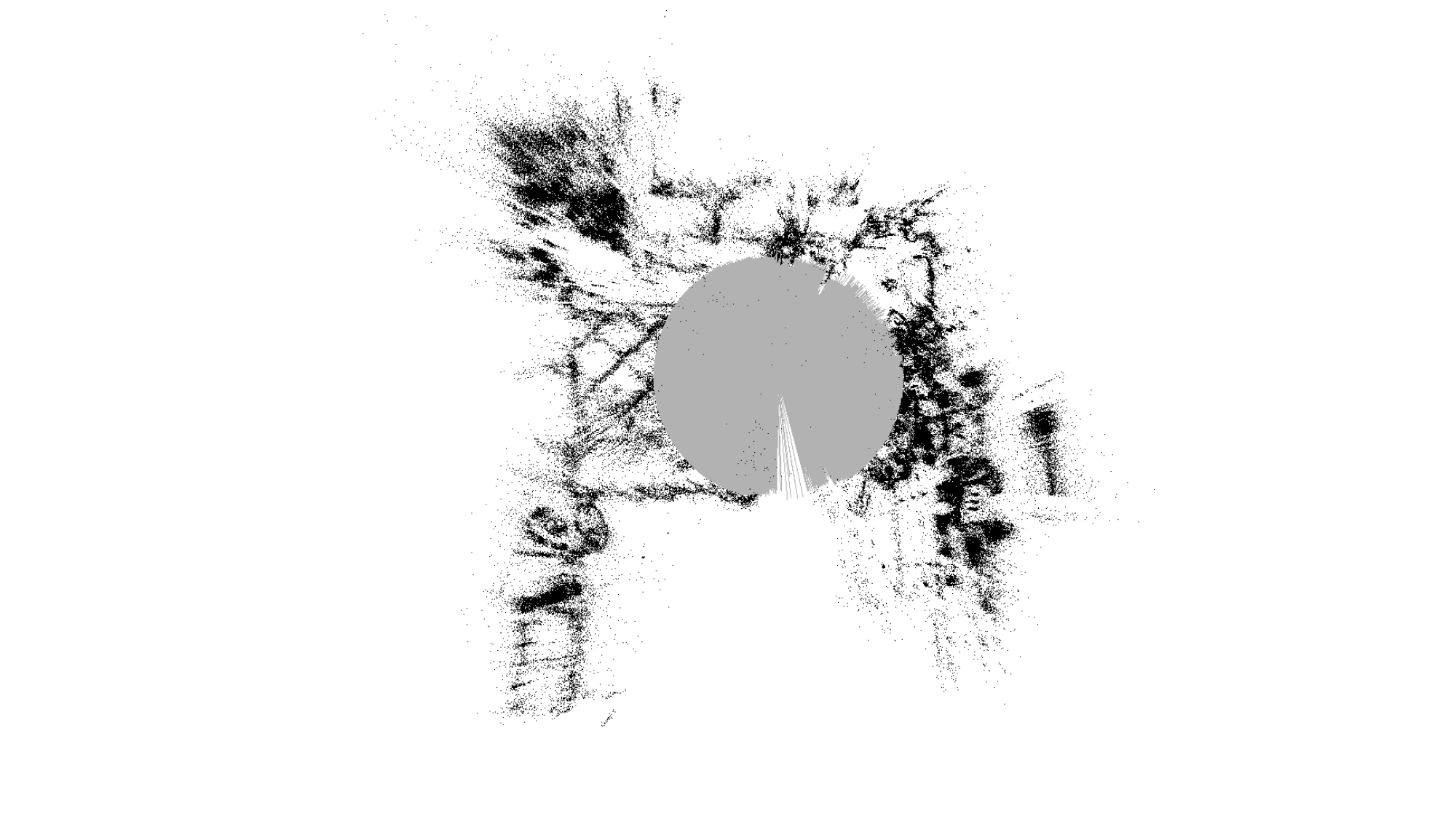}
    }
    \subfigure[][Centre \textbf{far from the scene}~\figlabel{fig:no_centroid_loc}]{
        \includegraphics[width=0.45\linewidth]{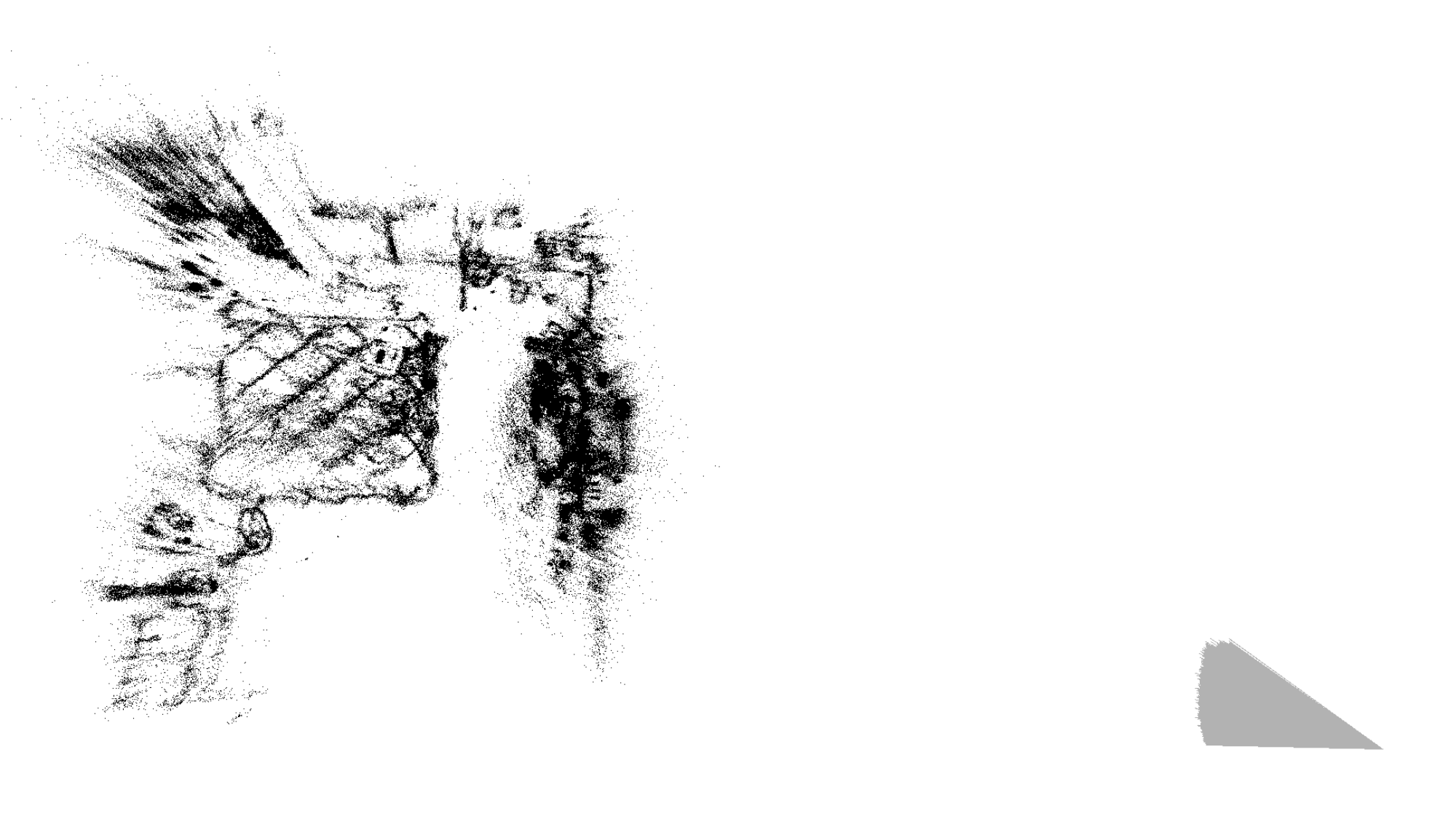}
    }
    \vspace{-1mm}
    \caption{
    Bird's-Eye-View(BEV) of 3D point cloud (black points) from \emph{apt2 luke} in 12-Scenes~\cite{valentin2016energy} and sphere cloud (gray lines) yielded from different locations of the sphere centre. As shown in (a), the distribution of lifted 3D lines is uniformly oriented where the centre is located at the centroid of the 3D points. In contrast, in (b), the intersection point is located far from the scene resulting in a quasi-parallel distribution of 3D lines.
    }
    \figlabel{fig:centre_location}
\end{figure}

While the location of the intersection point does not affect the neutralization of the geometry-inversion attack~\cite{chelani2021privacy}, it can affect the distribution of 3D line directions. \eg if the intersection point is far away from the scene then the lifted 3D lines become near-parallel (See Fig.~\figref{fig:no_centroid_loc}).
We empirically observed that this can unstabilize the localization accuracy as shown in Table~\ref{tab:location_results}.
Hence, we simply set the sphere centre as a centroid to avoid undesirable cases in localization and to yield uniformly distributed 3D lines.

\begin{table}[!ht]
\vspace{-2mm}
\centering
\fontsize{8}{10}\selectfont
\renewcommand{\arraystretch}{1.0}
\setlength{\tabcolsep}{3.0pt} 
\caption{
Comparison of localization performance according to different locations of the sphere centre.
For ablation, we set the proportion of true positive sphere points $\eta=100\%$. 
}
\vspace{+1mm}
\begin{tabular}{c|cc|cc|cc|cc}
\hline\hline
Centre location & \multicolumn{4}{c|}{Map centroid (Fig.~\figref{fig:centroid_loc})}  & \multicolumn{4}{c}{Distant from the scene (Fig.~\figref{fig:no_centroid_loc})} \\
\hline
\multirow{2}{*}{Localization metrics} & \multicolumn{2}{c|}{Rotation error [$^{\circ}$]} & \multicolumn{2}{c|}{Translation error [m]} & \multicolumn{2}{c|}{Rotation error [$^{\circ}$]} & \multicolumn{2}{c}{Translation error [m]} \\
\cline{2-9}
 & mean & median & mean & median & mean & median & mean & median \\
\hline
12-Scenes\cite{valentin2016energy} 
& 0.323 & 0.262 & 0.015 & 0.012 & 0.434 & 0.303 & 0.787 & 0.701\\
\hline
7-Scenes\cite{shotton2013scene} 
& 0.682 & 0.349 & 0.038 & 0.019 & 1.387 & 0.732 & 4.099 & 2.679 \\
\hline\hline
\end{tabular}
\vspace{-6mm}
\label{tab:location_results}
\end{table}

\subsection{Details on the construction procedure of sphere clouds}
\label{sec:detailed_construction_procedure}
\begin{algorithm}[!ht]
\centering
        \footnotesize
	\caption{
	    Sphere cloud construction procedure.
	}
	\label{alg:sphere_cloud}
	\begin{algorithmic}[1]
            \Statex \textbf{Inputs:} 3D point cloud~ {$\m{X} \in \real^{(3+128+3)\times N}$} containing XYZ values, SIFT descriptors~\cite{lowe2004sift} and RGB values
            \Statex $\m X_{xyz} \in \mathbb{R}^{3 \times N} \gets \text{XYZ values of 3D points}$
            \Statex $\m X_{desc} \in \mathbb{R}^{128 \times N} \gets \text{SIFT Descriptors of 3D points}$
            \Statex $\m X_{rgb} \in \mathbb{R}^{3 \times N} \gets \text{RGB values of 3D points}$
            \Statex $\eta \gets$ Proportion of true positive sphere points 
            \Statex $\sigma^2 \gets$ Variance of 3D Gaussian noise 
            \Statex $\varepsilon \sim N(\v 0,\sigma^2 \m I) \gets $ 3D Gaussian noise
            \newline

            \LeftComment{\textcolor{gray}{Calculate the centroid of the sparse 3D point cloud}}
            \State $\v c \gets \mathtt{mean}(\m X_{xyz})$

            \LeftComment{\textcolor{gray}{Project all 3D points into the unit sphere centred at the centroid}}
            
            \State $\hat {\m X} = [~]$

            \For{$i = 1,\cdots, N$}
                \State $\hat {\m X}\text{.append}(\frac{(\m X_{(i)\_xyz} - \v{c})}{\l2{(\m X_{(i)\_xyz} - \v{c})}})$
            \EndFor

            \LeftComment{\textcolor{gray}{Randomly split $\eta\%$ of 3D points as true positive sphere points, $(100-\eta)\%$ as discarded points}}
            \State \m{random.shuffle}($\m{\hat{X}}$)
            \State $\hat {\m X}_{\text{true}}, \hat {\m X}_{\text{discarded}} \gets \m{\hat{X}}[:\text{int}(\eta \cdot N)], \m{\hat{X}}[\text{int}(\eta \cdot N):]$
            
            \LeftComment{\textcolor{gray}{Generate fake sphere points}}
            \State $\hat {\m X}_{\text{fake}} \gets [~]$
            \For{$k = 1,\cdots, \text{len}(\hat {\m X}_{\text{false}})$}
                \LeftComment{\textcolor{gray}{Calculate the division ($k/\text{len}(\hat {\m X}_{\text{true}})$) and return the quotient $j$, followed by the remainder $i$}}
                \State $j, i \gets \m{divmod}(k, ~\text{len}(\hat {\m X}_{\text{true}}))$
                
                \LeftComment{\textcolor{gray}{Assign a fake point location}}
                \State $\v z_{ij} \gets (\hat {\v x}_{\text{true}(i)\_xyz} + \varepsilon) / \norm{\hat {\v x}_{\text{true}(i)\_xyz} + \varepsilon}$
                \State $\hat {\v x}_{\text{fake}(k)\_xyz} \in \real^{3} \gets \v z_{ij}$
                
                \LeftComment{\textcolor{gray}{Recycle discarded descriptor and RGB values}}
                \State $\hat {\v x}_{\text{fake}(k)\_desc} \in \real^{128} \gets ~\hat {\m X}_{\text{discarded}(k)\_desc}$ 
                \State $\hat {\v x}_{\text{fake}(k)\_rgb} \in \real^{3} \gets ~\hat {\m X}_{\text{discarded}(k)\_rgb} $
                \State $\hat {\m X}_{\text{fake}}\text{.append}(\hat {\v x}_{\text{fake}(k)})$
 
            \EndFor

            \LeftComment{\textcolor{gray}{Concatenate true positive and fake sphere points, then shuffle}}
            \State $\hat {\m X}_{\text{sphere}} \gets \mathtt{concatenate}(\hat {\m X}_{\text{true}}, \hat {\m X}_{\text{fake}})$
            \State $\m{random.shuffle}(\hat {\m X}_{\text{sphere}})$
            \newline
            
            \Statex \textbf{Output:} Sphere cloud {$\hat {\m X}_{\text{sphere}}$}
	\end{algorithmic}
\end{algorithm}

Algo.~\ref{alg:sphere_cloud} shows the detailed process of constructing the sphere cloud, including our enhanced construction strategy.
For construction, the 3D point cloud, consisting of N points containing information of xyz coordinates, SIFT descriptors~\cite{lowe2004sift}, and RGB values, is needed for lifting into the sphere cloud.
Also, hyperparameters are required, \eg $\eta$ which is the ratio of true positive sphere points to total points controlling the augmentation of fake points, and $\sigma^2$ denoting variance of 3D Gaussian noise ($\varepsilon$).
Through the projection process, the basic sphere cloud is yielded where the sphere centre is the centroid of the 3D points map.
However, since the basic spherical clouds can be prone to direct image inversion attack at the centre of the sphere, we implemented sparsification of the sphere points based on \(\eta\), followed by a fake sphere points augmentation step.
This generates the final enhanced spherical clouds.
For the rest of details, please refer to Algo.~\ref{alg:sphere_cloud}.

\vspace{-4mm}
\subsection{Details on the localization algorithm for sphere clouds}
\label{sec:details_localization}
\vspace{-2mm}
Algo.~\ref{algo:sphere_initial} presents the detailed process of localization with the sphere cloud. The threshold values for the decision of inliers are also shown in Algo.~\ref{algo:sphere_initial}.
We utilized LO-RANSAC~\cite{chum2003loransac, PoseLib} for the pose estimation, and Levenberg-Marquardt (LM) optimization was used for the refinement of the pose.
Our implementations of calculating the Jacobian matrix can be found in the accompanying PoseLib~\cite{PoseLib} modification.

\subsection{Illustration of different inversion attack scenarios}
\label{sec:details_inversion}
In our experiments, we conducted two types of image inversion scenarios with InvSfM~\cite{pittaluga2019revealing} based on the input camera pose, \emph{i.e} (a) inversion at test camera pose (\textcolor{red}{red camera} in Fig.~\figref{fig:attack pipeline}), and (b) inversion at sphere centre (\textcolor{blue}{blue camera} in Fig.~\figref{fig:attack pipeline}).
In (a), we fed the 2D projection of the 3D points to InvSfM by utilizing the ground truth 6-DOF pose of the test camera, while in (b), we projected the 3D points to a virtual camera located at the sphere centre with varying rotation matrices. 

Fig.~\figref{fig:attack pipeline} shows the recovered scene details for each scenario.
In (a), where the inversion was conducted at the test camera pose,
it is evident that inverting the 3D point cloud is prone to privacy leaks. 
In contrast, the recovered image from the sphere cloud is blank as the sphere cloud is robust 
 to the geometry-revealing attack~\cite{chelani2021privacy}.

As stated above, the images in scenario (b) are the inversion results from the centroid of the 3D map but since there is no ground truth, we set the rotation matrix to align the image as close as possible to the nearest test image to make qualitative comparisons.
Again, the image reconstruction from the 3D point cloud recovers the scene details as in (a).
On the other hand, while the images recovered from the sphere cloud are not blank anymore, they are still difficult to recognize the scene details. 
This demonstrates our enhanced sphere cloud with fake points augmentation can neutralize this potential attack.
Further visualization of two different inversion scenarios for other scenes are shown in Fig.~\figref{fig:test camera inversion} and Fig.~\figref{fig:center inversion}.
\begin{figure}[t!]
    \centering
    \includegraphics[width=1.0\linewidth]{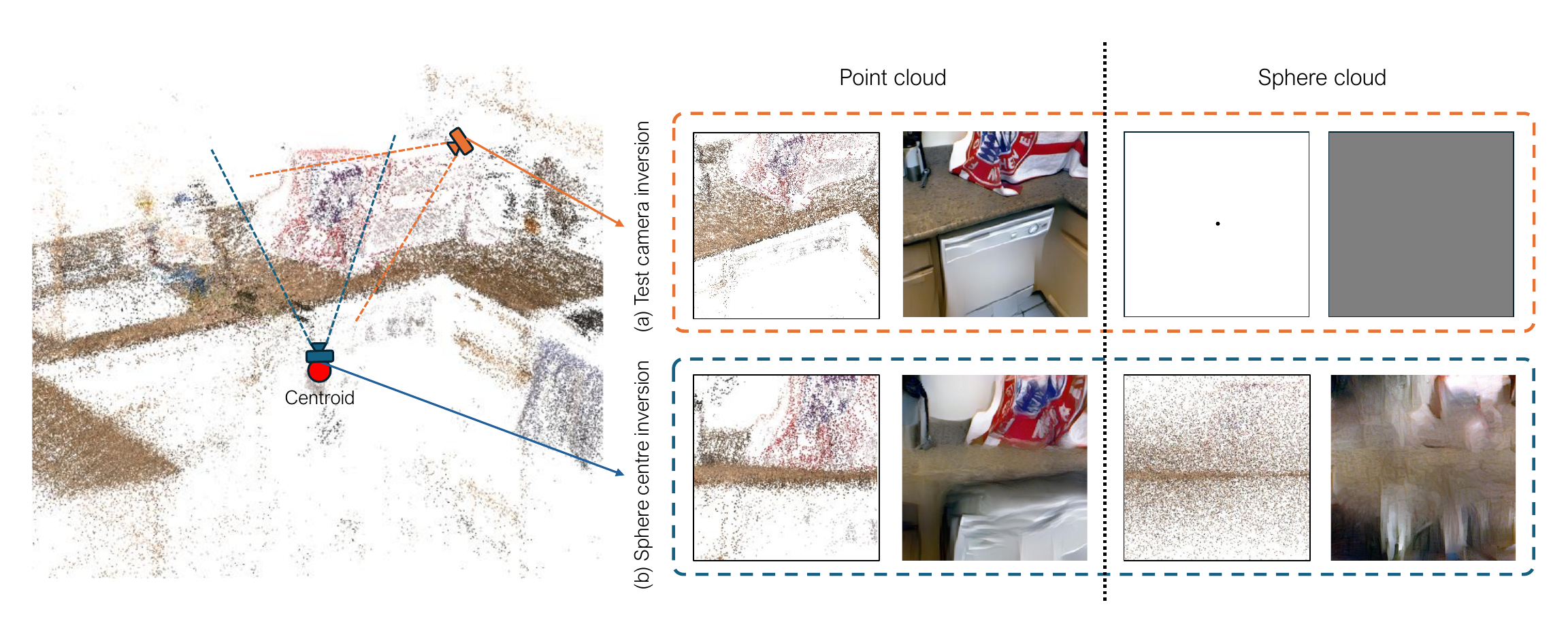}
    \vspace{-8mm}
    \caption{
    Comparison of different inversion attack~\cite{pittaluga2019revealing} scenarios for \emph{apt1 kitchen} in~\cite{valentin2016energy}.
    (a) indicates the inversion at the test camera's 6-DOF ground truth pose (\textcolor{red}{red camera}), where the 2D projections of 3D point cloud and recovered 3D points from sphere cloud ($\eta$$=$$25\%$) with~\cite{chelani2021privacy} were fed to InvSfM~\cite{pittaluga2019revealing}.
    In (b), the 3D points are projected to a virtual camera (\textcolor{blue}{blue camera}) located at the sphere centre and the rotation matrix was set to align the image of (a) as close as possible. Then, projected 2D points were fed to InvSfM to yield the scene detailed images.
    }
    \figlabel{fig:attack pipeline}
\end{figure}

\vspace{-4mm}
\subsection{Details on the depth regularization loss function}
\label{sec:details_depth_reguralization}
In this part, we illustrate more detailed about the derivation of depth regularization in Eq.(2) of our main paper~\cite{authors24bmvcmain}.

As mentioned in~\cite{authors24bmvcmain}, ideally the $i$th 3D keypoint $\v p_{i} \in \real^3$ should lie along its corresponding $i$th 3D line(ray) originating from the centre of the sphere and passing through the sphere point $\hat{\v x}_i \in S^2$.
Also due to the enforced cheirality constraint~\cite{hartley1998cheirality}, $\v p_i$ should remain on the positive direction of the 3D ray($\hat{\v x}_i$) and as we set the sphere centre as an origin of the world coordinate, the below equation can be derived:
\vspace{-2mm}
\begin{align}
\alpha_i \hat{\v x}_i = \m R \v p_i + \v t,
\eqlabel{eq:intersection_eq1}
\end{align} 
where $[\m R|\v t]$ denotes query-to-world pose and $\alpha_i \in \real^+$ is a positive scale factor.
In terms of the query camera's coordinates, we can reformulate Eq.~\eqref{eq:intersection_eq1} as 
\begin{align}
\v p_i = \alpha_i \m R\tr\hat{\v x}_i - \m R\tr\v t.
\eqlabel{eq:intersection_eq1_query}
\end{align}

Due to noise in the depth observations from the TOF sensor, the initial estimated pose from Eq.~\eqref{eq:intersection_eq1} can be noisy.
To mitigate the negative effect of depth noise, we employ depth regularization leveraging the constraint that 3D keypoint $\v p_i$ should lie on the 3D line $\hat{\v x}_i$ when projected onto the XZ-plane of the query camera.

We project a ray from the origin of the query camera to the 3D keypoint $\v p_{i}$ and its corresponding 3D line($\hat{\v x}_i$) into the \emph{XZ} plane of the query camera coordinate.
Leveraging Eq.~\eqref{eq:intersection_eq1_query}, we find an intersection point between the projected ray and the projected 3D line on the \emph{XZ} plane of the query camera as shown below,

\begin{align}
\beta_i \begin{pmatrix} [x_i^{TOF}] \\ [z_i^{TOF}] \end{pmatrix}
= 
\alpha_i \begin{pmatrix} [\m {R}\tr~ \hat{\v x}_i]_{X} \\ [\m {R}\tr~ \hat{\v x}_i]_{Z} \end{pmatrix}
\eqlabel{eq:intersection_eq3}
- 
\begin{pmatrix} [\m R\tr \v t]_{X} \\ [\m R\tr \v t]_{Z} \end{pmatrix}
=
\begin{pmatrix} x_i(\m R, \v t) \\ z_i(\m R, \v t) \end{pmatrix}
\end{align}
where $x_i^{TOF}$, $z_i^{TOF} \in \real$ denote X and Z element of the 3D keypoint $\v p_{i}$,
and $\alpha_i, \beta_i \in \real$ denote scale factors.
We also note that $\alpha_i, \beta_i > 0$, as we enforced cheirality constraints on both sides of the query image and sphere cloud (See sec.~\ref{sec:cheirlatiy}).
We expect that if the depth noise of the 3D keypoint is minimal, $\beta_i$($=z_i(\m R, \v t)/z_i^{TOF}$) will be close to 1.
Conversely, if the 3D keypoint will be far from the 3D line due to the significant depth noise, then the $\beta_i$ will also be far from 1.
This intuition leads to our depth regularization loss as below:
\begin{align}
L_i^d &= (\beta_i -1)^2
\eqlabel{depth_error}
\end{align}
Combined with epipolar cost as a re-projection constraint, the total loss is defined as the summation of the epipolar cost and the depth cost (Eq.~\eqref{depth_error}) multiplied by $\lambda$.

\section{Ablation study}
\label{sec:ablation}

\begin{figure}[ht]
    \vspace{-2mm}
    \centering
    \subfigure[12-Scenes~\cite{valentin2016energy}~\figlabel{cdf_oracle_12}]{
        \includegraphics[width=0.22\linewidth]{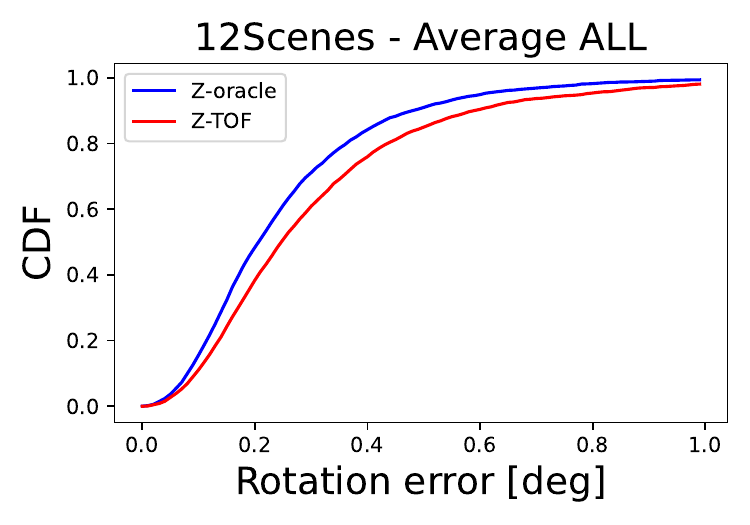}    
        \includegraphics[width=0.22\linewidth]{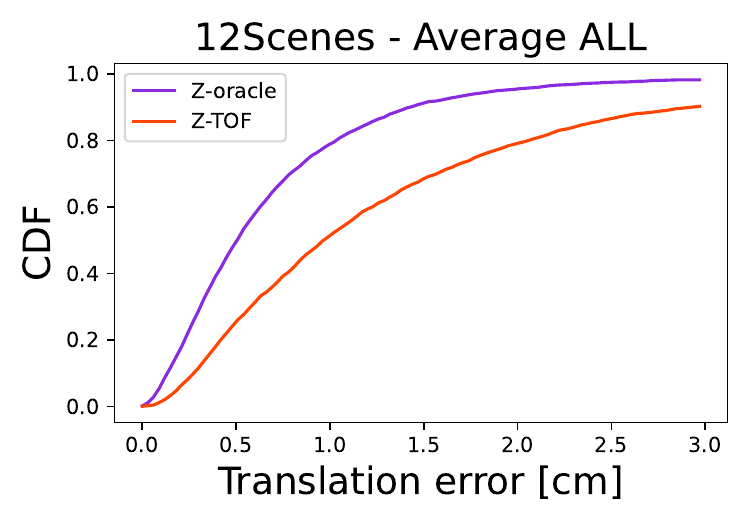}
    }
    \subfigure[7-Scenes~\cite{shotton2013scene}~\figlabel{cdf_oracle_7}]{
        \includegraphics[width=0.22\linewidth]{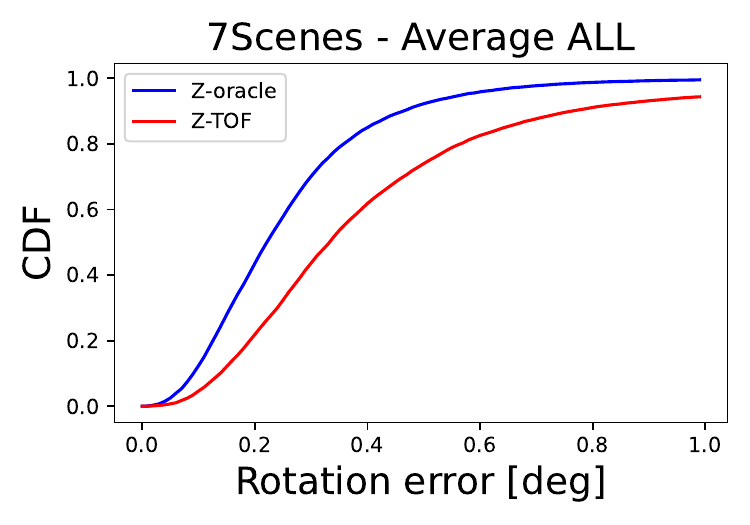}    
        \includegraphics[width=0.22\linewidth]{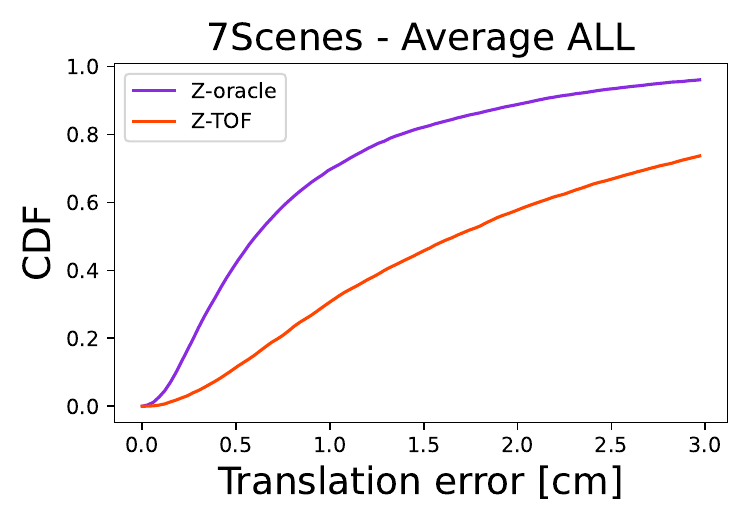}
    }    
    \vspace{-1mm}
    \caption{Cumulative distributions of rotational and translation error according to presences of depth noises (a) 12-Scenes~\cite{valentin2016energy} and (b) 7-Scenes~\cite{shotton2013scene}. 
    }
    \figlabel{fig:cdf_depth_noises}
\end{figure}

\subsection{Effect of depth measurement noise on localization accuracy}
\label{sec:depth_oracle}

Since we used depth priors from TOF sensors without any denoising pre-processing, the noises in depth measurement can negatively impact localization accuracy.
In Fig.~\figref{fig:cdf_depth_noises}, when the depth values estimated from 3D points reconstructed via SfM~\cite{schoenberger2016sfm} are used as depth priors (referred to as \emph{Z-oracle}) instead of TOF values (Z-TOF), we can confirm that noise in the depth measurement adversely affects pose accuracy, leading to degradation.

\vspace{-2mm}
\subsection{Effect of level of Gaussian noise $\sigma$ on fake point generation}
\label{sec:std_ratio}
\vspace{-2mm}
\begin{table}[t!]
\centering
\fontsize{8}{9.5}\selectfont
\renewcommand{\arraystretch}{1.0}
\setlength{\tabcolsep}{1.8pt} 
\caption{
Comparisons of recovered image quality from the direct inversion at the sphere centre with different levels of 3D Gaussian noise variance ($\sigma^2$). \textbf{Bold} indicates the best privacy-preserving performance and \underline{underline} denotes second.}
\vspace{1mm}
\begin{tabular}{c|c|cccc|cccc}
\hline\hline
\multirow{2}{*}{$\eta$} & \multirow{2}{*}{$\sigma^2$} 
& \multicolumn{4}{c|}{ 12-Scenes~\cite{valentin2016energy} } 
& \multicolumn{4}{c}{ 7-Scenes~\cite{shotton2013scene} } \\
\cline{3-10}
& & PSNR($\downarrow$) & LPIPS($\uparrow$) & SSIM($\downarrow$) & MAE($\uparrow$) & PSNR($\downarrow$) & LPIPS($\uparrow$) & SSIM($\downarrow$) & MAE($\uparrow$)
\\ \hline

\multirow{4}{*}{$\begin{tabular}{c}$25\%$\end{tabular}$}
    & 0.00 & \underline{13.31} & 0.490 & 0.481 & \underline{46.59} & 13.23 & 0.542 & 0.392 & 43.44 \\
    & 0.01 & 14.13 & 0.490 & 0.483 & 41.53 & \textbf{11.38} & \textbf{0.567} & \textbf{0.366} & \textbf{54.57} \\
    & 0.10 & \textbf{12.40} & \textbf{0.560} & \textbf{0.398} & \textbf{48.80} & \underline{12.98} & \underline{0.551} & \underline{0.378} & \underline{45.51} \\
    & 1.00 & 13.36 & \underline{0.540} & \underline{0.424} & 43.70 & 13.49 & 0.535 & 0.400 & 42.67 \\
    \cline{1-10} 

\multirow{4}{*}{ $\begin{tabular}{c}$33\%$\end{tabular}$}
    & 0.00 & 14.34 & 0.466 & 0.510 & 40.93 & \textbf{12.24} & \textbf{0.550} & \textbf{0.387} & \textbf{48.61} \\
    & 0.01 & 15.06 & 0.462 & 0.516 & 36.79 & 14.11 & 0.522 & 0.422 & 38.82 \\
    & 0.10 & \textbf{13.07} & \textbf{0.538} & \textbf{0.430} & \textbf{44.82} & \underline{13.63} & \underline{0.534} & \underline{0.403} & \underline{41.78} \\
    & 1.00 & \underline{13.86} & \underline{0.521} & \underline{0.451} & \underline{41.22} & 14.03 & 0.517 & 0.426 & 39.82 \\
    \cline{1-10}

\multirow{4}{*}{ $\begin{tabular}{c}$50\%$\end{tabular}$}
    & 0.00 & \underline{13.31} & 0.490 & 0.481 & \underline{46.59} & \textbf{14.16} & \textbf{0.511} & \textbf{0.439} & \textbf{37.95} \\
    & 0.01 & 14.13 & 0.490 & 0.483 & 41.53 & 15.74 & 0.482 & 0.482 & 31.71 \\
    & 0.10 & \textbf{12.40} & \textbf{0.560} & \textbf{0.398} & \textbf{48.80} & \underline{15.10} & \underline{0.496} & \underline{0.460} & \underline{34.68} \\
    & 1.00 & 13.36 & \underline{0.540} & \underline{0.424} & 43.70 & 15.20 & 0.492 & 0.464 & 34.42 \\
    \cline{1-10}
    
\end{tabular}
\label{tab:std_ablation}
\vspace{-4mm}
\end{table}

In order to assess the quality of images recovered from the sphere clouds, we employed Gaussian noise with varying levels of variance $\sigma^2$ and a true positive ratio $\eta$ of sphere points. As shown in Table~\ref{tab:std_ablation}, we empirically found that the sphere cloud with $\sigma^2=0.10$ has the best privacy-preserving performance in 12-Scenes~\cite{valentin2016energy}.
For 7-Scenes~\cite{shotton2013scene}, $\sigma^2=0.01$ and $\sigma^2=0.00$ performed best when $\eta=25\%$ and $\eta=33\%, 50\%$ each.
However, as demonstrated in Fig.~4 in our main paper~\cite{authors24bmvcmain}, sphere clouds with small variances (\eg $\eta=0.00, 0.01$) have potential risk revealing the geometric information of the scene as the true positions of sphere points are preserved.
Hence, we utilized $\sigma^2=0.10$ in also 7-Scenes, showing the second best scene concealing performance.
Therefore, we empirically set $\sigma^2=0.10$ for all of our experiments considering the performance of hiding scene details.

\subsection{Effect of changing the weight ($\lambda$) of the depth regularization term}
\label{sec:depth_reguralrization}

\begin{figure}[!ht]
    \centering
    \subfigure{
        \includegraphics[width=0.45\linewidth]{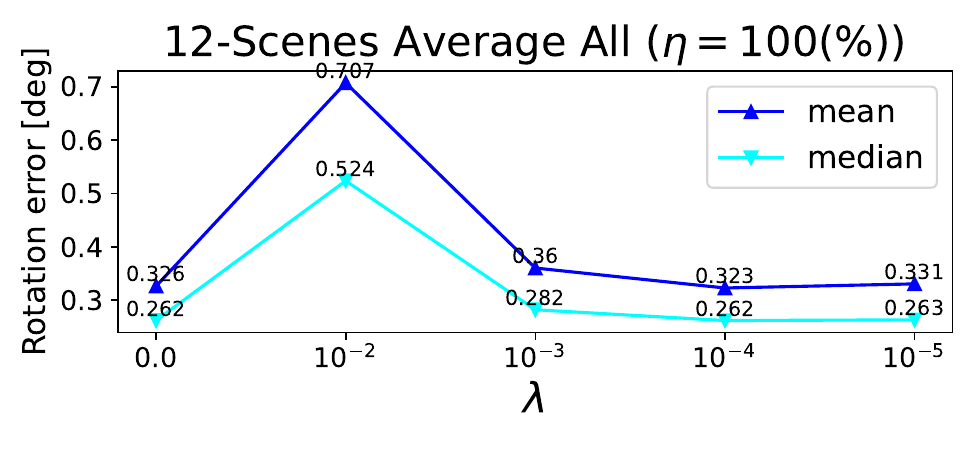}
        \includegraphics[width=0.45\linewidth]{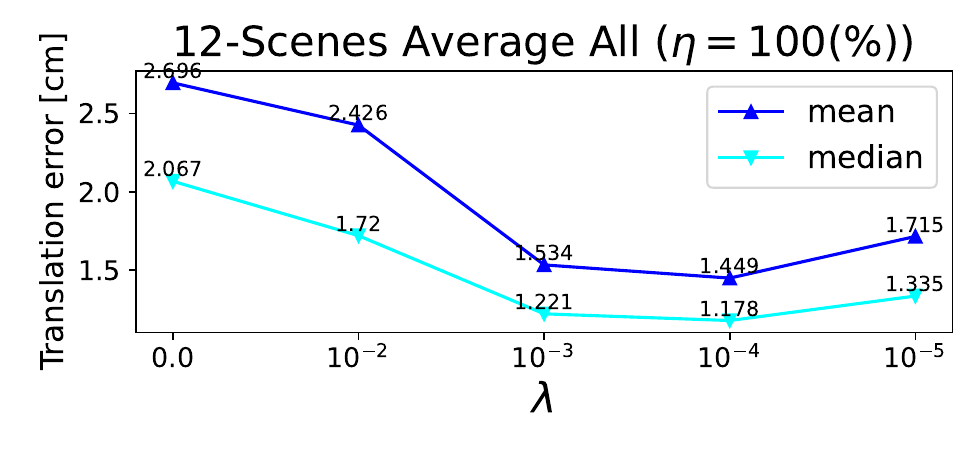}
    } 
    \\
    \vspace{-6mm}
    \subfigure{
        \includegraphics[width=0.45\linewidth]{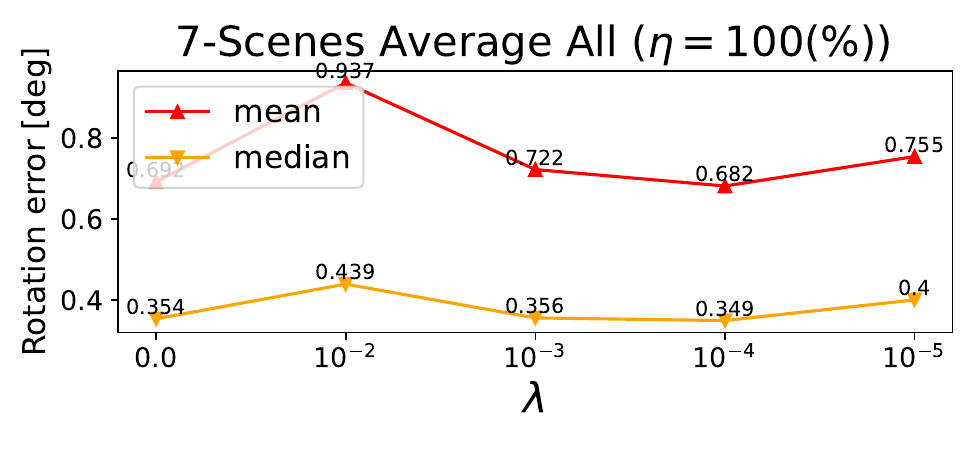} 
        \includegraphics[width=0.45\linewidth]{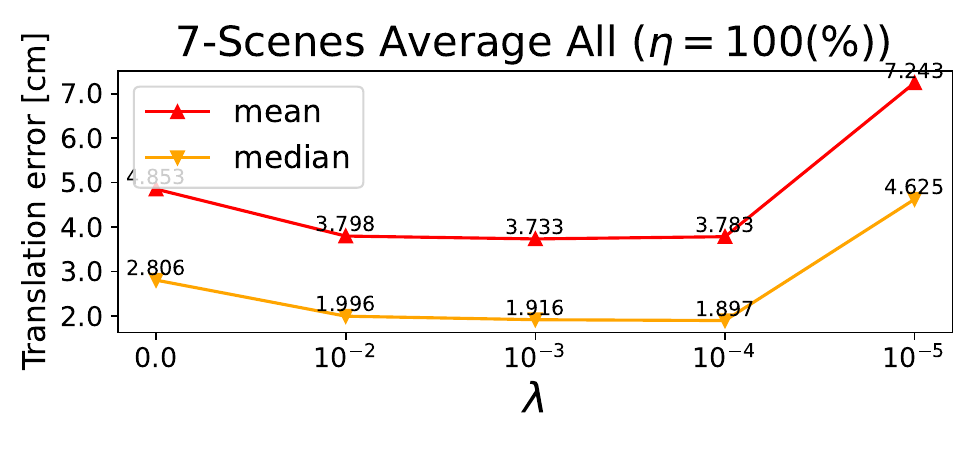}
    }
    \vspace{-4mm}
    \caption{
    The pose errors of the sphere cloud measured on two public datasets (\emph{Top:} 12-Scenes~\cite{valentin2016energy}, \emph{Bottom:} 7-Scenes~\cite{shotton2013scene}) \emph{w.r.t} the different depth regularization weight (\textbf{$\lambda$}).
    We empirically found that \textbf{$\lambda$$=$$10^{-4}$} serves as the \emph{"sweet spot"} on both datasets. 
    }
    \figlabel{fig:lambda_ablation}
    \vspace{-2mm}
\end{figure}
As shown in Fig.~\figref{fig:lambda_ablation}, we measured rotational and translation error with different weight of depth regularization (\textbf{$\lambda$}) during the refinement of the initial pose, where \textbf{$\lambda$} is the hyperparameter.
Empirically, we found that $\lambda=10^{-4}$ provides the \emph{sweet spot} for the localization accuracy.
Consequently, $\lambda=10^{-4}$ was used for all localization experiments.
Note, for the ablation, we set the proportion of the true positive sphere points $\eta=100\%$ during the measurements.

\subsection{Localization results with different ratios ($\eta$) of true positive points}
\label{sec:eta_ablation}

Table~\ref{tab:eta_comparision} shows the localization performance with different proportions of the true positive sphere points $\eta$.
When \(\eta = 100\%\), where no fake points exist in the sphere cloud, the localization accuracy and runtimes outperform the other cases.
Although the localization accuracy declines as the $\eta$ decreases, notably, the accuracy of $\eta=25\%$ does not degrade significantly compared to $\eta=100\%$.
We cautiously think this is due to the LO-RANSAC algorithm~\cite{chum2003loransac, PoseLib} being robust to outlier correspondences and allowing minimal degradation of pose accuracy but at the cost of increased runtime for sampling inliers.

\begin{table}[!ht]
\centering
\fontsize{8}{10}\selectfont
\renewcommand{\arraystretch}{1.0}
\setlength{\tabcolsep}{3.0pt} 
\caption{
Comparison of localization performance with different true positive ratios ($\eta$).
}

\begin{tabular}{c|c|cccc}
\hline\hline
Dataset & Metric 
& $\begin{tabular}{c} Sphere\\($\eta$$=$25\%)\end{tabular}$
& $\begin{tabular}{c} Sphere\\($\eta$$=$33\%)\end{tabular}$ 
& $\begin{tabular}{c} Sphere\\($\eta$$=$50\%)\end{tabular}$
& $\begin{tabular}{c} Sphere\\ ($\eta$$=$100\%)\end{tabular}$
\\
\hline

\multirow{5}{*}{$\begin{tabular}{c}12-Scenes\\\cite{valentin2016energy}\end{tabular}$}

    & $\Delta{\m R} (^{\circ})$~($\downarrow$) & 0.300 & 0.288 & 0.279 & \textbf{0.262}\\
    & \textbf{$\Delta{\v t} $} (cm)~($\downarrow$) & 1.310 & 1.282 & 1.243 & \textbf{1.178} \\
    \cline{2-6}
    
    & $\Delta{\m R}$$<$$3^{\circ}$ (\%)~($\uparrow$) & 99.00 & 99.33 & 99.45 & \textbf{99.87} \\

    & $\Delta{\v t}$$<$3cm (\%)~($\uparrow$) & 86.97 & 87.84 & 88.67 & \textbf{90.24} \\
    \cline{2-6}

    & \textrm{Runtime(ms)} ($\downarrow$) & 48 & 24 & 13 & \textbf{12} \\    
    \cline{1-6} 

\multirow{5}{*}{ $\begin{tabular}{c} 7-Scenes\\\cite{shotton2013scene}\end{tabular}$}

    & $\Delta{\m R} (^{\circ})$~($\downarrow$) & 0.438 & 0.404 & 0.387 & \textbf{0.349} \\ 
    & $\Delta{\v t}$ (cm)~($\downarrow$) & 2.119 & 2.051 & 1.937 & \textbf{1.897}\\
    \cline{2-6}
    
    & $\Delta{\m R}$$<$$3^{\circ}$ (\%)~($\uparrow$) & 97.00 & 97.57 & 98.21 & \textbf{99.02} \\

    & $\Delta{\v t}$$<$3cm (\%)~($\uparrow$) & 69.75 & 70.93 & 73.09 & \textbf{73.68} \\
    \cline{2-6}

    & \textrm{Runtime (ms)} ($\downarrow$) & 52 & 25 & 11 & \textbf{9} \\
\hline\hline
\end{tabular}
\label{tab:eta_comparision}
\end{table}

\begin{figure}[t]
    \centering
    \subfigure{
        \includegraphics[width=0.14\linewidth]{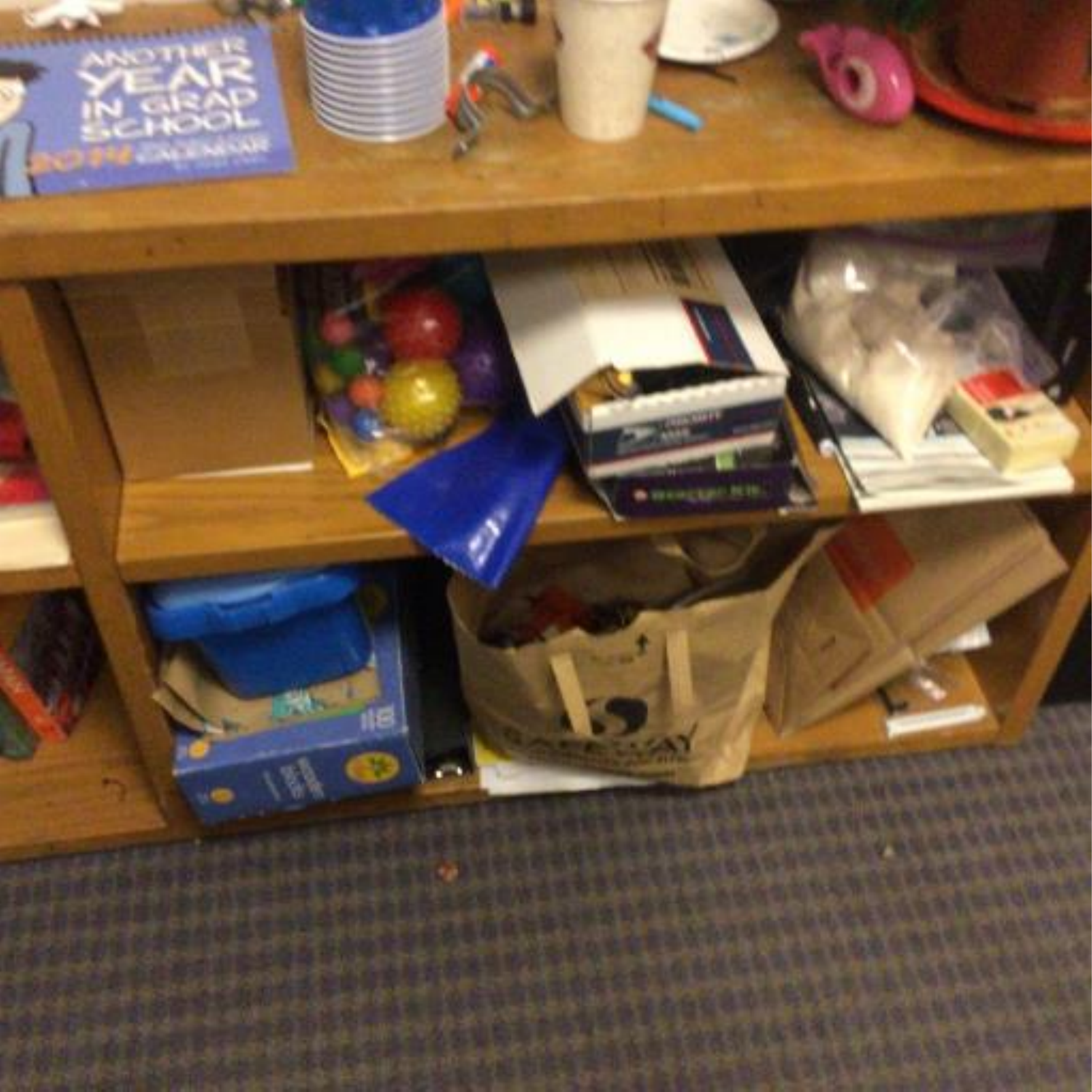}
    }
    \hspace{-3mm}
    \subfigure{
        \includegraphics[width=0.14\linewidth]{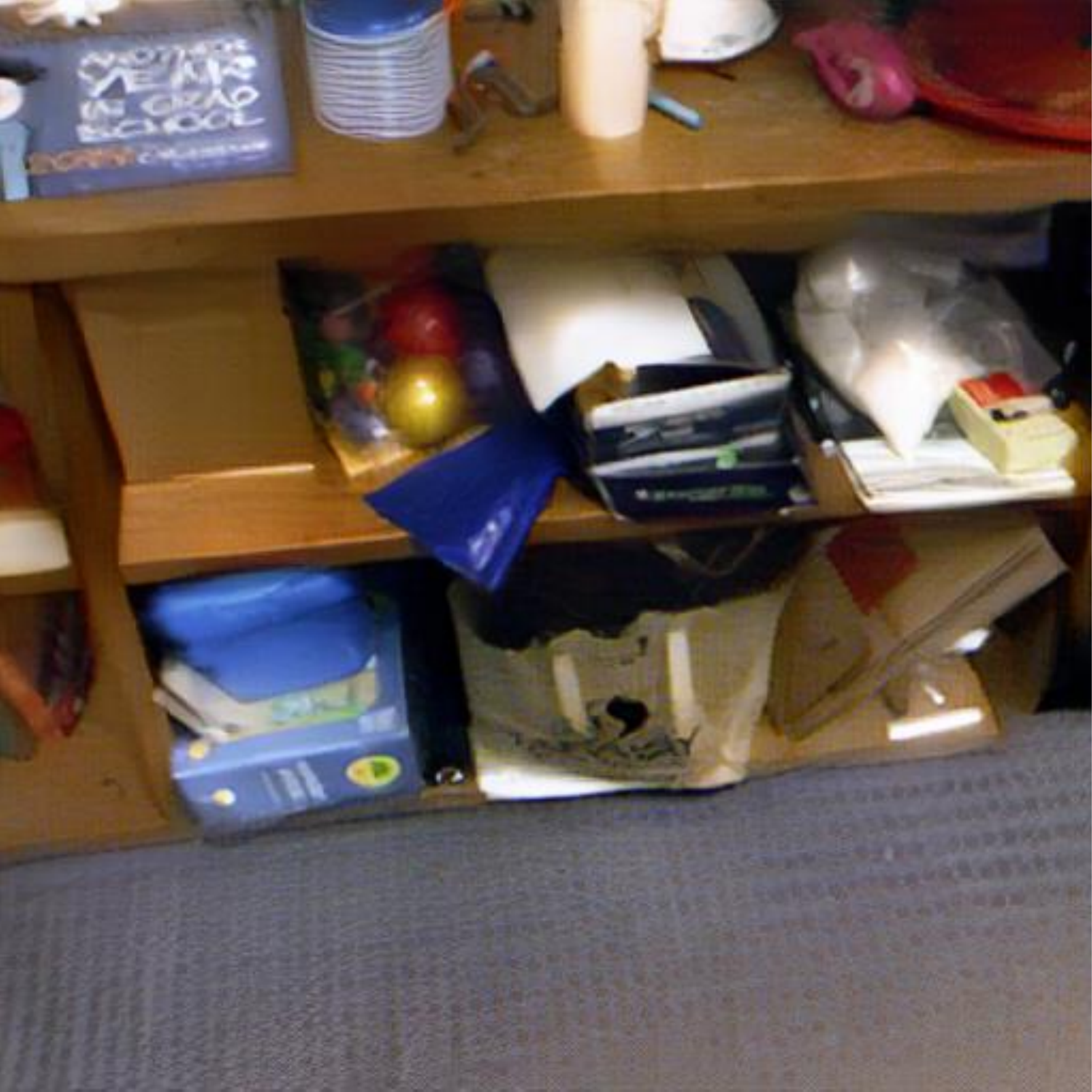}
    }
    \hspace{-3mm}
    \subfigure{
        \includegraphics[width=0.14\linewidth]{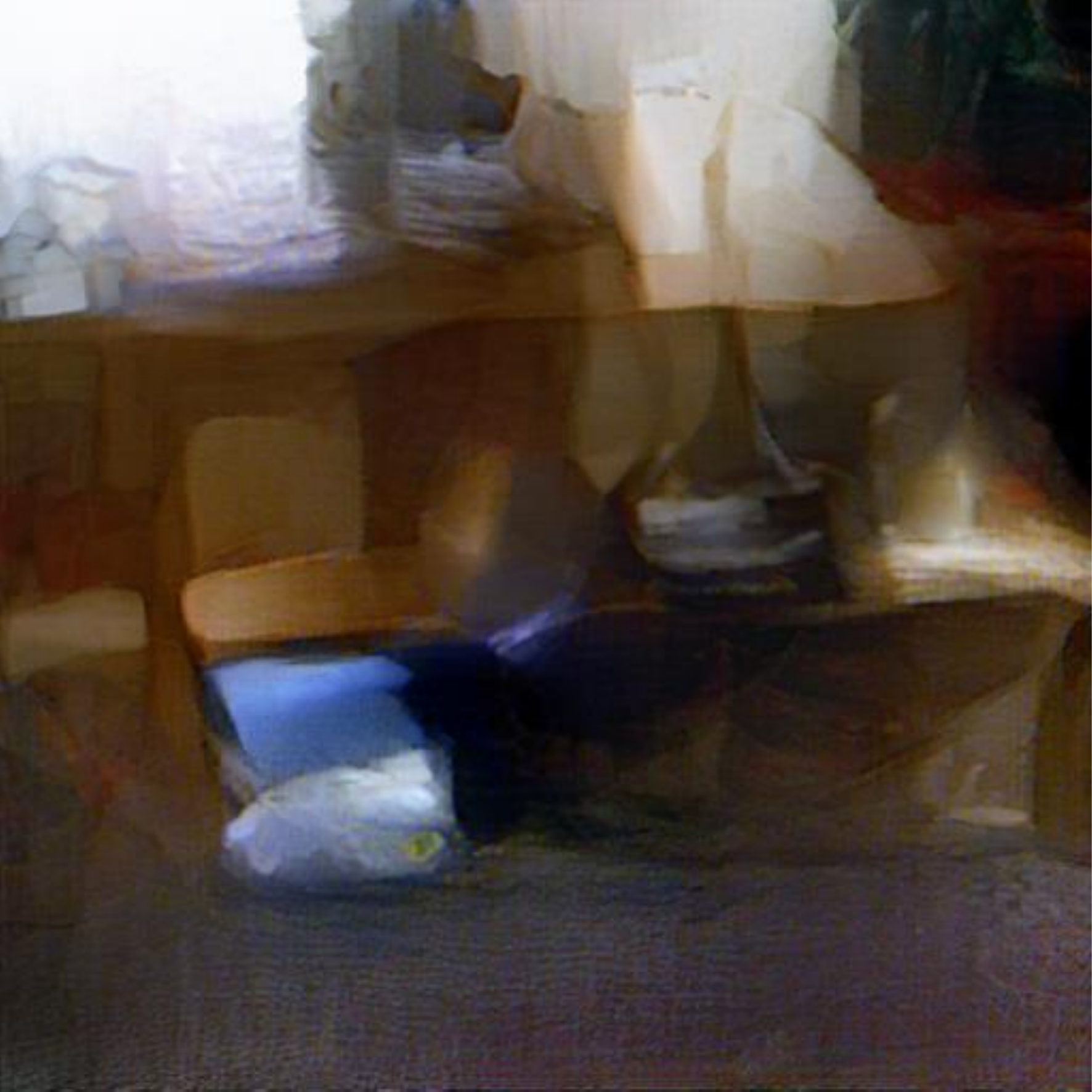}
    }
    \hspace{-3mm}
    \subfigure{
        \includegraphics[width=0.14\linewidth]{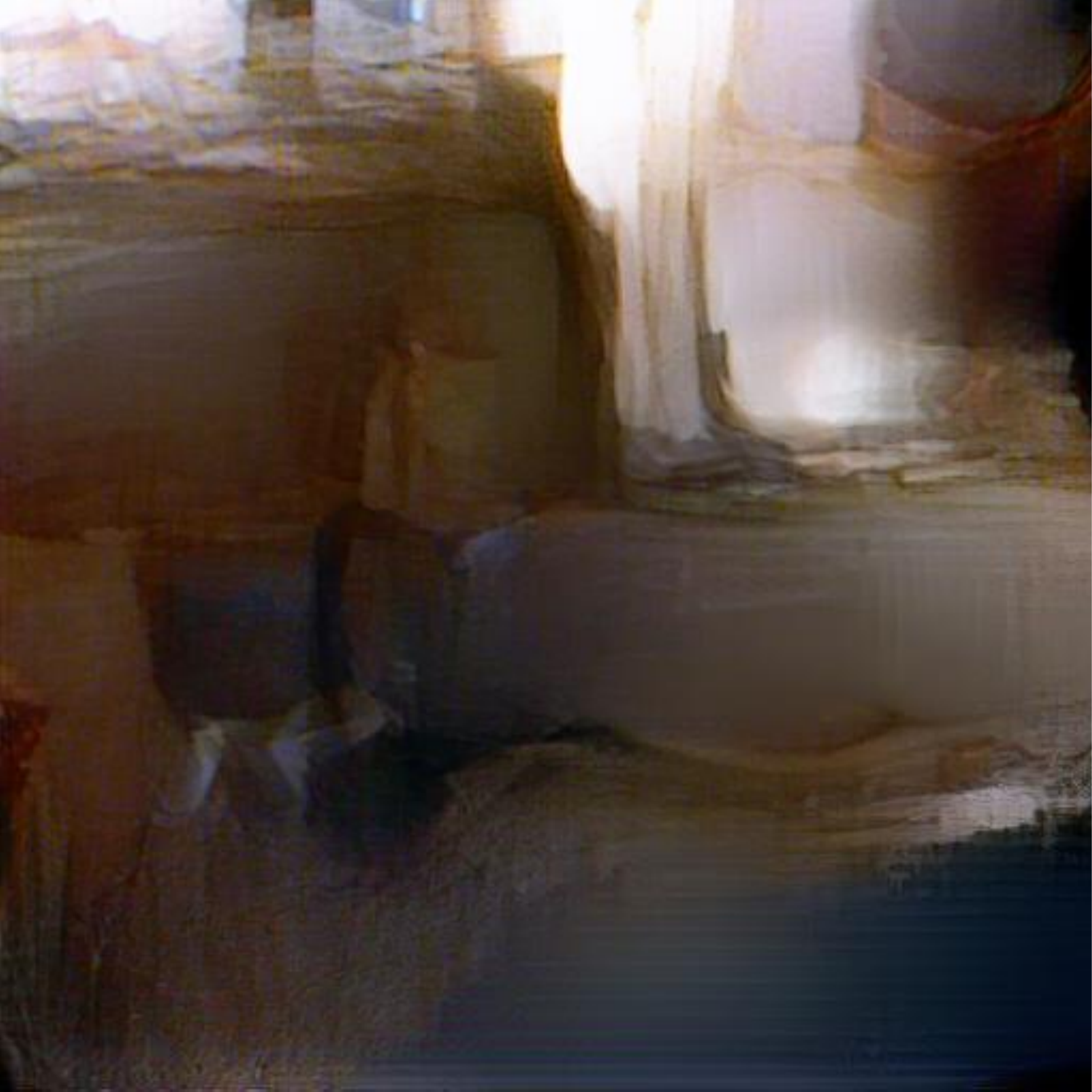}
    }
    \hspace{-3mm}
    \subfigure{
        \includegraphics[width=0.14\linewidth]{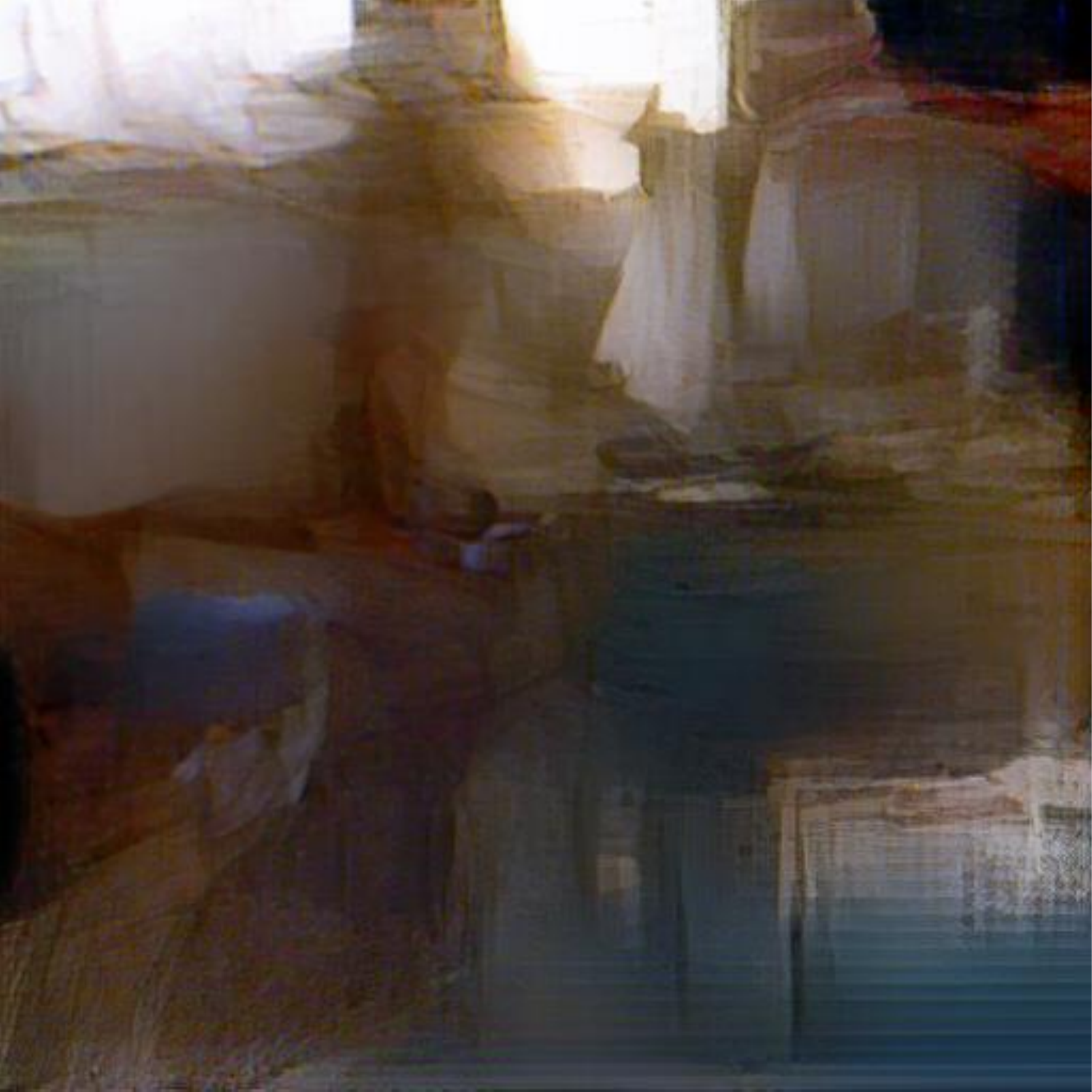}
    }
    \hspace{-3mm}
    \subfigure{
        \includegraphics[width=0.14\linewidth]{fig/office1_gates362/Sphere-frame-000034.color.pdf}
    }
    \\
    \vspace{-3mm}
    \subfigure{
        \includegraphics[width=0.14\linewidth]{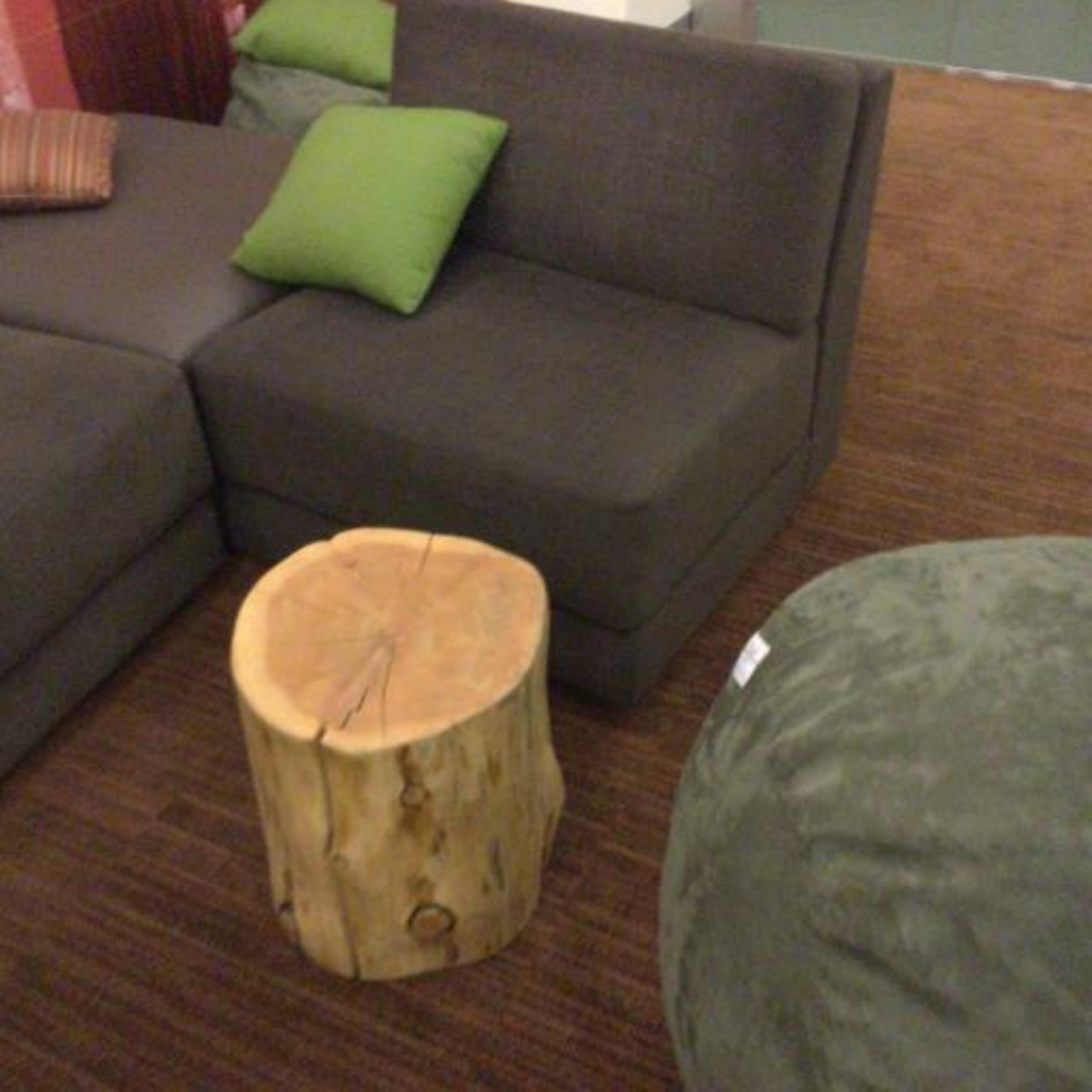}
    }
    \hspace{-3mm}
    \subfigure{
        \includegraphics[width=0.14\linewidth]{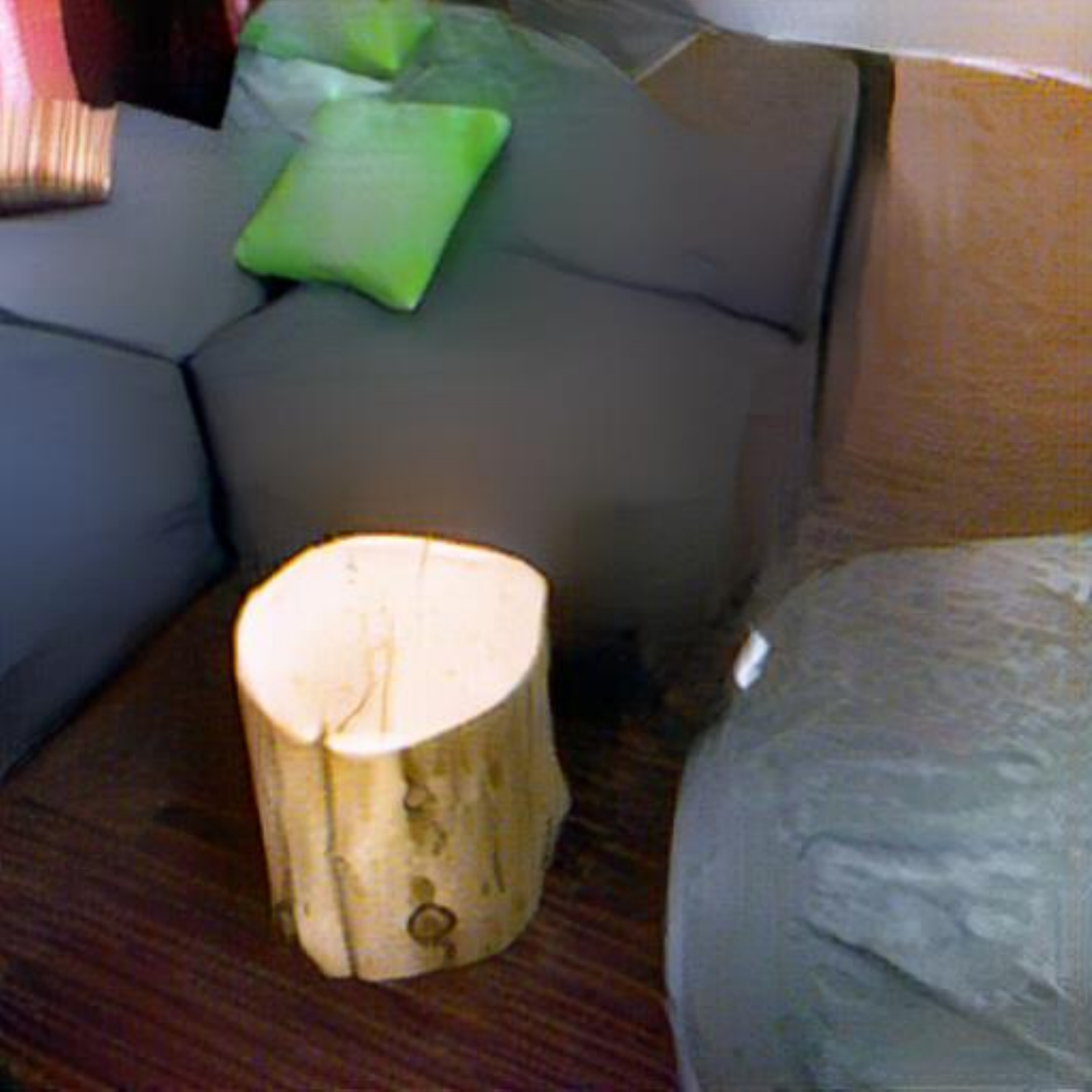}
    }
    \hspace{-3mm}
    \subfigure{
        \includegraphics[width=0.14\linewidth]{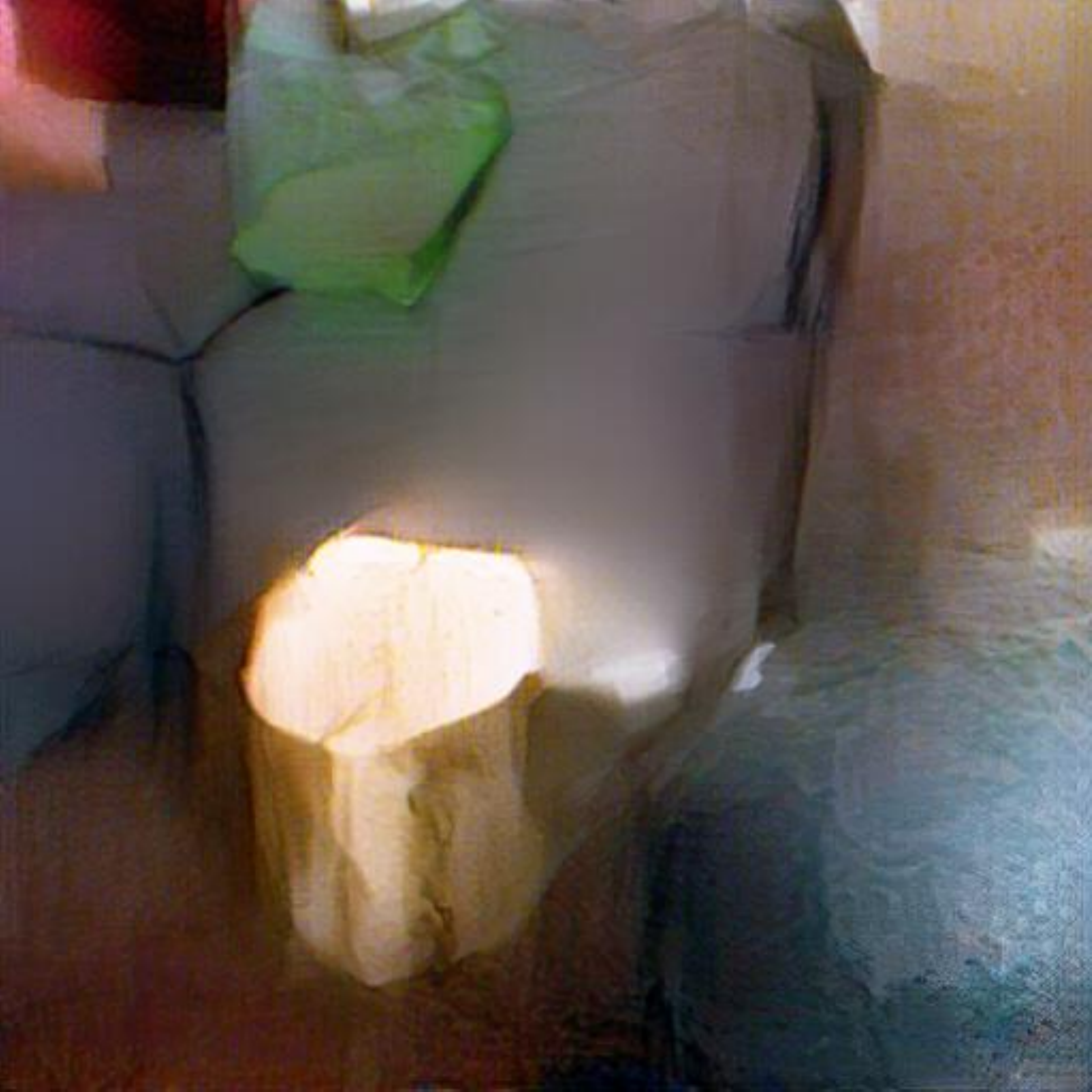}
    }
    \hspace{-3mm}
    \subfigure{
        \includegraphics[width=0.14\linewidth]{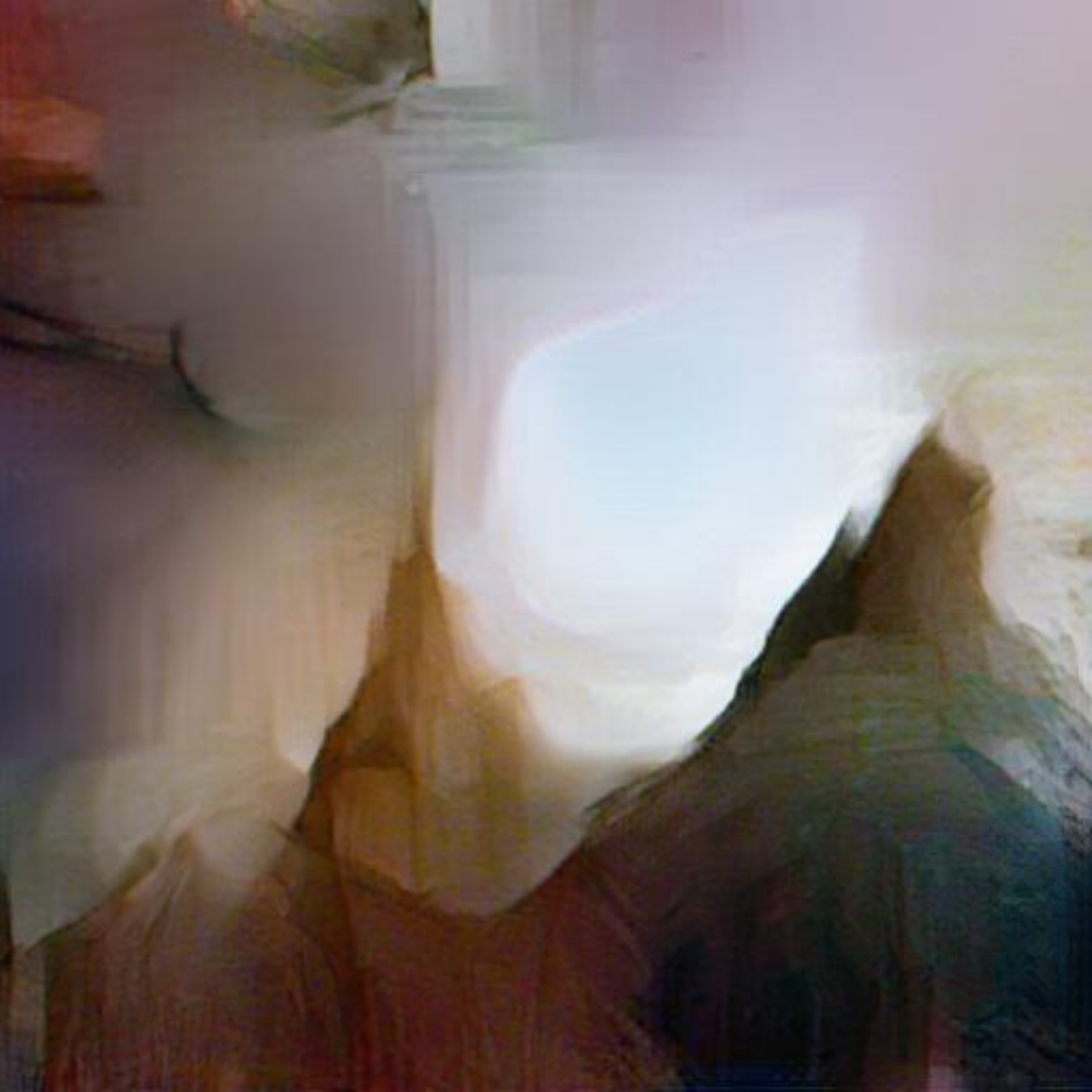}
    }
    \hspace{-3mm}
    \subfigure{
        \includegraphics[width=0.14\linewidth]{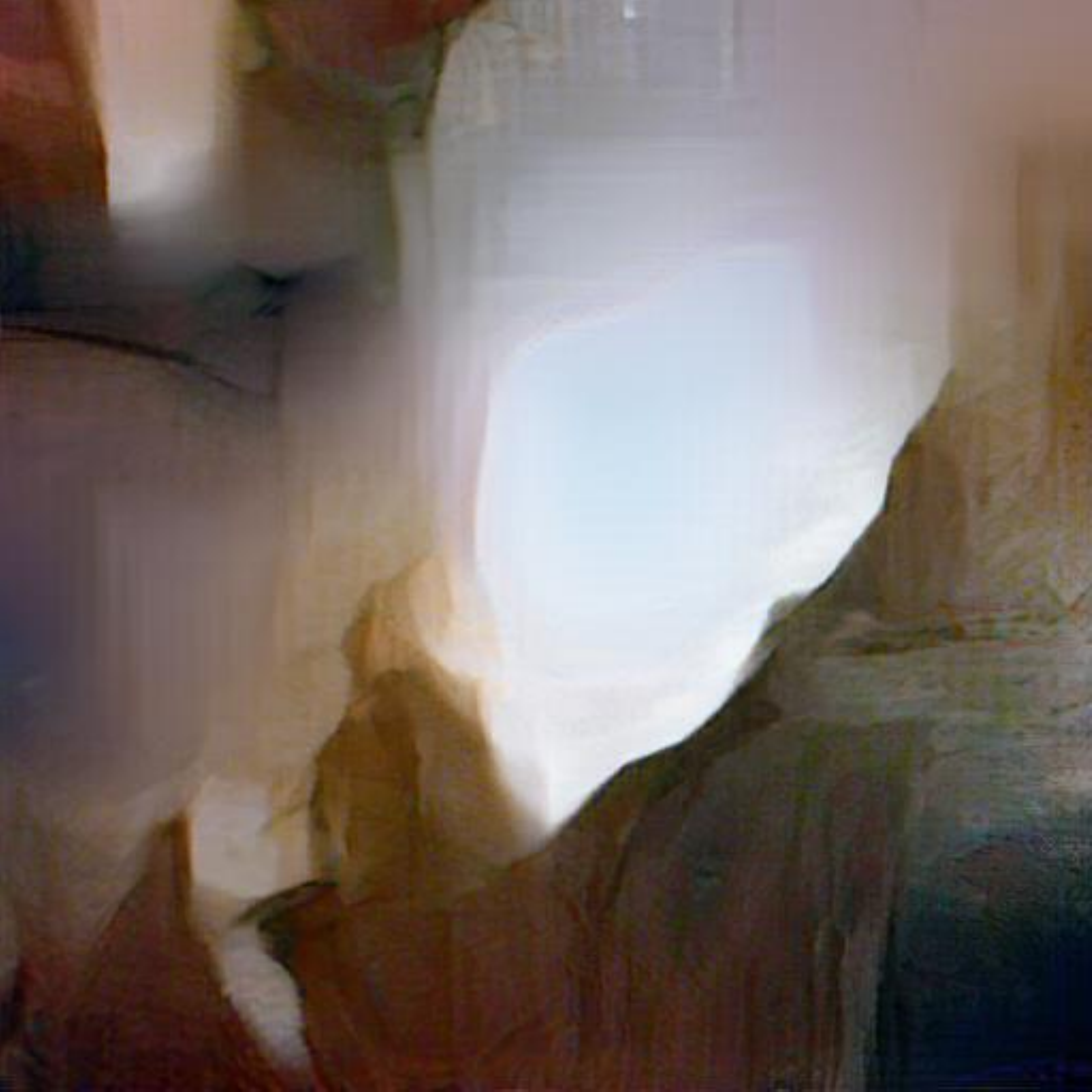}
    }
    \hspace{-3mm}
    \subfigure{
        \includegraphics[width=0.14\linewidth]{fig/heads/Sphere-seq-01_frame-000565.color.pdf}
    }
    \\
        \vspace{-3mm}
    \subfigure{
        \includegraphics[width=0.14\linewidth]{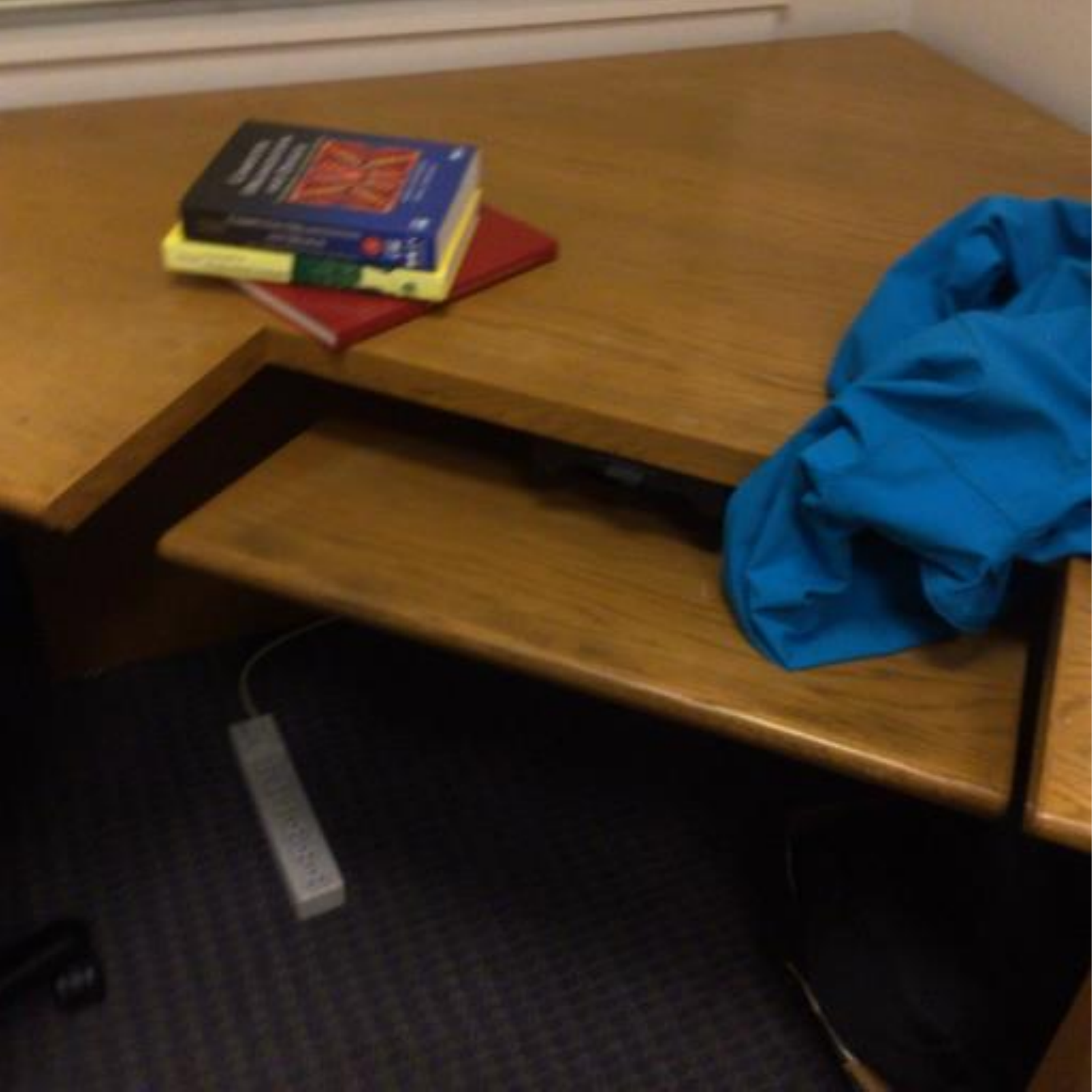}
    }
    \hspace{-3mm}
    \subfigure{
        \includegraphics[width=0.14\linewidth]{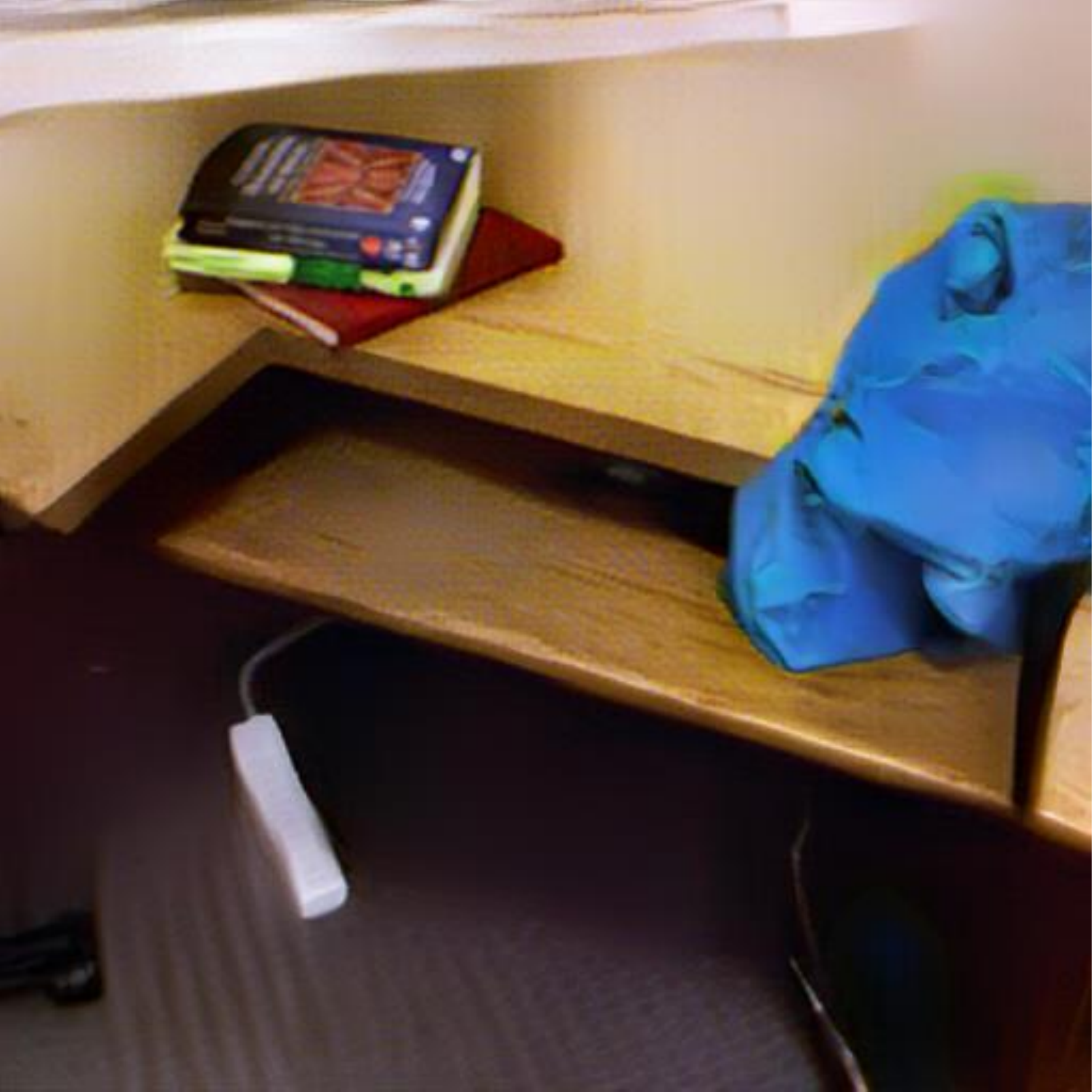}
    }
    \hspace{-3mm}
    \subfigure{
        \includegraphics[width=0.14\linewidth]{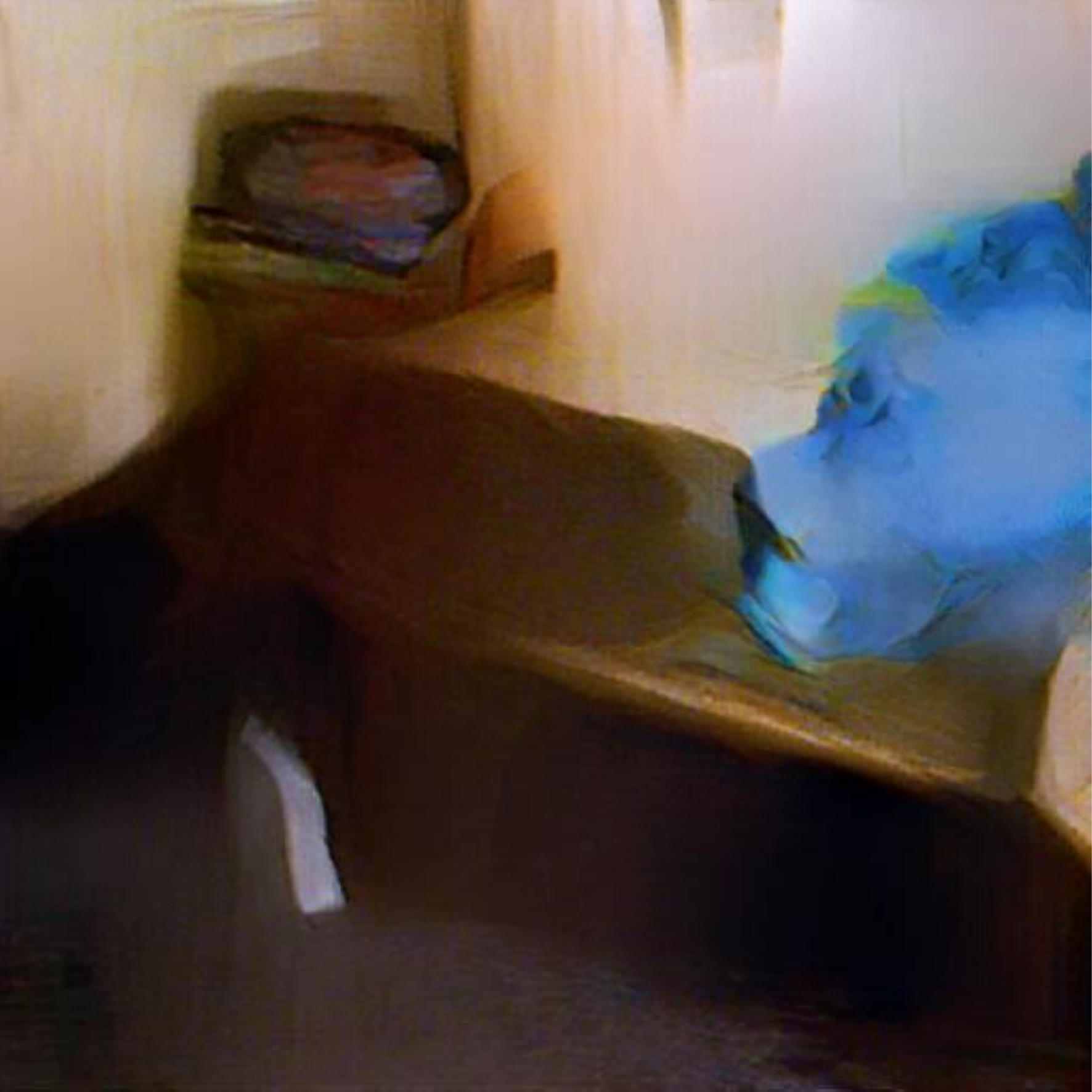}
    }
    \hspace{-3mm}
    \subfigure{
        \includegraphics[width=0.14\linewidth]{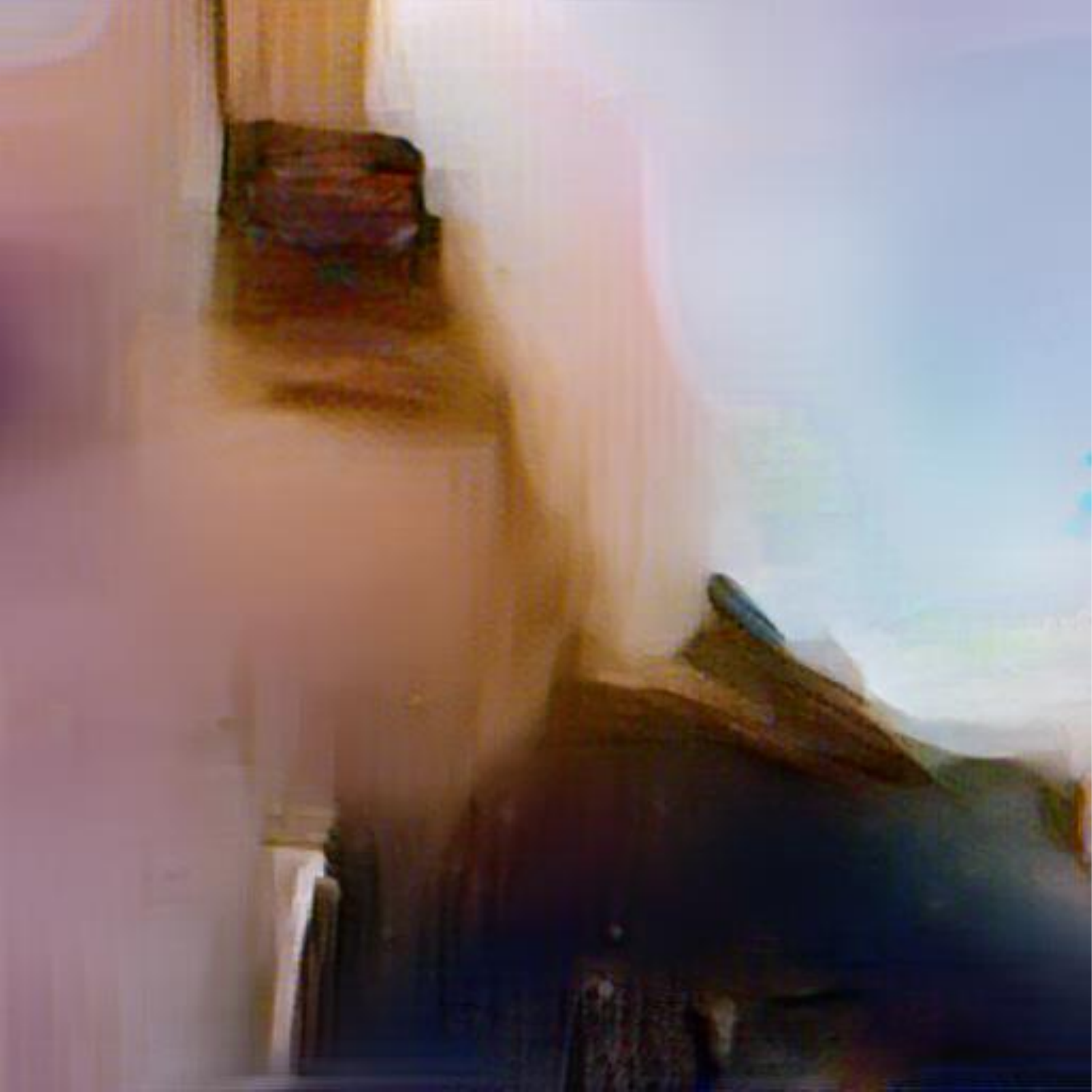}
    }
    \hspace{-3mm}
    \subfigure{
        \includegraphics[width=0.14\linewidth]{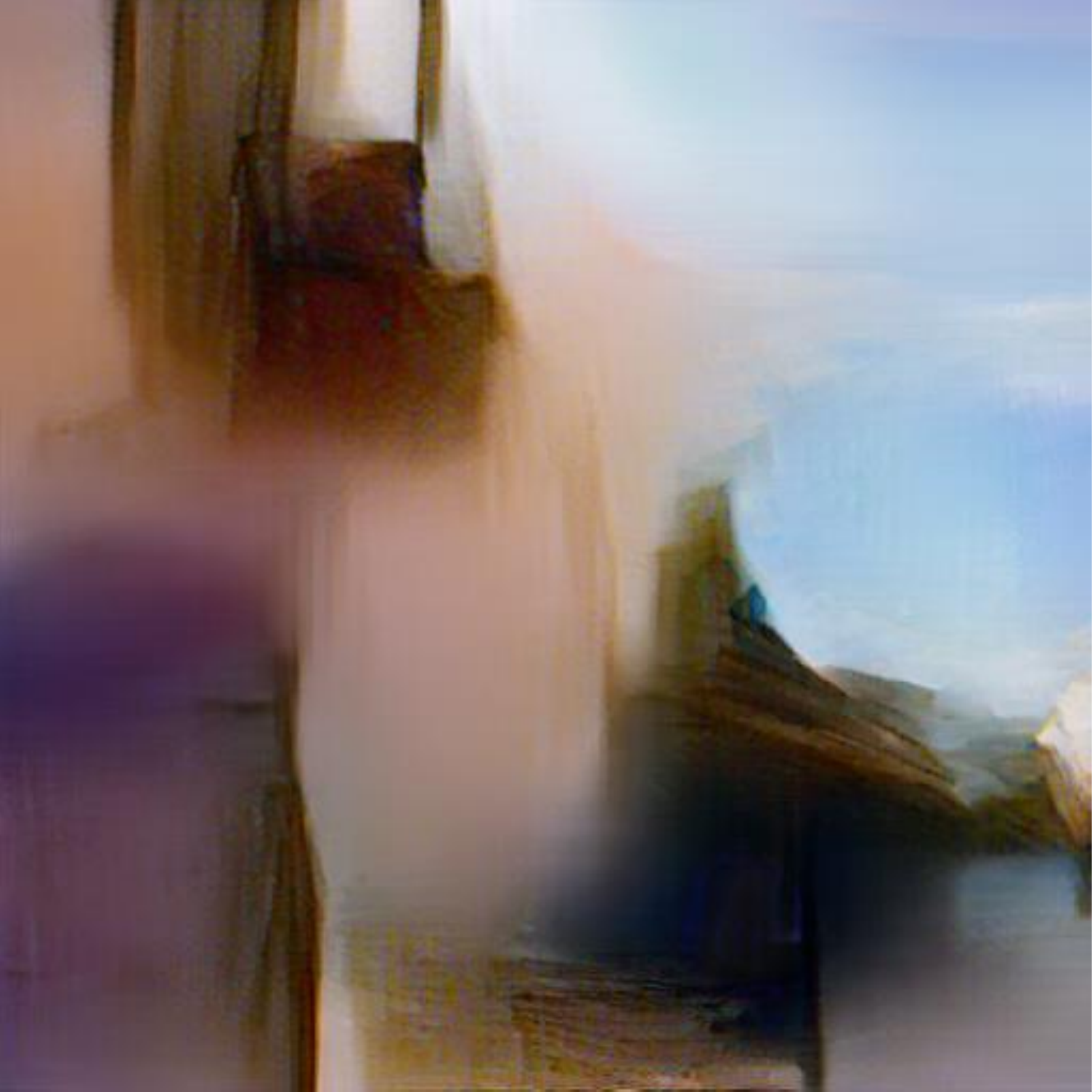}
    }
    \hspace{-3mm}
    \subfigure{
        \includegraphics[width=0.14\linewidth]{fig/heads/Sphere-seq-01_frame-000565.color.pdf}
    }
    \\
        \vspace{-3mm}
    \subfigure{
        \includegraphics[width=0.14\linewidth]{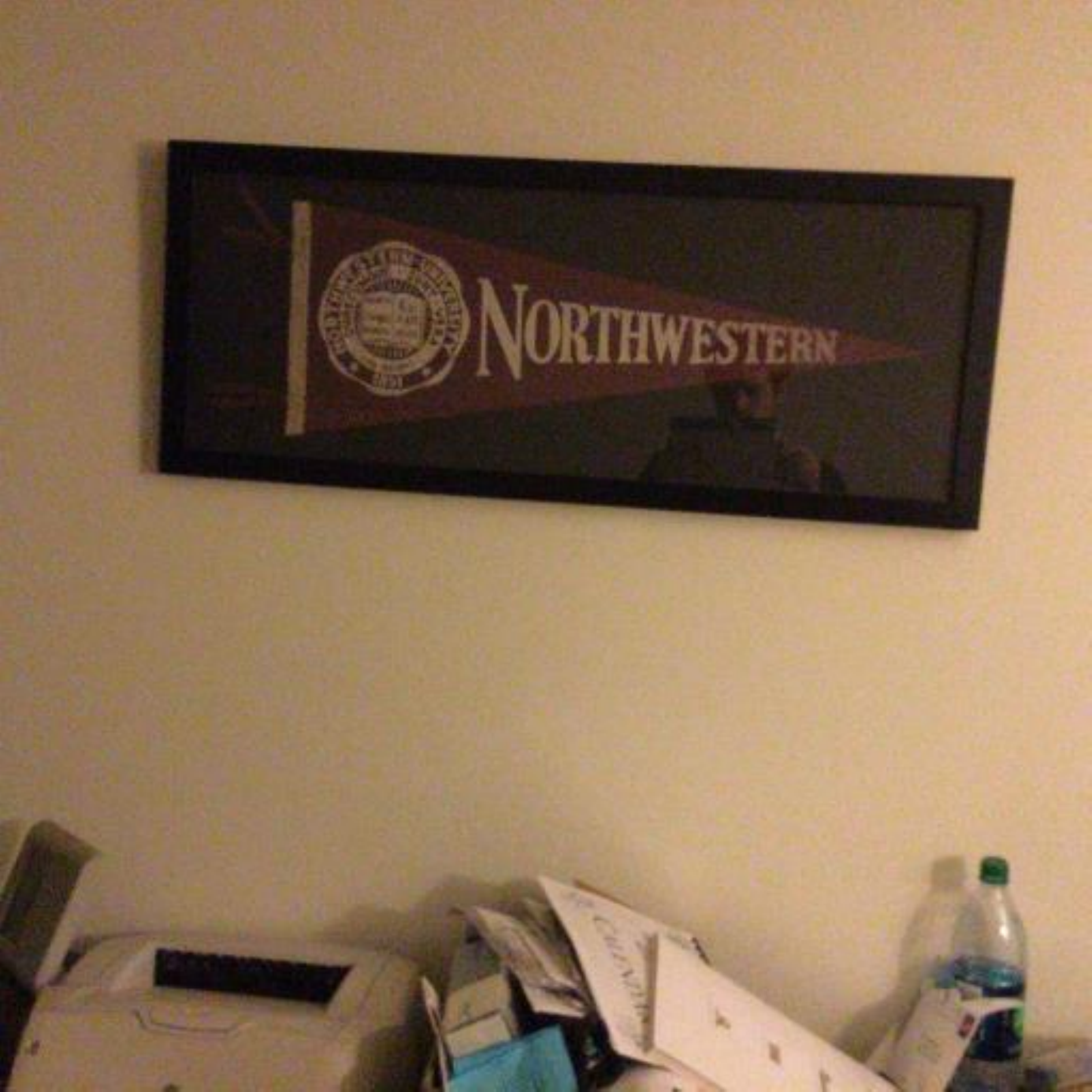}
    }
    \hspace{-3mm}
    \subfigure{
        \includegraphics[width=0.14\linewidth]{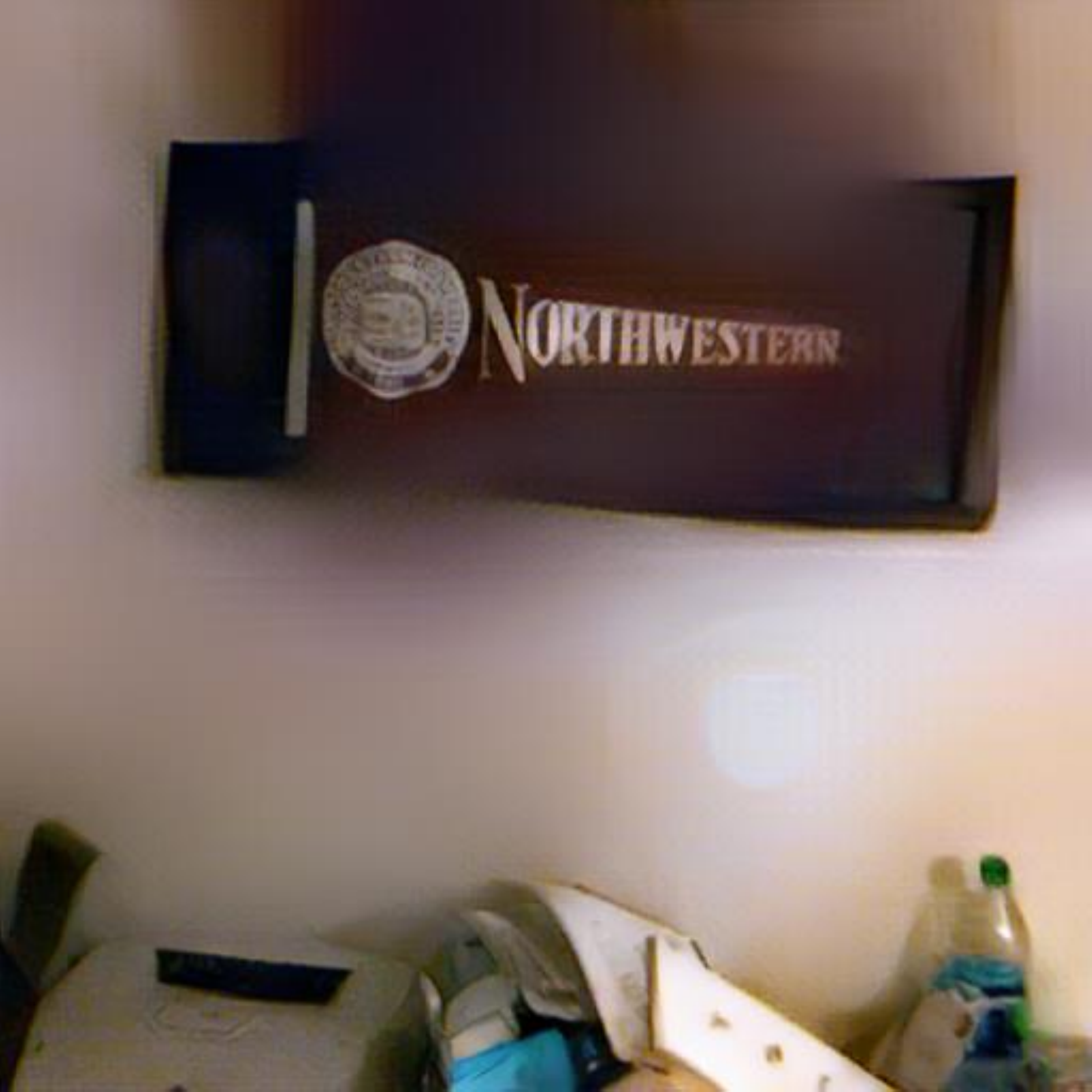}
    }
    \hspace{-3mm}
    \subfigure{
        \includegraphics[width=0.14\linewidth]{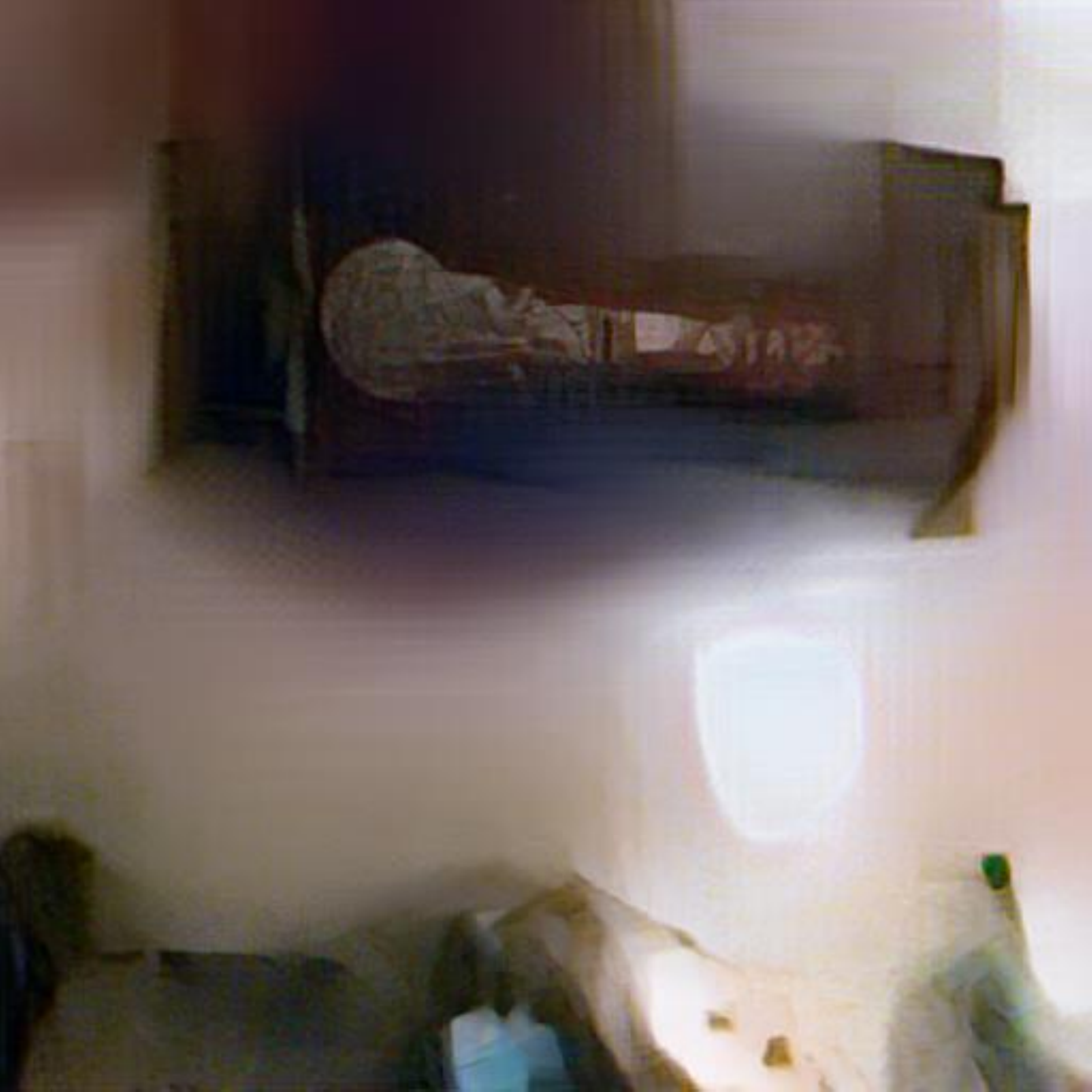}
    }
    \hspace{-3mm}
    \subfigure{
        \includegraphics[width=0.14\linewidth]{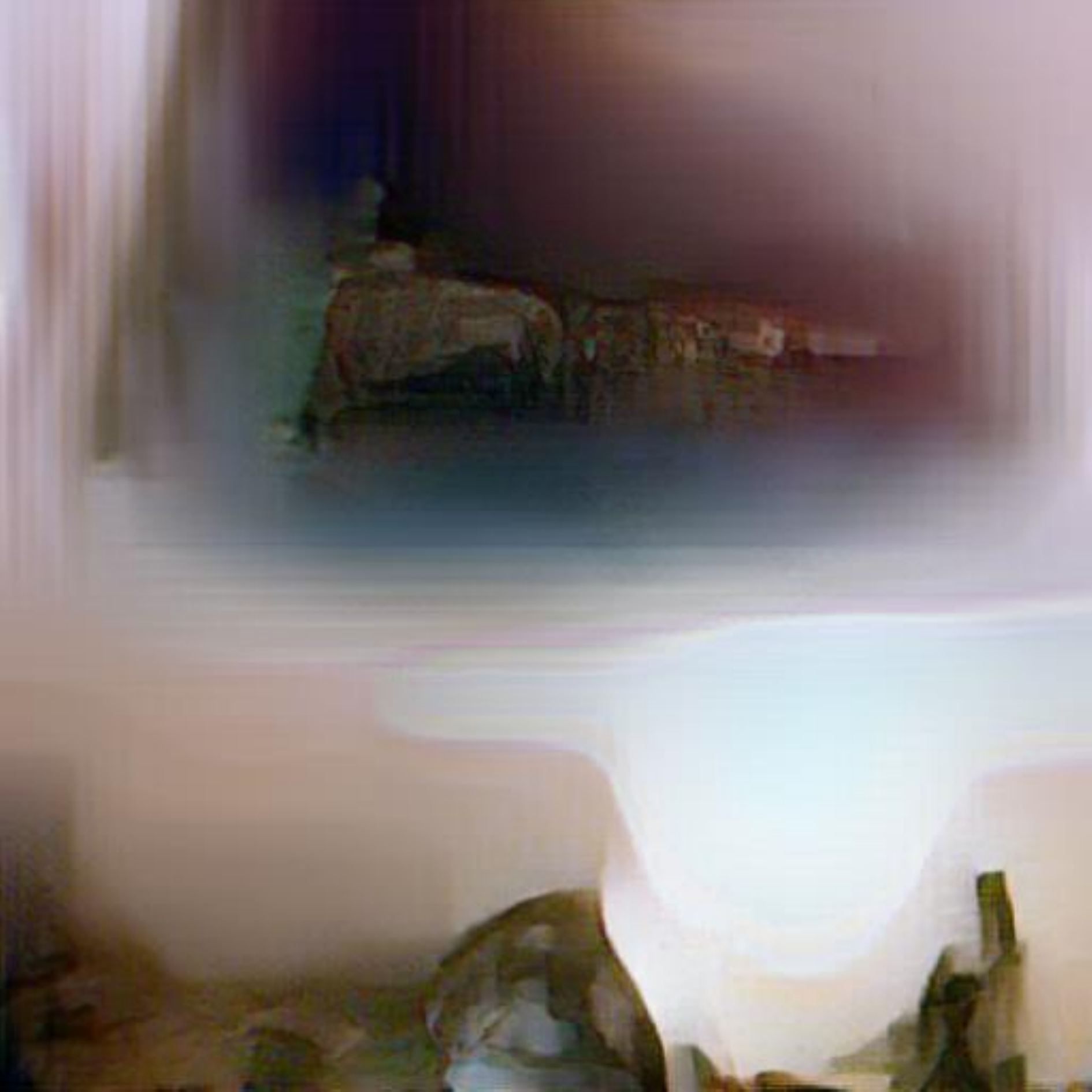}
    }
    \hspace{-3mm}
    \subfigure{
        \includegraphics[width=0.14\linewidth]{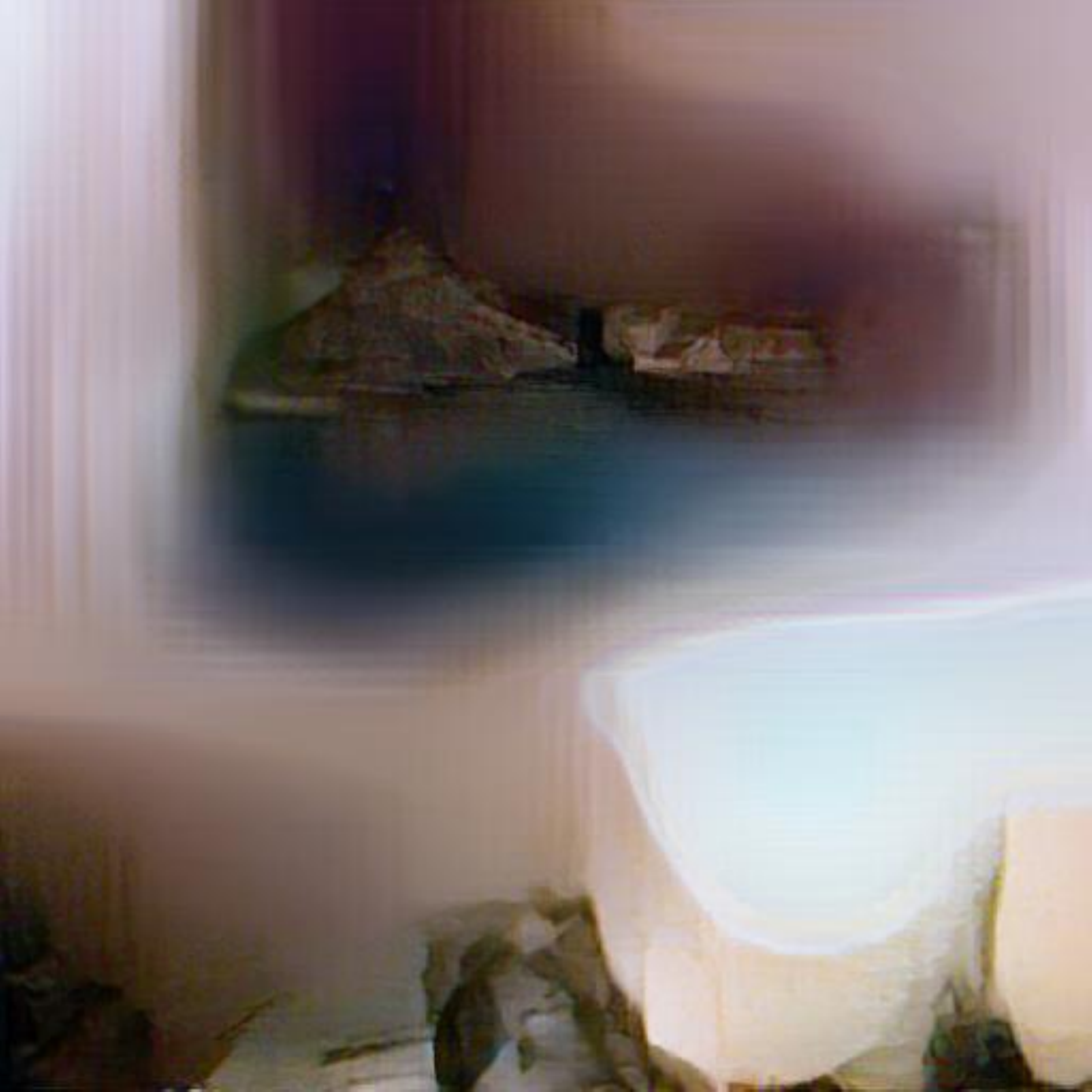}
    }
    \hspace{-3mm}
    \subfigure{
        \includegraphics[width=0.14\linewidth]{fig/heads/Sphere-seq-01_frame-000565.color.pdf}
    }
    \\

    \subfigure{
        \includegraphics[width=0.14\linewidth]{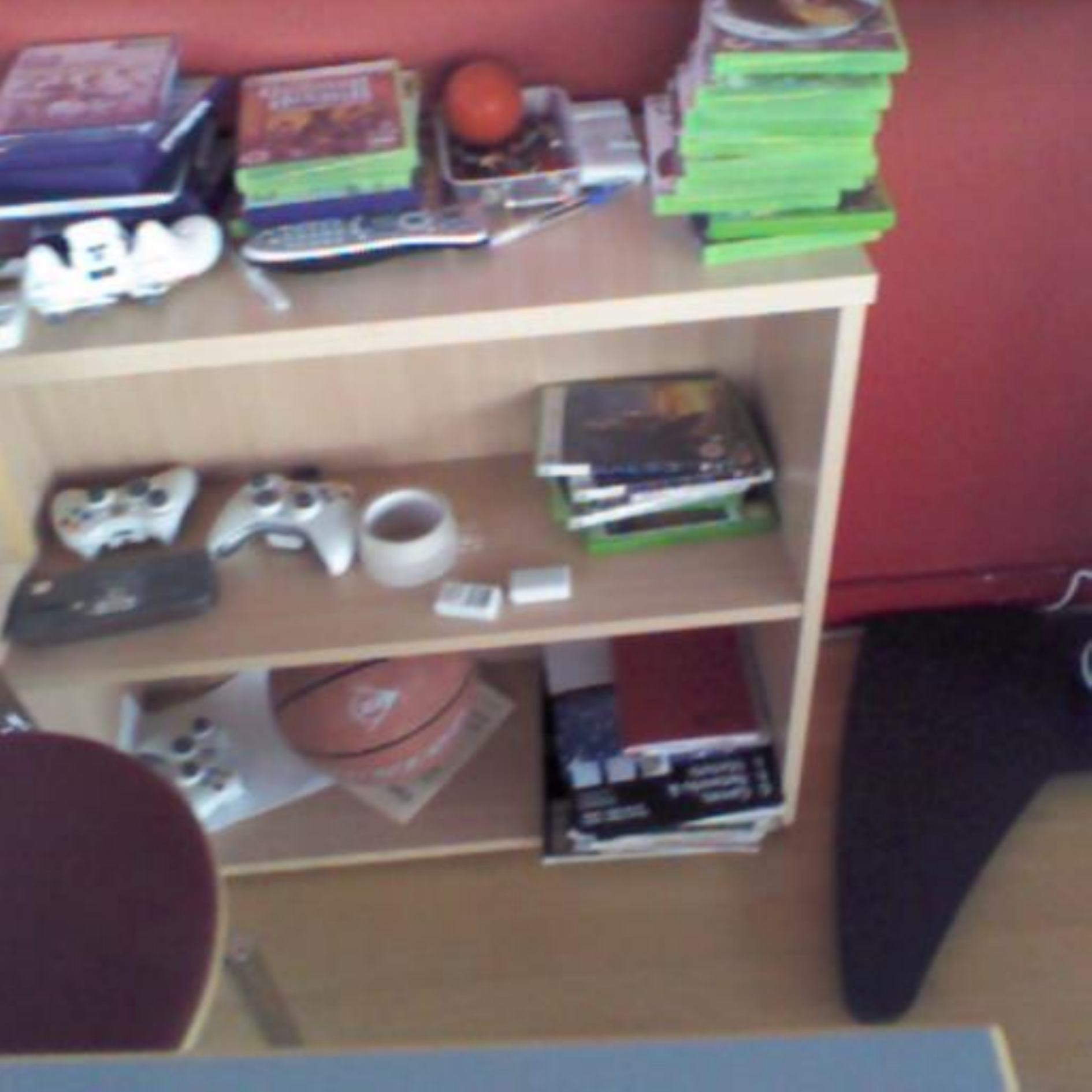}
    }
    \hspace{-3mm}
    \subfigure{
        \includegraphics[width=0.14\linewidth]{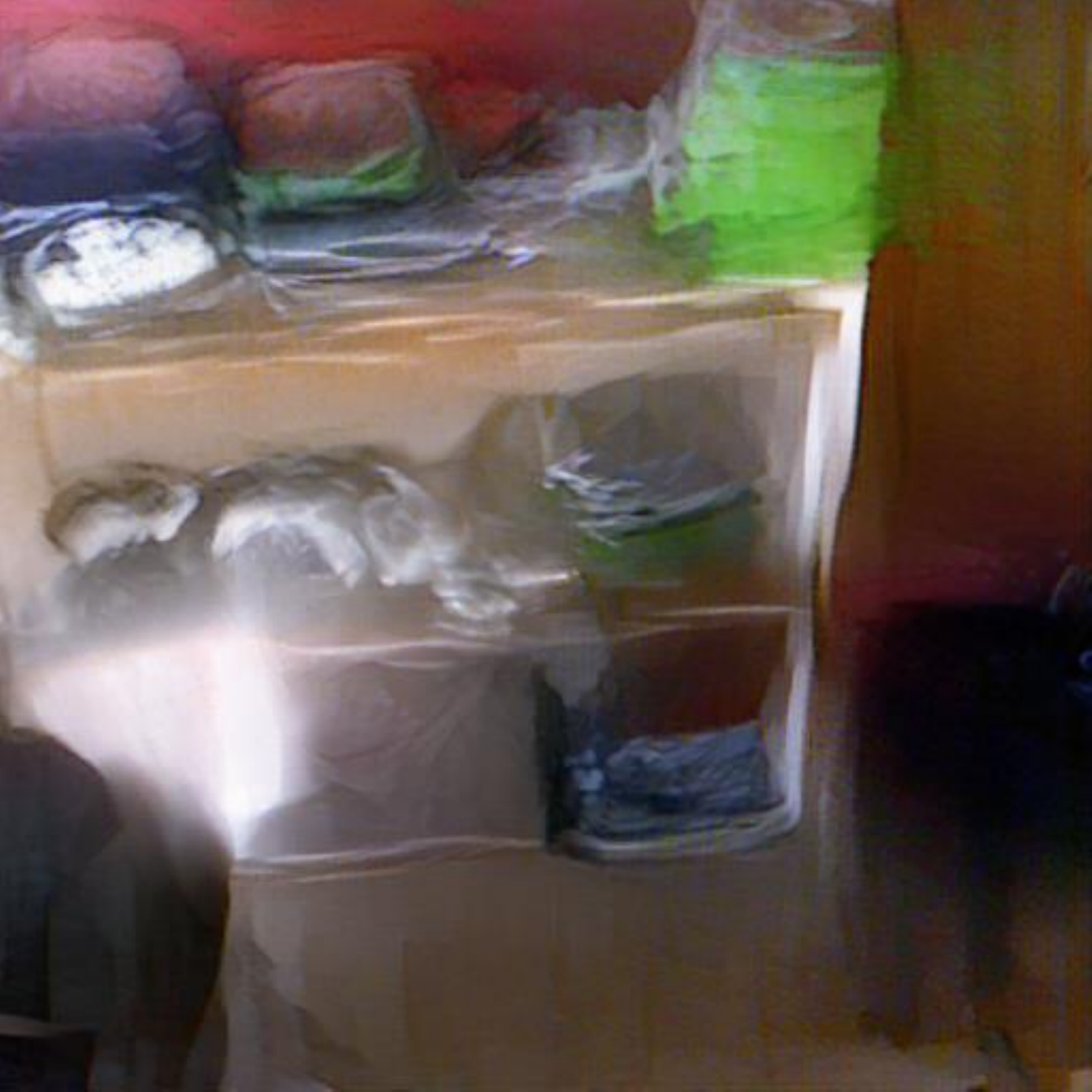}
    }
    \hspace{-3mm}
    \subfigure{
        \includegraphics[width=0.14\linewidth]{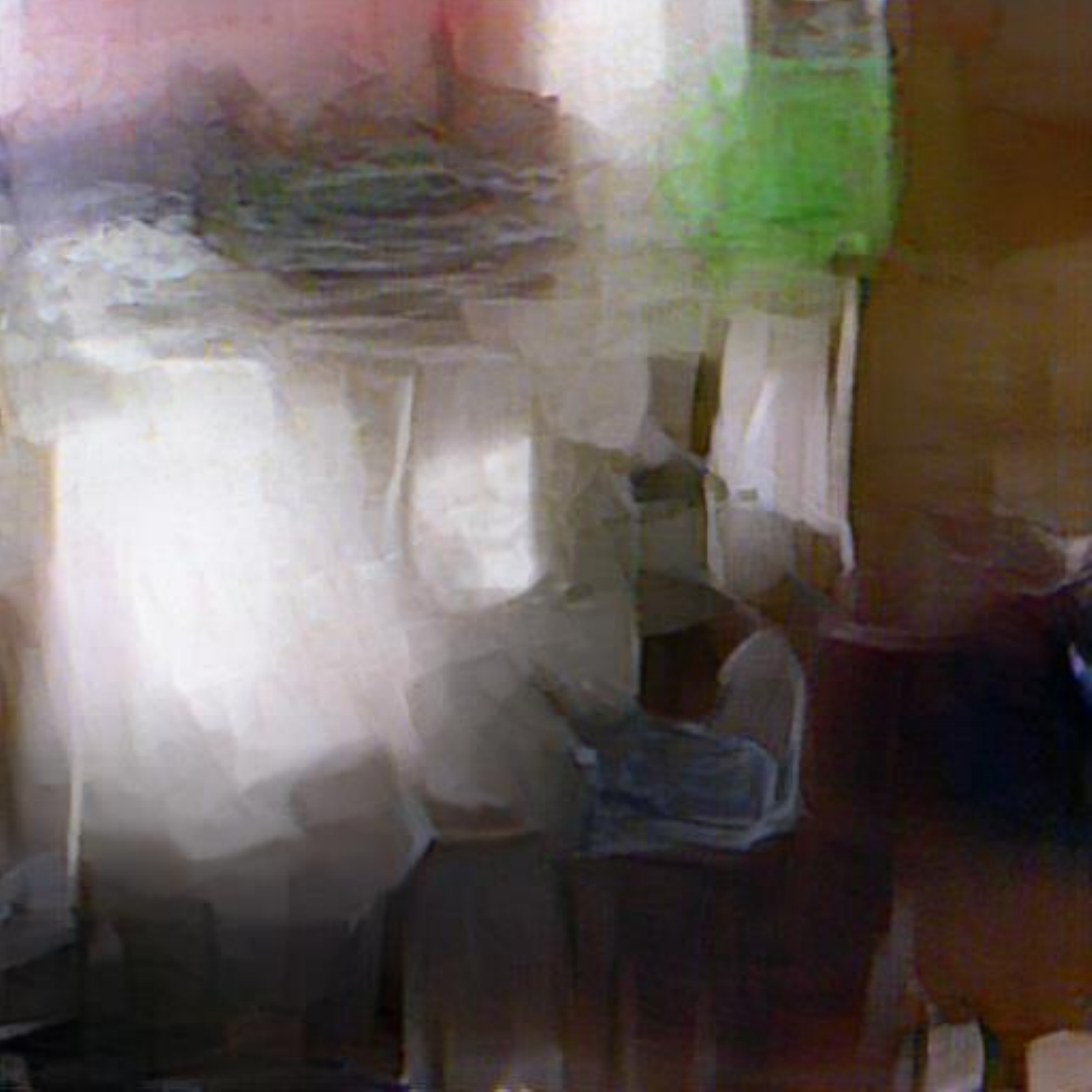}
    }
    \hspace{-3mm}
    \subfigure{
        \includegraphics[width=0.14\linewidth]{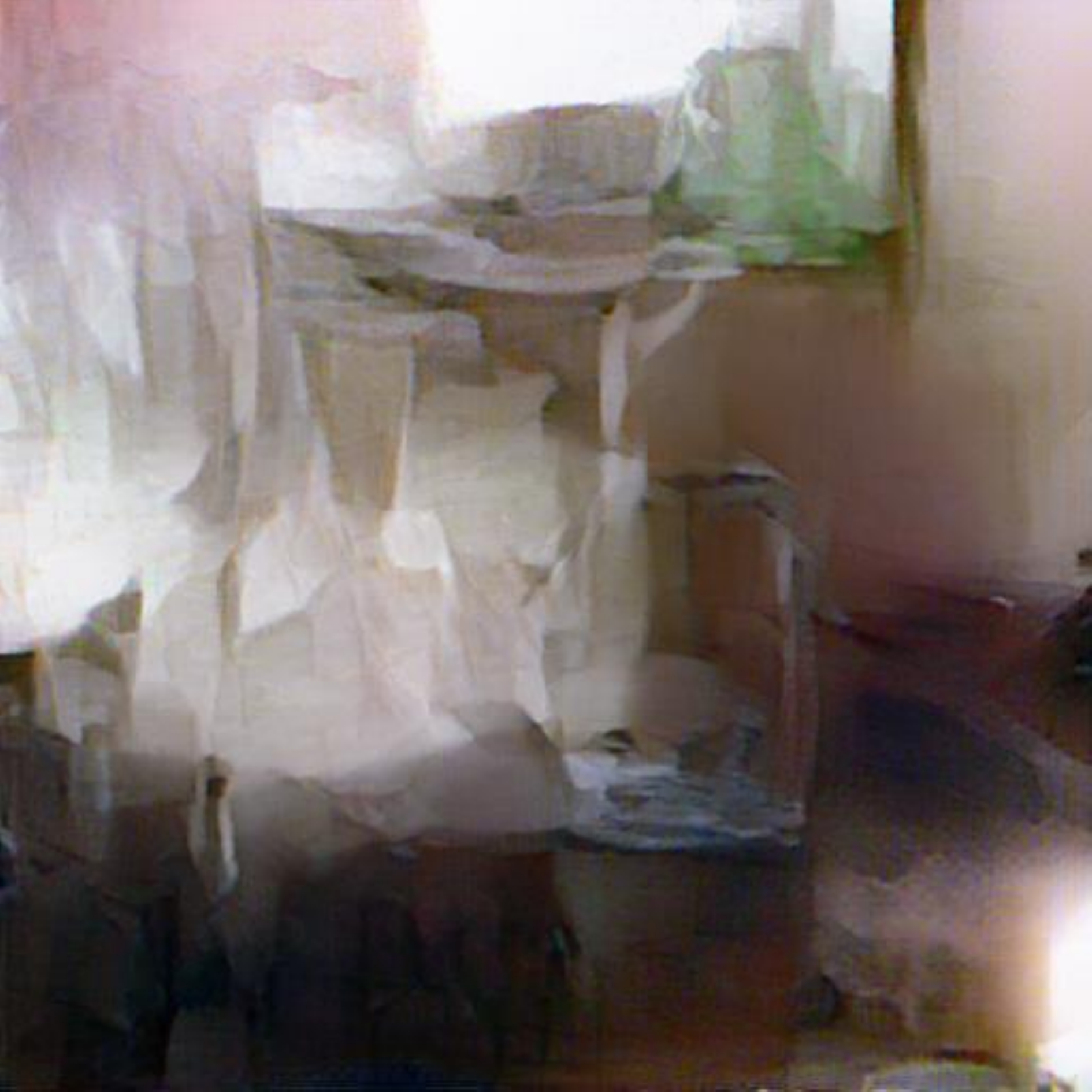}
    }
    \hspace{-3mm}
    \subfigure{
        \includegraphics[width=0.14\linewidth]{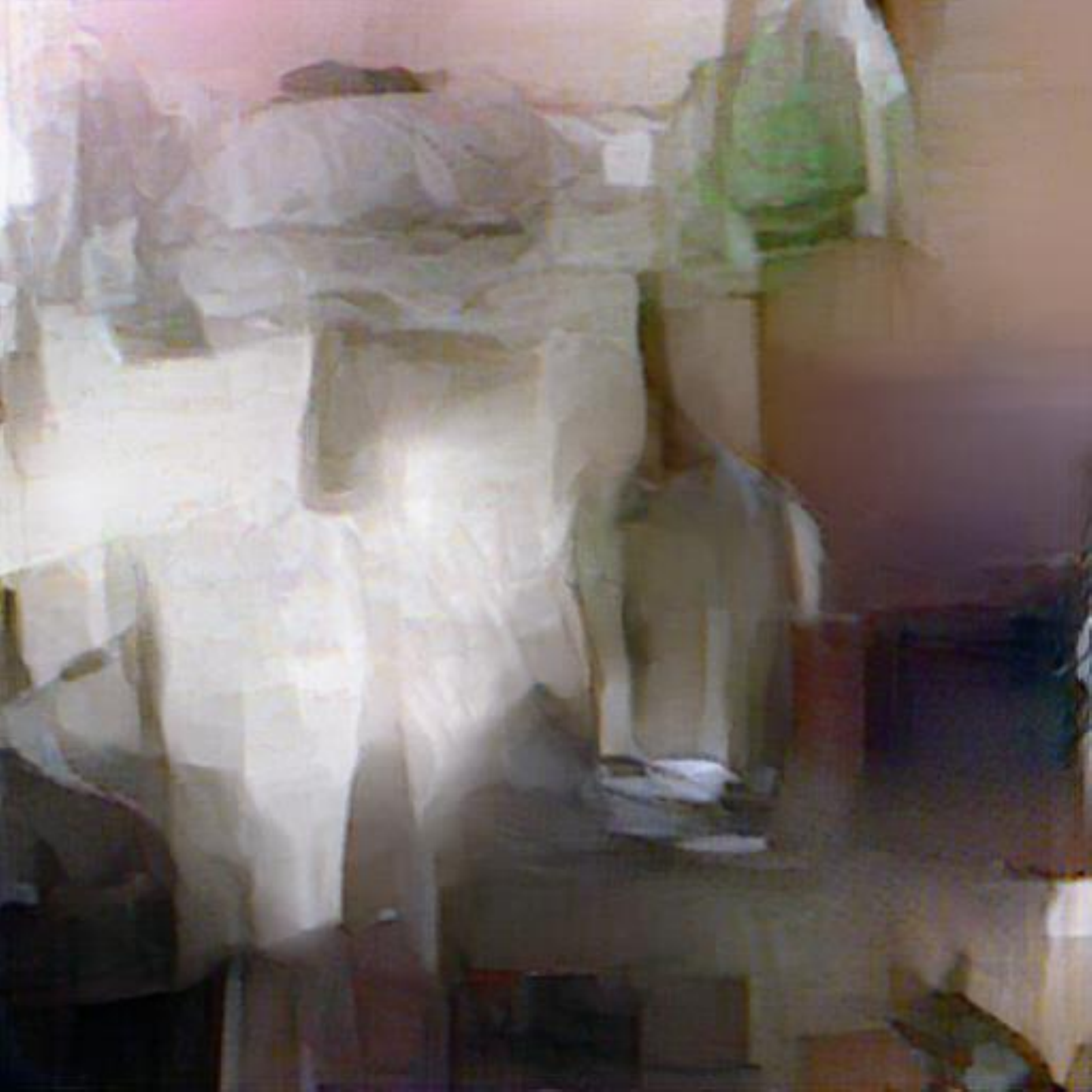}
    }
    \hspace{-3mm}
    \subfigure{
        \includegraphics[width=0.14\linewidth]{fig/heads/Sphere-seq-01_frame-000565.color.pdf}
    }
    \\
        \vspace{-3mm}
    \subfigure{
        \includegraphics[width=0.14\linewidth]{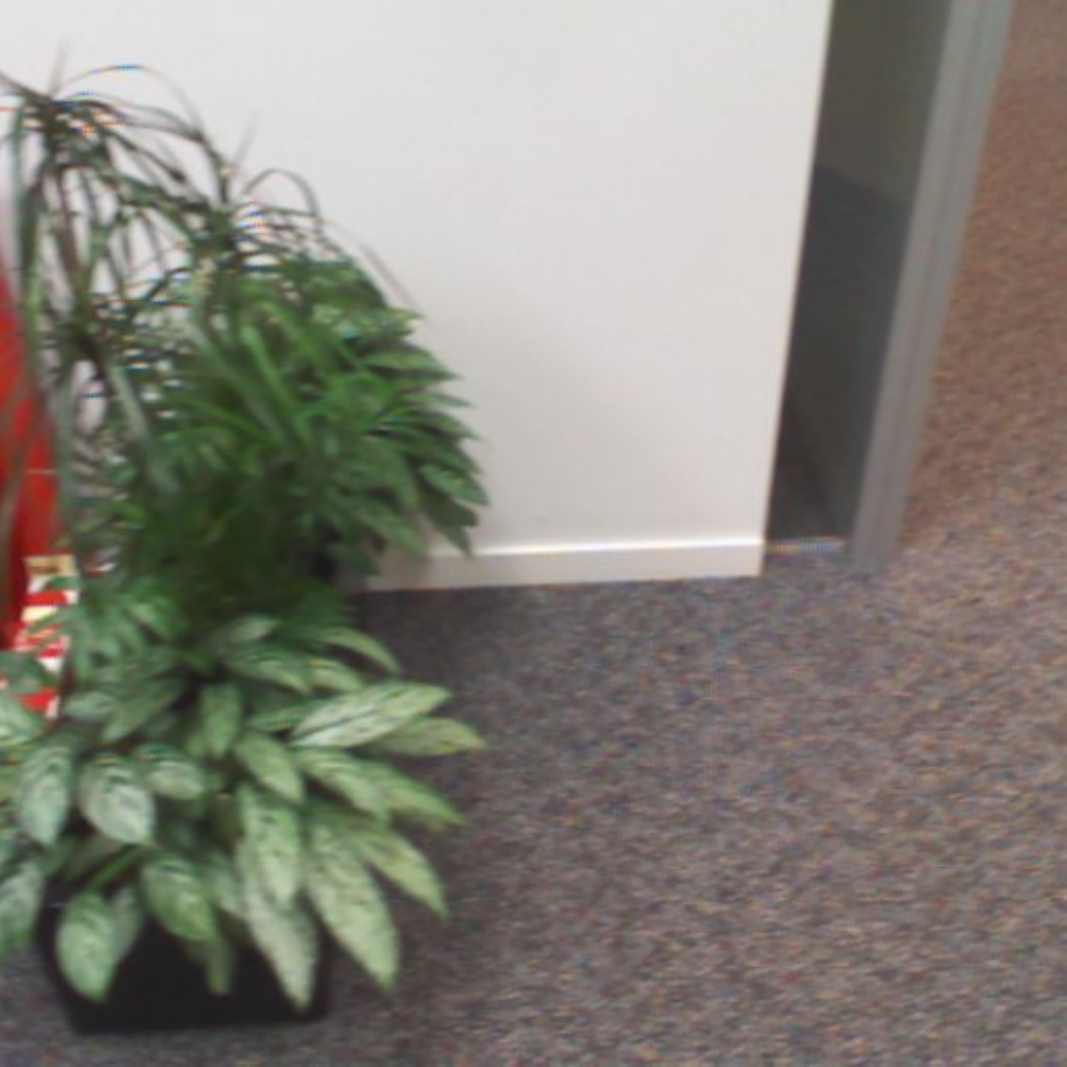}
    }
    \hspace{-3mm}
    \subfigure{
        \includegraphics[width=0.14\linewidth]{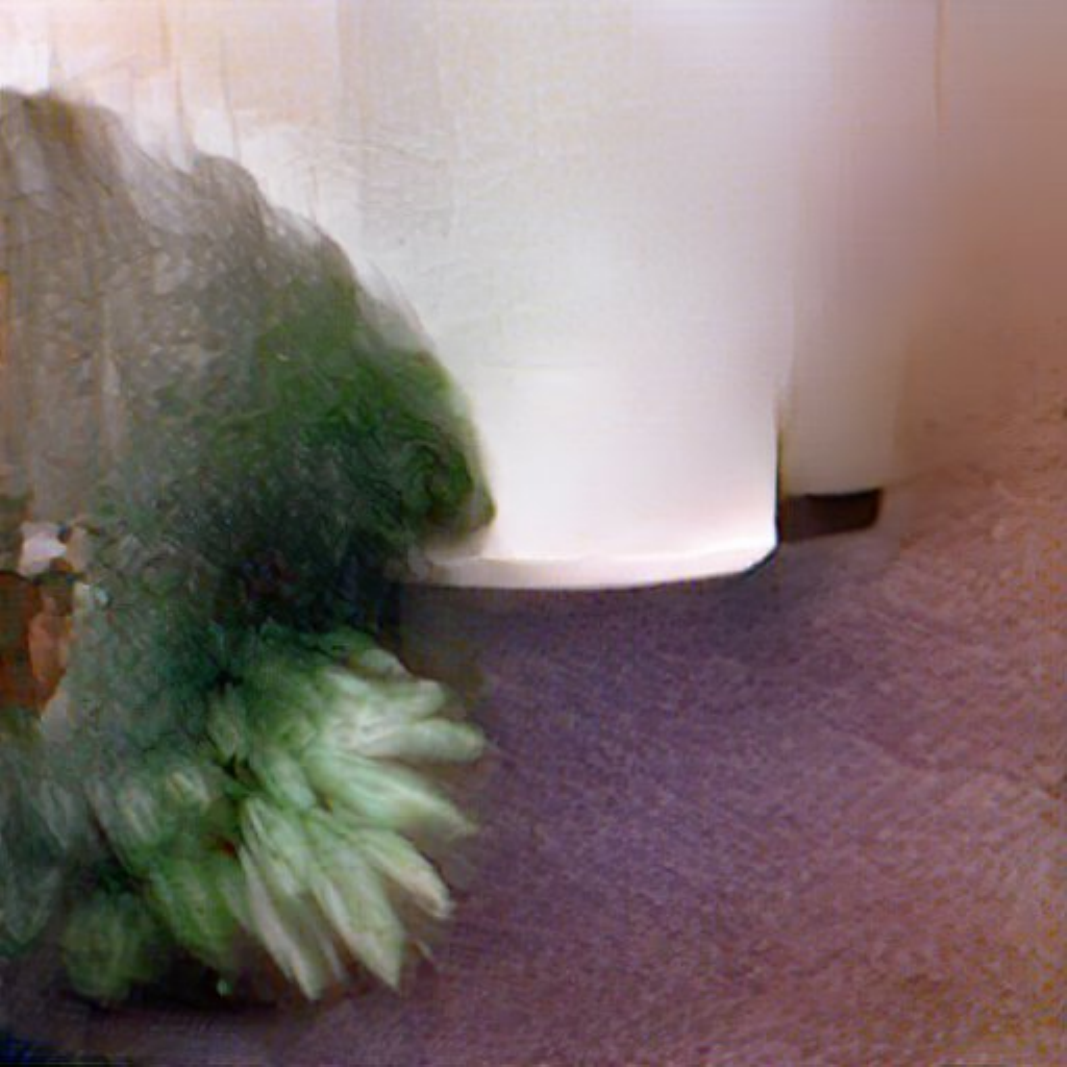}
    }
    \hspace{-3mm}
    \subfigure{
        \includegraphics[width=0.14\linewidth]{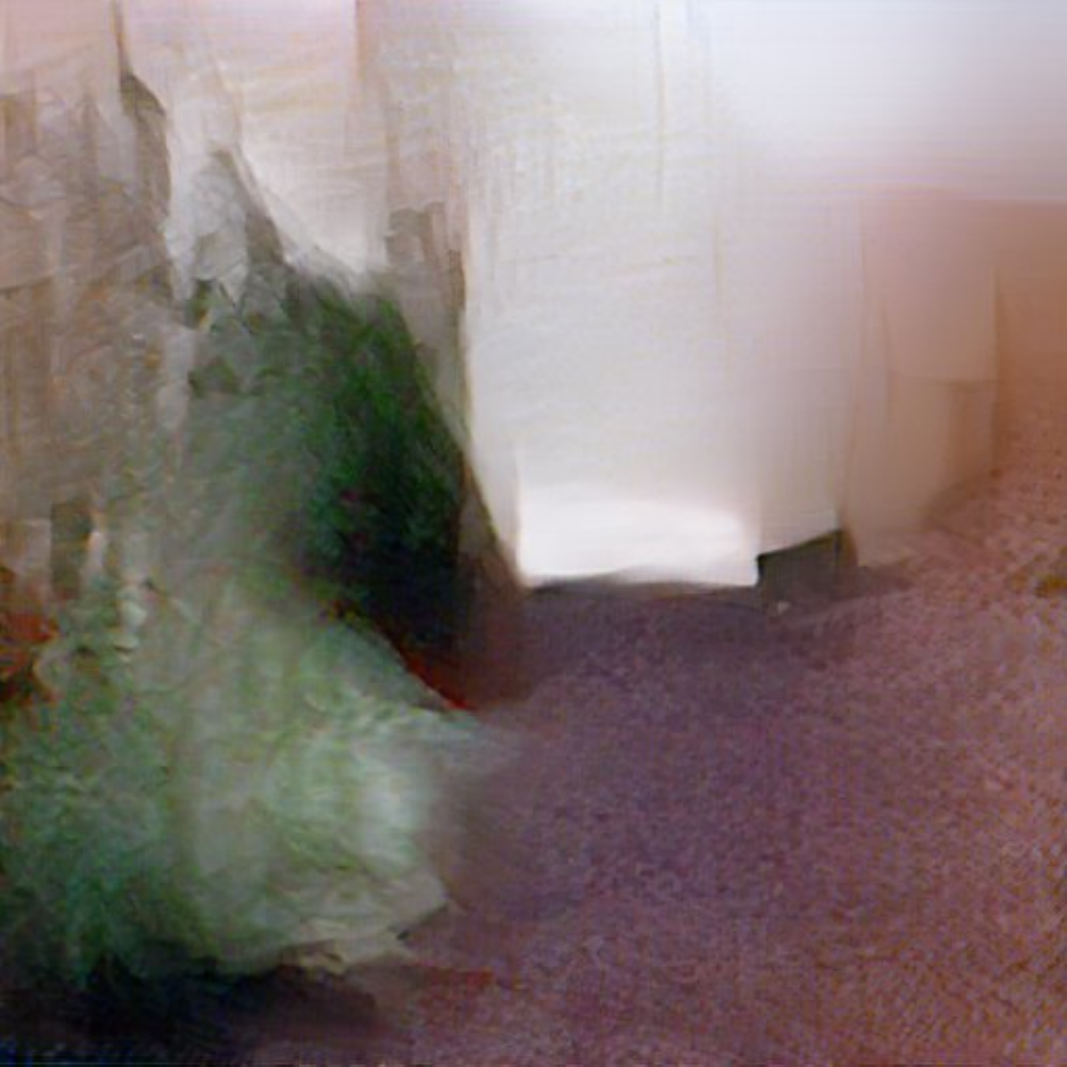}
    }
    \hspace{-3mm}
    \subfigure{
        \includegraphics[width=0.14\linewidth]{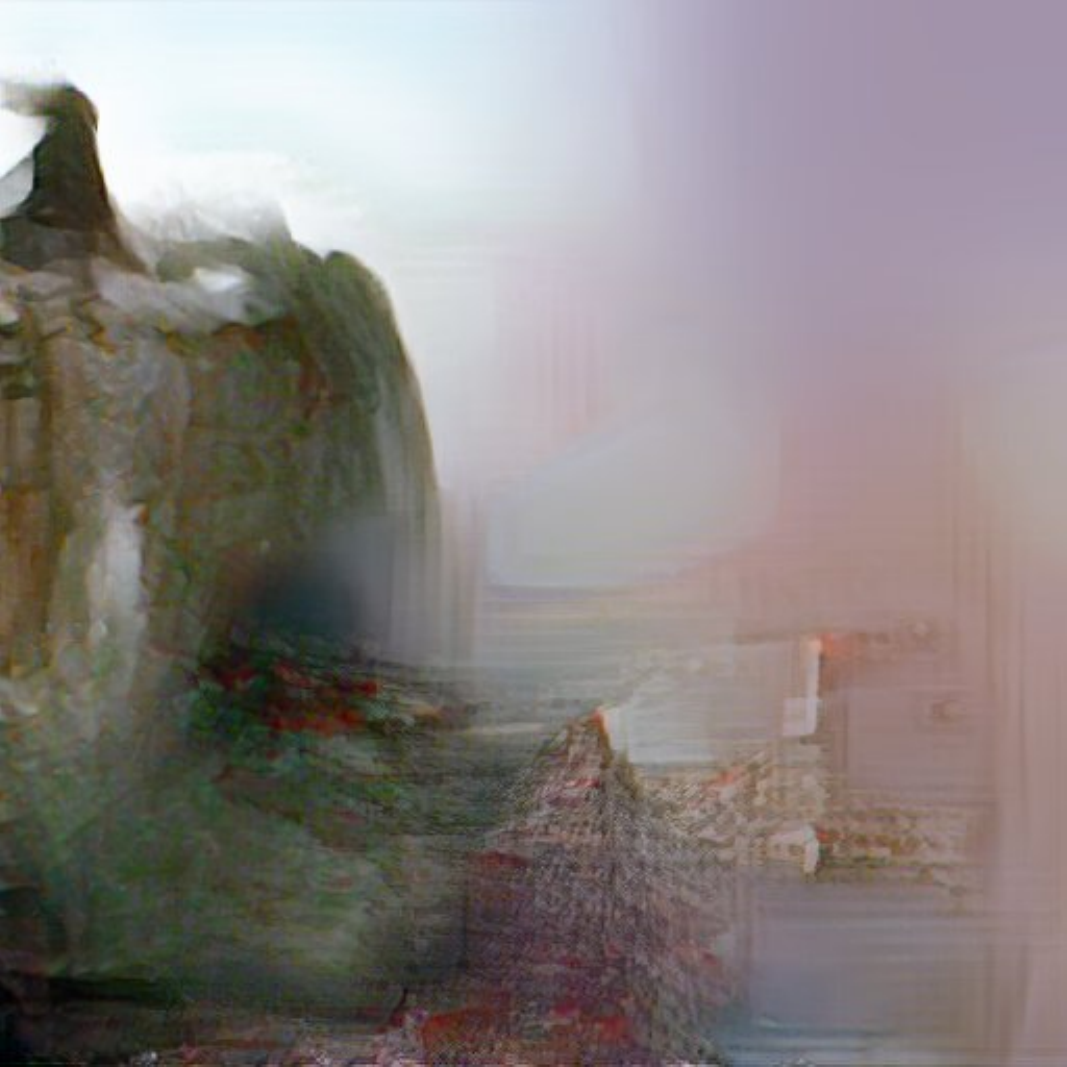}
    }
    \hspace{-3mm}
    \subfigure{
        \includegraphics[width=0.14\linewidth]{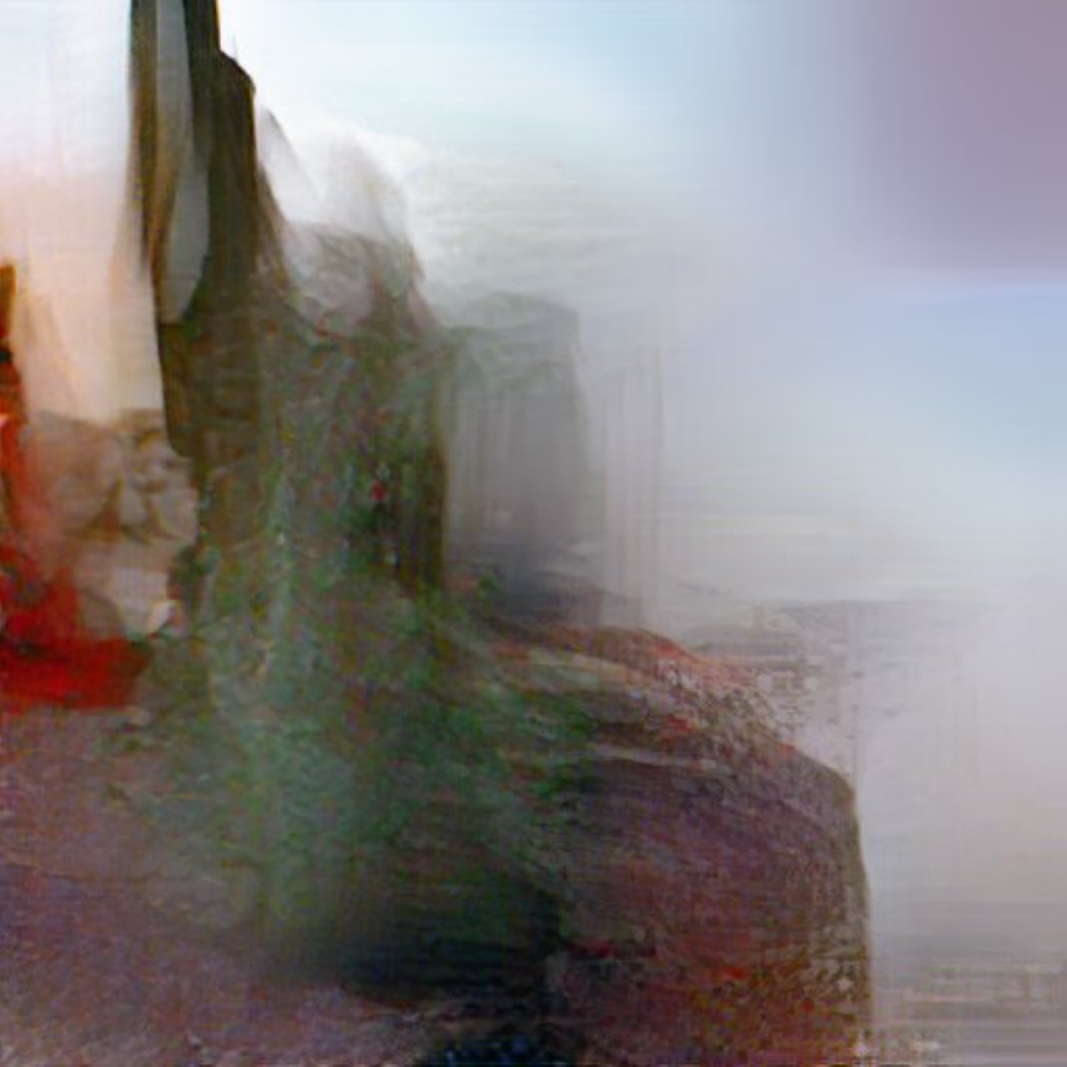}
    }
    \hspace{-3mm}
    \subfigure{
        \includegraphics[width=0.14\linewidth]{fig/heads/Sphere-seq-01_frame-000565.color.pdf}
    }
    \\
        \vspace{-3mm}
    \subfigure{
        \includegraphics[width=0.14\linewidth]{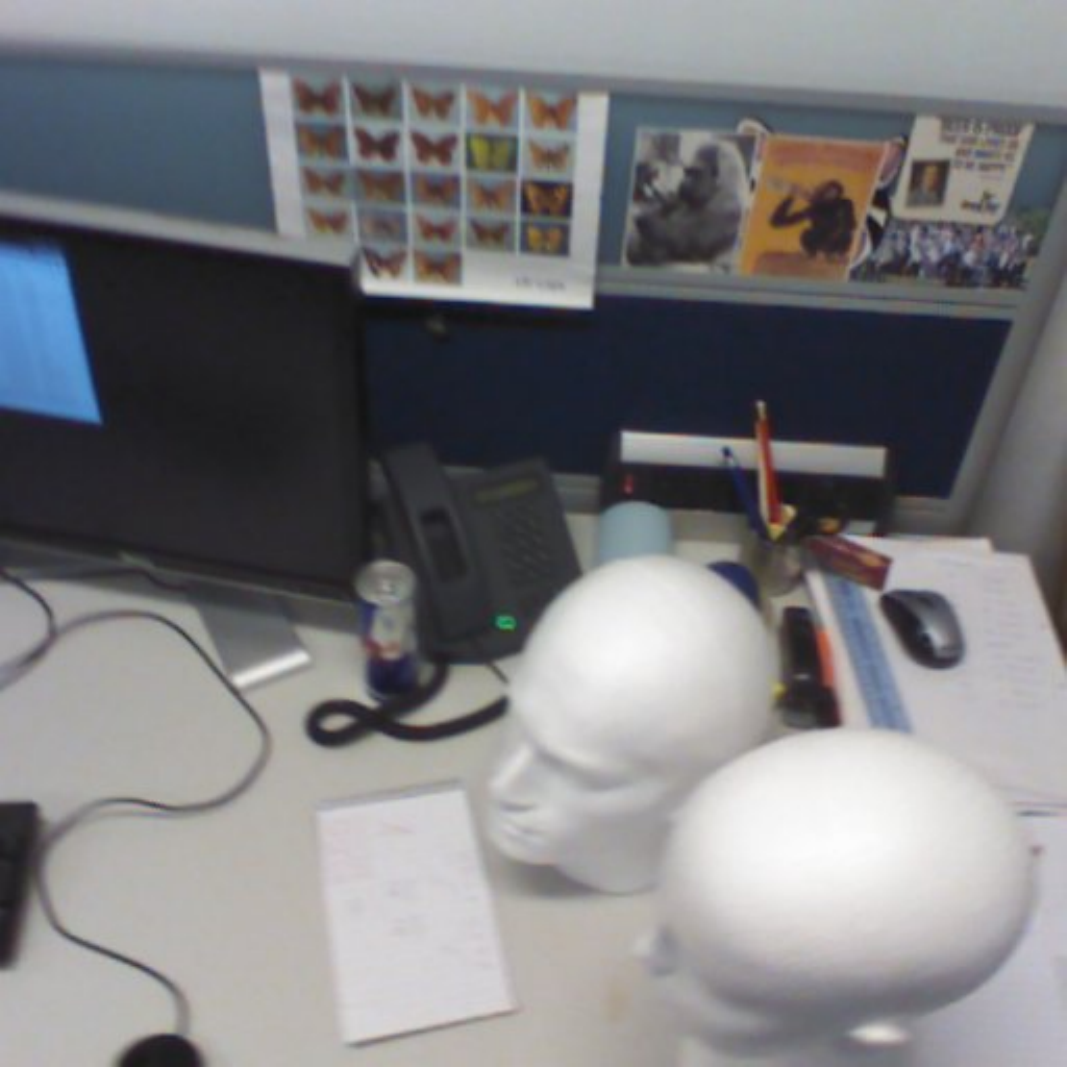}
    }
    \hspace{-3mm}
    \subfigure{
        \includegraphics[width=0.14\linewidth]{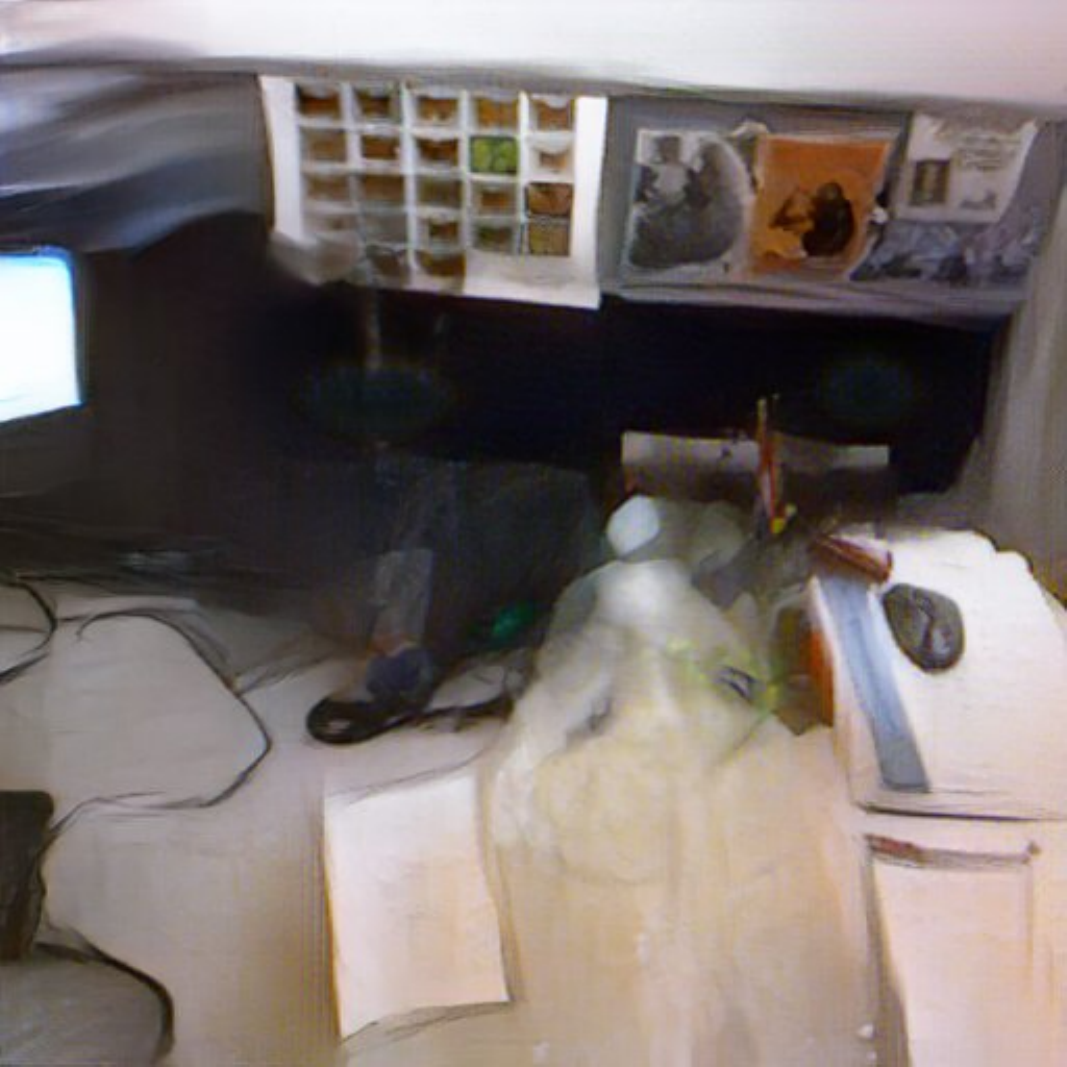}
    }
    \hspace{-3mm}
    \subfigure{
        \includegraphics[width=0.14\linewidth]{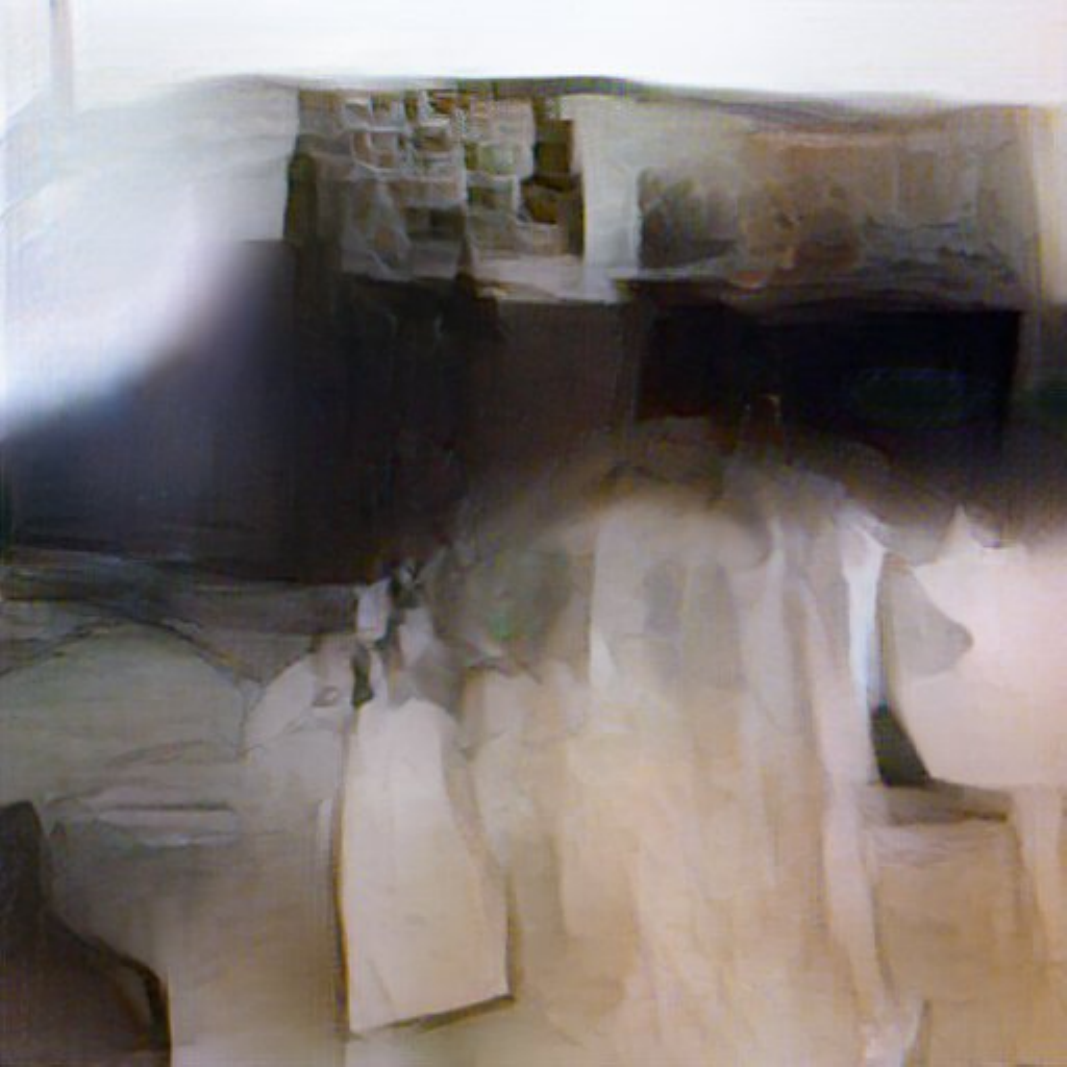}
    }
    \hspace{-3mm}
    \subfigure{
        \includegraphics[width=0.14\linewidth]{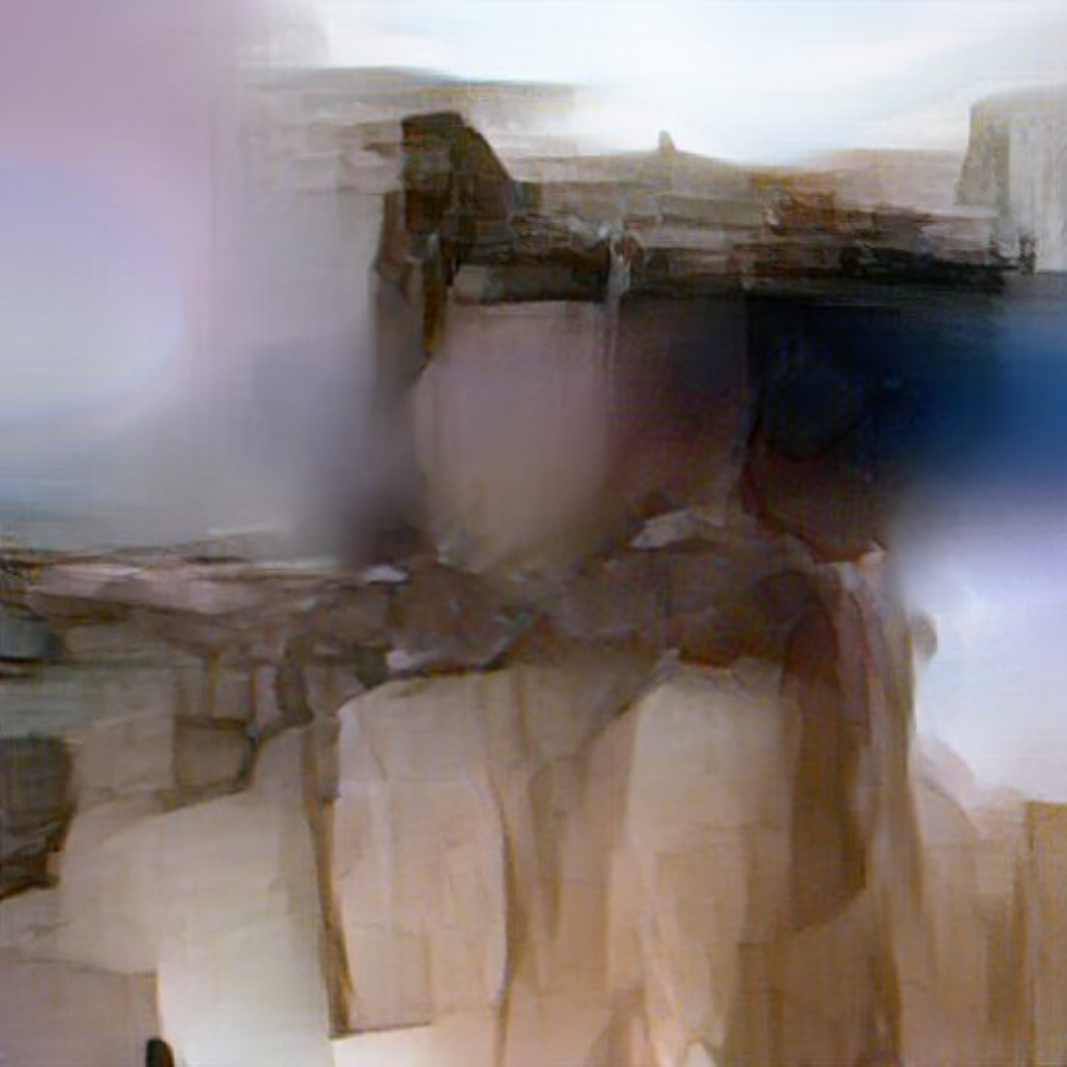}
    }
    \hspace{-3mm}
    \subfigure{
        \includegraphics[width=0.14\linewidth]{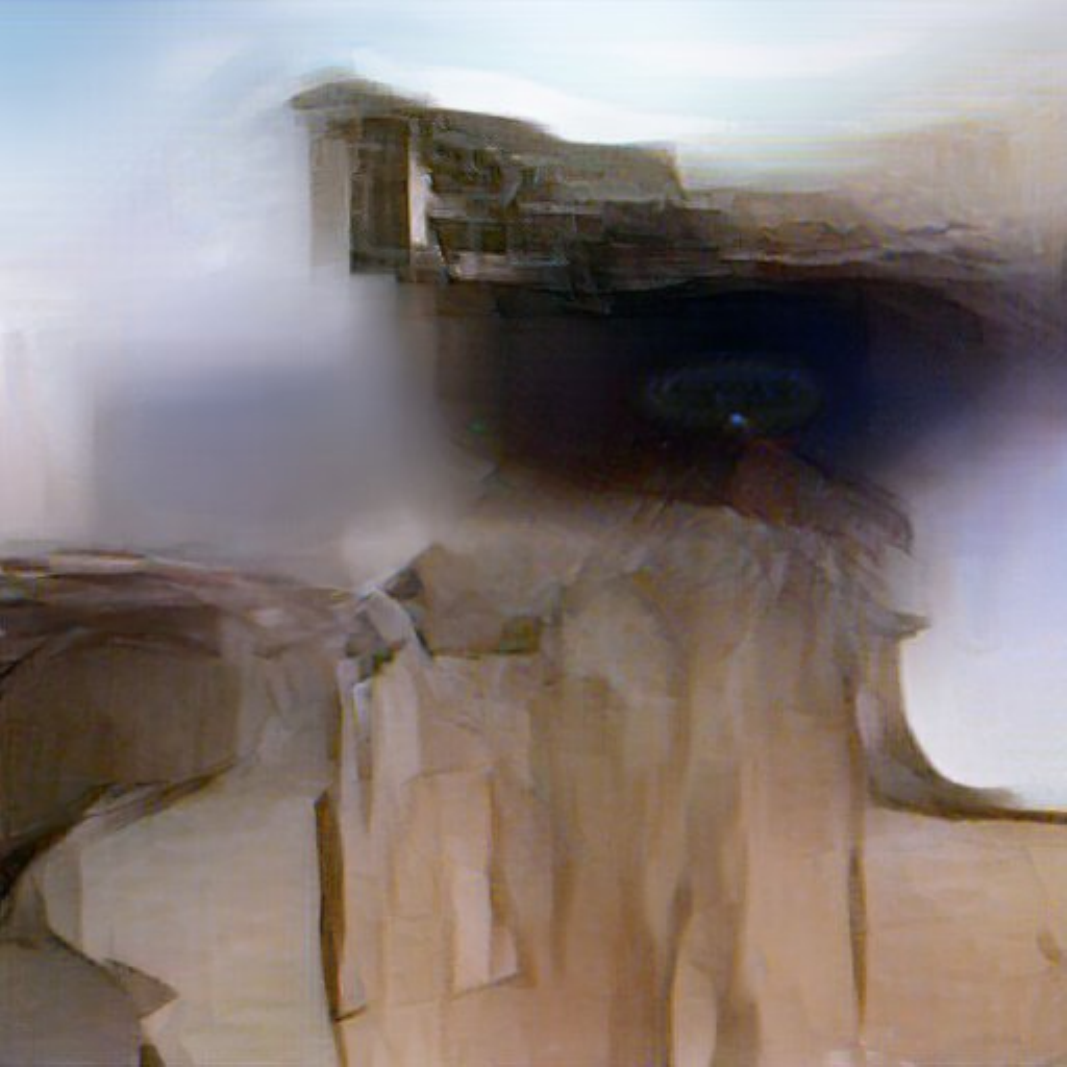}
    }
    \hspace{-3mm}
    \subfigure{
        \includegraphics[width=0.14\linewidth]{fig/heads/Sphere-seq-01_frame-000565.color.pdf}
    }
    \\    
    \vspace{-3mm}
    \setcounter{subfigure}{0}
    \hspace{0.1mm}
    \subfigure[Ground truth]{
        \includegraphics[width=0.14\linewidth]{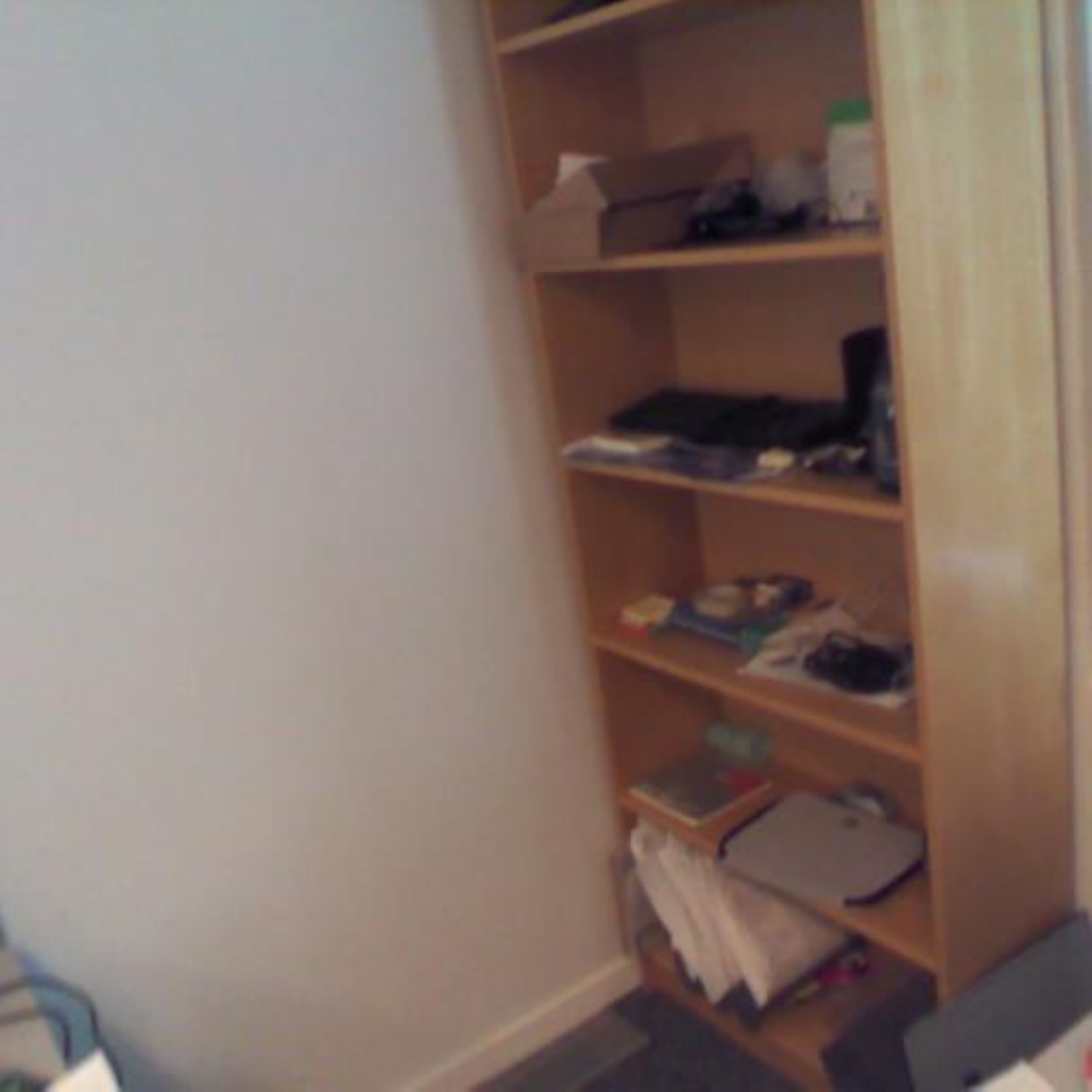}
    }
    \hspace{-3mm}
    \subfigure[Point cloud]{
        \includegraphics[width=0.14\linewidth]{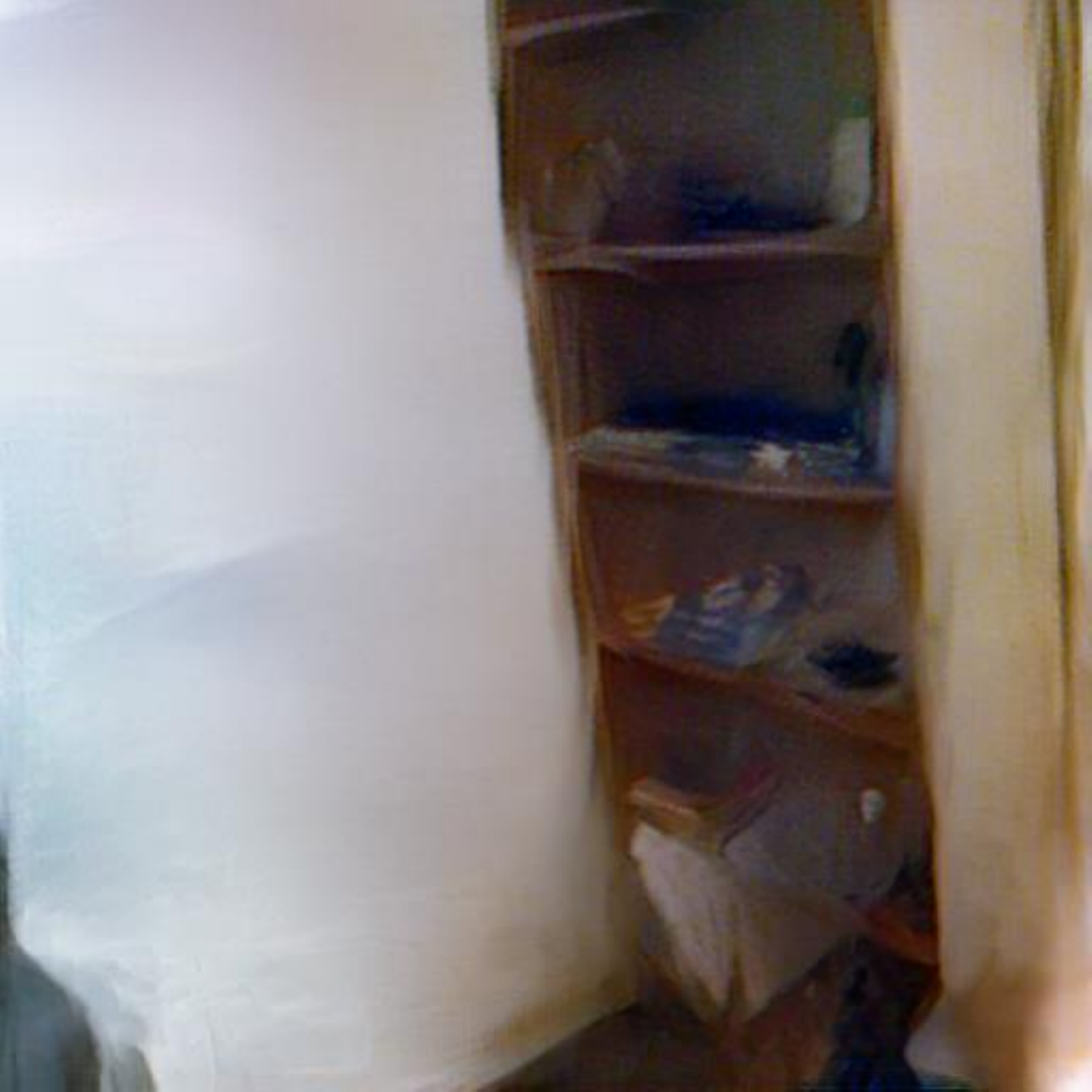}
    }
    \hspace{-3mm}
    \subfigure[ULC~\cite{speciale2019privacy}]{
        \includegraphics[width=0.14\linewidth]{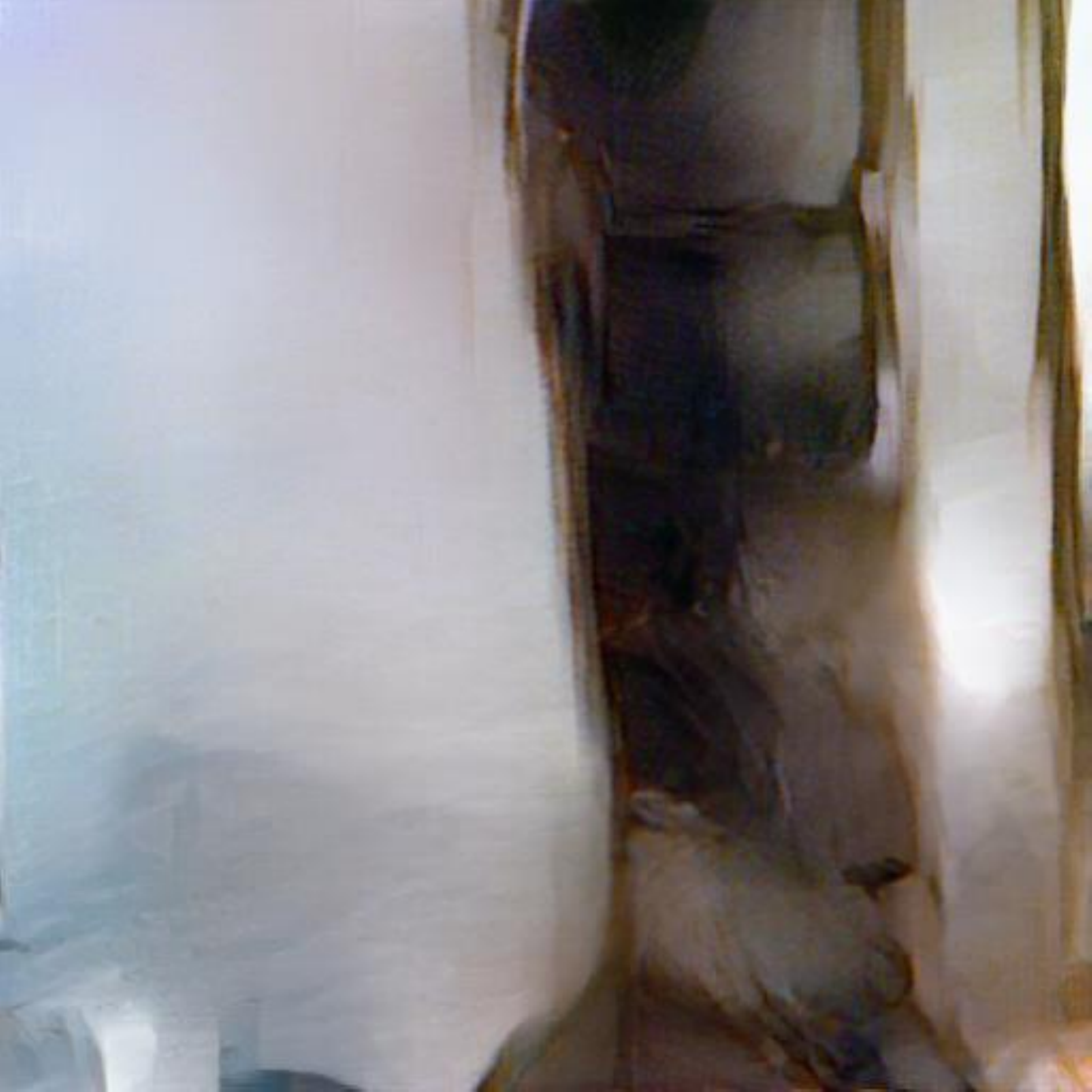}
    }
    \hspace{-3mm}
    \subfigure[PPL~\cite{lee2023ppl}]{
        \includegraphics[width=0.14\linewidth]{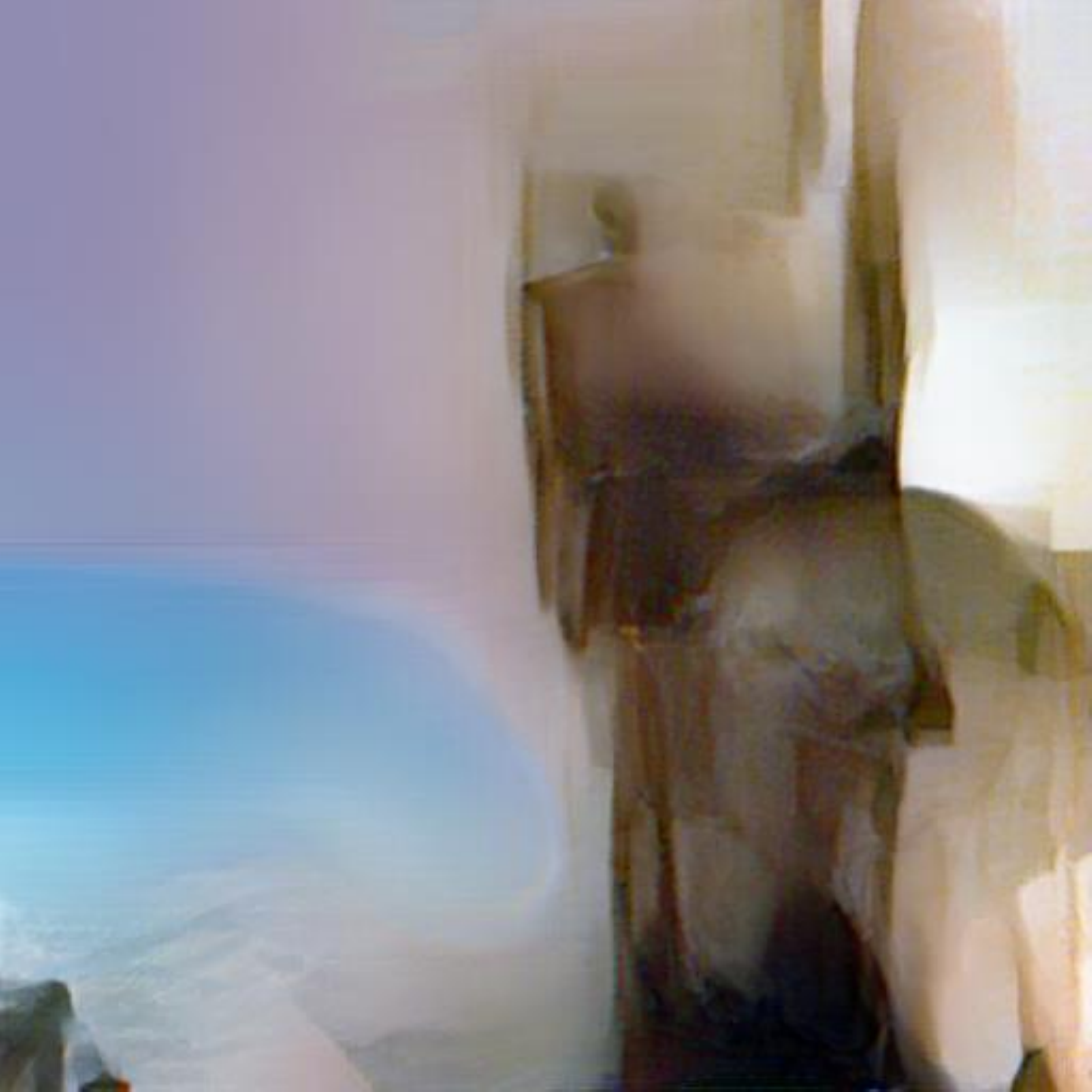}
    }
    \hspace{-3mm}
    \subfigure[PPL+~\cite{lee2023ppl}]{
        \includegraphics[width=0.14\linewidth]{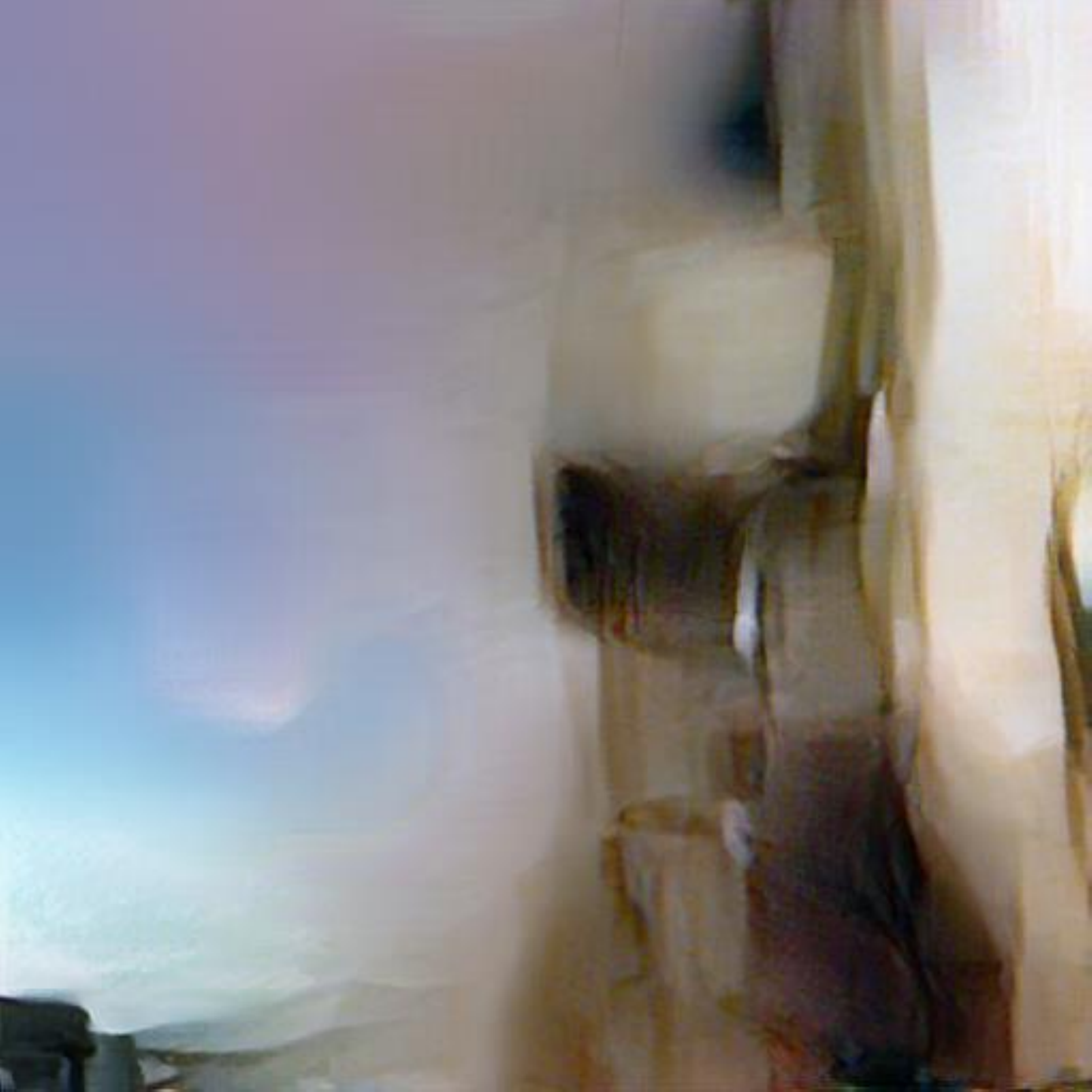}
    }
    \hspace{-3mm}
    \subfigure[Sphere cloud~\figlabel{sphere_invsfm}]{
        \includegraphics[width=0.14\linewidth]{fig/heads/Sphere-seq-01_frame-000565.color.pdf}
    }    
    \vspace{-1mm}
    \caption{Images revealed from some test camera poses across different scene representation via 
    InvSfM~\cite{pittaluga2019revealing} in 12-Scenes~\cite{valentin2016energy} (Top 4 rows) and 7-Scenes~\cite{shotton2013scene} (Bottom 4 rows).
    }
    \vspace{-6mm}
    \figlabel{fig:test camera inversion}
\end{figure}

\begin{figure*}[t]
    \centering
    \hspace{1.5mm}
    \subfigure{
        \includegraphics[width=0.13\linewidth]{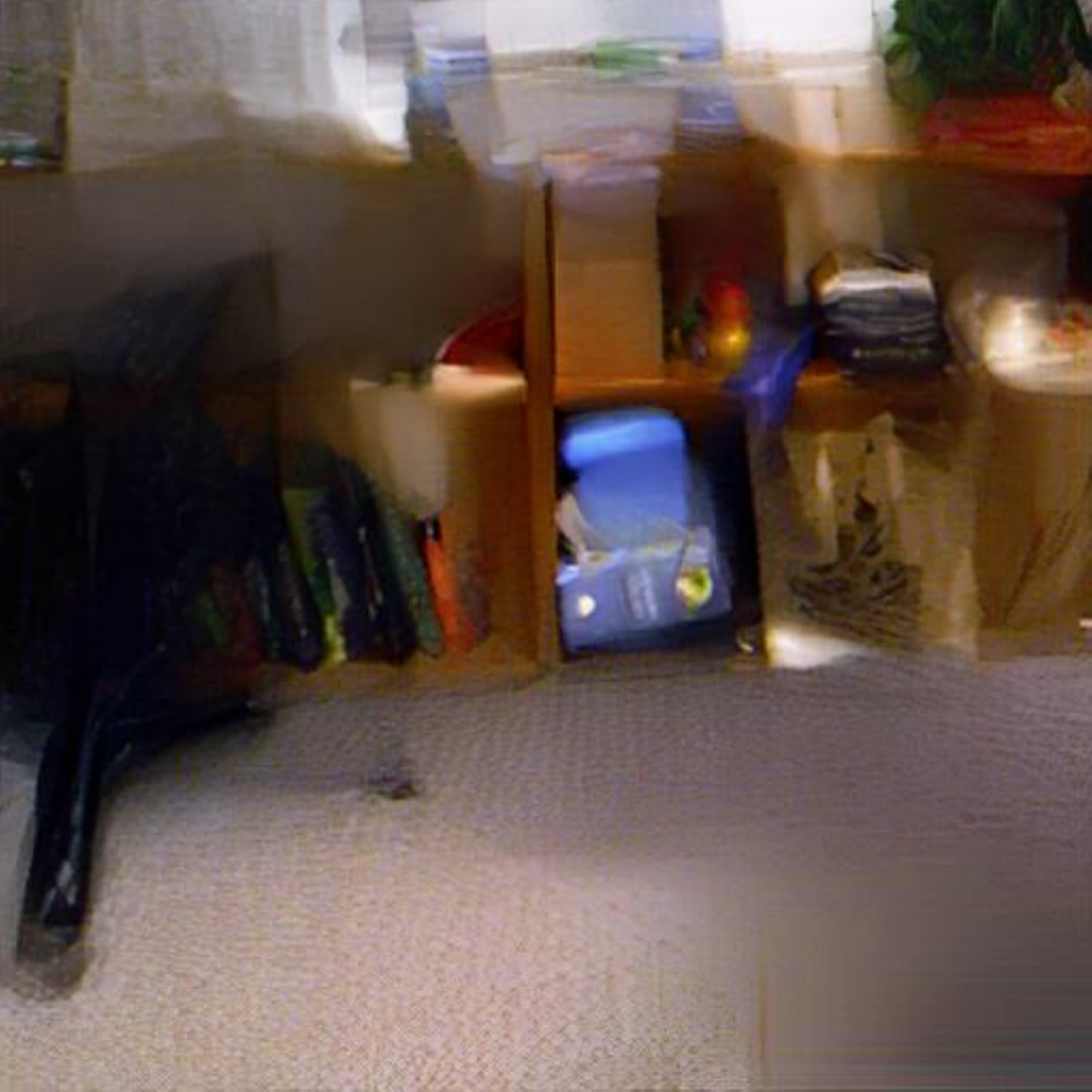}
    }
    \hspace{-3mm}
    \subfigure{
        \includegraphics[width=0.13\linewidth]{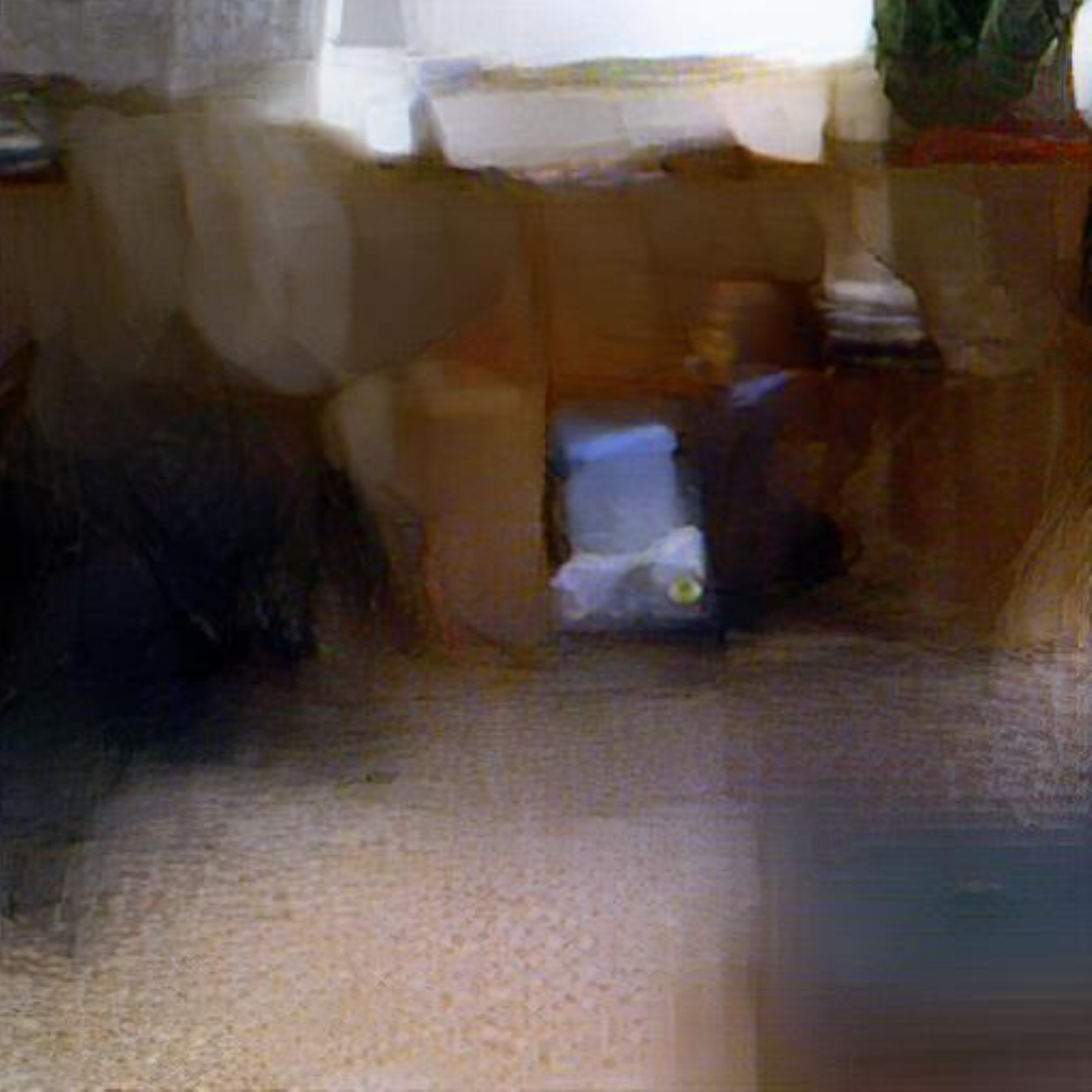}
    }
    \hspace{-3mm}
    \subfigure{
        \includegraphics[width=0.13\linewidth]{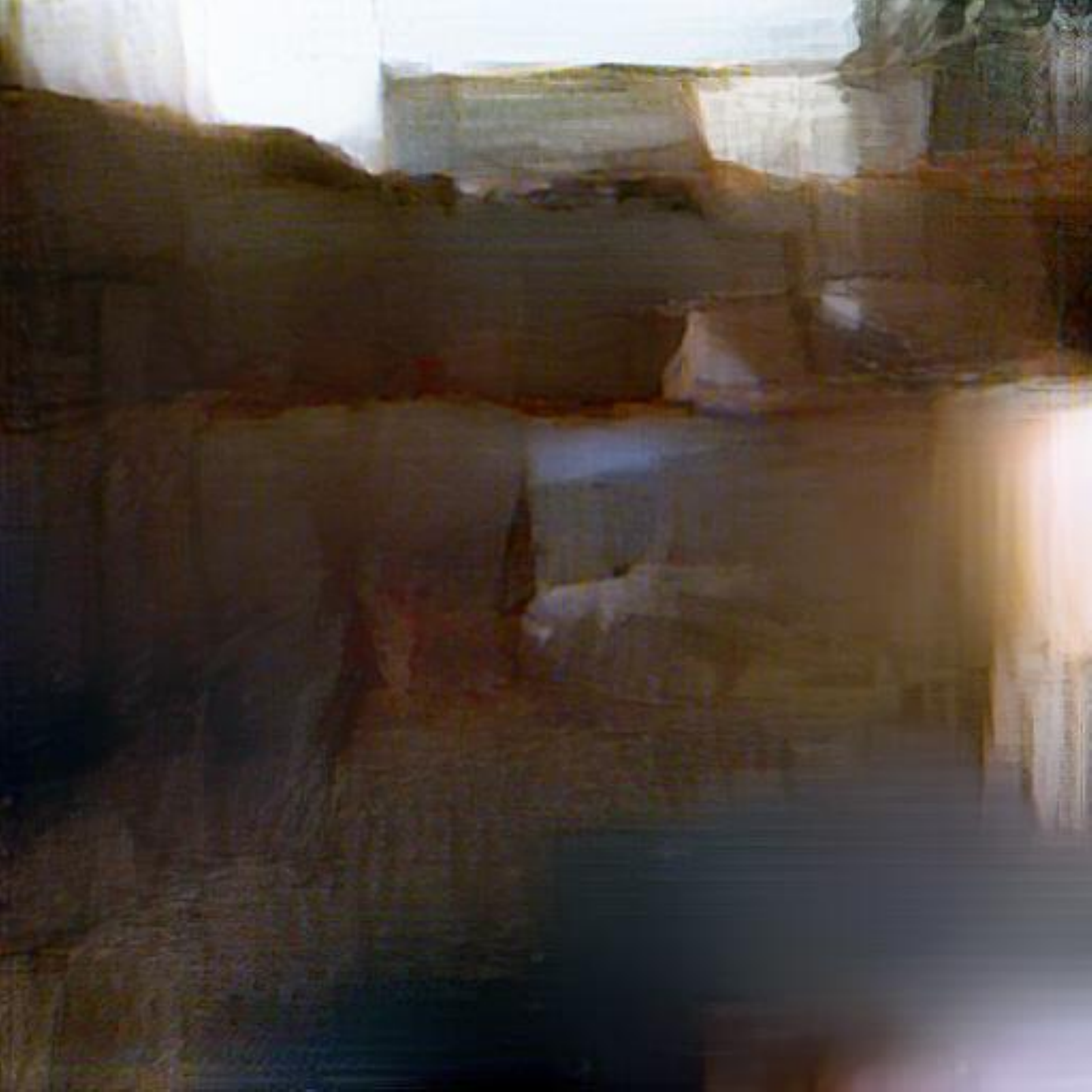}
    }
    \hspace{-3mm}
    \subfigure{
        \includegraphics[width=0.13\linewidth]{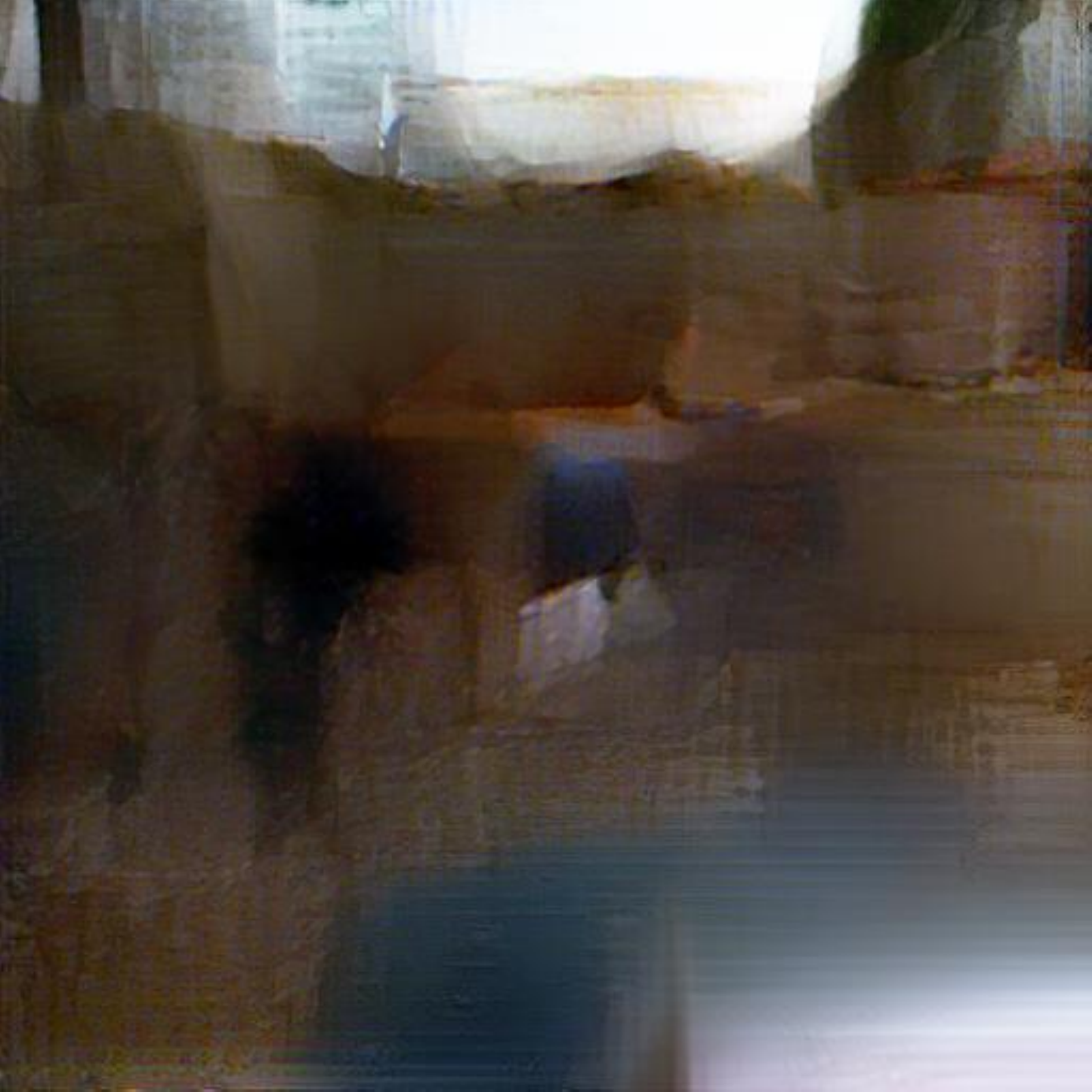}
    }
    \hspace{-3mm}
    \subfigure{
        \includegraphics[width=0.13\linewidth]{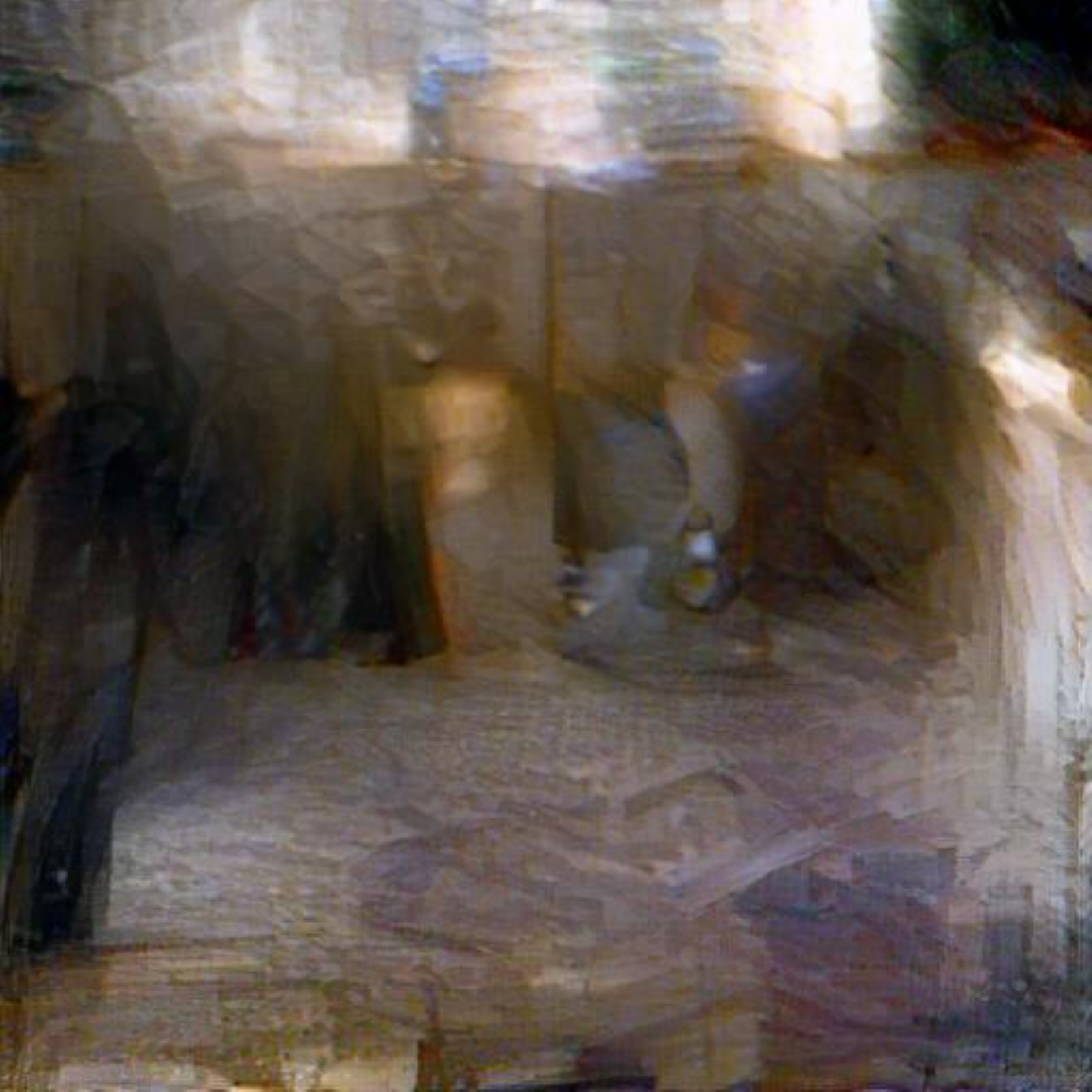}
    }
    \hspace{-3mm}
    \subfigure{
        \includegraphics[width=0.13\linewidth]{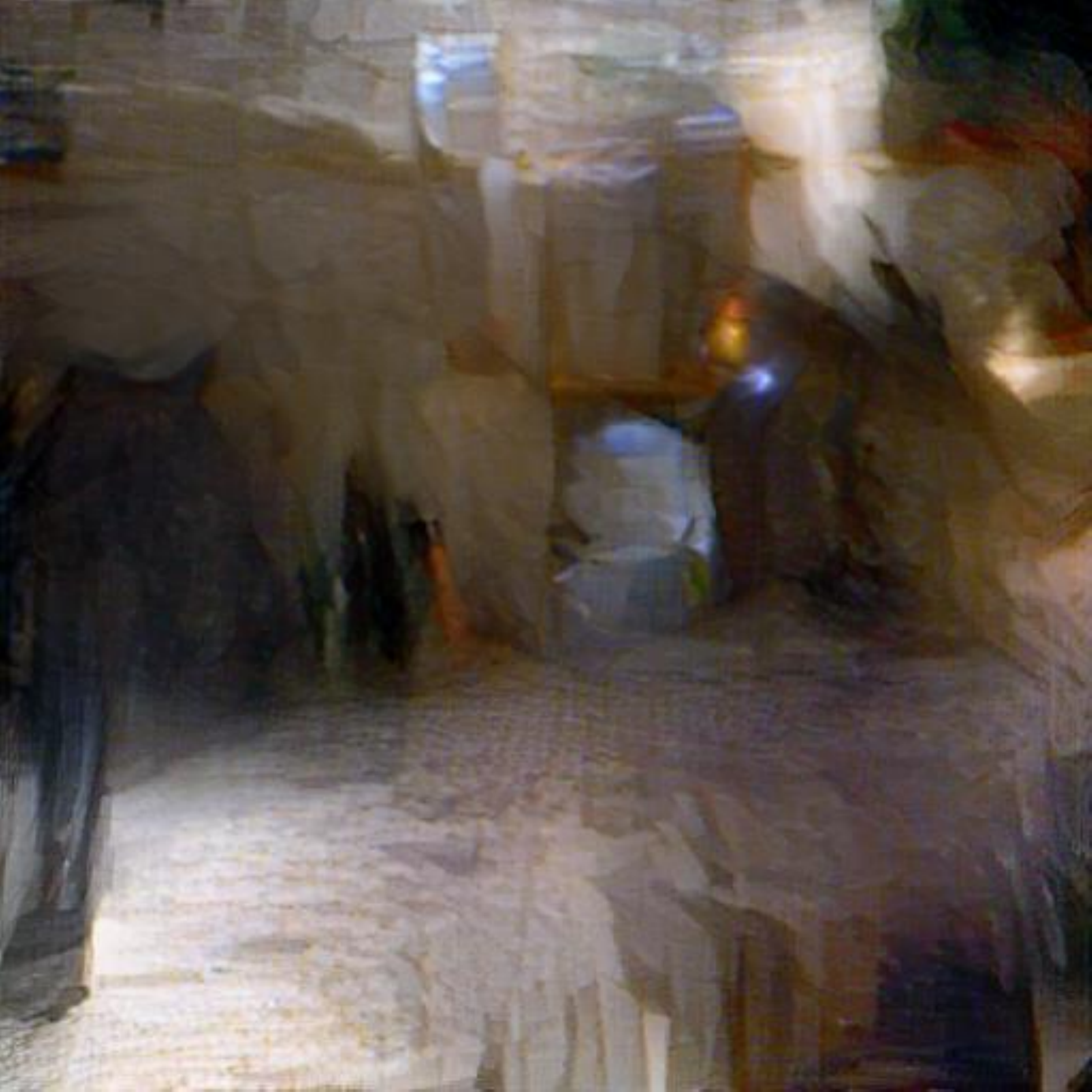}
    }
    \hspace{-3mm}
    \subfigure{
        \includegraphics[width=0.13\linewidth]{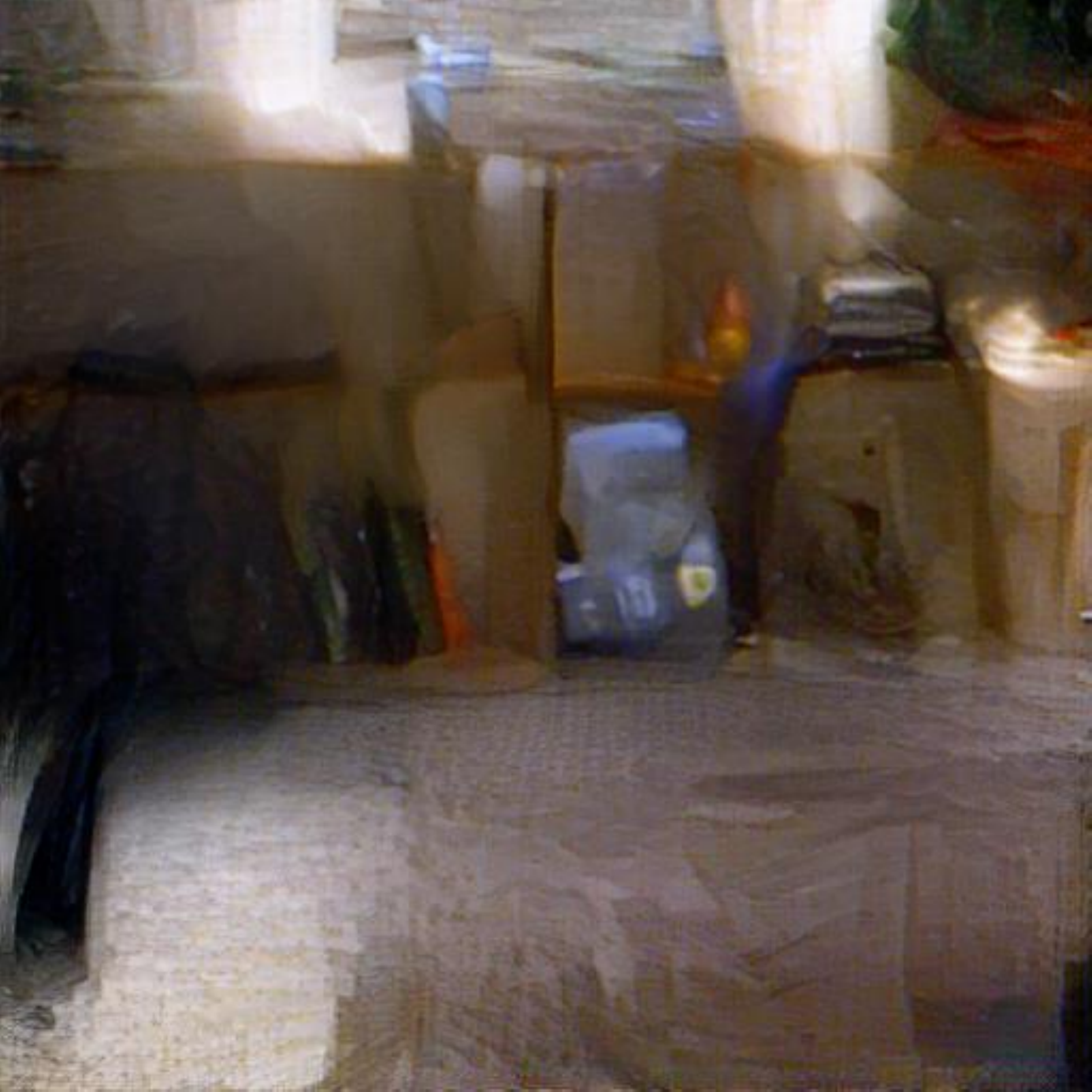}
    }
    \\
    \vspace{-3.5mm}
    \hspace{1.5mm}
    \subfigure{
        \includegraphics[width=0.13\linewidth]{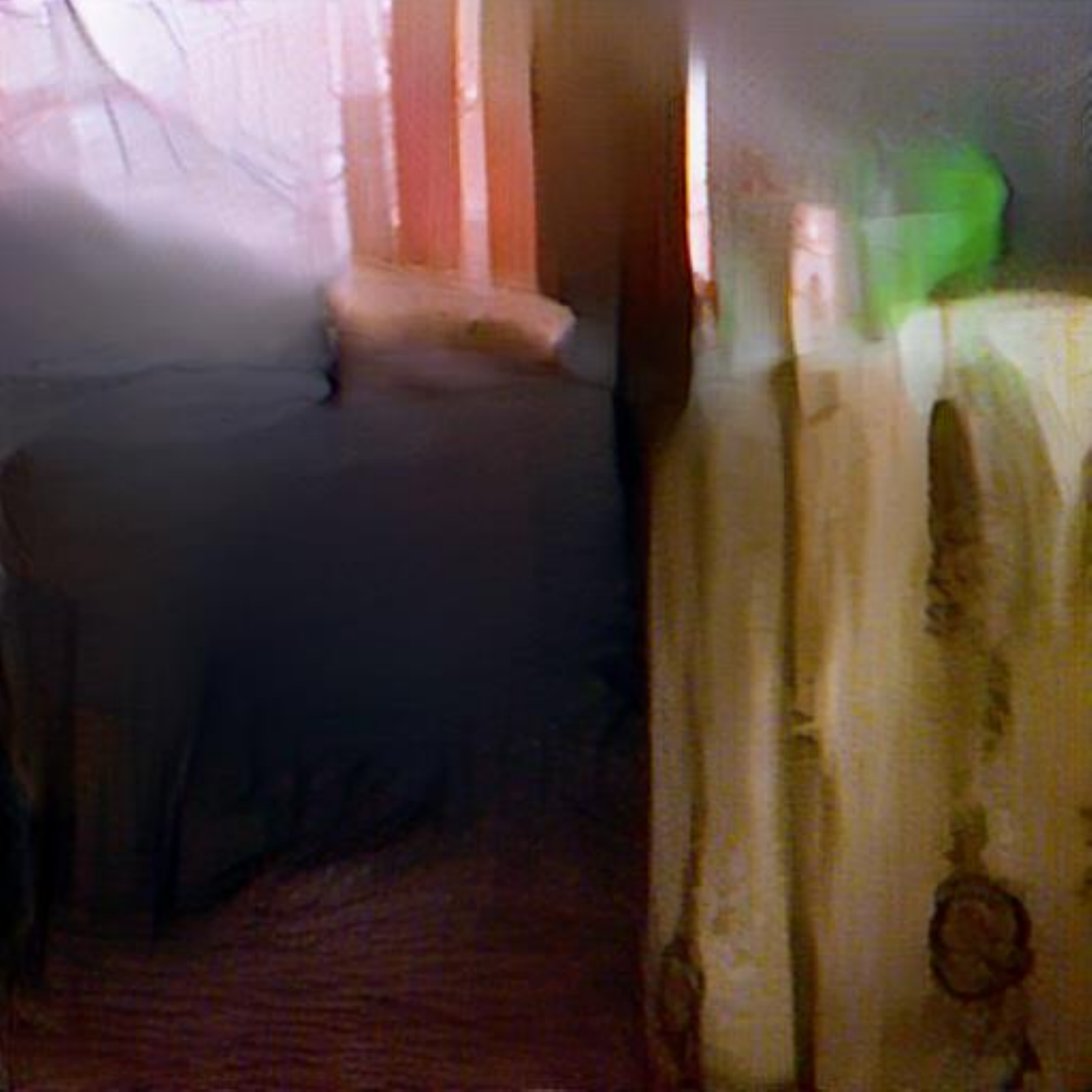}
    }
    \hspace{-3mm}
    \subfigure{
        \includegraphics[width=0.13\linewidth]{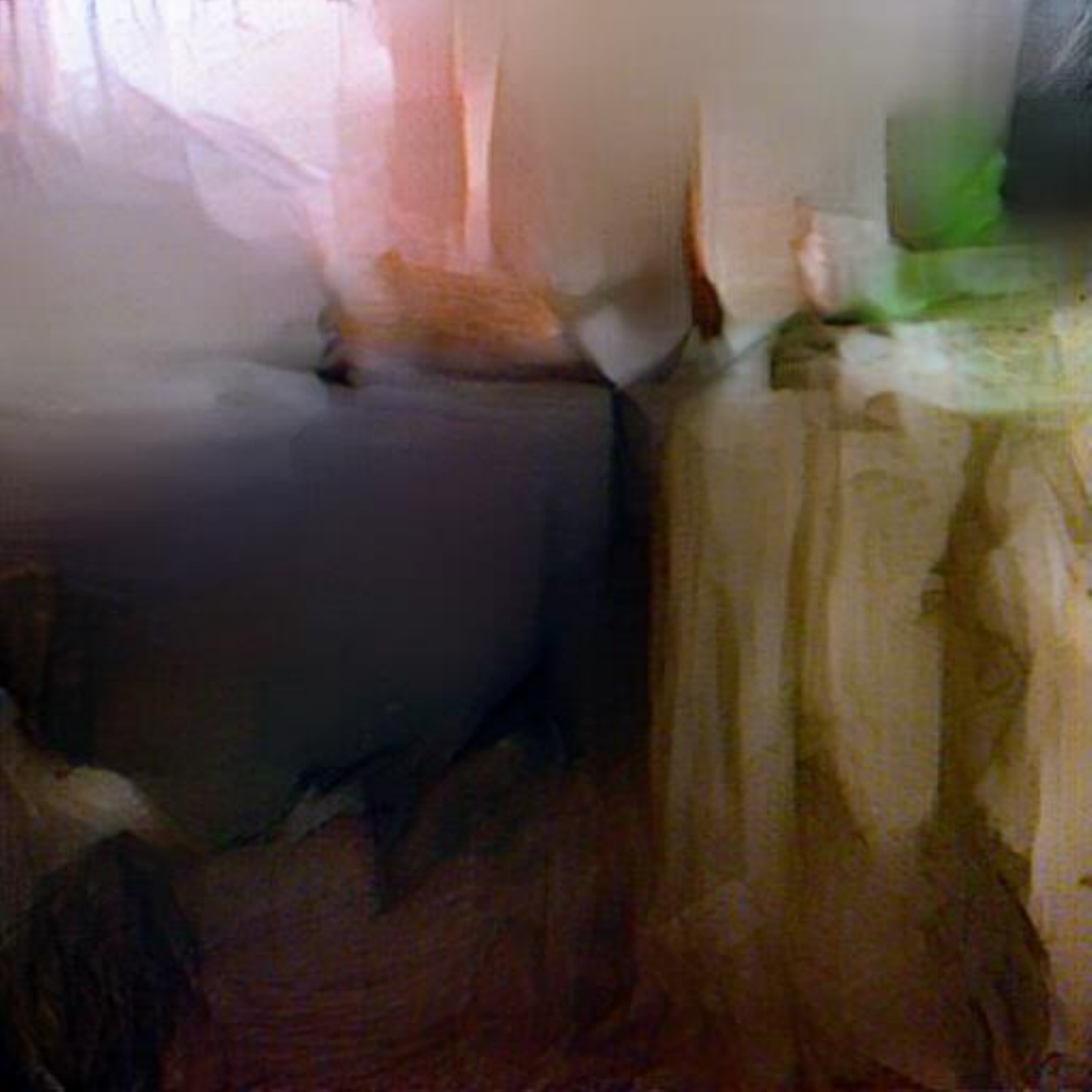}
    }
    \hspace{-3mm}
    \subfigure{
        \includegraphics[width=0.13\linewidth]{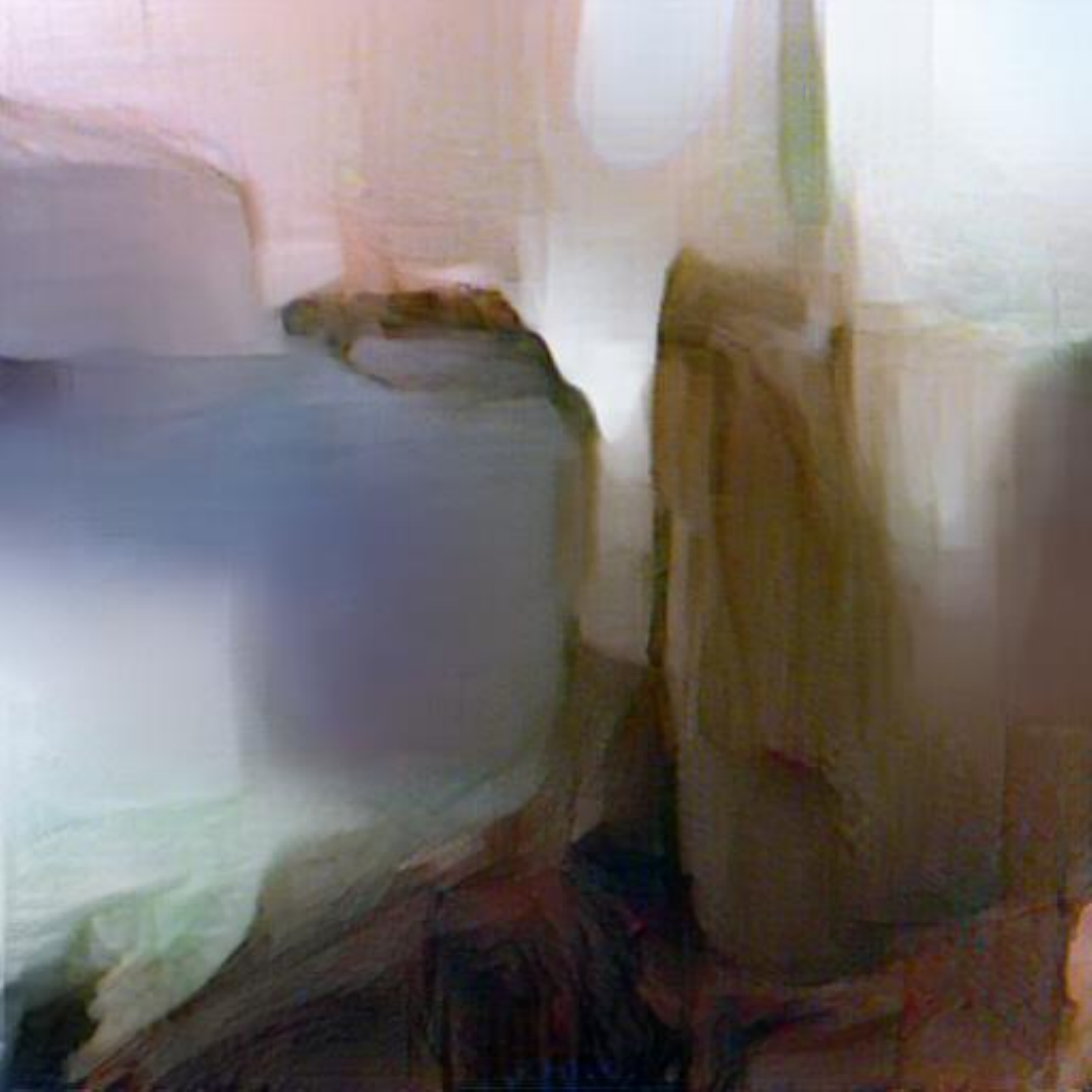}
    }
    \hspace{-3mm}
    \subfigure{
        \includegraphics[width=0.13\linewidth]{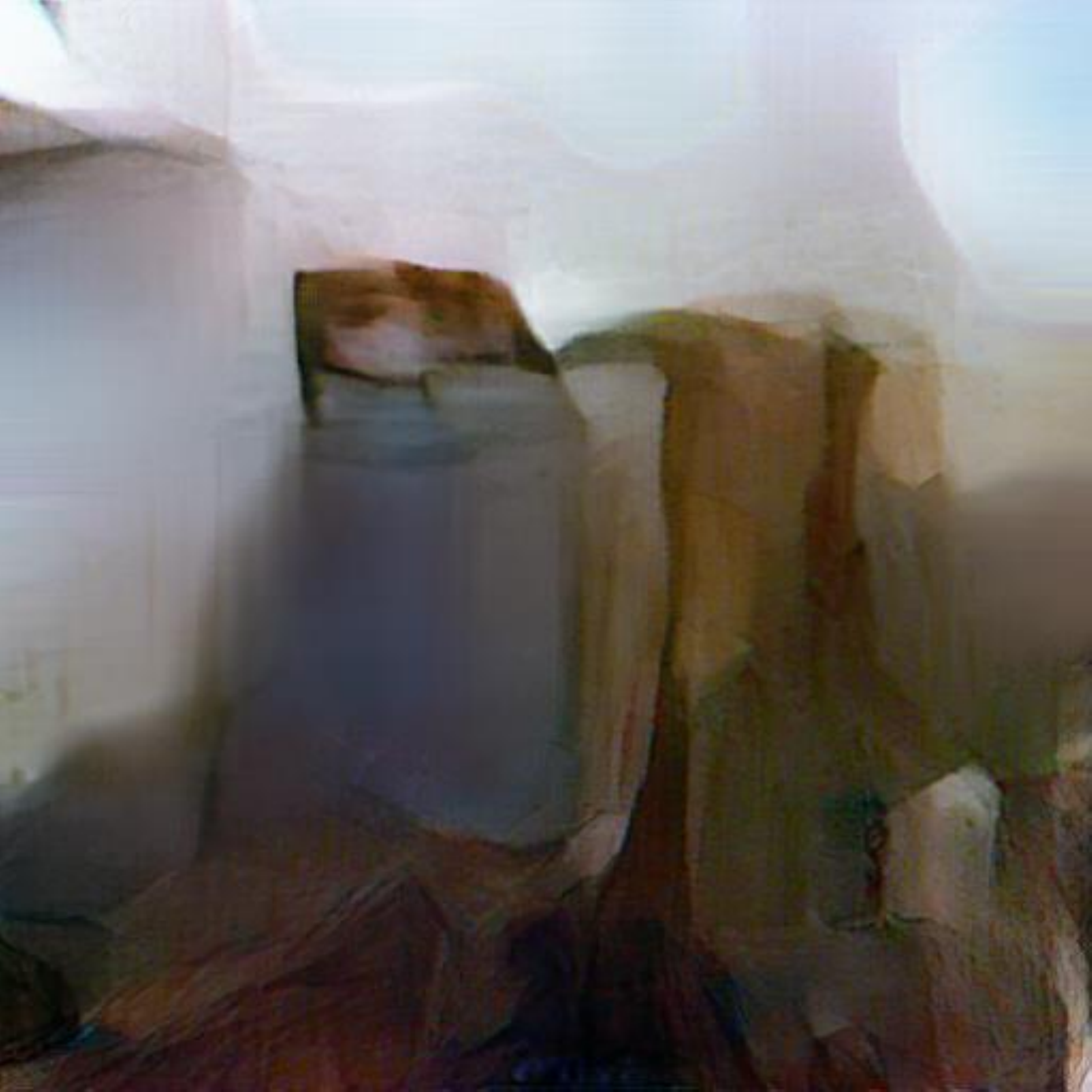}
    }
    \hspace{-3mm}
    \subfigure{
        \includegraphics[width=0.13\linewidth]{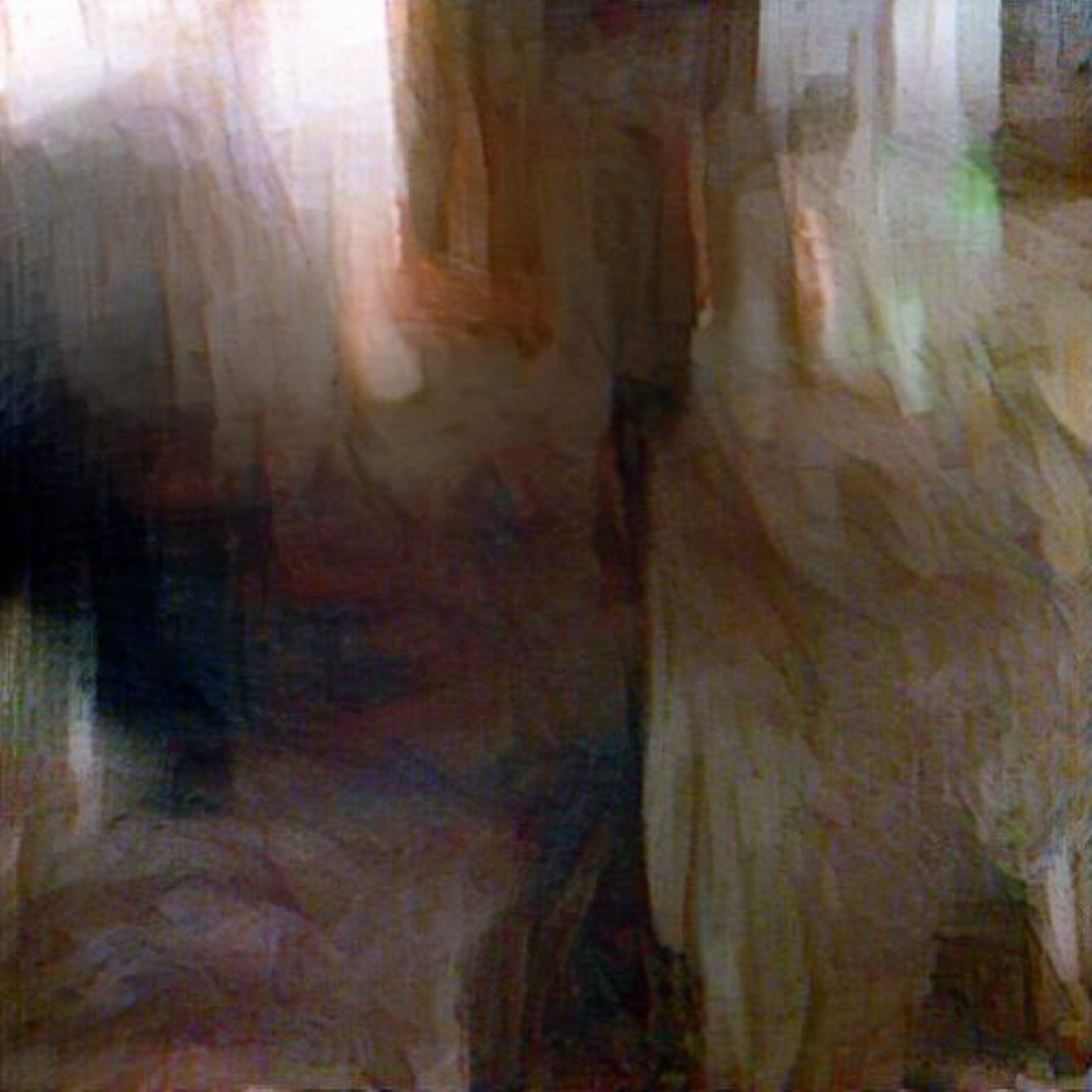}
    }
    \hspace{-3mm}
    \subfigure{
        \includegraphics[width=0.13\linewidth]{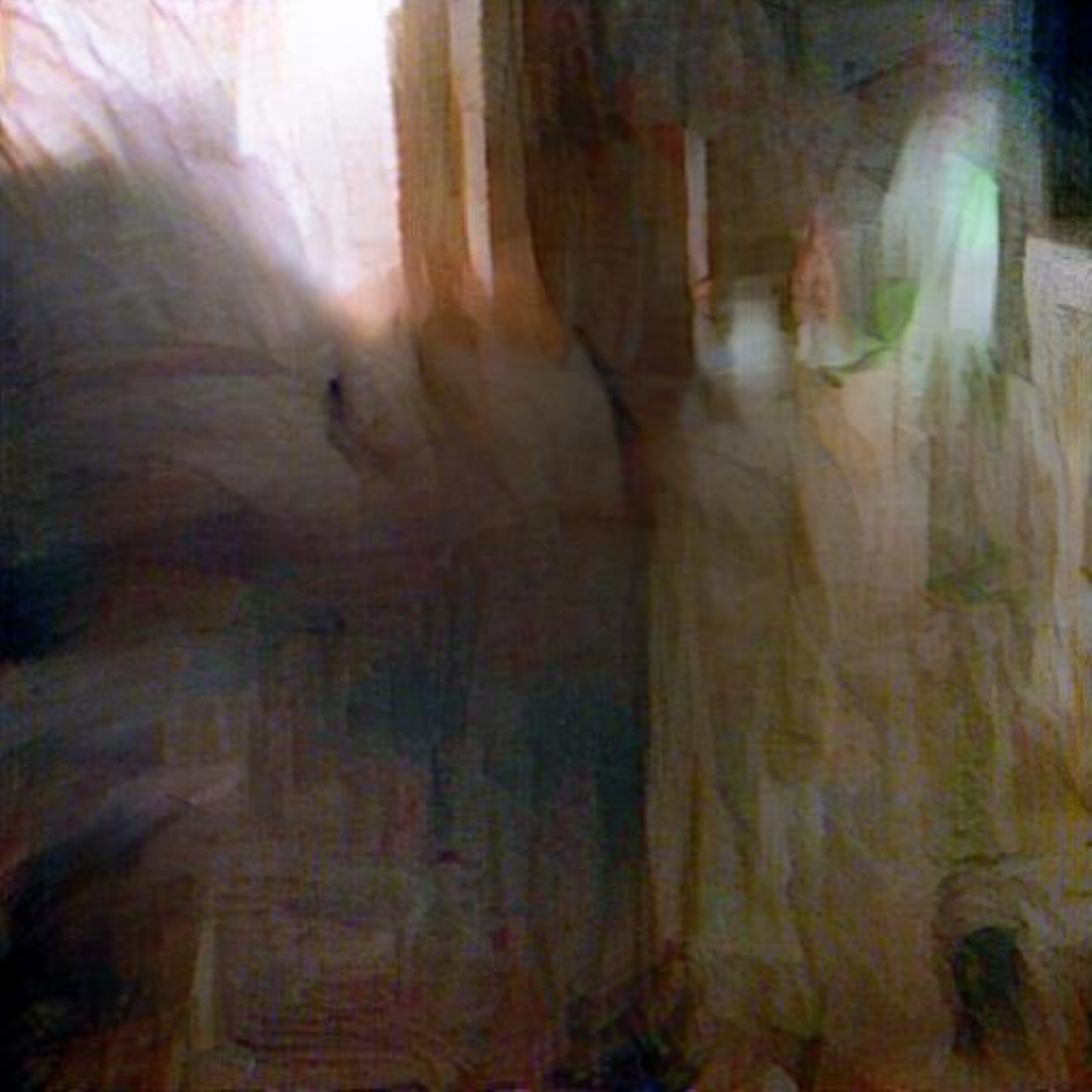}
    }
    \hspace{-3mm}
    \subfigure{
        \includegraphics[width=0.13\linewidth]{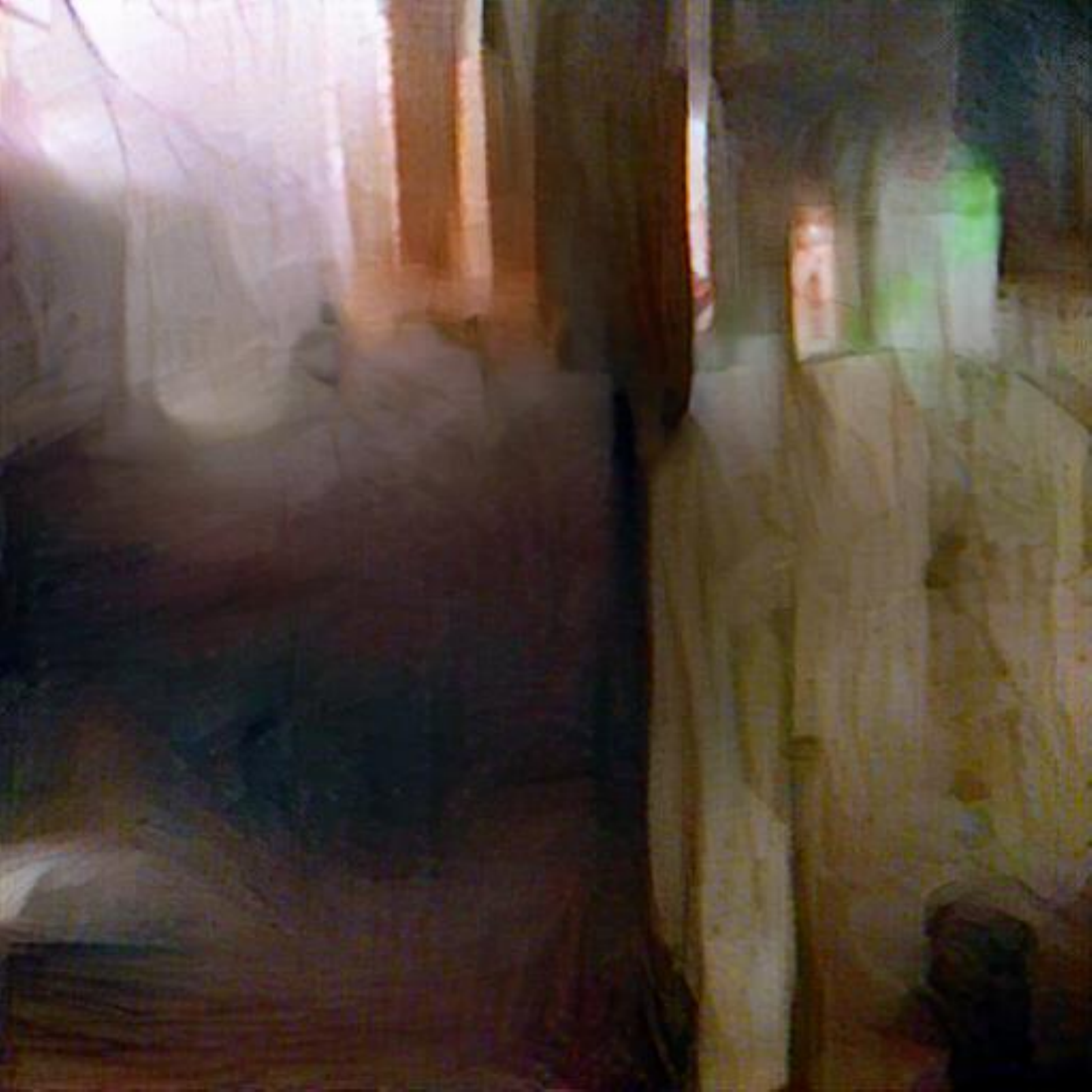}
    }
    \\
    \vspace{-3.5mm}
    \hspace{1.5mm}
    \subfigure{
        \includegraphics[width=0.13\linewidth]{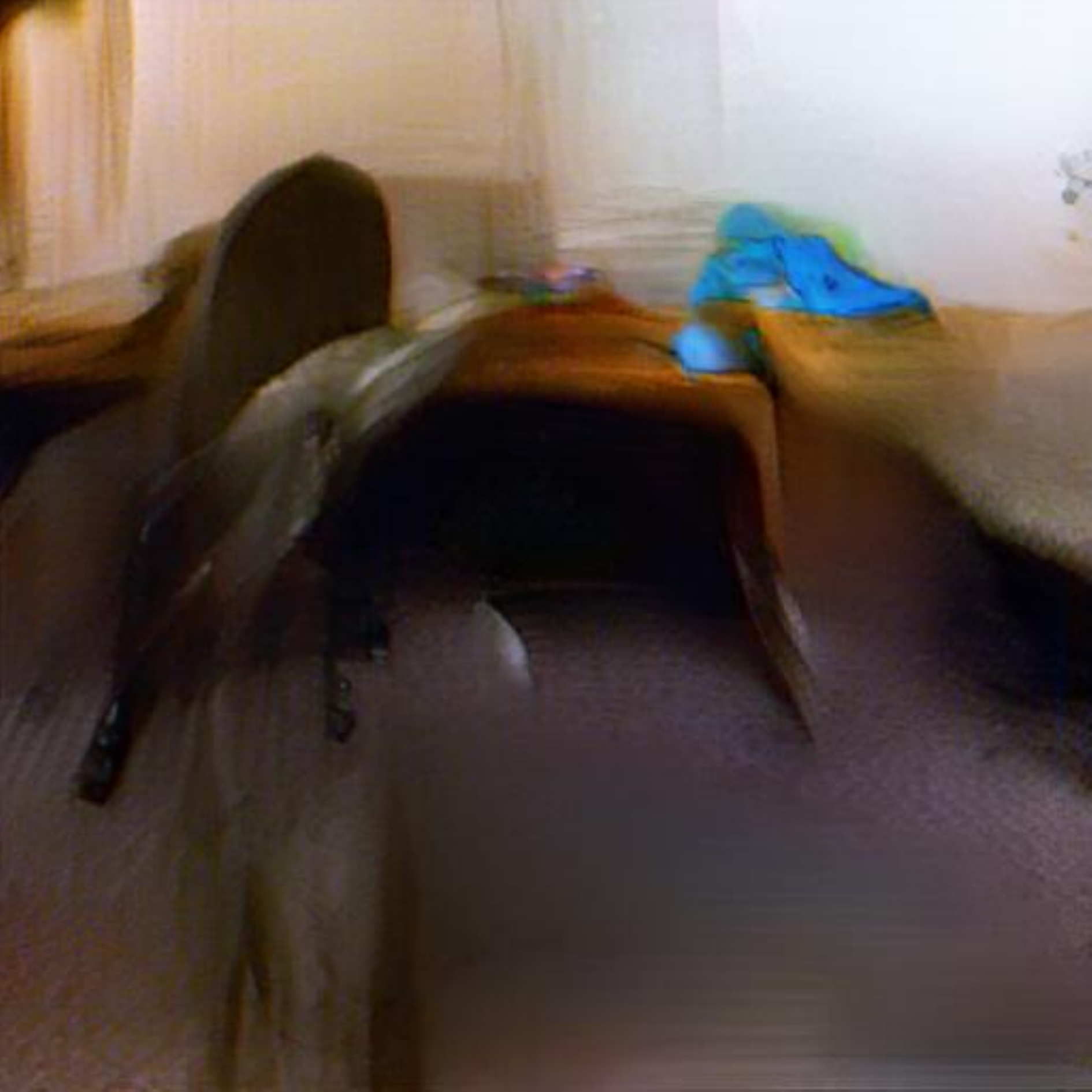}
    }
    \hspace{-3mm}
    \subfigure{
        \includegraphics[width=0.13\linewidth]{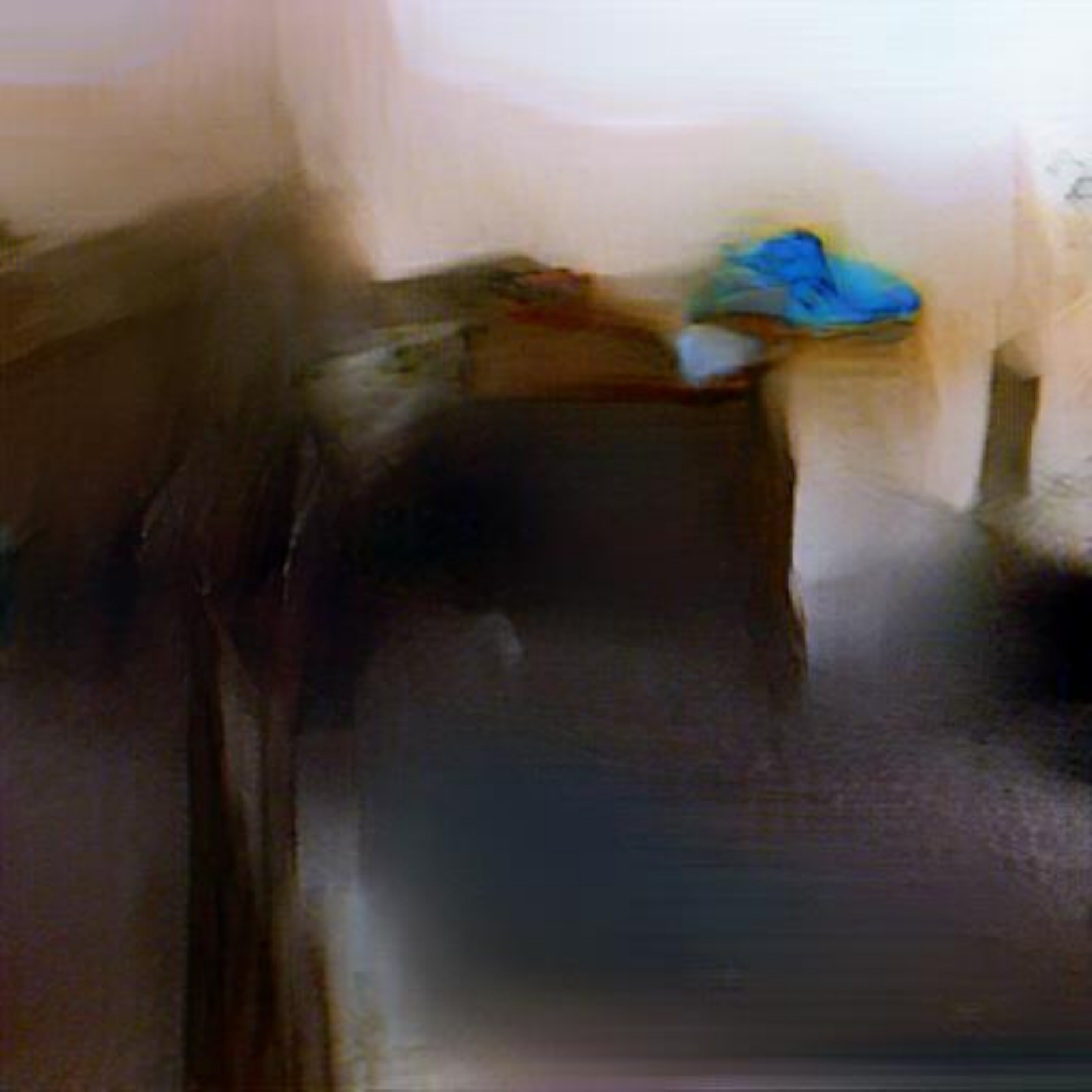}
    }
    \hspace{-3mm}
    \subfigure{
        \includegraphics[width=0.13\linewidth]{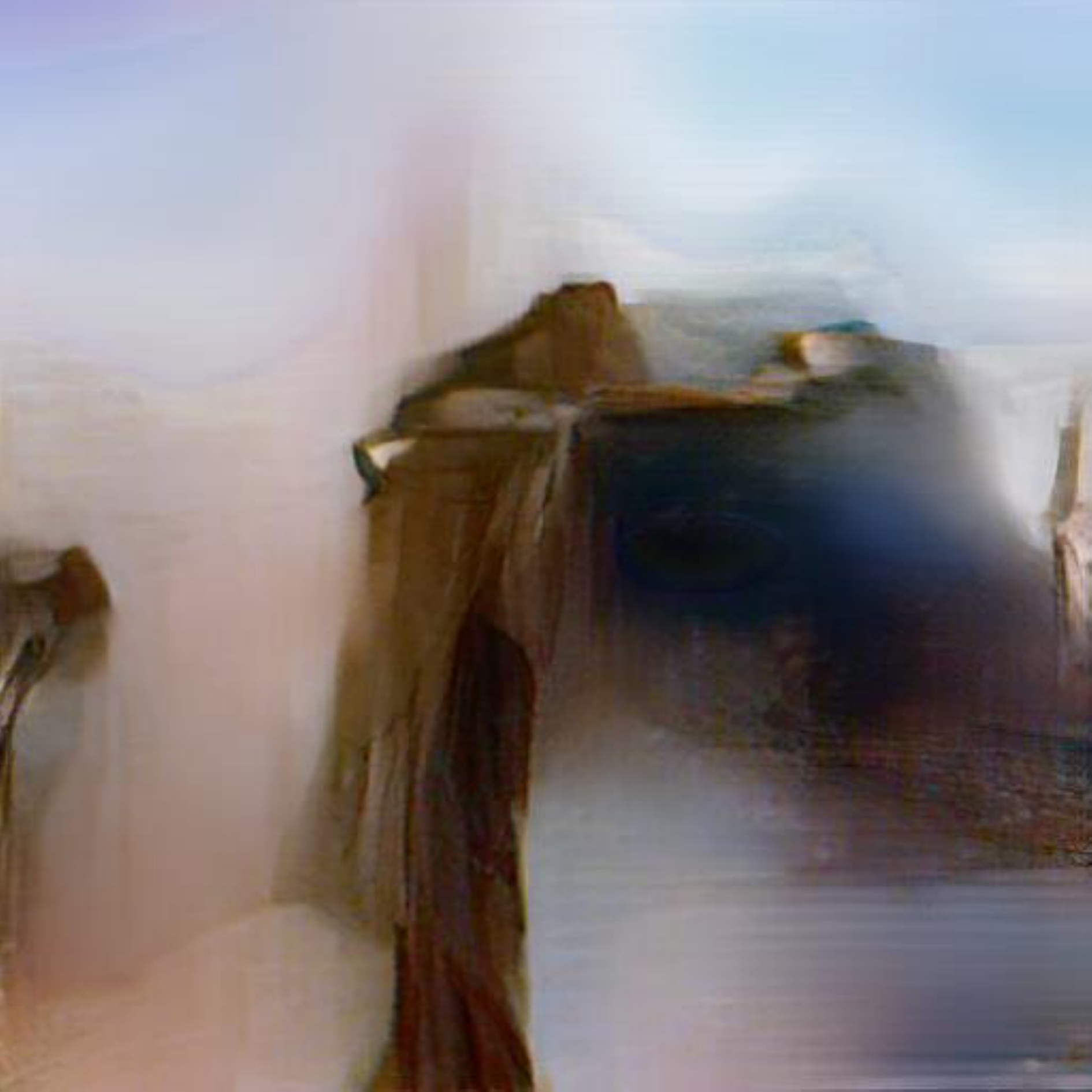}
    }
    \hspace{-3mm}
    \subfigure{
        \includegraphics[width=0.13\linewidth]{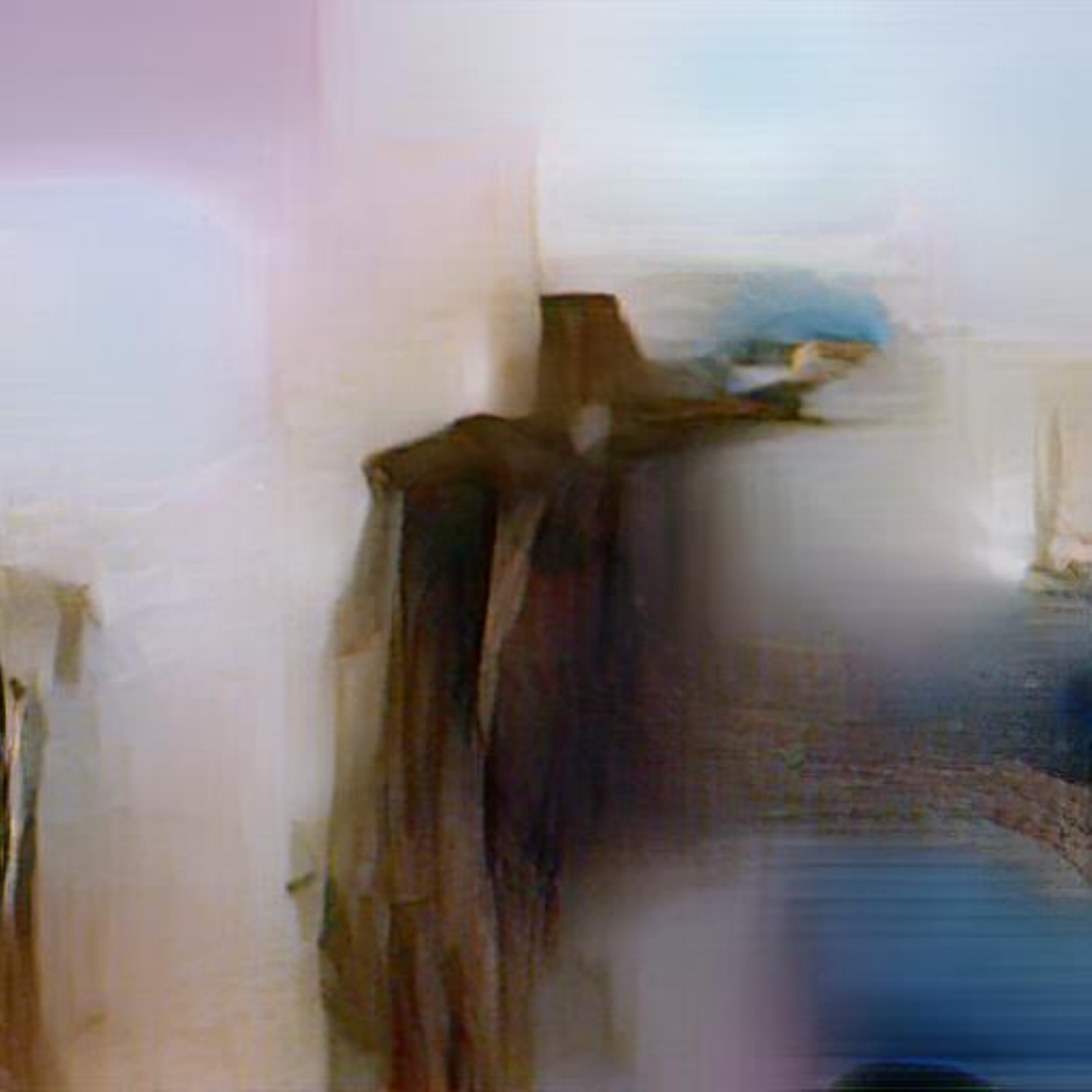}
    }
    \hspace{-3mm}
    \subfigure{
        \includegraphics[width=0.13\linewidth]{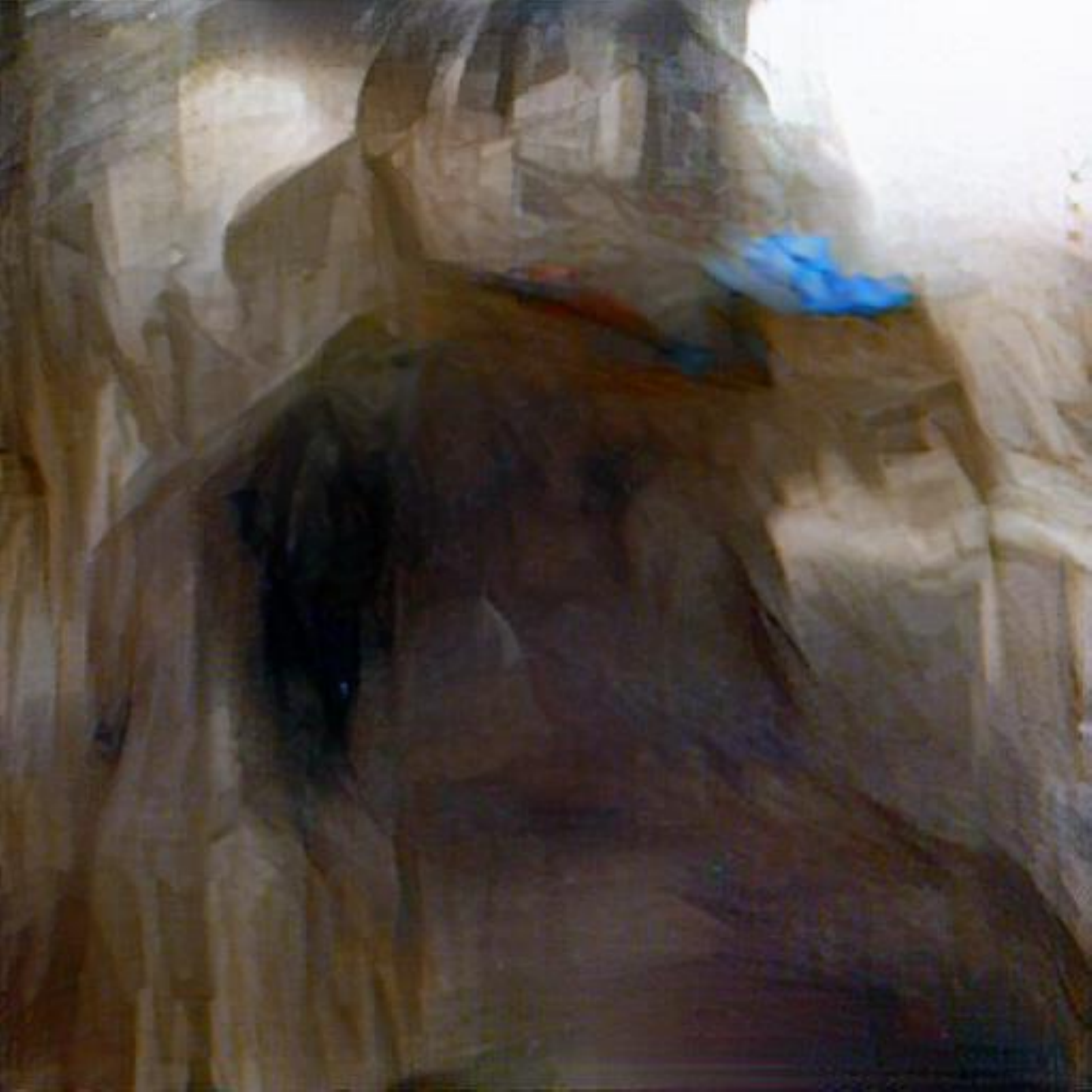}
    }
    \hspace{-3mm}
    \subfigure{
        \includegraphics[width=0.13\linewidth]{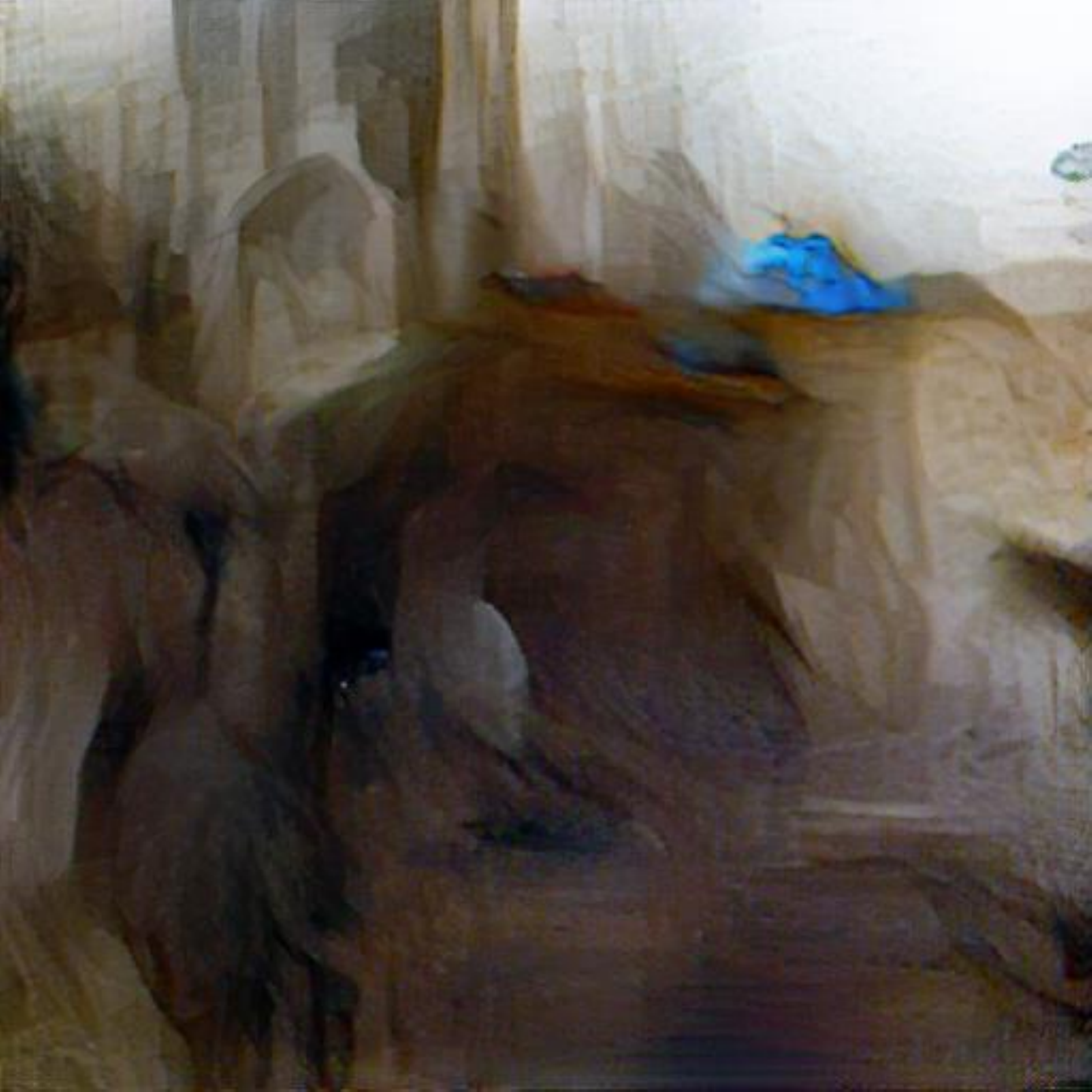}
    }
    \hspace{-3mm}
    \subfigure{
        \includegraphics[width=0.13\linewidth]{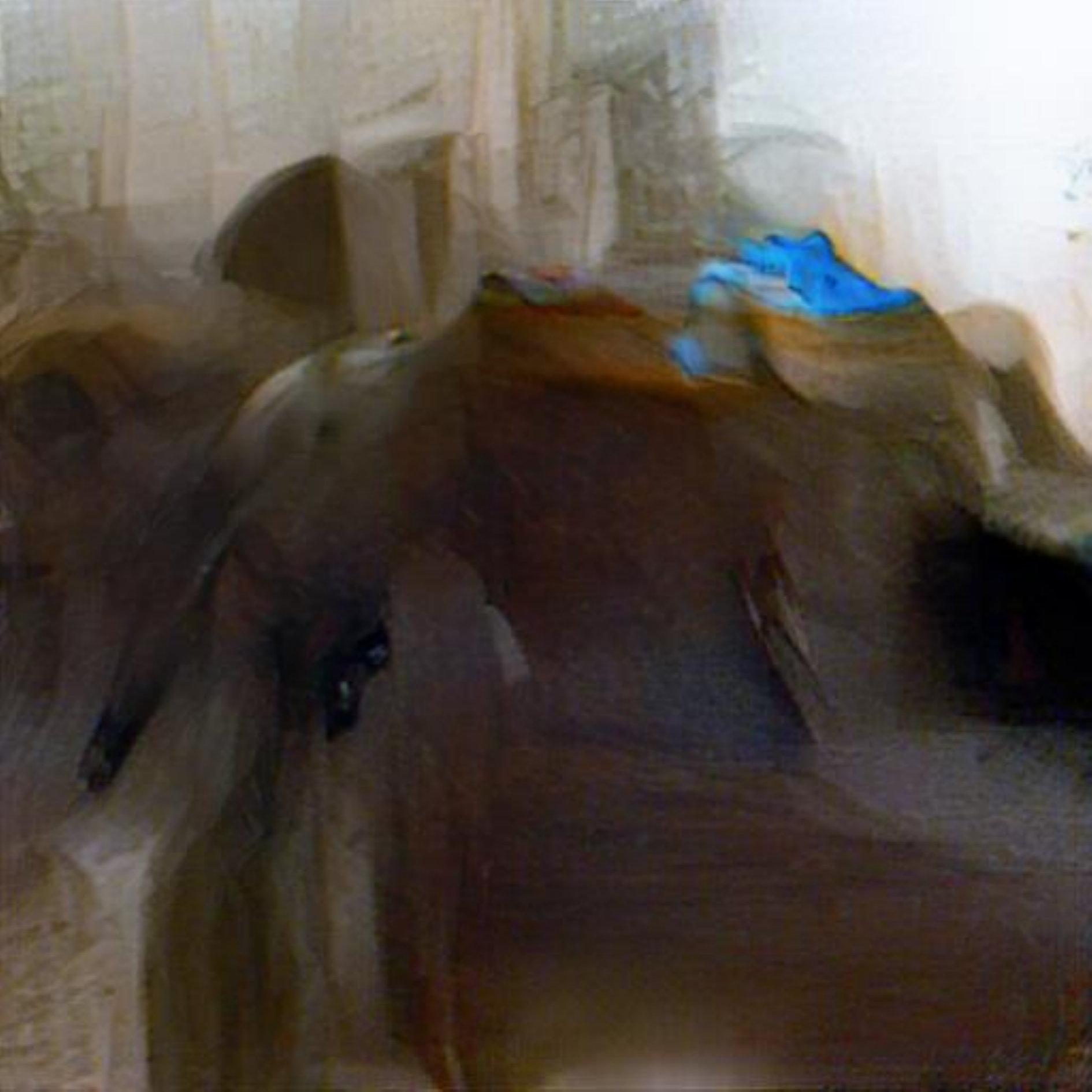}
    }
    \\
    \vspace{-3.5mm}
    \hspace{1.5mm}
    \subfigure{
        \includegraphics[width=0.13\linewidth]{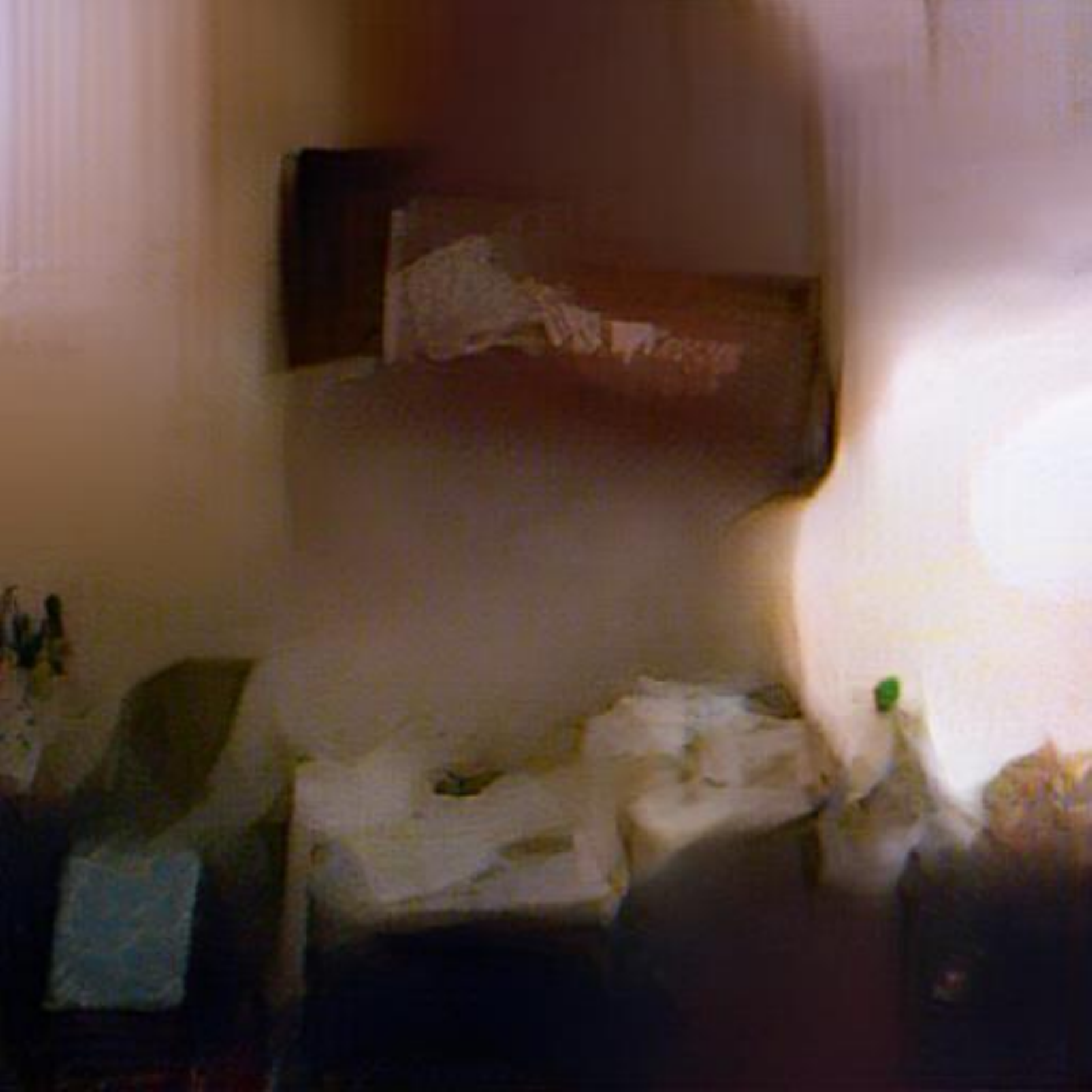}
    }
    \hspace{-3mm}
    \subfigure{
        \includegraphics[width=0.13\linewidth]{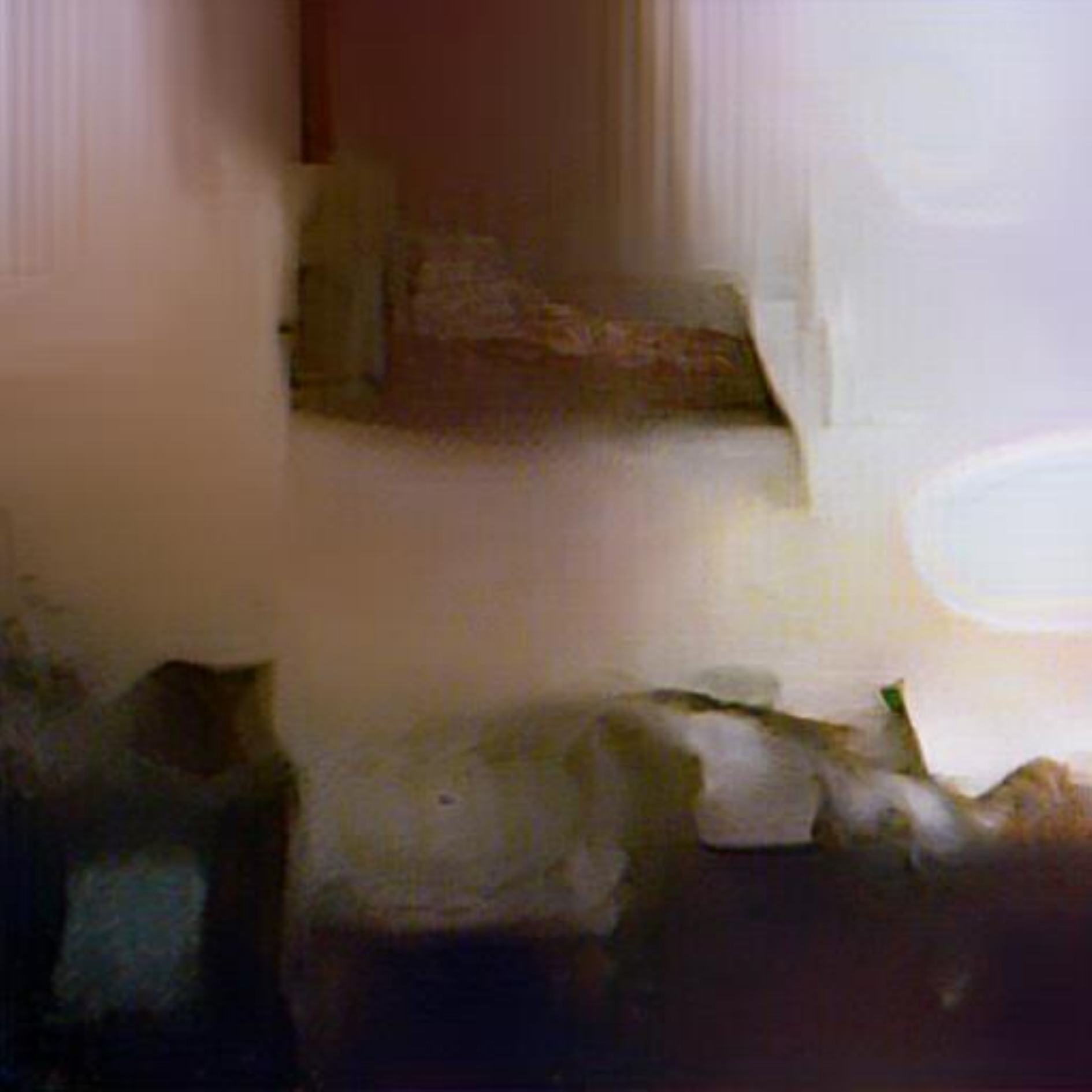}
    }
    \hspace{-3mm}
    \subfigure{
        \includegraphics[width=0.13\linewidth]{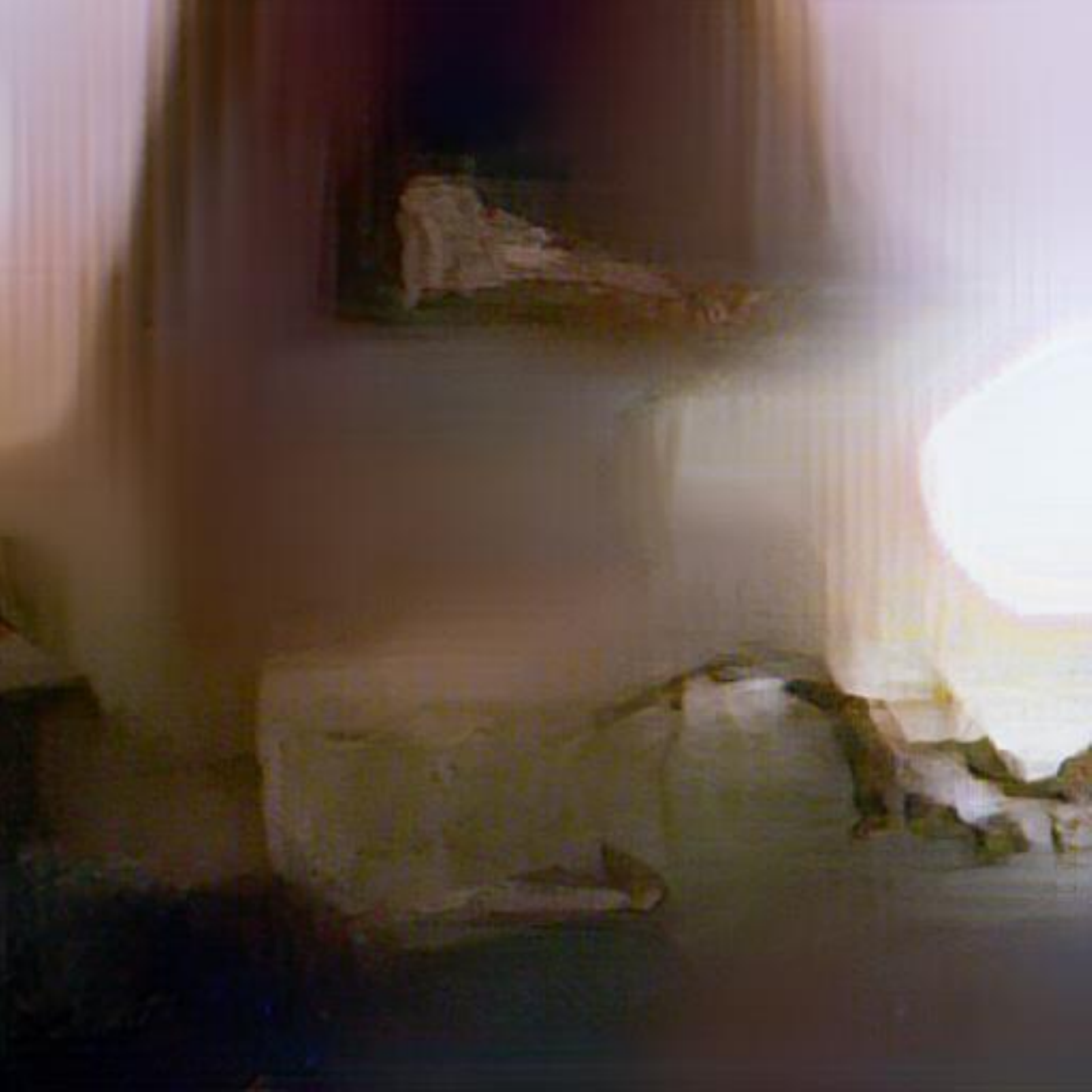}
    }
    \hspace{-3mm}
    \subfigure{
        \includegraphics[width=0.13\linewidth]{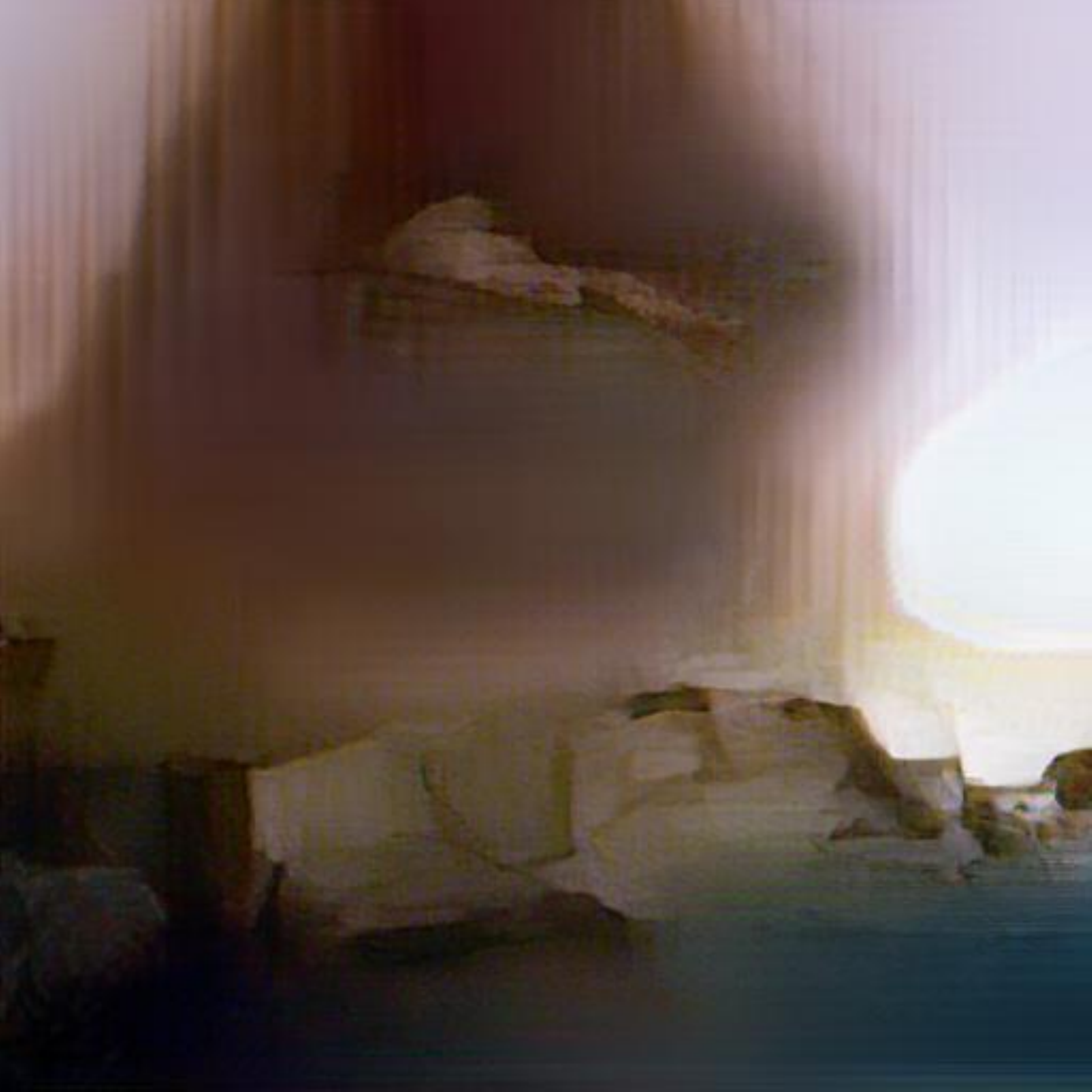}
    }
    \hspace{-3mm}
    \subfigure{
        \includegraphics[width=0.13\linewidth]{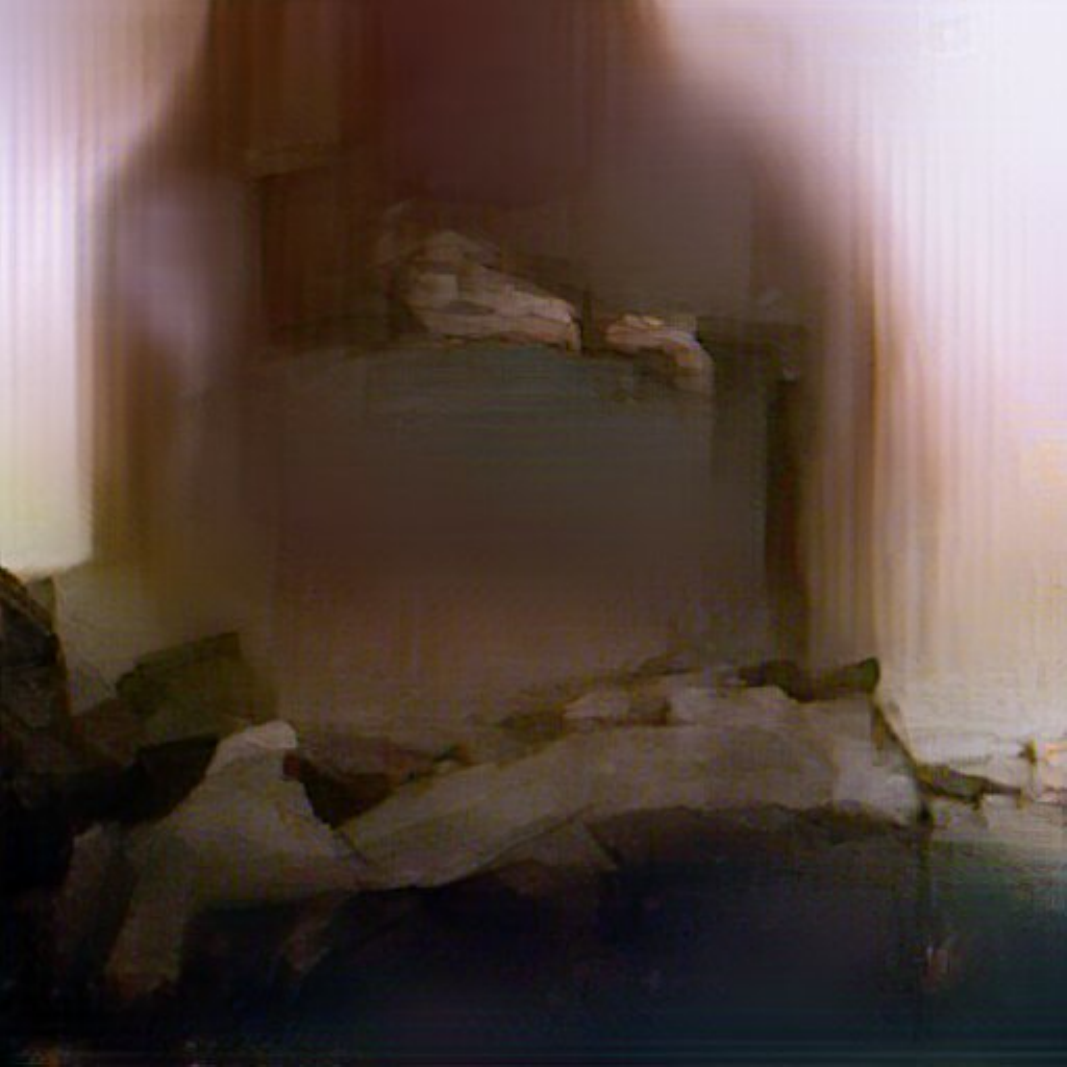}
    }
    \hspace{-3mm}
    \subfigure{
        \includegraphics[width=0.13\linewidth]{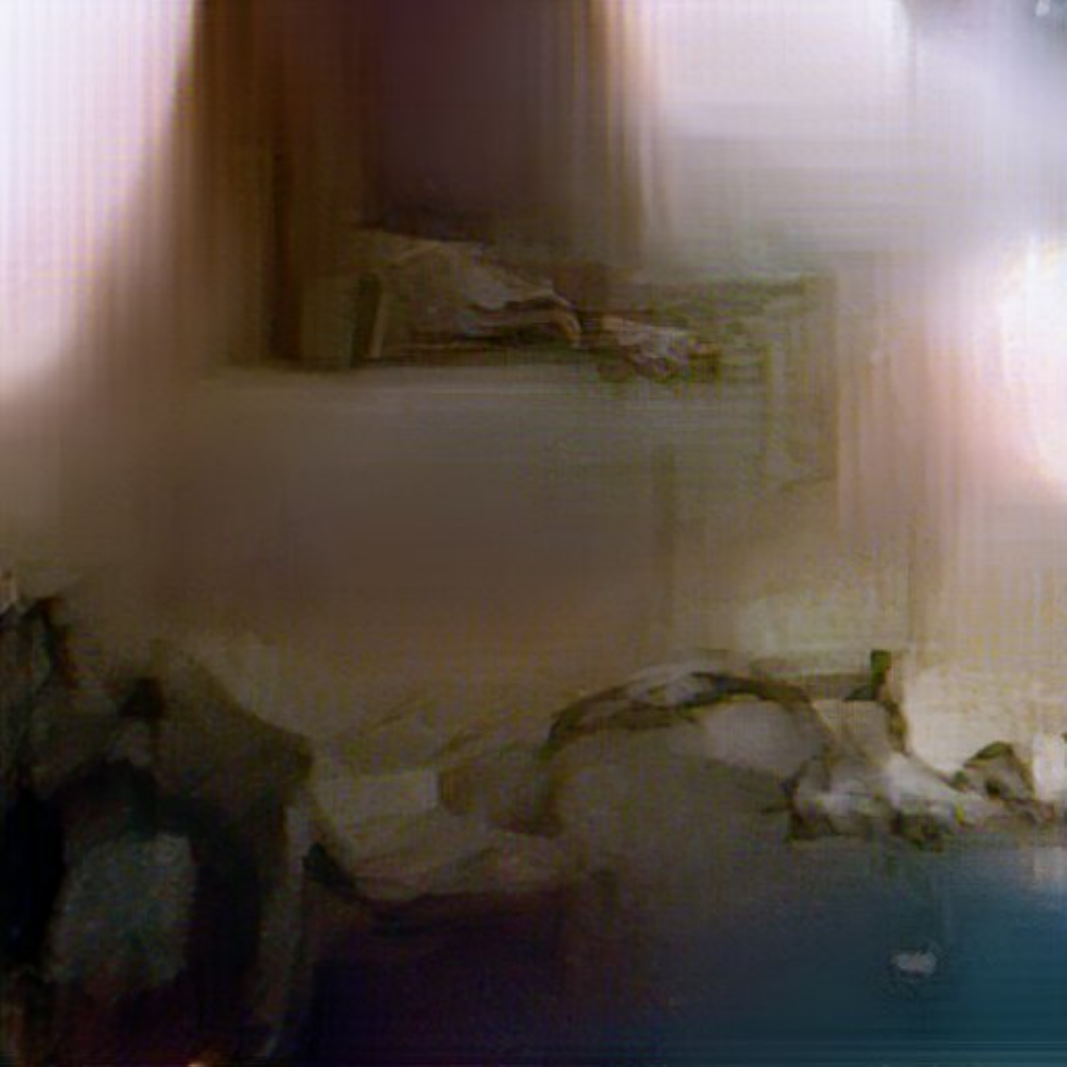}
    }
    \hspace{-3mm}
    \subfigure{
        \includegraphics[width=0.13\linewidth]{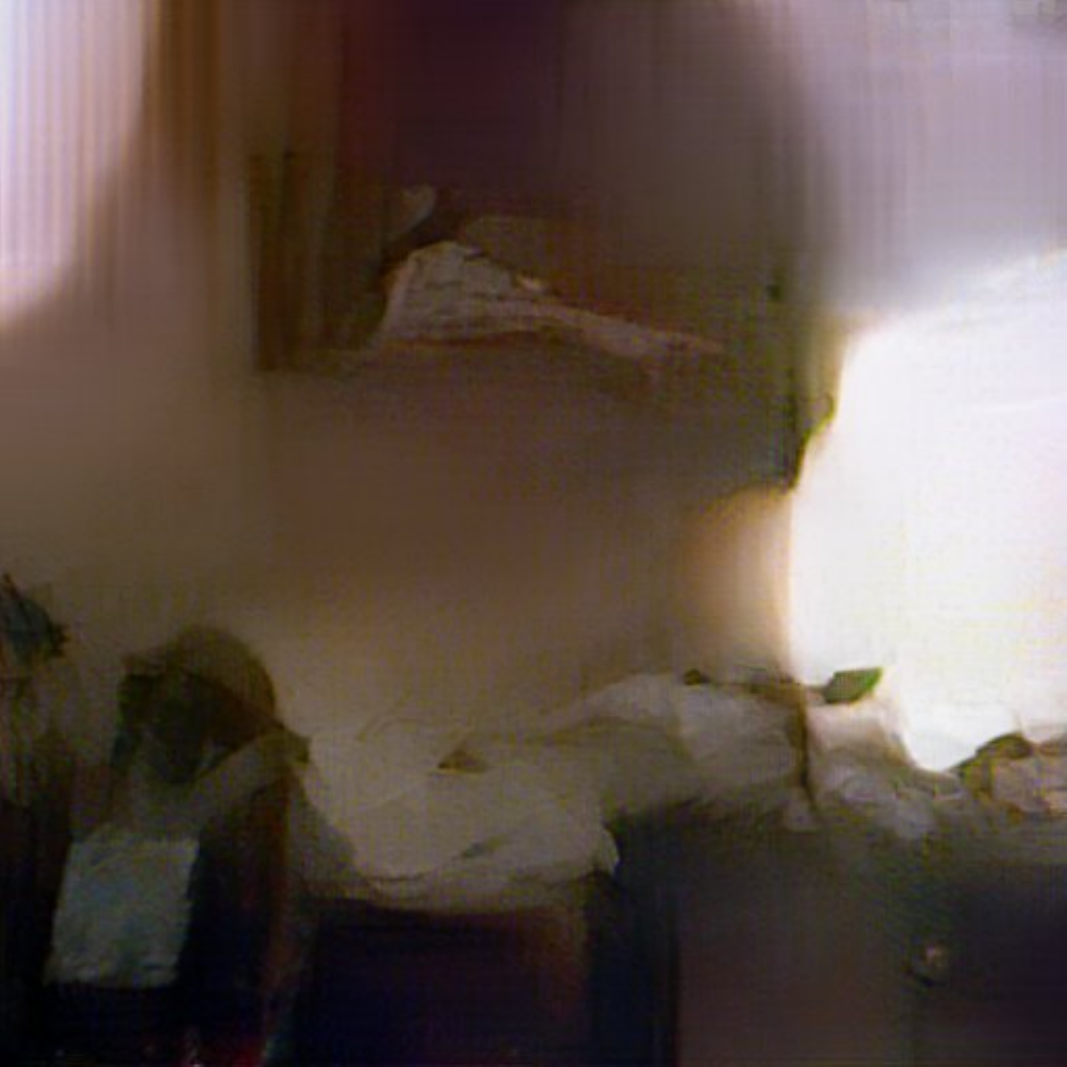}
    }
    \\
    \hspace{1.5mm}
    \subfigure{
        \includegraphics[width=0.13\linewidth]{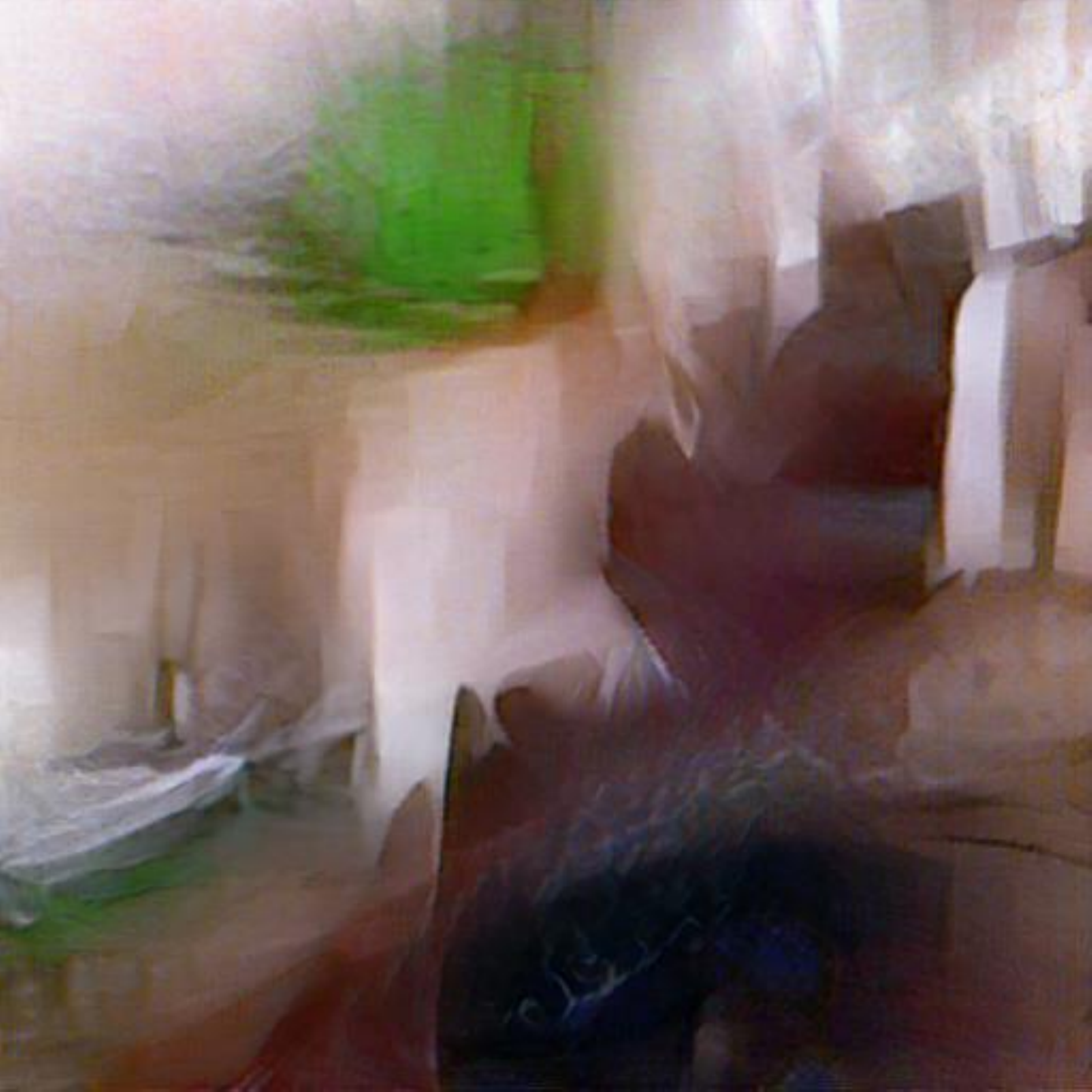}
    }
    \hspace{-3mm}
    \subfigure{
        \includegraphics[width=0.13\linewidth]{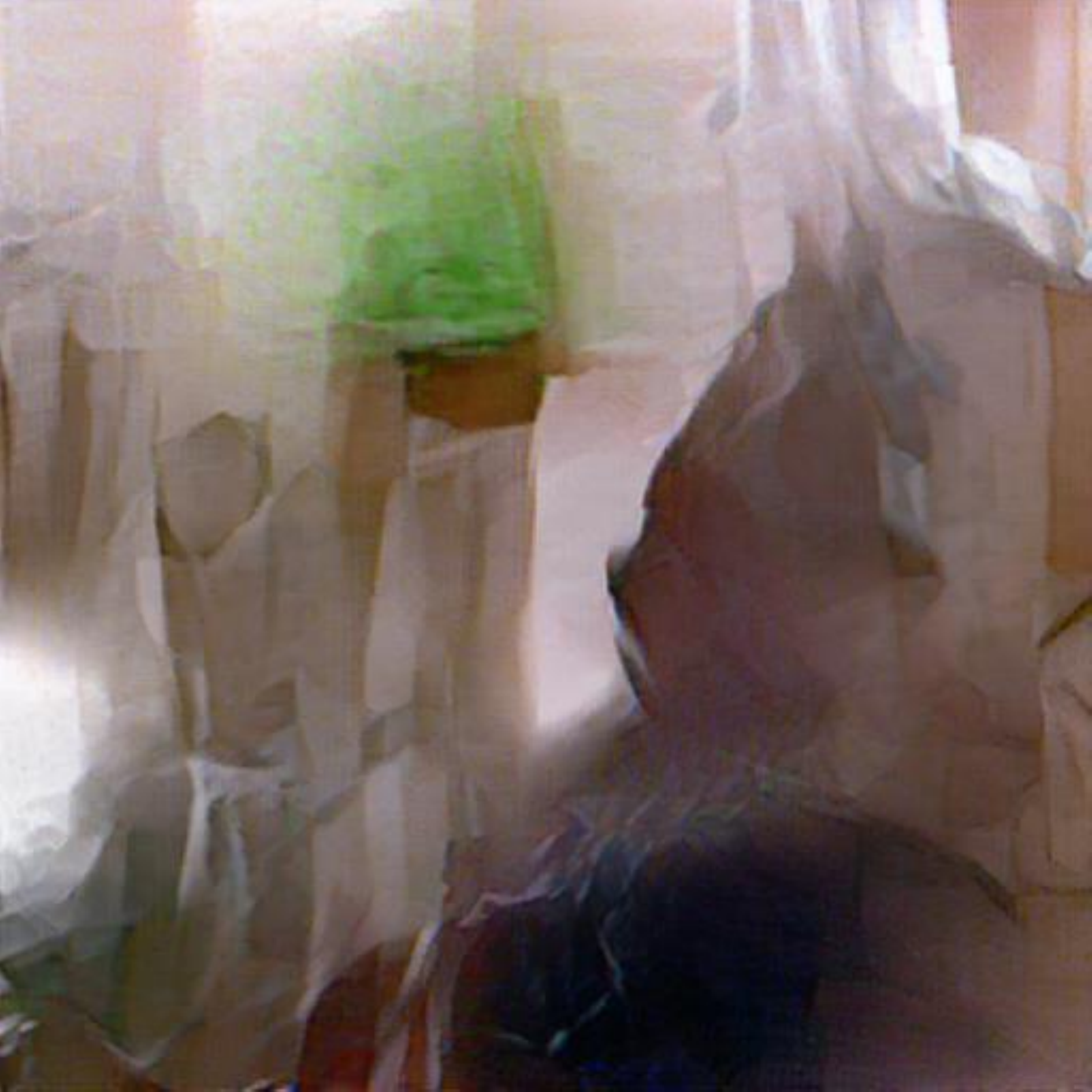}
    }
    \hspace{-3mm}
    \subfigure{
        \includegraphics[width=0.13\linewidth]{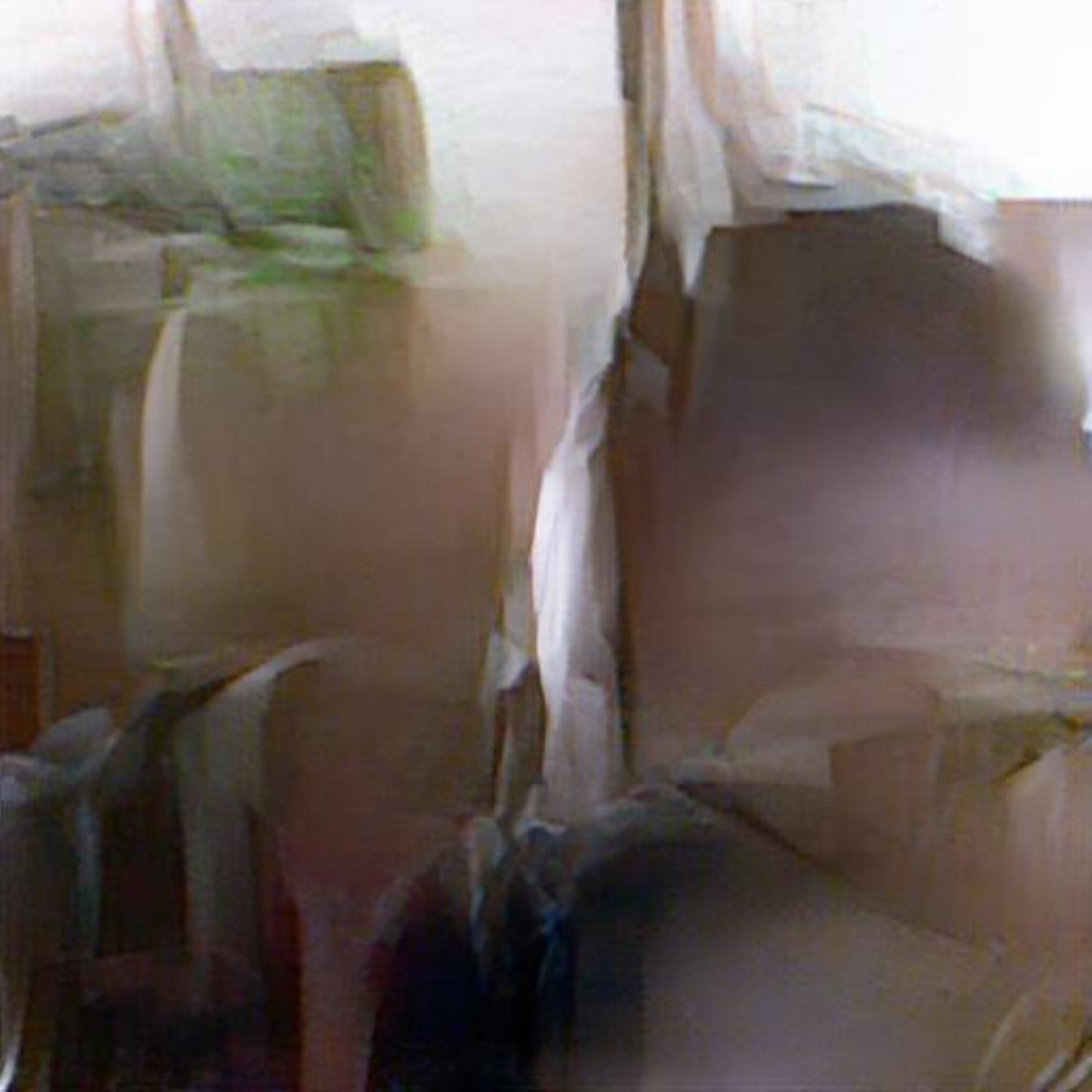}
    }
    \hspace{-3mm}
    \subfigure{
        \includegraphics[width=0.13\linewidth]{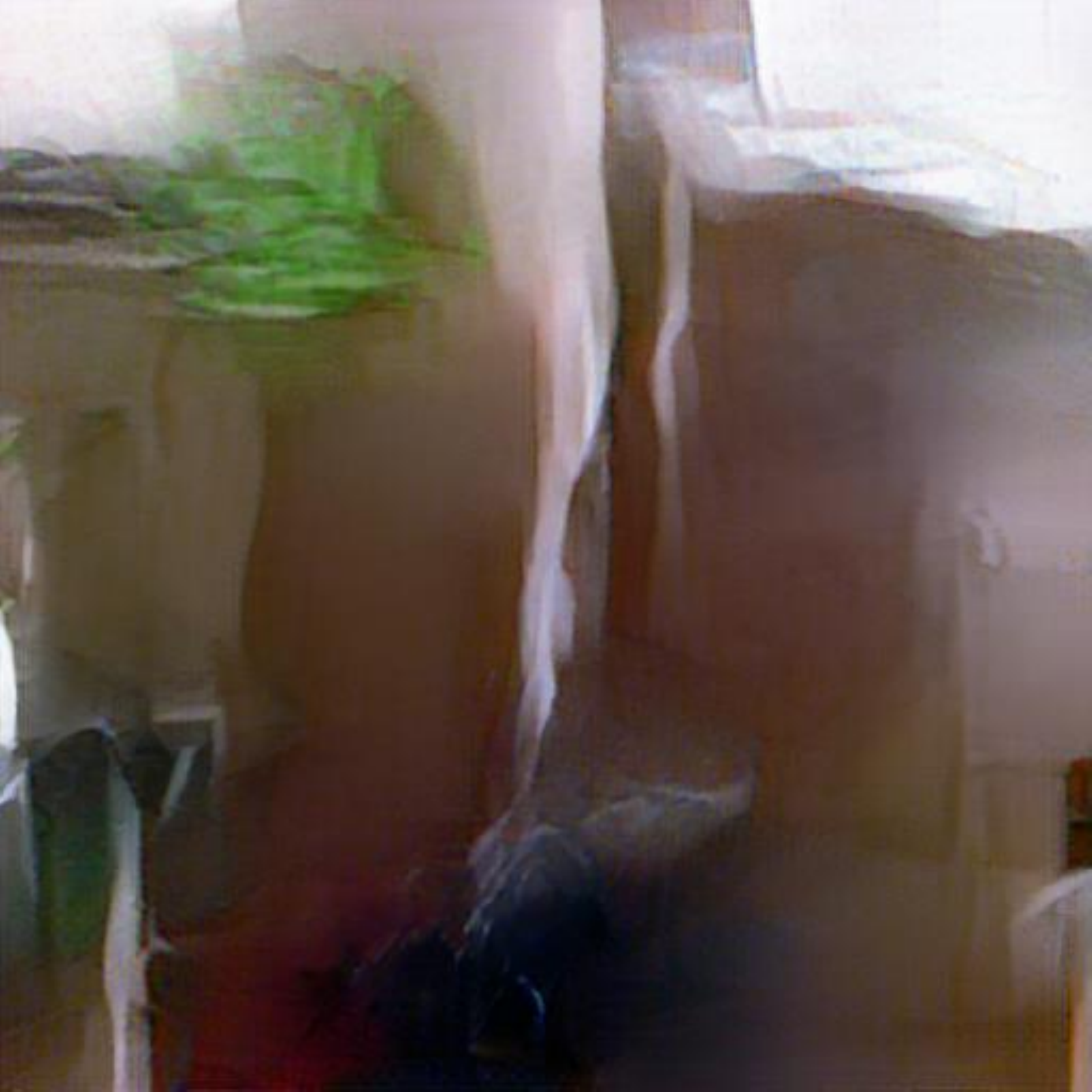}
    }
    \hspace{-3mm}
    \subfigure{
        \includegraphics[width=0.13\linewidth]{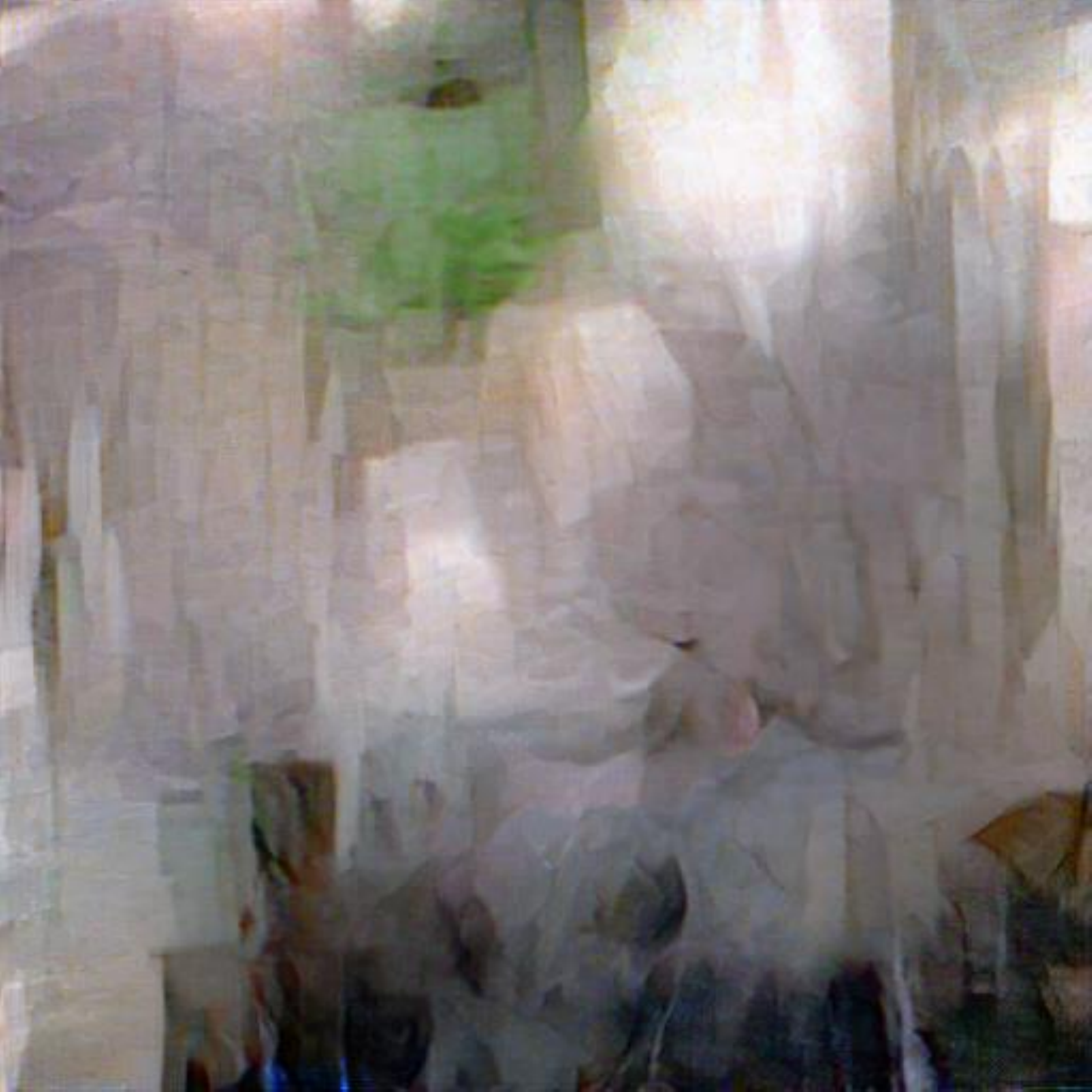}
    }
    \hspace{-3mm}
    \subfigure{
        \includegraphics[width=0.13\linewidth]{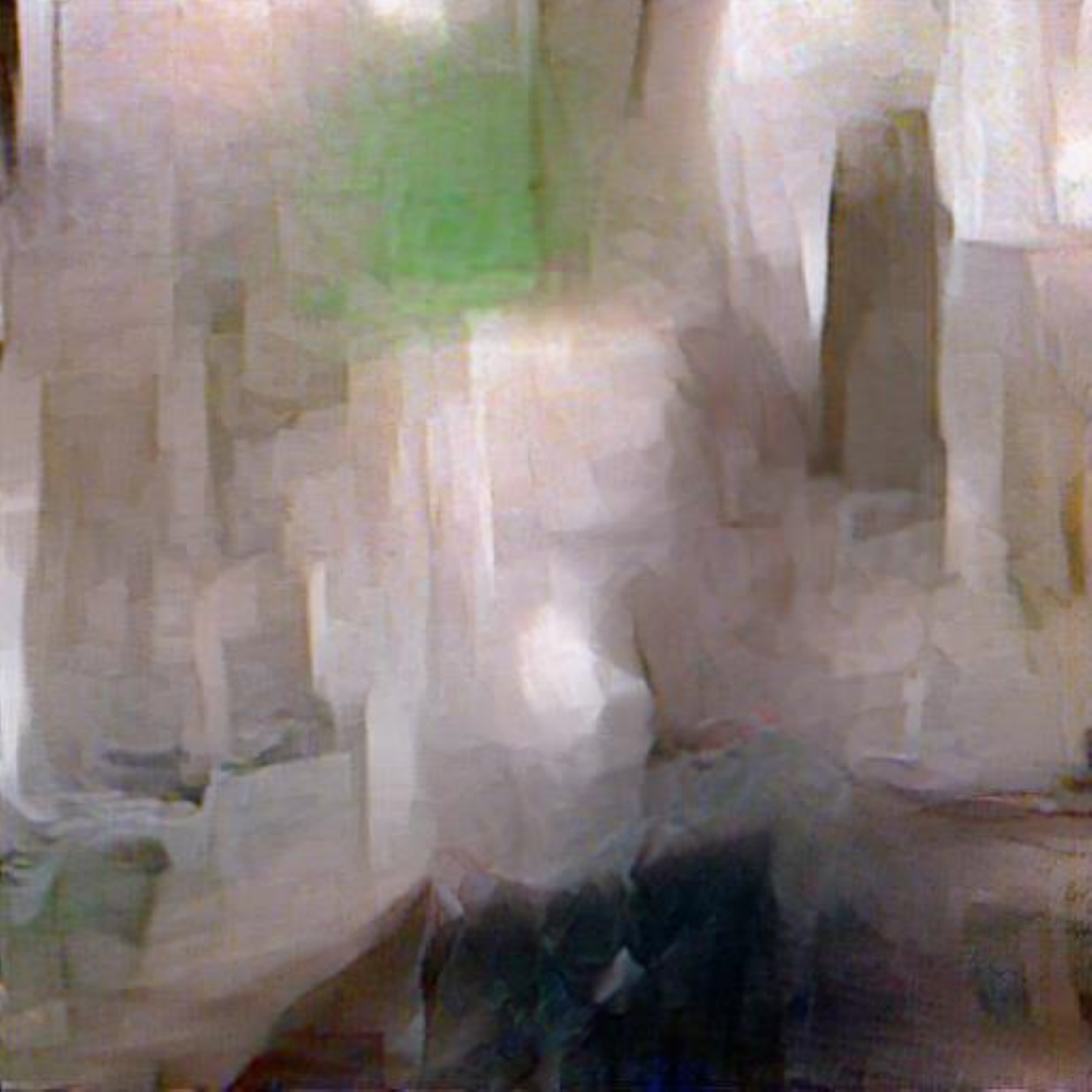}
    }
    \hspace{-3mm}
    \subfigure{
        \includegraphics[width=0.13\linewidth]{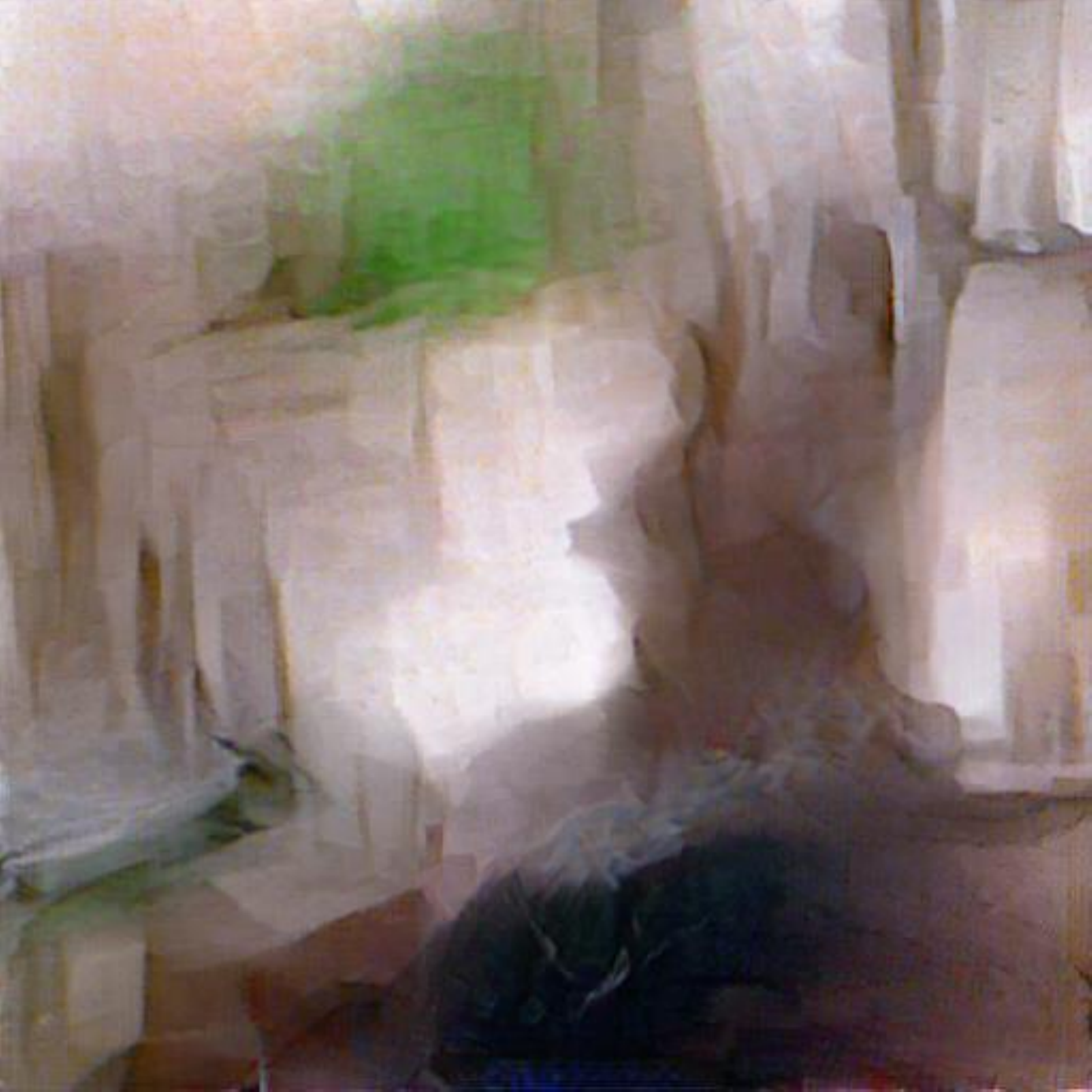}
    }
    \\
    \vspace{-3.5mm}
    \hspace{1.5mm}
    \subfigure{
        \includegraphics[width=0.13\linewidth]{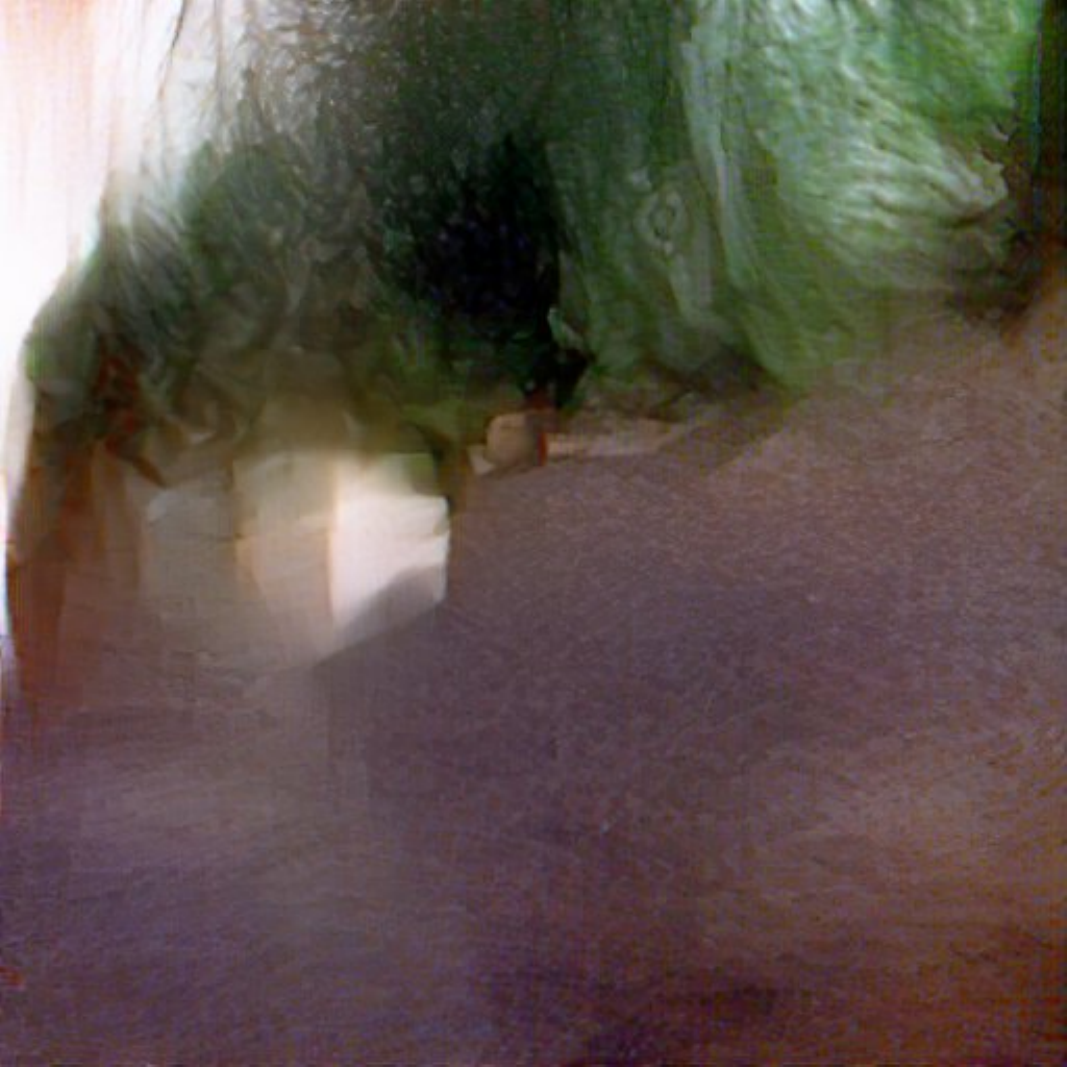}
    }
    \hspace{-3mm}
    \subfigure{
        \includegraphics[width=0.13\linewidth]{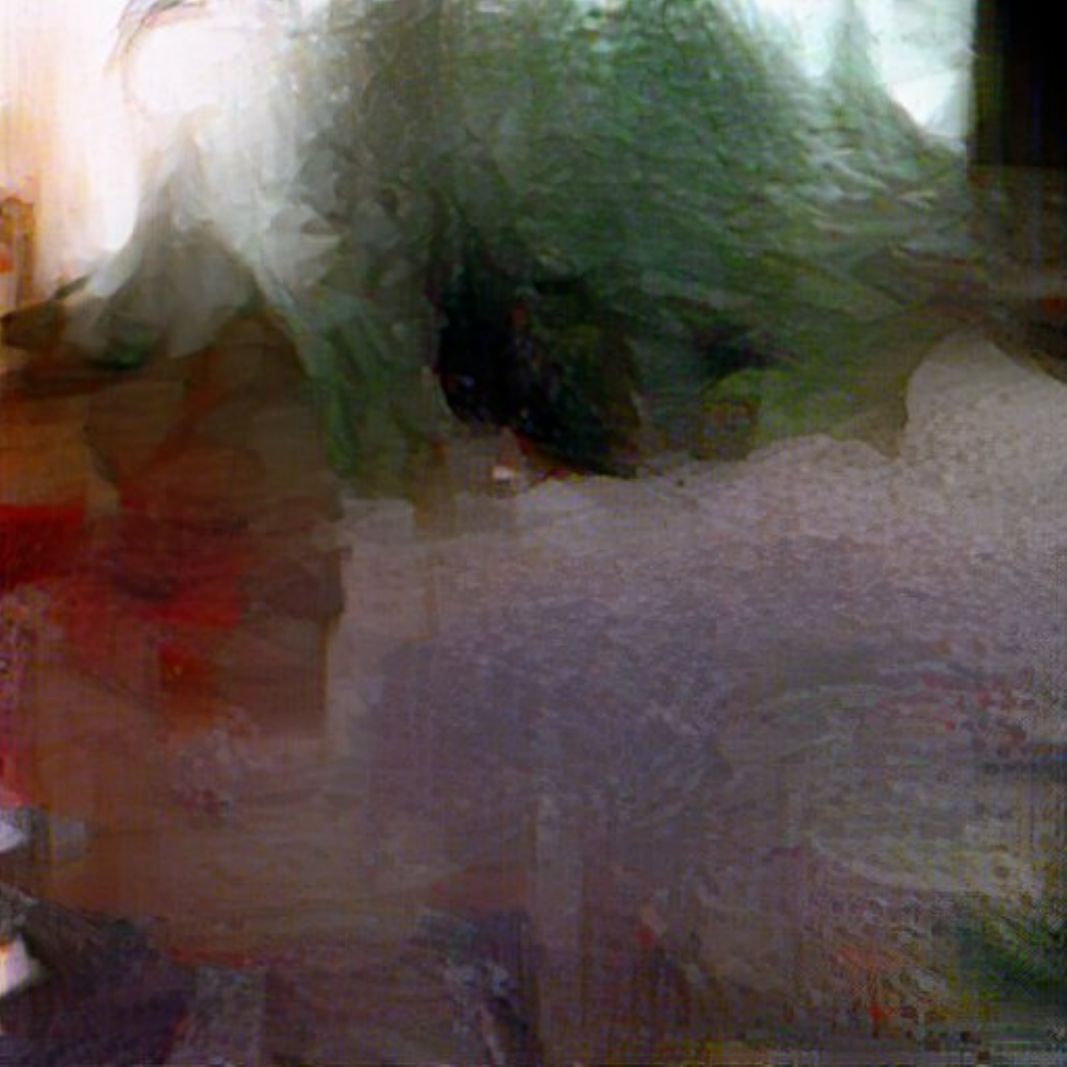}
    }
    \hspace{-3mm}
    \subfigure{
        \includegraphics[width=0.13\linewidth]{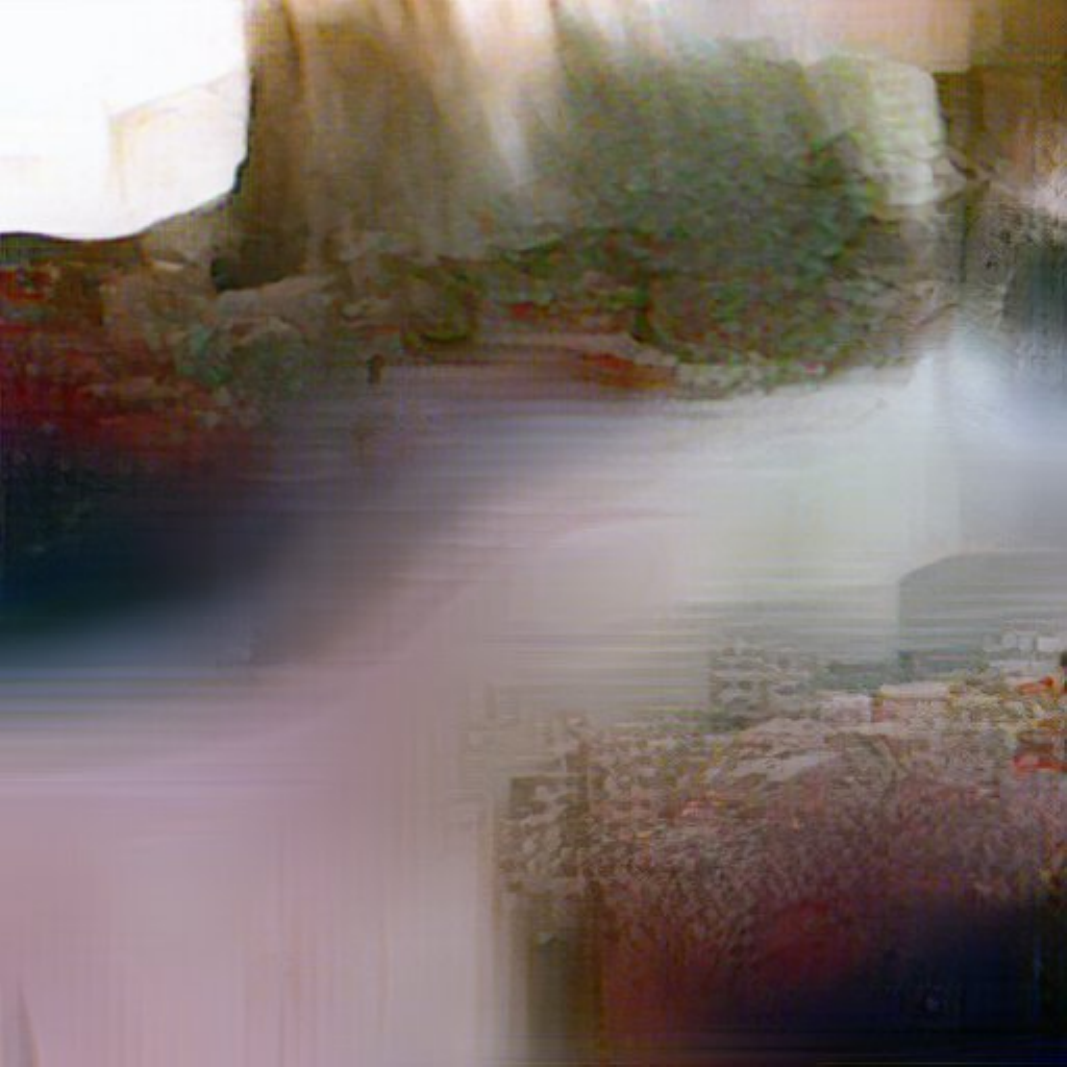}
    }
    \hspace{-3mm}
    \subfigure{
        \includegraphics[width=0.13\linewidth]{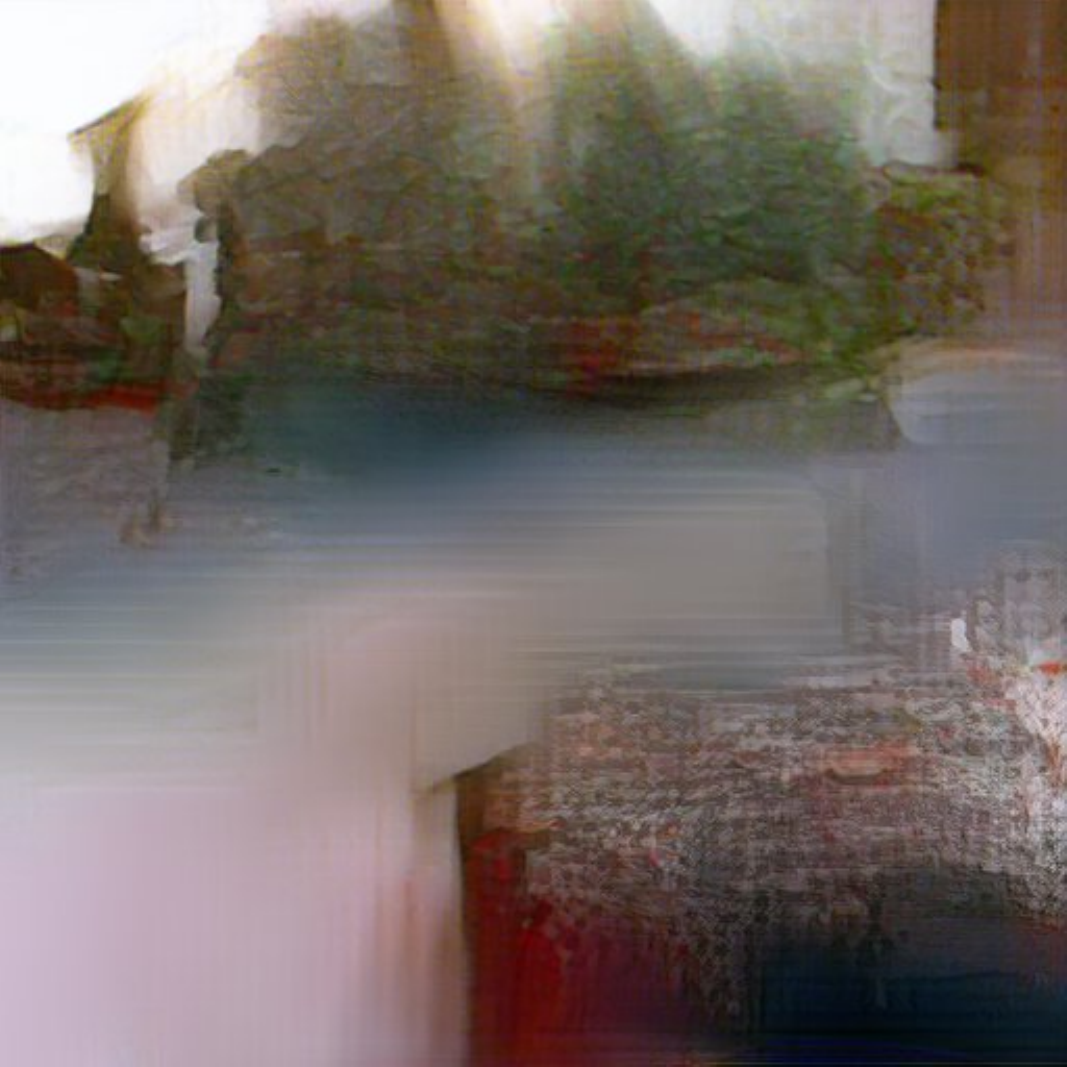}
    }
    \hspace{-3mm}
    \subfigure{
        \includegraphics[width=0.13\linewidth]{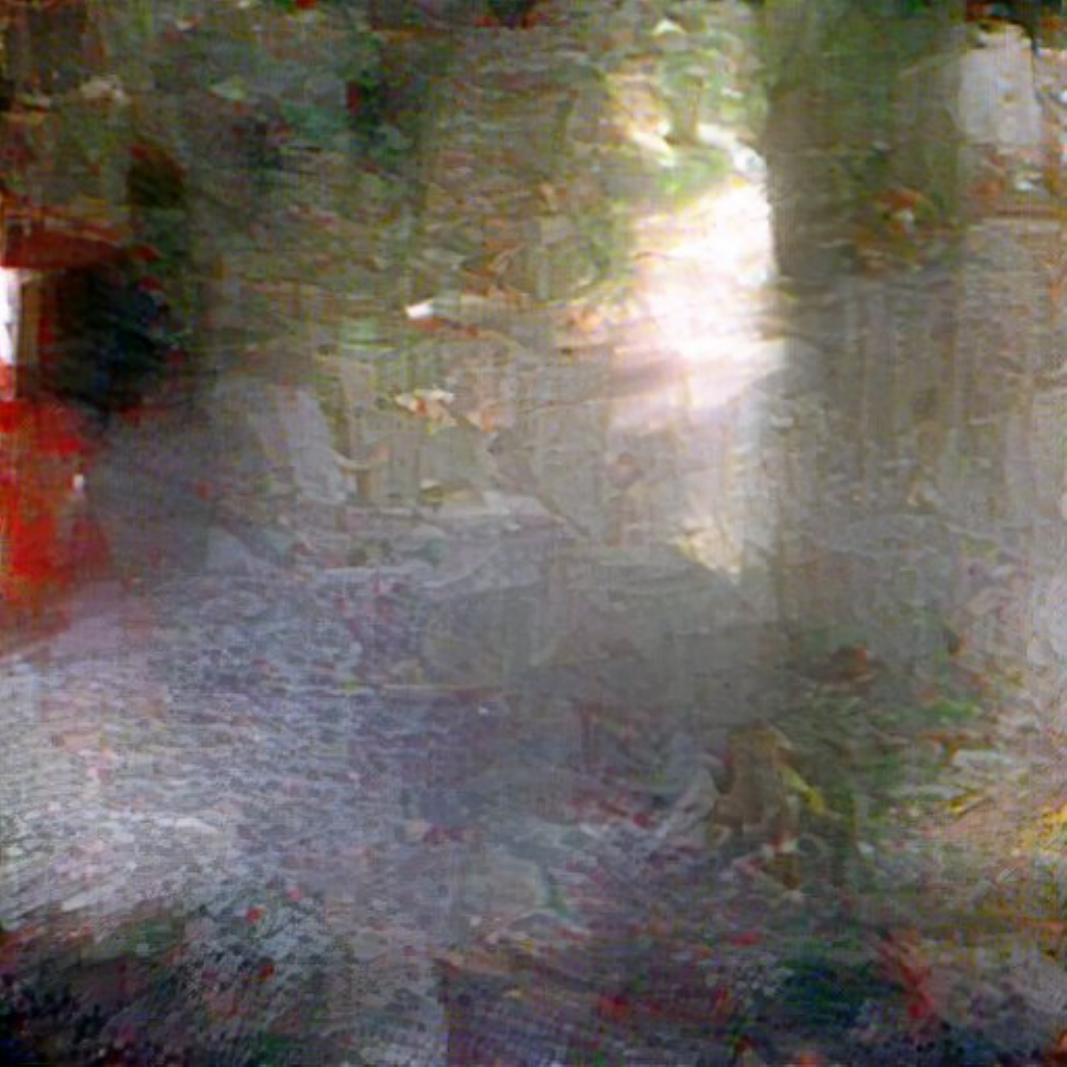}
    }
    \hspace{-3mm}
    \subfigure{
        \includegraphics[width=0.13\linewidth]{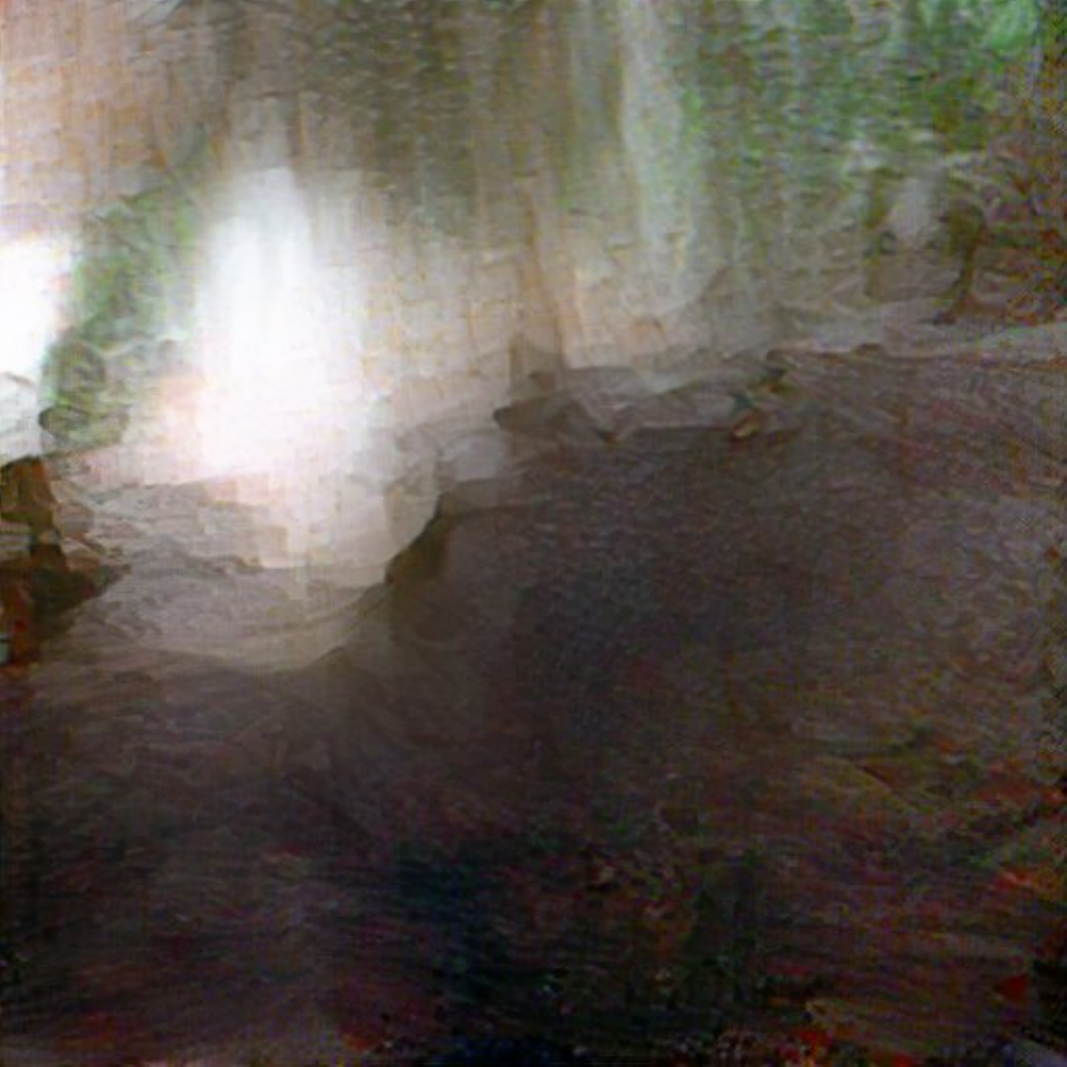}
    }
    \hspace{-3mm}
    \subfigure{
        \includegraphics[width=0.13\linewidth]{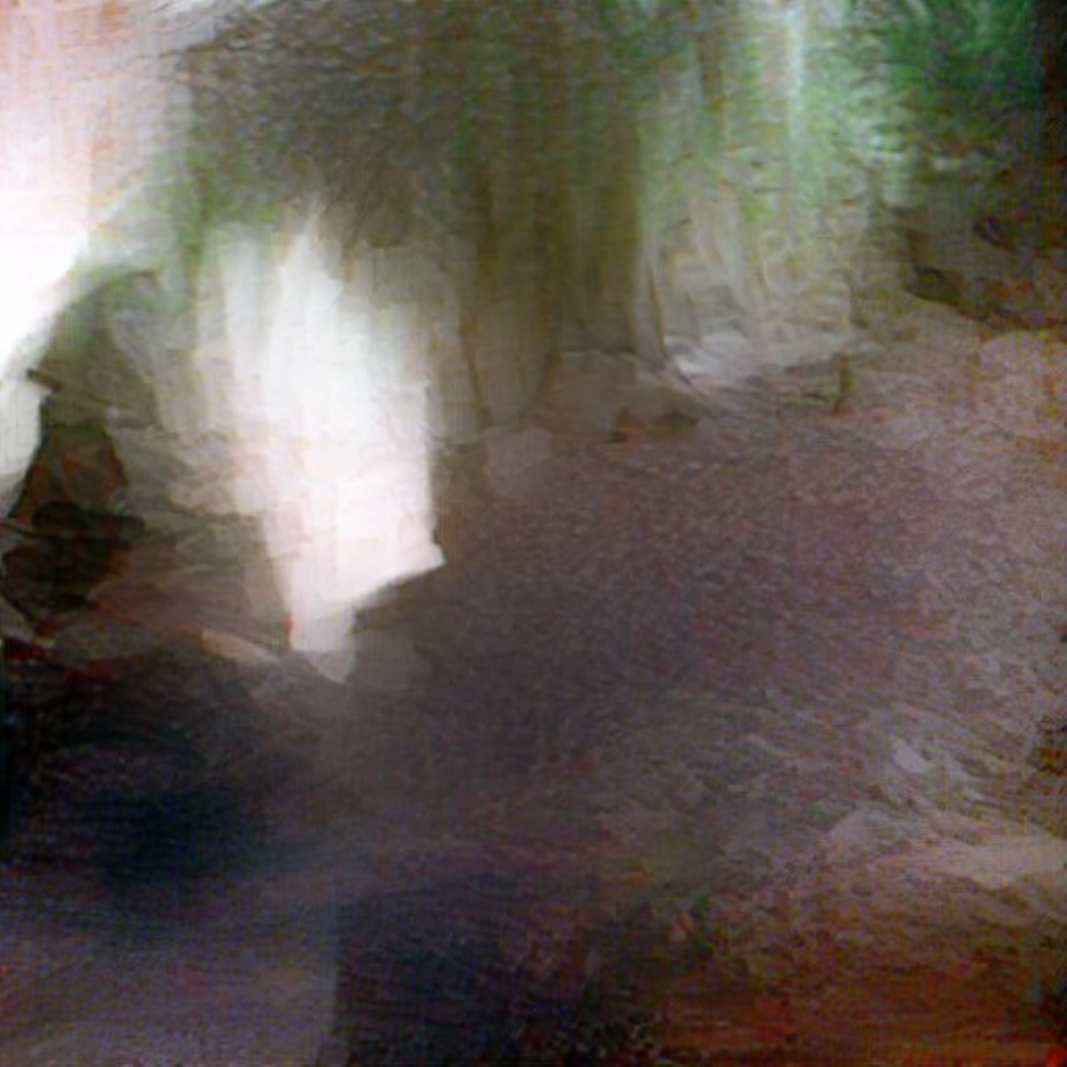}
    }
    \\
    \vspace{-3.5mm}
    \hspace{1.5mm}
    \subfigure{
        \includegraphics[width=0.13\linewidth]{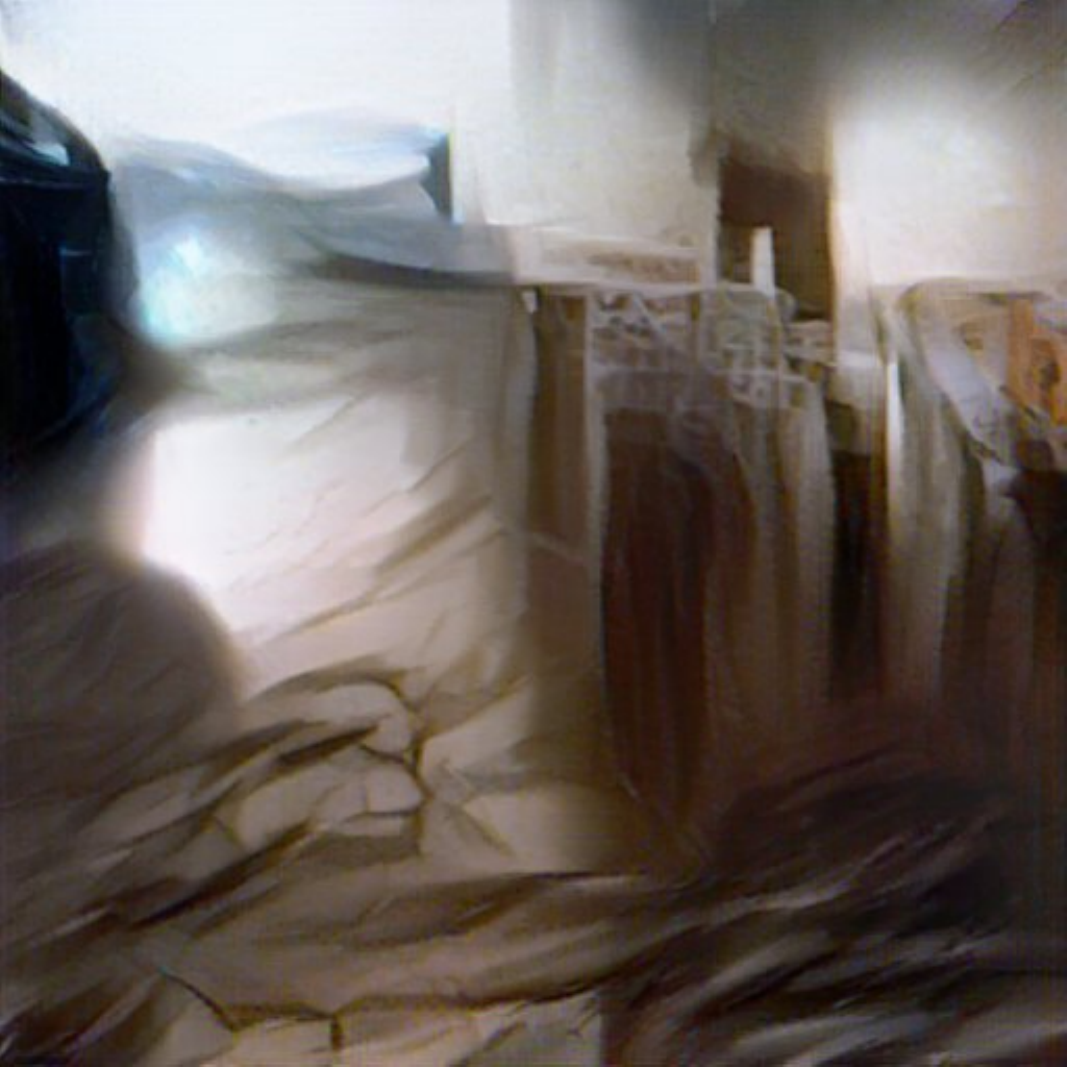}
    }
    \hspace{-3mm}
    \subfigure{
        \includegraphics[width=0.13\linewidth]{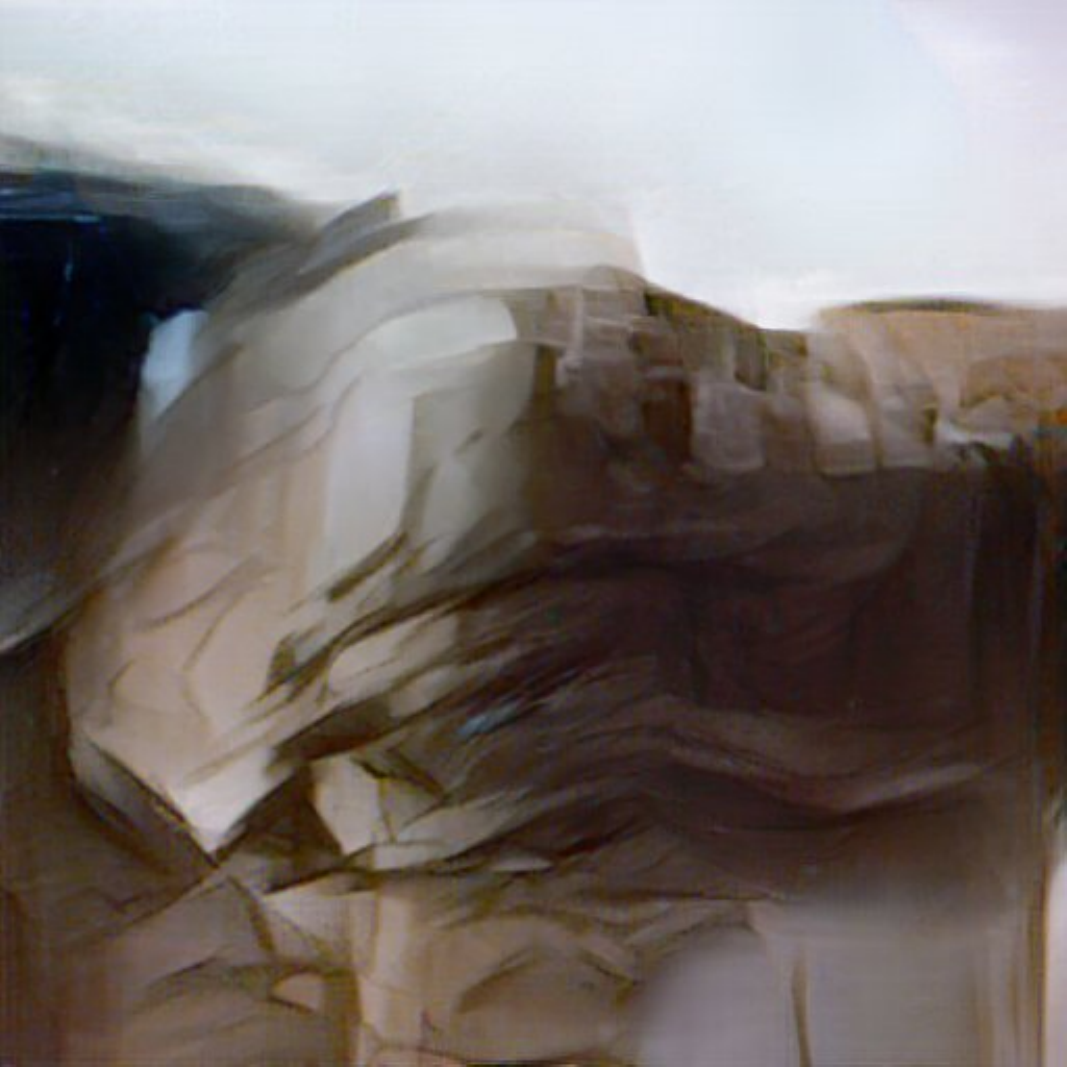}
    }
    \hspace{-3mm}
    \subfigure{
        \includegraphics[width=0.13\linewidth]{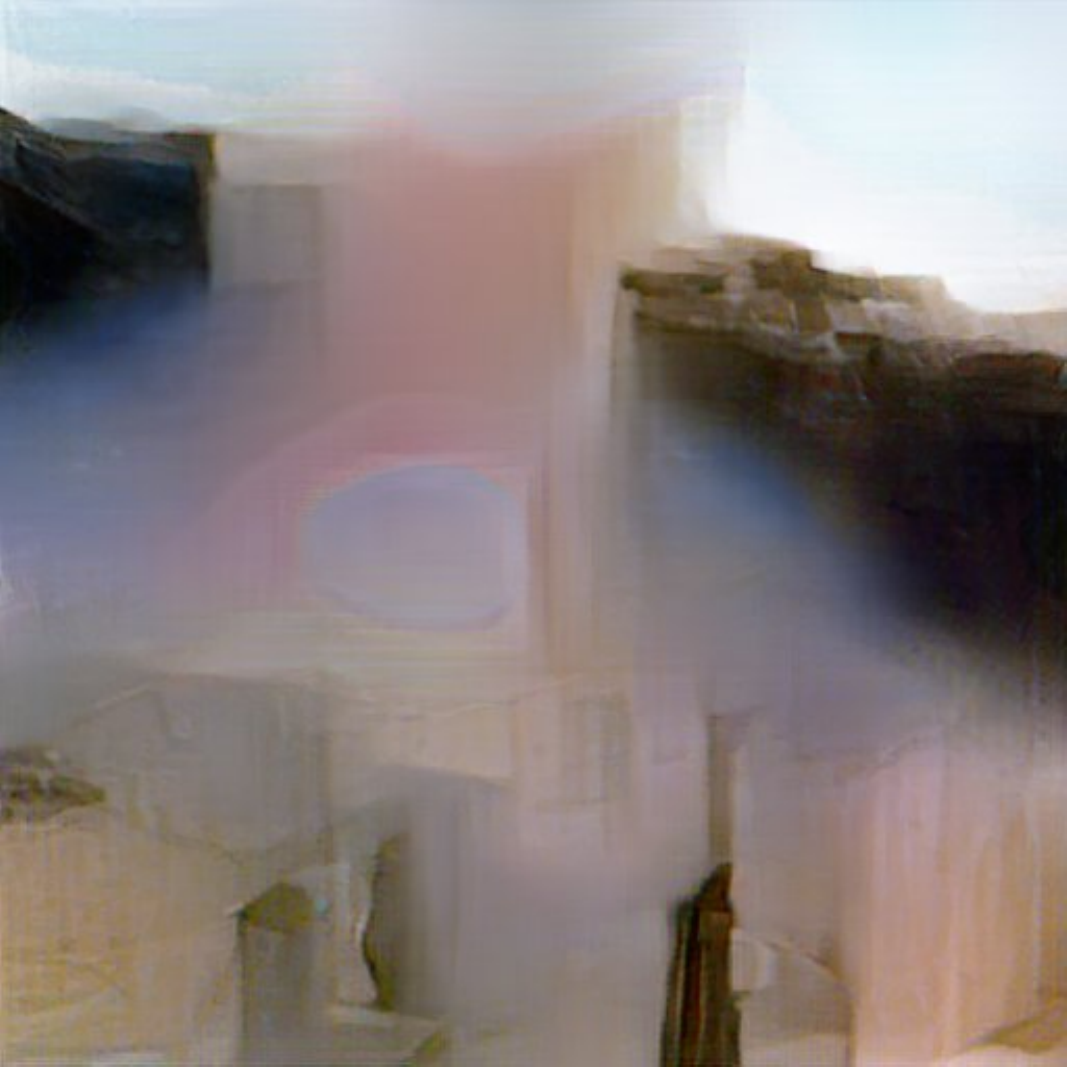}
    }
    \hspace{-3mm}
    \subfigure{
        \includegraphics[width=0.13\linewidth]{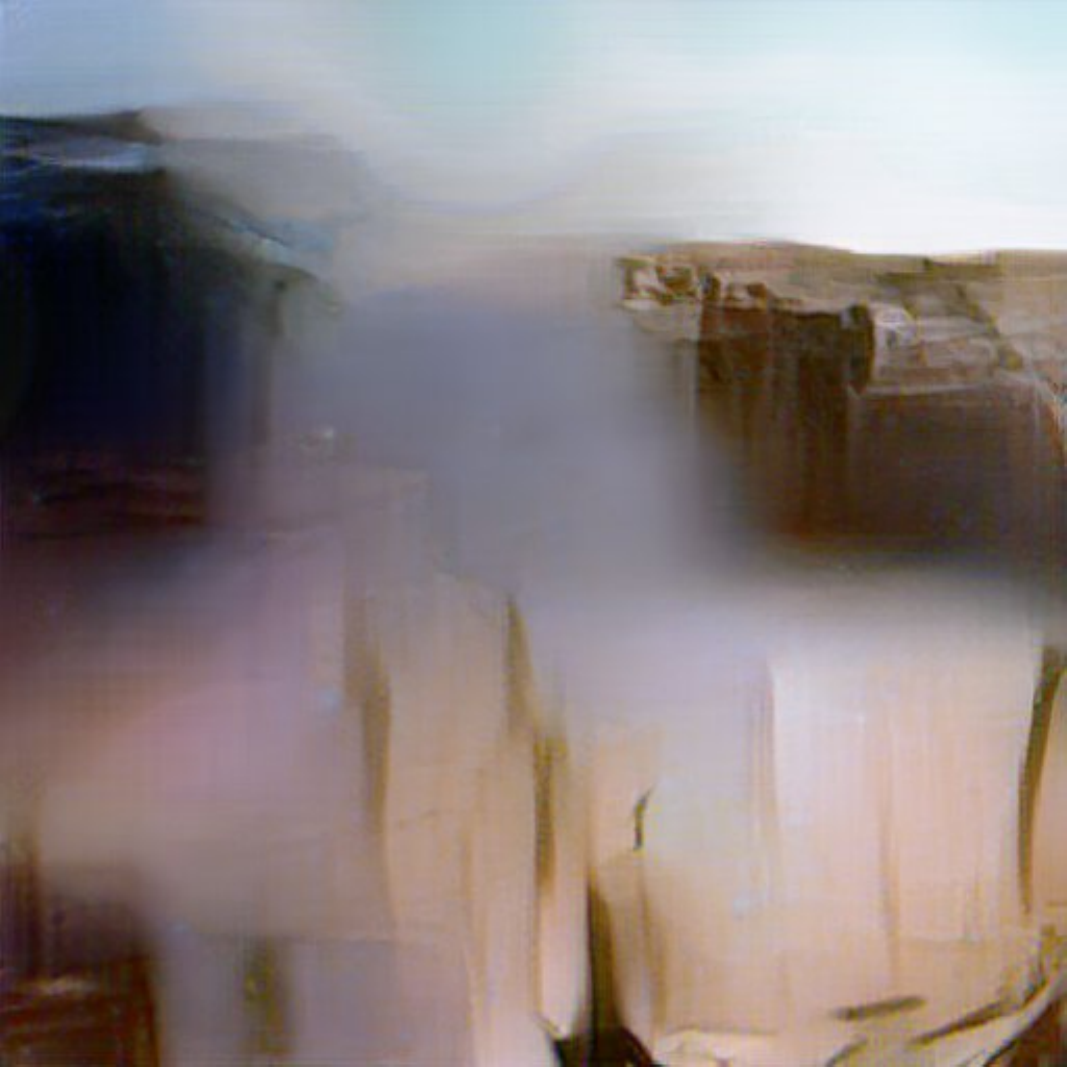}
    }
    \hspace{-3mm}
    \subfigure{
        \includegraphics[width=0.13\linewidth]{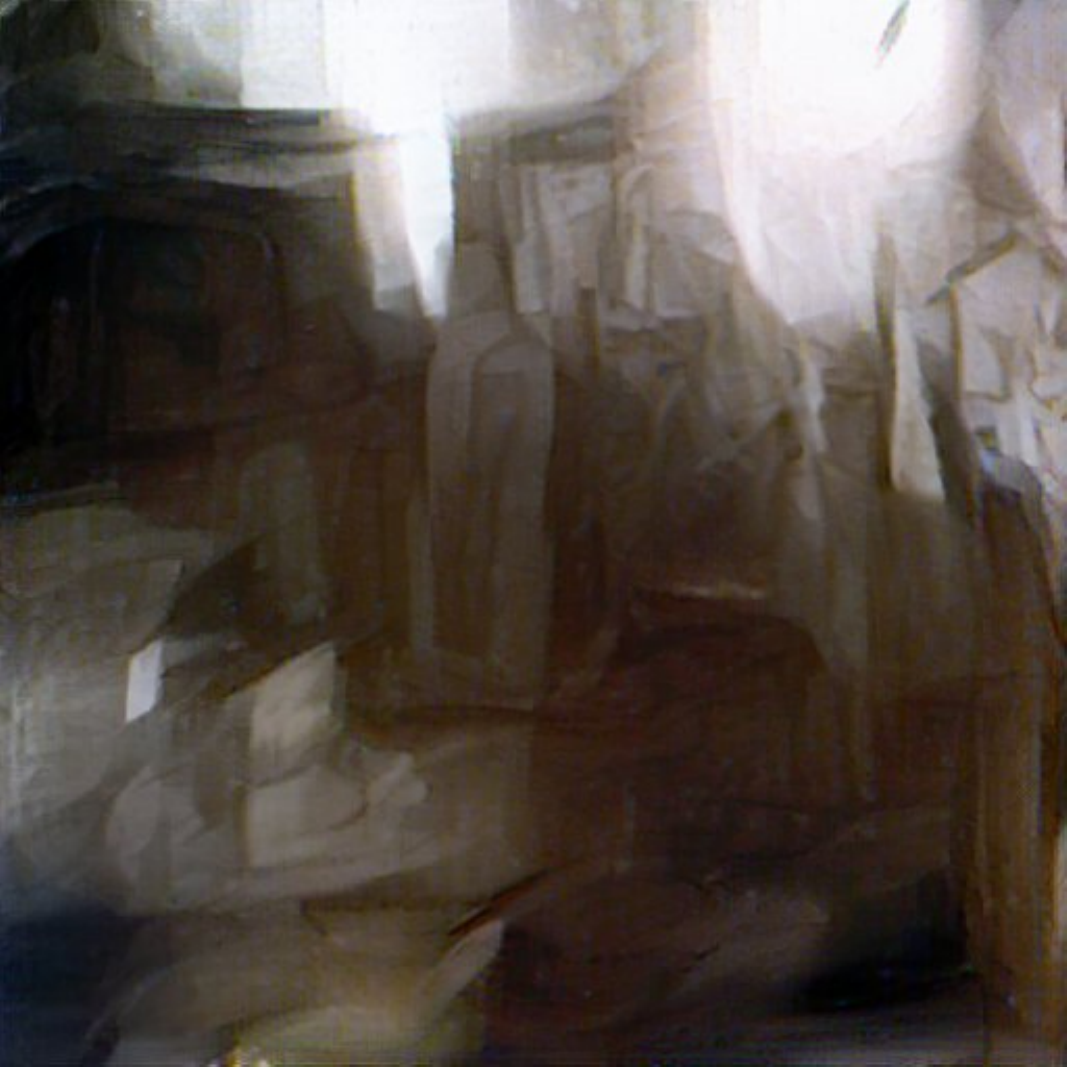}
    }
    \hspace{-3mm}
    \subfigure{
        \includegraphics[width=0.13\linewidth]{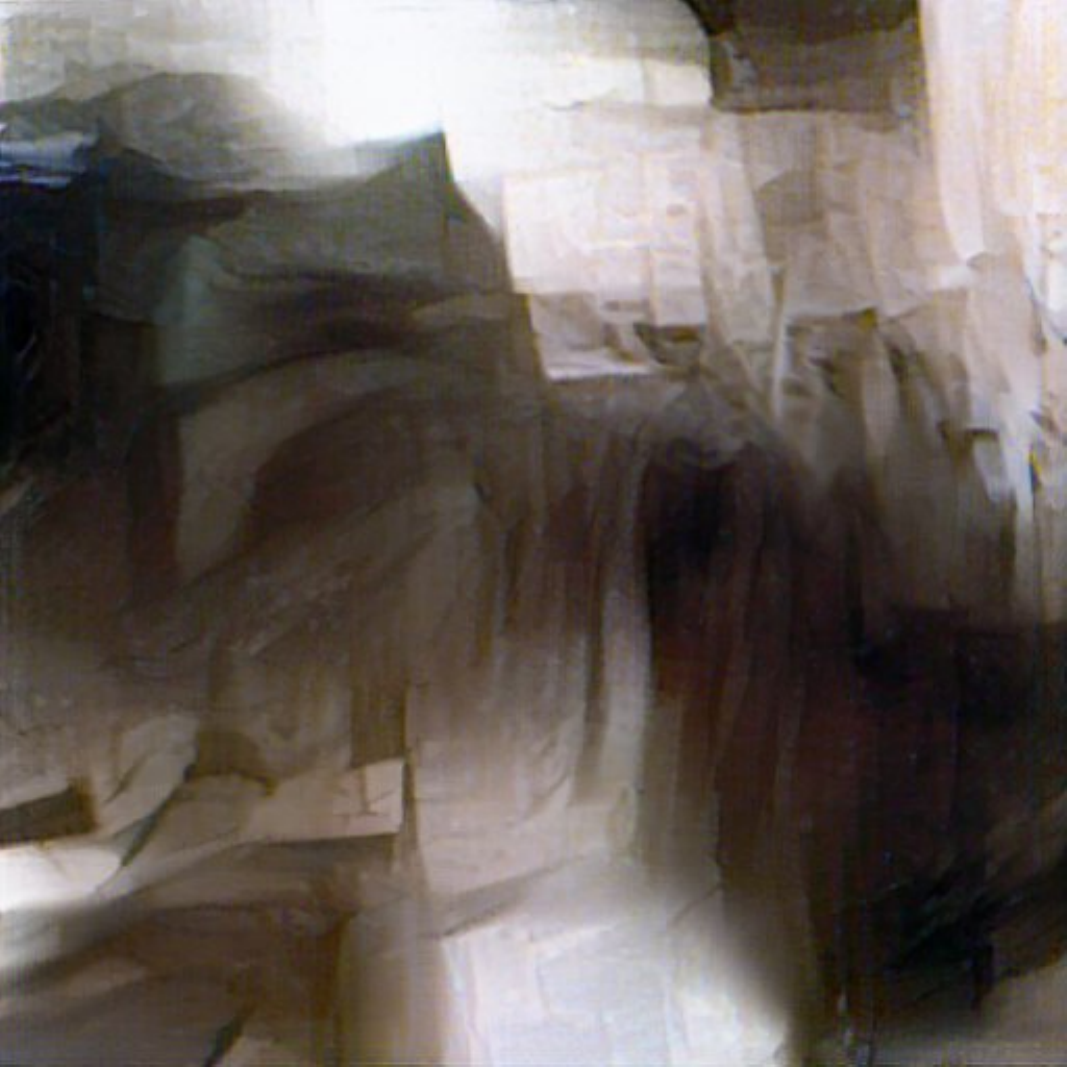}
    }
    \hspace{-3mm}
    \subfigure{
        \includegraphics[width=0.13\linewidth]{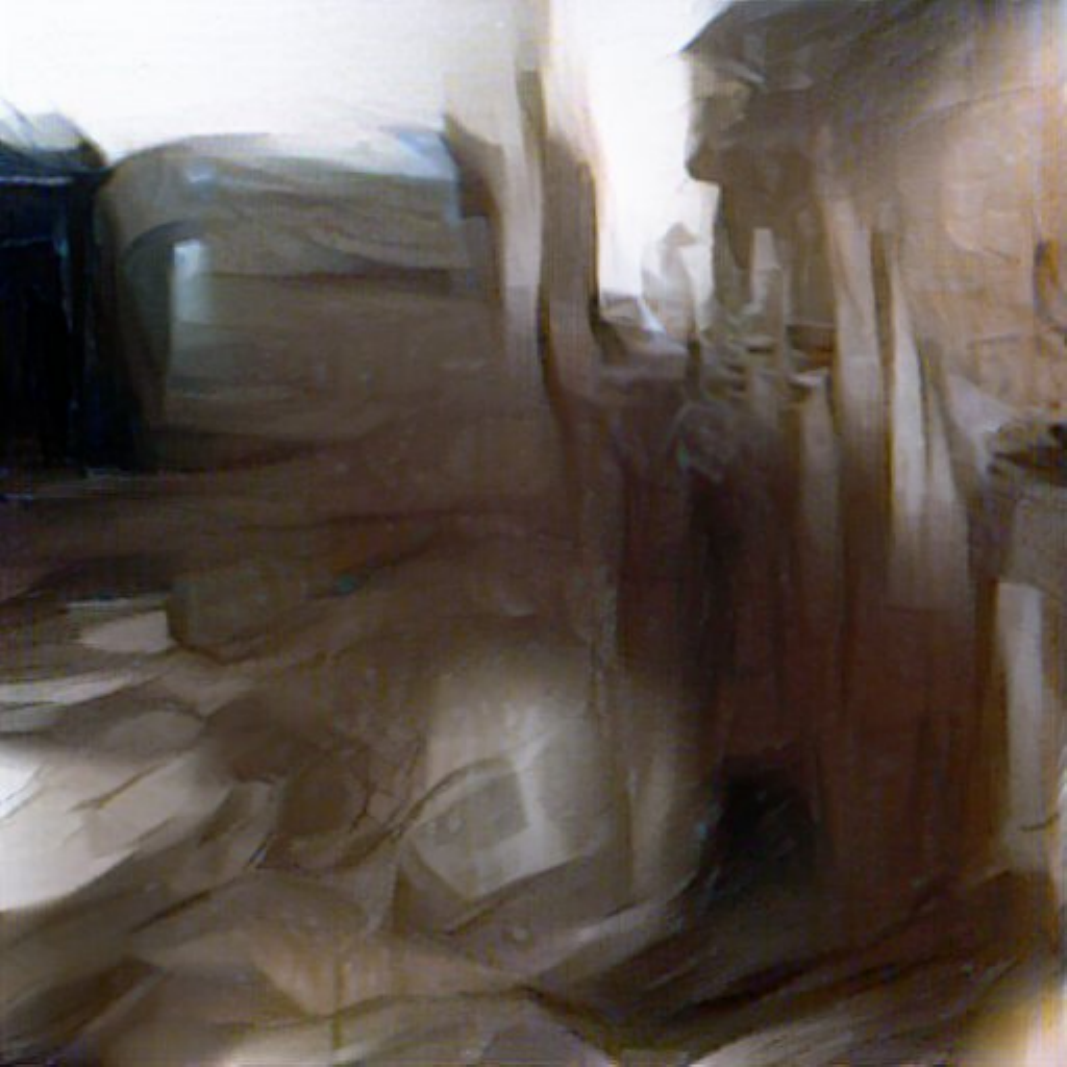}
    }
    \\
    \vspace{-3.5mm}
    \setcounter{subfigure}{0}
    \subfigure[Pseudo-GT~\figlabel{fig:pseudoGT}]{
        \includegraphics[width=0.13\linewidth]{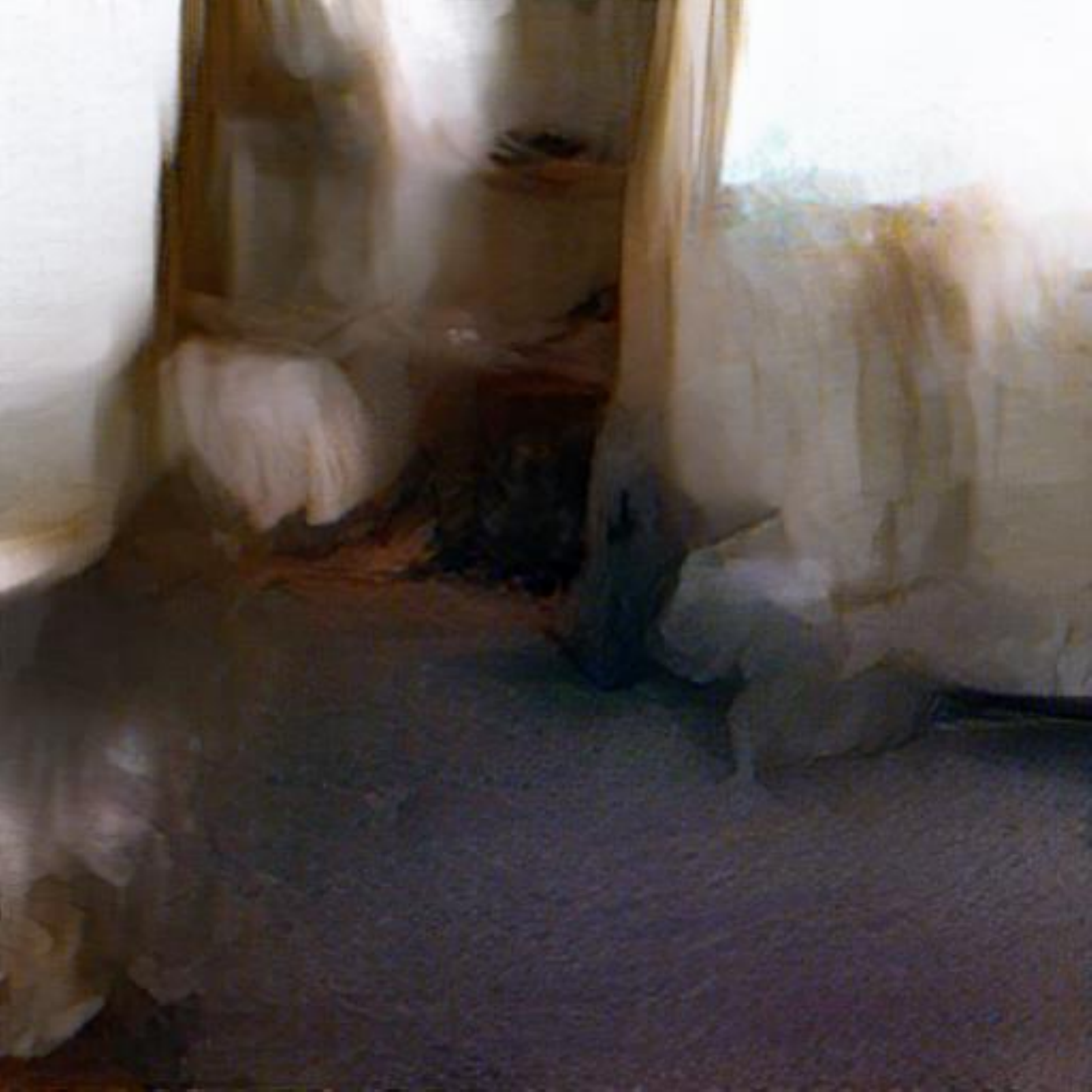}
    }
    \hspace{-3mm}
    \subfigure[ULC~\cite{speciale2019privacy}]{
        \includegraphics[width=0.13\linewidth]{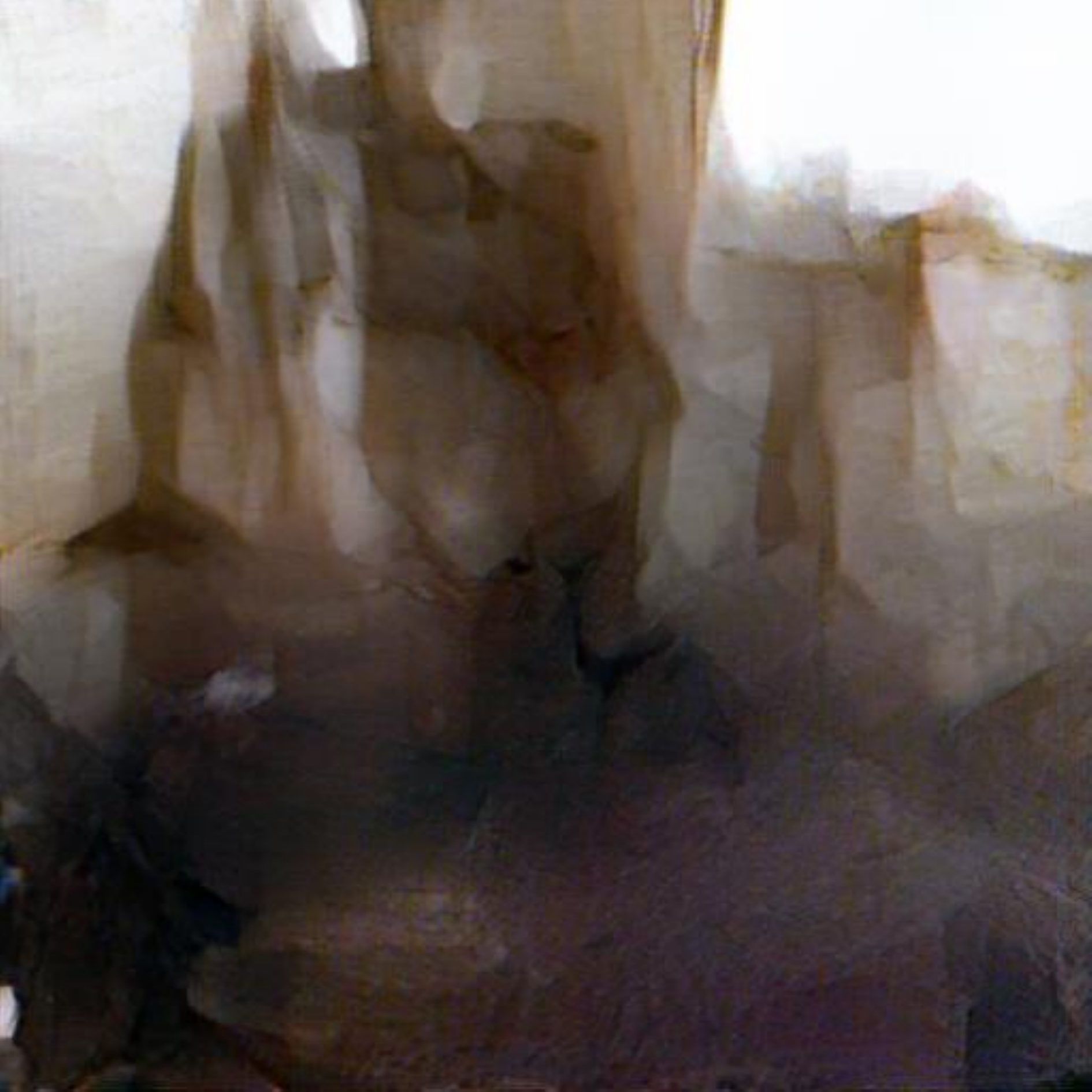}
    }
    \hspace{-3mm}
    \subfigure[PPL~\cite{lee2023ppl}]{
        \includegraphics[width=0.13\linewidth]{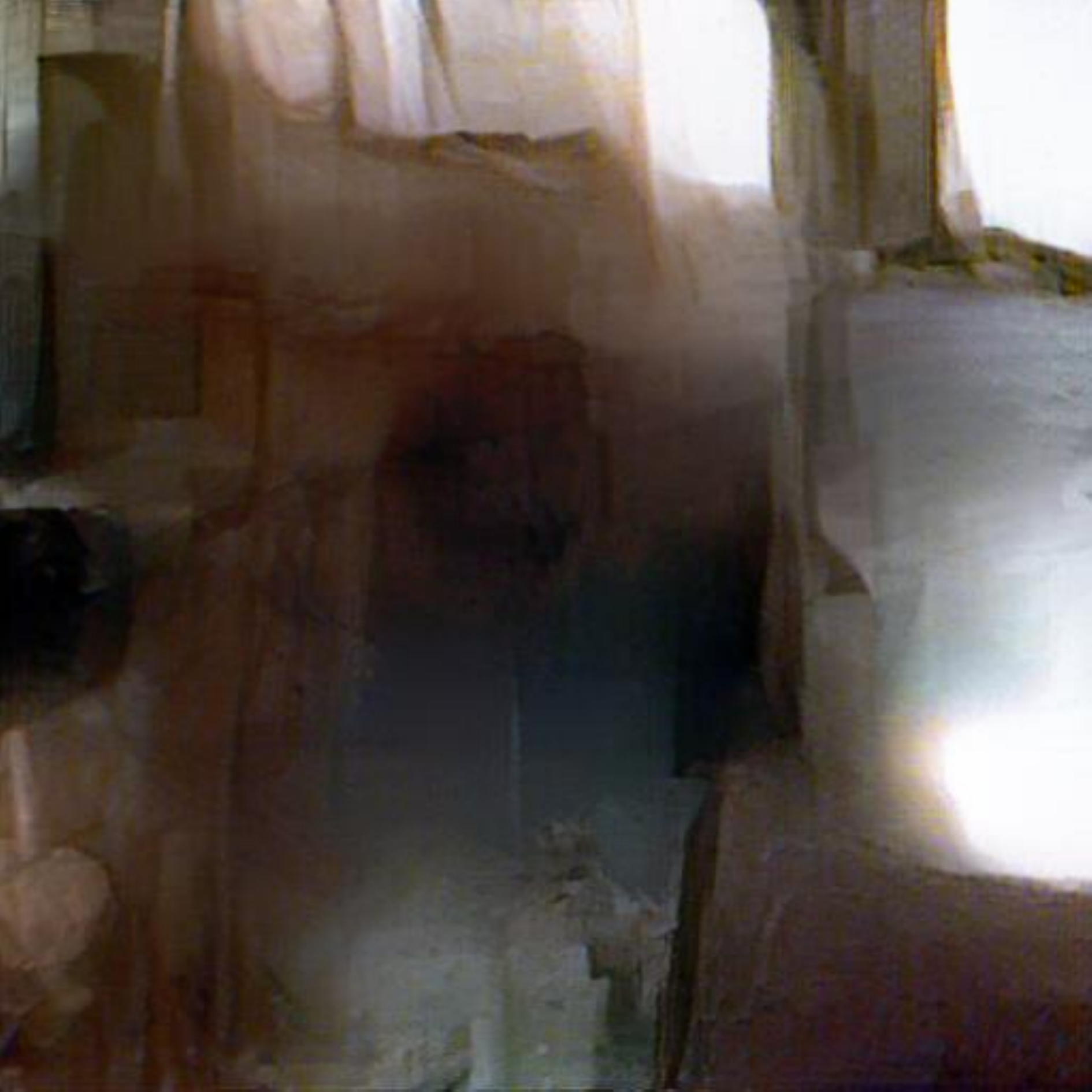}
    }
    \hspace{-3mm}
    \subfigure[PPL+~\cite{lee2023ppl}]{
        \includegraphics[width=0.13\linewidth]{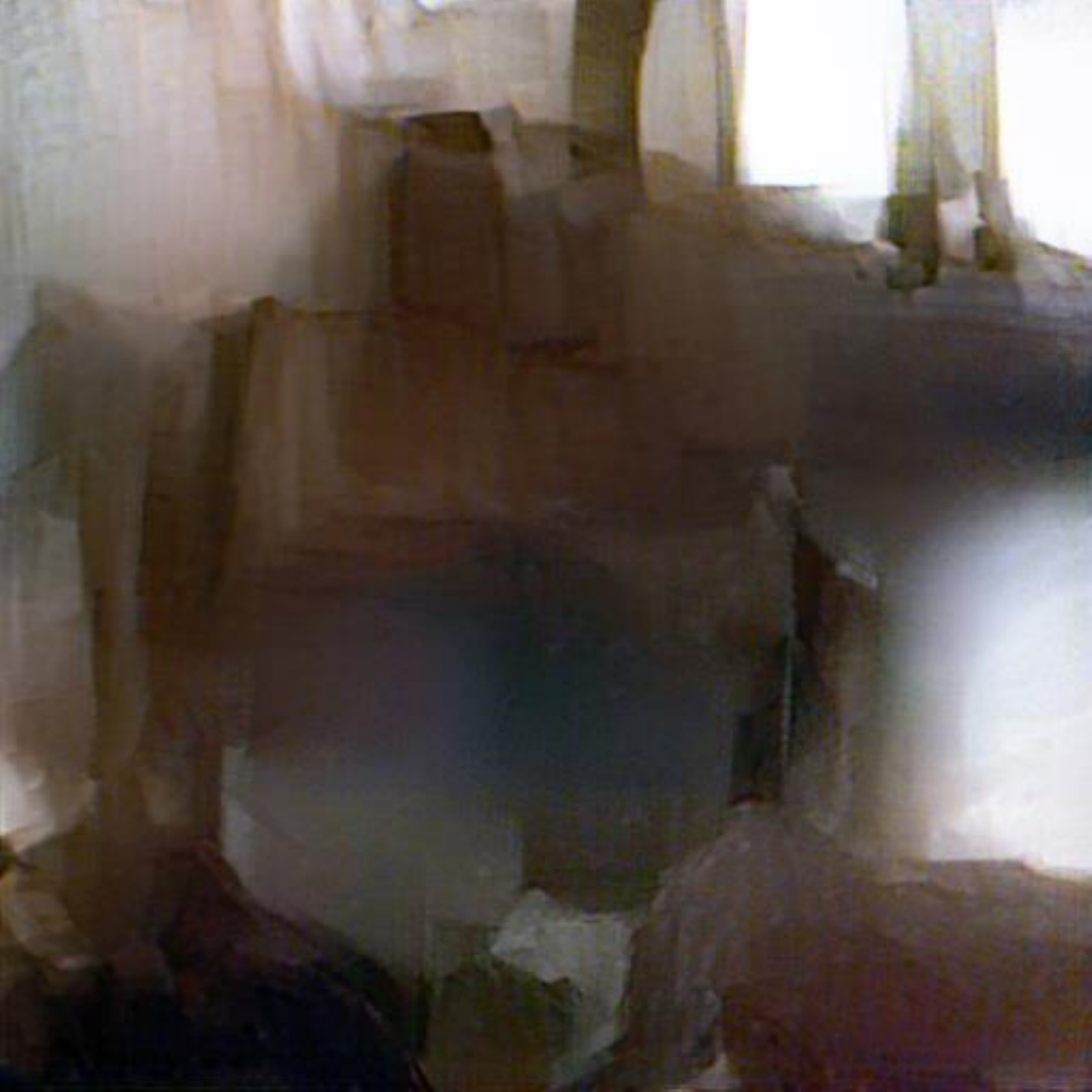}
    }
    \hspace{-3mm}
    \subfigure[Ours~(25\%)]{
        \includegraphics[width=0.13\linewidth]{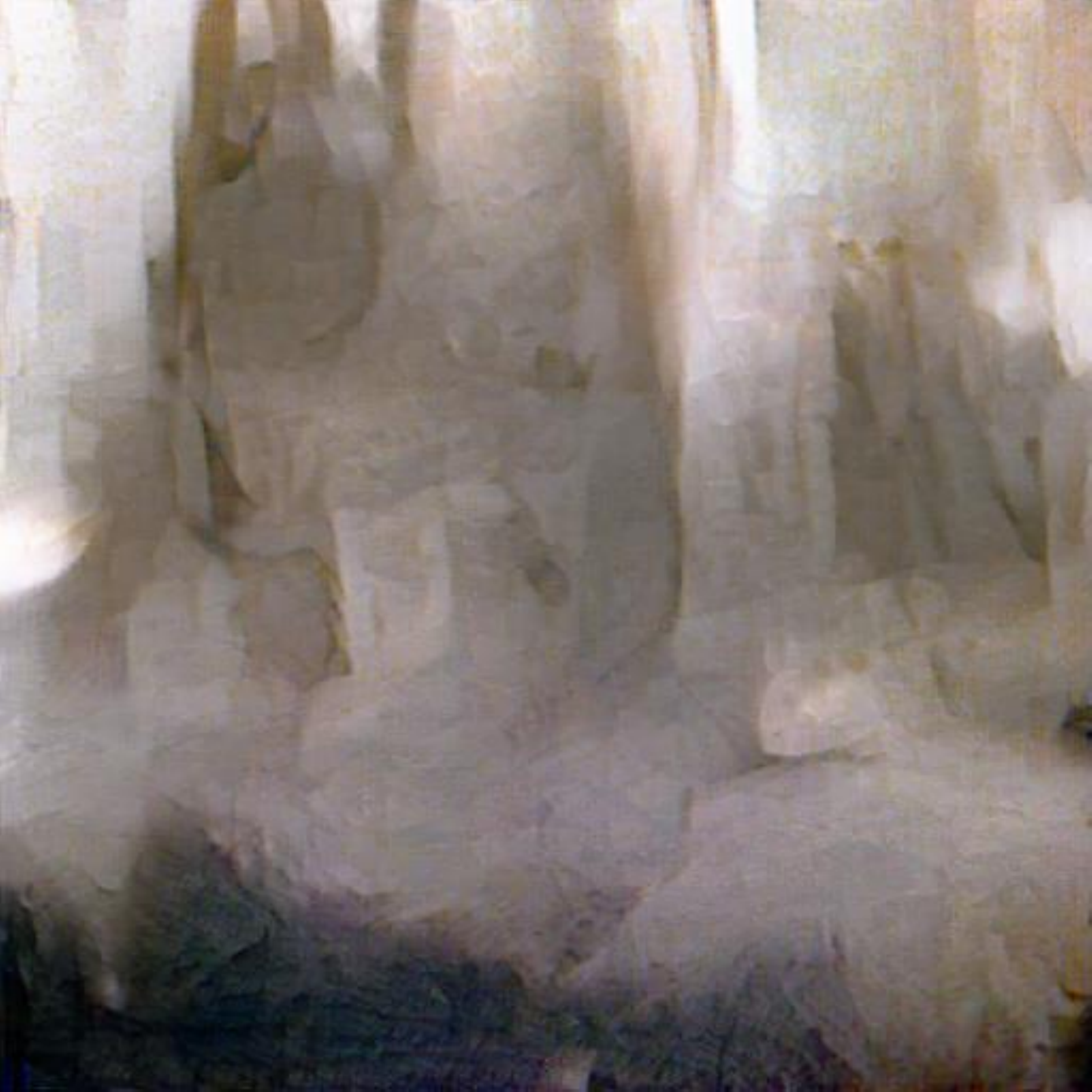}
    }
    \hspace{-3mm}
    \subfigure[Ours~(33\%)]{
        \includegraphics[width=0.13\linewidth]{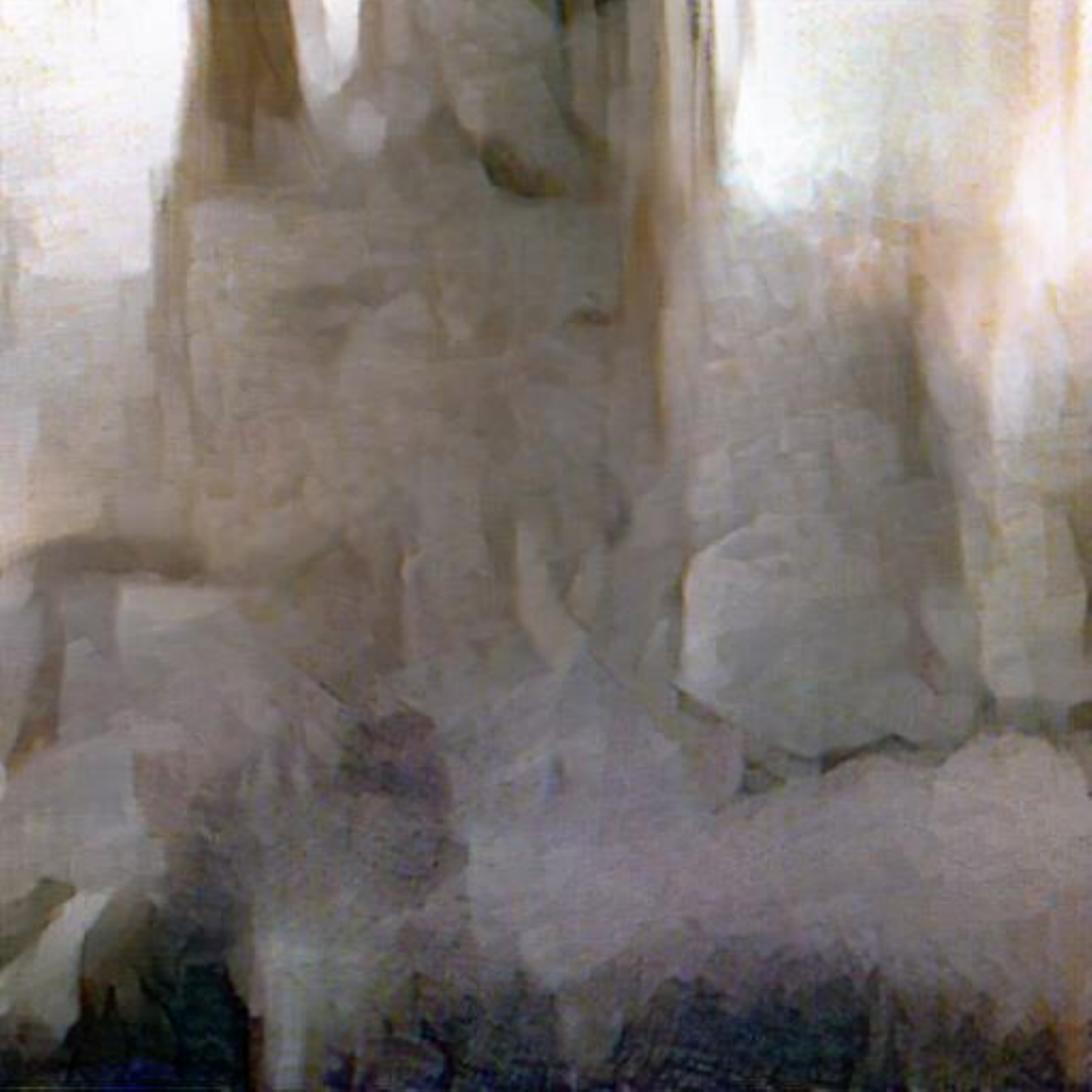}
    }
    \hspace{-3mm}
    \subfigure[Ours~(50\%)]{
        \includegraphics[width=0.13\linewidth]{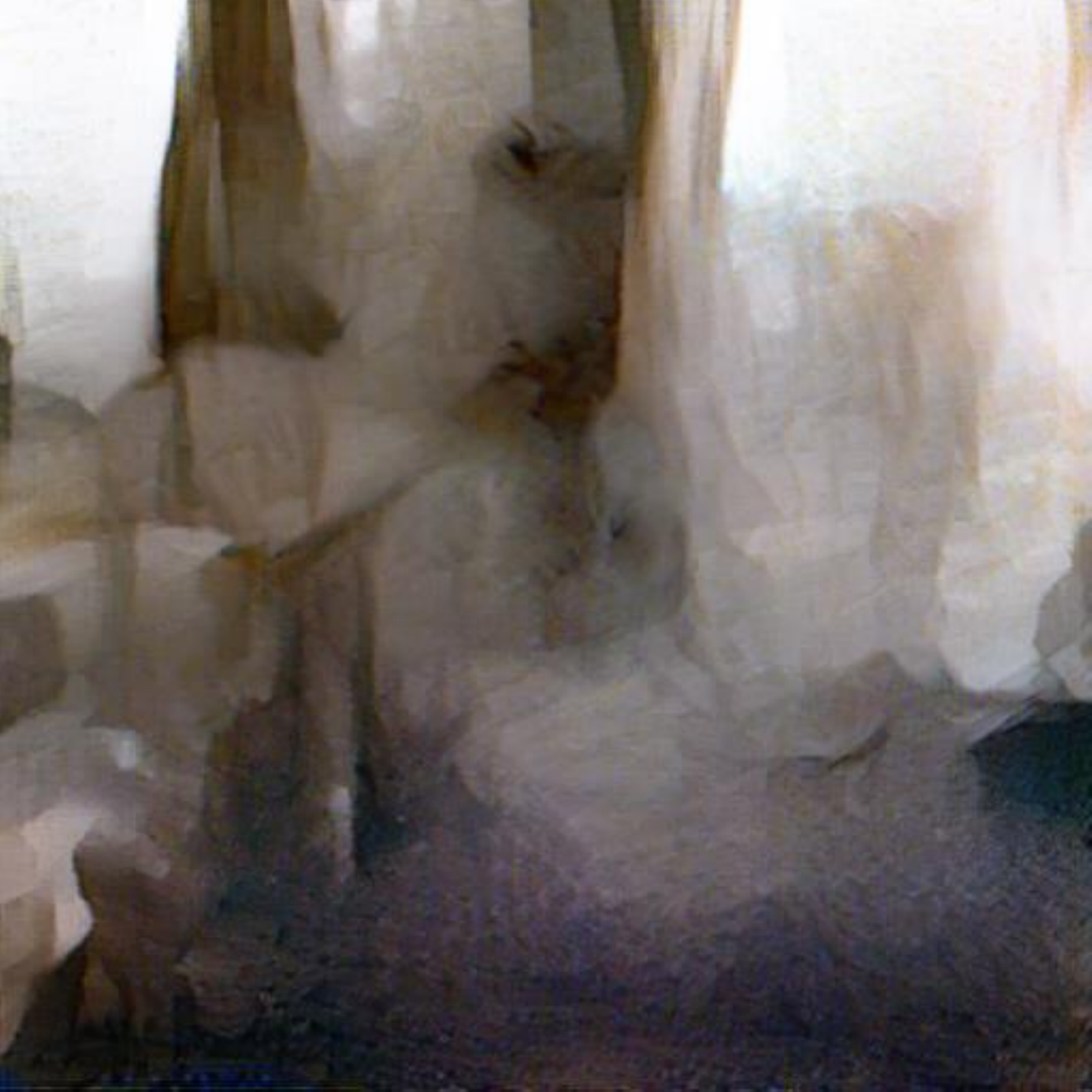}
    }
    \hspace{-4mm}
    \vspace{-1mm}
    
    \caption{Visualization of images directly reconstructed from sphere clouds about the sphere centre in 12-Scenes~\cite{valentin2016energy} (Top 4 rows) and 7-Scenes~\cite{shotton2013scene} (Bottom 4 rows).
    }
    \vspace{-2mm}
    \figlabel{fig:center inversion}
    \vspace{-4mm}
\end{figure*}


\begin{algorithm*}[!ht]
    \scriptsize
    \caption{
    Absolute pose estimation using a sphere cloud based on LO-RANSAC~\cite{chum2003loransac, PoseLib}}
	\begin{algorithmic}
        \State \textbf{Input}: Set of all correspondences $\Omega$ between 2D keypoint and 3D line from sphere cloud
        \State \textbf{Output}: Best absolute pose of query camera $\m R_{world\rightarrow query}^{*}, \v t_{world\rightarrow query}^{*}$
        \State \textbf{Given}: 
        \State $\v u \in \real^2$ $\gets$ 2D keypoints from the query image
        \State $\hat{\v x} \in S^2 \gets$ Points on the surface of unit sphere, where sphere centre is the origin of the world coordinate
        \State $\m K \in \real^{3\times3}$ $\gets$ Intrinsic of query camera
        \State $z^{TOF} \in \real$ $\gets$ Depth measurements from TOF sensor
        \State $\v p \in \real^3$ $\gets z^{TOF} \m {K}^{-1} [\v u\tr, 1]\tr:$ 3D keypoints lifted from 2D keypoints $\v u$
        \State Inlier threshold of epipolar error $\tau_{epipolar} \gets 1.5 (px) /$ Query camera focal length
        \State Inlier threshold of depth error $\tau_{depth} \gets 0.1$
        \State Depth regularization weight $\lambda$
        \State Total upper bound $\tau_{total}^2 \gets \tau_{epipolar}^2 + \lambda*\tau_{depth}^2$
        \newline
        
        \LeftComment{\textcolor{gray}{Set initial state}}        
        \State $\mathtt{POSE}_{best} \gets \emptyset$
        \State $\mathtt{num\_inliers}_{best} \gets 0$,~ $\mathtt{num\_inliers}_{LO} \gets 0$
        \State $\mathtt{MSAC\_SCORE}_{best} \gets \inf$, ~$\mathtt{MSAC\_SCORE}_{LO} \gets \inf$

        \For{$iter \gets 1~to~\mathtt{MAX\_ITER}$ \do}
            \State $\mathtt{CUR\_POSE} = [~] $
            \LeftComment{\textcolor{gray}{Draw 3 correspondences $\Omega_{i_3}$ between sphere cloud and query image}}
            \For{$\Omega_{i_3} \in \Omega$ \do}
            \State {Calculate $\mathtt{POSE}_{query\rightarrow world}$ from perspective-3-point algorithm~\cite{persson2018lambda} using $\Omega_{i_3}$}
            \State $\alpha_i \hat{\v x}_i = \m R_{q\rightarrow w} \v p_i + \v t_{q\rightarrow w}$
                \For{$[\m R_{q\rightarrow w} | \v t_{q\rightarrow w}] \in \mathtt{POSE}_{q\rightarrow w}$  \do}
                \State {Absolute pose of query camera $[\m R_{w\rightarrow q} | \v t_{w\rightarrow q}] \gets [\m R_{q\rightarrow w}\tr | -\m R_{q\rightarrow w}\tr \v t_{q\rightarrow w}]$}
                \State $\mathtt{CUR\_POSE}\text{.append}([\m R_{w\rightarrow q} | \v t_{w\rightarrow q}])$
                \EndFor
            \EndFor
            \newline
            
            \LeftComment{\textcolor{gray}{Find best local model}}
            \State Index of $\mathtt{POSE}_{best}:~{best\_index} \gets -1$
            \For{$k \gets 0$ to $\mathtt{CUR\_POSE}$.size()-1 \do}
            \State {$\v p_i \in\real^3 \gets$ $i^{th}$ 3D keypoint the in query image}
            \State {$\tilde{\v x}_i \in\real^3 \gets \hat{\v x}_i / |\hat{\v x}_{iz}|$ }
            \State {$\beta_i \in\real \gets {\v z}_i(\m R, \v t) / \v z^{TOF}_{i}$} (detailed derivation of ${\v z}_i(\m R, \v t)$ is available in Sec.~\ref{sec:details_depth_reguralization})

            \State {$\m E:=[\v e_{1}, \v e_{2}, \v e_{3}]\tr \gets$ Essential matrix between from $\mathtt{CUR\_POSE}[k]$}
            \State $\mathtt{MSAC\_SCORE}_{k} \gets
            \sum_{i\in\Omega}
            \min\left( \frac{([\v u_i\tr, 1]~\m {K}^{-\top} \m E~ \tilde{\v x}_i)^2}
            {(\v e_{1}\tr \tilde{\v x}_i)^2 + (\v e_{2}\tr \tilde{\v x}_i)^2}
            +
            \lambda*(\beta_i-1)^2
            ,\tau_{total}^2\right)$ 
            \State $\mathtt{num\_inliers}_k \gets \sum_{i\in\Omega} \mathds{1}_{\m {Inlier(i)}}$, where $\m {Inlier(i)}:=\{True~| (\frac{([\v u_i\tr, 1]~\m {K}^{-\top} \m E~ \tilde{\v x}_i)^2}
            {(\v e_{1}\tr \tilde{\v x}_i)^2 + (\v e_{2}\tr \tilde{\v x}_i)^2}
            \leq 
            \tau_{epipolar}^2)
            \cap
            ((\beta_i-1)^2 \leq \tau_{depth}^2) \}$
            \newline
            
            \LeftComment{\textcolor{gray}{Update best model status}}
            \If{$\mathtt{num\_inliers}_{LO} \leq \mathtt{num\_inliers}_{k} \textbf{ or } \mathtt{MSAC\_SCORE}_{k} \leq \mathtt{MSAC\_SCORE}_{LO}$}
            \State ${best\_index} \gets k$ 
            \State $\mathtt{num\_inliers}_{LO} \gets \mathtt{num\_inliers}_{k},~~ \mathtt{MSAC\_SCORE}_{LO} \gets \mathtt{MSAC\_SCORE}_{k}$
                \If{$\mathtt{MSAC\_SCORE}_{k} < \mathtt{MSAC\_SCORE}_{best}$}
                \State $\mathtt{num\_inliers}_{best} \gets \mathtt{num\_inliers}_{k}, ~~\mathtt{MSAC\_SCORE}_{best} \gets \mathtt{MSAC\_SCORE}_{k},
                ~~\mathtt{POSE}_{best} \gets \mathtt{CUR\_POSE}[k]$
                \EndIf
            \EndIf
            \EndFor
            \newline
            
            \LeftComment{\textcolor{gray}{If no best model is found, skip the refinement}}
            \If{$best\_index == -1$}
                \State continue
                \EndIf
            \newline

            \LeftComment{\textcolor{gray}{Non-linear (LM) local refinement}}
            \State $\mathtt{POSE}_{best}$ $\gets$ $\mathtt{CUR\_POSE}[{best\_index}]$
            \State $\mathtt{POSE}_{refined} \gets$ Non-linear (LM) optimization from $\mathtt{POSE}_{best}$
            \If{ $\mathtt{MSAC\_SCORE}_{refined} < \mathtt{MSAC\_SCORE}_{best}$}
            \State $\mathtt{num\_inliers}_{best} \gets \mathtt{num\_inliers}_{refined},~~\mathtt{MSAC\_SCORE}_{best} \gets \mathtt{MSAC\_SCORE}_{refined},~~\mathtt{POSE}_{best} \gets \mathtt{POSE}_{refined}$
            \EndIf
        \EndFor
        \LeftComment{\textcolor{gray}{Final non-linear refinement with only inliers}}
        \State $\mathtt{POSE}_{refined} \gets$ Non-linear (LM) optimization from $\mathtt{POSE}_{best}$
        \If{$\mathtt{MSAC\_SCORE}_{refined} < \mathtt{MSAC\_SCORE}_{best}$}
            \State $\mathtt{num\_inliers}_{best} \gets \mathtt{num\_inliers}_{refined},~~\mathtt{MSAC\_SCORE}_{best} \gets \mathtt{MSAC\_SCORE}_{refined},~~\mathtt{POSE}_{best} \gets \mathtt{POSE}_{refined}$
        \EndIf
        \State {\textbf{Return} $\m R_{world\rightarrow query}^{*}, \v t_{world\rightarrow query}^{*} \gets \mathtt{POSE}_{best}$}
	\end {algorithmic}
        \label{algo:sphere_initial}
\end{algorithm*}